\documentclass[thesis]{thesis-umich}
 \pdfoutput=1

\usepackage{blindtext} 
\usepackage[export]{adjustbox}
\usepackage{tabularx}
\usepackage{float}
\usepackage{wrapfig}
\usepackage{lscape}
\usepackage[inline]{enumitem}
\usepackage{boldline}
\usepackage{geometry}
\usepackage{pdfpages}
\usepackage{graphicx}
\usepackage{fullpage}
\usepackage{algpseudocode}
\usepackage{algorithm}
\usepackage{hyperref}
\usepackage{comment}
\usepackage{amsmath}
\usepackage{bm}
\usepackage{amsfonts}
\usepackage{dsfont}
\usepackage{bbm}
\usepackage{lmodern}
\usepackage{natbib}
\usepackage{amssymb}
\usepackage{appendix}
\usepackage{multirow}
\usepackage{booktabs}
\usepackage{algorithm}
\usepackage{algpseudocode}
\usepackage{hhline}

\usepackage{setspace}
\usepackage{scalerel}
\usepackage{makecell}
\usepackage{bbding}
\usepackage{pgf}
\usepackage{mdframed}
\usepackage{csquotes}
\usepackage{epigraph}
\usepackage{textcomp}
\usepackage{tikz}
\usetikzlibrary{calc}
\usepackage{xcolor}
\usepackage{colortbl}
\usepackage[most]{tcolorbox}
\usepackage{tablefootnote}
 
\usepackage{pifont}

\DeclareMathOperator*{\argmax}{\arg\!\max}

\newcommand{\fig}{Figure}
\newcommand{\figs}{Figures}
\newcommand{\tab}{Table}
\newcommand{\tabs}{Tables}
\newcommand{\sect}{Section}
\newcommand{\sects}{Sections}
\newcommand{\eq}{Equation}
\newcommand{\eqs}{Equations}

\newcommand{\chap}{Chapter}
\newcommand{\chaps}{Chapters}

\newcommand{\cmark}{\ding{51}}%
\newcommand{\xmark}{\ding{55}}%
\newcommand{\ccaintra}{{\it CCA-intra}}
\newcommand{\ccafbank}{{\it CCA-fbank}}
\newcommand{\ccaphone}{{\it CCA-phone}}
\newcommand{\ccaword}{{\it CCA-word}}
\newcommand{\ccaagwe}{{\it CCA-AGWE}}
\newcommand{\ccasyn}{{\it CCA-PTB}}
\newcommand{\ccasem}{{\it CCA-SemCor}}

\newcommand{\etal}{{\it et al.}}

\newcommand{\sfm}{SFM}
\newcommand{\sfms}{SFMs}
\newcommand{\wavtovec}{{\it wav2vec2.0}}
\newcommand{\wavtovecl}{{\it wav2vec2.0-large}}

\newcommand{\fastvgs}{{\it  FaST-VGS}}
\newcommand{\fastvgsp}{{\it  FaST-VGS+}}
\newcommand{\avhubert}{{\it AV-HuBERT}}
\newcommand{\hubert}{{\it HuBERT}}
\newcommand{\hubertl}{{\it HuBERT-large}}

\newcommand{\xlsr}{{\it XLSR-53}}
\newcommand{\wavlm}{{\it WavLM}}
\newcommand{\wavlml}{{\it WavLM-large}}
\newcommand{\datatovec}{{\it data2vec}}
\newcommand{\vghubert}{{\it  VG-HuBERT}}
\newcommand{\whisper}{{\it  Whisper}}
\newcommand{\owsm}{{\it  OWSM 3.1}}
\newcommand{\baseM}{{\it Base}}
\newcommand{\mediumM}{{\it Medium}}
\newcommand{\largeM}{{\it Large}}
\newcommand{\voxM}{{\it Vox}}

\newcommand{\randinit}{{\it rand-init}}

\newcommand{\sfl}{{\it single-frozen}}
\newcommand{\afl}{{\it weighted-frozen}}
\newcommand{\tff}{{\it top-finetune}}

\newcommand{\cka}{CKA}
\newcommand{\op}{Procrustes distance}
\newcommand{\mi}{MI}
\newcommand{\linear}{linear}
\newcommand{\cca}{CCA}
\newcommand{\vcca}{vanilla-CCA}
\newcommand{\svcca}{SVCCA}
\newcommand{\pwcca}{PWCCA}
\newcommand{\phone}{{\it phone}}
\newcommand{\word}{{\it word}}
\newcommand{\sem}{{\it semantic}}

\definecolor{lightcream}{rgb}{1.0, 1.0, 0.92}

\definecolor{gradientstart}{RGB}{255,250,240} 
\definecolor{gradientend}{RGB}{139,69,19}      

\newcommand{\cellgrad}[1]{%
  \pgfmathsetmacro{\colorgrad}{int((#1-0.56)/(1.0-0.56)*100)}%
  \edef\temp{\noexpand\cellcolor{gradientstart!\colorgrad!gradientend}%
              \noexpand\pgfmathprintnumber[fixed, precision=2, fixed zerofill]{#1}}\temp
}

\definecolor{gradientstart}{RGB}{255,255,237} 
\definecolor{gradientend}{RGB}{0, 150, 255}      

\newcommand{\maxvaluesc}{88.3}
\newcommand{\minvaluesc}{73.0}
\newcommand{\cellgradsc}[1]{%
  \pgfmathsetmacro{\colorgrad}{int((\maxvaluesc-#1)/(\maxvaluesc-\minvaluesc)*100)}%
  \edef\temp{\noexpand\cellcolor{gradientstart!\colorgrad!gradientend}%
              \noexpand\pgfmathprintnumber[fixed, precision=1, fixed zerofill]{#1}}\temp
}

\newcommand{\maxvaluesd}{74.1}
\newcommand{\minvaluesd}{57.0}
\newcommand{\cellgradsd}[1]{%
  \pgfmathsetmacro{\colorgrad}{int((\maxvaluesd-#1)/(\maxvaluesd-\minvaluesd)*100)}%
  \edef\temp{\noexpand\cellcolor{gradientstart!\colorgrad!gradientend}%
              \noexpand\pgfmathprintnumber[fixed, precision=1, fixed zerofill]{#1}}\temp
}

\newcommand{\maxvaluese}{18.2}
\newcommand{\minvaluese}{9.0}
\newcommand{\cellgradse}[1]{%
  \pgfmathsetmacro{\colorgrad}{int((\maxvaluese-#1)/(\maxvaluese-\minvaluese)*100)}%
  \edef\temp{\noexpand\cellcolor{gradientend!\colorgrad!gradientstart}%
              \noexpand\pgfmathprintnumber[fixed, precision=1, fixed zerofill]{#1}}\temp
}

\newcommand{\maxvaluesf}{75.0}
\newcommand{\minvaluesf}{56.0}
\newcommand{\cellgradsf}[1]{%
  \pgfmathsetmacro{\colorgrad}{int((\maxvaluesf-#1)/(\maxvaluesf-\minvaluesf)*100)}%
  \edef\temp{\noexpand\cellcolor{gradientstart!\colorgrad!gradientend}%
              \noexpand\pgfmathprintnumber[fixed, precision=1, fixed zerofill]{#1}}\temp
}

\definecolor{gradientstartnew}{RGB}{220,233,213} 
\definecolor{gradientendnew}{RGB}{120,166,90}      

\definecolor{verylightgreen}{RGB}{245, 250, 244}  

\newcommand{\maxvaluesg}{75.5}
\newcommand{\minvaluesg}{68.0}
\newcommand{\cellgradsg}[1]{%
  \pgfmathsetmacro{\colorgrad}{int((\maxvaluesg-#1)/(\maxvaluesg-\minvaluesg)*100)}%
  \edef\temp{\noexpand\cellcolor{gradientstartnew!\colorgrad!gradientendnew}%
              \noexpand\pgfmathprintnumber[fixed, precision=1, fixed zerofill]{#1}}\temp
}

\newcommand{\maxvaluesh}{88.6}
\newcommand{\minvaluesh}{62.7}
\newcommand{\cellgradsh}[1]{%
  \pgfmathsetmacro{\colorgrad}{int((\maxvaluesh-#1)/(\maxvaluesh-\minvaluesh)*100)}%
  \edef\temp{\noexpand\cellcolor{gradientstartnew!\colorgrad!gradientendnew}%
              \noexpand\pgfmathprintnumber[fixed, precision=1, fixed zerofill]{#1}}\temp
}

\newif\ifdraft
\drafttrue

\newif\ifdraftc
\draftctrue

\definecolor{dkgreen}{RGB}{0,179,36}
\definecolor{dkred}{RGB}{240,0,0}
\definecolor{dkblue}{RGB}{0,100,240}
\definecolor{dkorange}{RGB}{230,115,0}
\definecolor{lightorange}{RGB}{250,219,183}
\definecolor{pink}{RGB}{255,0,247}

\ifdraft

\newcommand{\hiddennote}[1]{\textcolor{red}{[{\it Hidden note: #1}]}}

\newcommand{\apremove}[1]{\textcolor{pink}{{\sout{#1}}}}
\newcommand{\klremove}[1]{\textcolor{blue}{{\sout{#1}}}}
\else

\newcommand{\apremove}[1]{{}}
\newcommand{\klremove}[1]{{}}
\newcommand{\hiddennote}[1]{}
\fi

\title{\large What do Speech Foundation Models Learn? Analysis and Applications}
\author{Ankita Pasad}
\date{Toyota Technological Institute at Chicago}

\email{ankitap@ttic.edu}

\frontpagestyle{7} 

\abstract{
Speech foundation models (SFMs) are designed to serve as general-purpose representations for a wide range of speech-processing tasks. The last five years have seen an influx of increasingly successful self-supervised and supervised pre-trained models with impressive performance on various downstream tasks.

Although the zoo of SFMs continues to grow, our understanding of the knowledge they acquire lags behind. This thesis presents a lightweight analysis framework using statistical tools and training-free tasks to investigate the acoustic and linguistic knowledge encoded in SFM layers. We conduct a comparative study across multiple SFMs and statistical tools. Our study also shows that the analytical insights have concrete implications for downstream task performance. 

The effectiveness of an SFM is ultimately determined by its performance on speech applications. Yet it remains unclear whether the benefits extend to spoken language understanding (SLU) tasks that require a deeper understanding than widely studied ones, such as speech recognition. The limited exploration of SLU is primarily due to a lack of relevant datasets. To alleviate that, this thesis contributes tasks, specifically spoken named entity recognition (NER) and named entity localization (NEL), to the Spoken Language Understanding Evaluation benchmark. We develop SFM-based approaches for NER and NEL, and find that end-to-end (E2E) models leveraging SFMs can surpass traditional cascaded (speech recognition followed by a text model) approaches. Further, we evaluate E2E SLU models across SFMs and adaptation strategies to assess the impact on task performance. 

Collectively, this thesis tackles previously unanswered questions about SFMs, providing tools and datasets to further our understanding and to enable the community to make informed design choices for future model development and adoption.

\newpage

}

\acknowledgments[7]{This thesis has been made possible through the support, encouragement, teaching, and kindness of many, from the TTIC community and my collaborators to internship mentors, early research guides, friends, and family. I take this opportunity to thank everyone who has been part of this journey, and if I have missed a name, please know that your impact has been felt and is deeply appreciated.

I am deeply grateful to my advisor, Karen Livescu, whose meticulous guidance, deep expertise, and remarkable work ethic have shaped me both as a researcher and as a person. Karen has been my most constructive critic while always having my back.
I am thankful for the freedom she gave me to explore and the confidence she placed in me as I stumbled and gradually found my footing, especially in the initial years. I have often been inspired by her ability to see the big picture while also giving attention to detail and technical finesse, and I aspire to carry these qualities forward. I have learned a great deal from Karen, from coursework and research to leading with focus, composure, and kindness. Beyond the academic, she has made me feel welcome in countless ways, from hosting lovely summer get-togethers and giving challah-making tutorials to ensuring I was wilderness-ready before my first camping trip. 

I am also grateful to my committee members---Michael Auli, Allyson Ettinger, and Kevin Gimpel---for their time, expertise, and thought-provoking discussions.
Michael’s pioneering contributions have helped advance the field of speech foundation models, the very topic at the heart of this thesis.
Allyson’s passion for uncovering how text models understand language and the mechanisms that shape their linguistic capabilities has been an inspiration ever since I got to know her as a research assistant professor at TTIC, and it sparked my interest in this area and shaped my choice of thesis topic. Kevin has been a valuable presence throughout my time at TTIC, from serving on my qualifying committee to teaching two of the most engaging courses I have taken, to being my go-to person for any NLP-related queries. I am grateful to all three for the many ways they have helped shape this work.

I am thankful to my internship mentors and collaborators---Ariel Gordon, Tsung-Yi Lin, and Anelia Angelova at Google, and Suwon Shon, Felix Wu, Pablo Brusco, and Kyu J Han at ASAPP---for sharing their expertise and guidance, which helped me make the most of my short internship experiences. I had the pleasure of collaborating with Suwon and Felix on multiple projects that ultimately became a part of my thesis. I have also been fortunate to work with many brilliant collaborators---Bowen Shi, Herman Kamper, Ju-Chieh Chou, Shuo Xie, Hongyuan Mei, Siddhant Arora, Roshan Sharma, Chung-Ming Chien, Shane Settle, and Kwanghee Choi. Thank you for all the insightful discussions and for sharing both the “aha” moments and the “hmm, what could be going wrong?” moments. 

Beyond these collaborations, I have greatly benefited from the warm and inspiring intellectual community at TTIC. I am grateful to the professors for their teaching, mentorship, and encouragement, both in and out of the classroom: Madhur Tulsiani, Mesrob Ohannessian, Greg Shakhnarovich, Yury Makarychev, David McAllester, Kevin Gimpel, Julia Chuzhoy, Matt Walter, Ben Zhao (University of Chicago), Nati Srebro, Avrim Blum, Matthew Turk, and Sadaoki Furui. I will always remember Sadaoki with great fondness and feel deeply fortunate to have experienced his warmth, wisdom, and high-energy spirit; as one of the pioneers of speech research and a former president of TTIC, his legacy and inspiring presence continue to resonate with me. 

TTIC’s administrative staff has kept the institute a remarkably bureaucracy-free place. Special thanks to Adam Bohlander, who was always just a message away for hardware and cluster-related issues. I have also learned a great deal from Adam about being a good SLURM citizen. I am grateful to Erica Cocom and Chrissy Coleman—my go-tos for administrative questions—for consistently making things easy. I appreciate Mary Marre for always going the extra mile for the student community, often checking in and making sure we were well fed. And thank you to Amy Minick for making immigration-related processes as stress-free as possible. Thanks to the administrative staff, logistical hurdles were few and far between and never became research hurdles.

I am grateful to my fellow colleagues from my cohort and senior cohorts who helped make my transition to graduate school smooth---Chip, Hai, Hao, Harris, Lifu, Lingyu, Mingda, Nick, Omar, Pedro, Rachit, Reza, Ruotian, Somaye, Sudarshan, Srinadh, Suriya, and Takeshi. From coursework TA sessions and group studies to late-night problem-solving, and from answering my PyTorch questions to helping me find my way around Hyde Park, I always found the TTIC student and RAP community welcoming and approachable. I have also enjoyed the camaraderie and valuable discussions with fellow speech and language researchers, Bowen, Chung-Ming, David, Freda, Jiawei, Ju-Chieh, Karl, Kartik, Puyuan, Qingming, Sameer, Shane, Shubham, Shuning, Yanhong, and Yushi. These exchanges have been invaluable for day-to-day problem solving, refining paper drafts, and practicing dry runs. Special thanks to Shubham for bringing to my attention the initial related work that sparked the analysis direction in my thesis. I also extend special thanks to Jiawei for stepping in as an interim advisor while Karen was away on her sabbatical; I appreciated his genuine interest in my projects and enjoyed our discussions on research and beyond. 

Before TTIC, my time at IIT Bombay gave me invaluable opportunities to learn and ultimately discover my passion for speech processing, for which I am deeply grateful. Alongside the excellent course lecturers, I especially thank Prof. Preeti Rao for introducing me to the world of speech research and guiding me through my Master’s thesis. I am grateful to my DAP Lab colleagues---Hitesh, Kamini, Divyansh, Prakhar, Kaustuv, Vinutha, and Shruti---for the stimulating discussions that guided my early journey in this field, for their steady support along the way, and for the friendships that have endured long past our time at DAP Lab. I am also indebted to Prof. Preethi Jyothi for her timely mentorship, which played a significant role in my decision to pursue a Ph.D. at TTIC. I would also like to thank my high school teachers---Ms. Dani, Mr. Parulekar, and Mr. Moghe---who taught me many interesting fundamentals in a way that sparked my interest in science and math. Their belief in me meant a great deal to my thirteen-year-old self. 

Over the years, I’ve been fortunate to build friendships that have brought joy, perspective, and balance to my journey.
Whether it was weekend hikes, board game nights, sharing home-cooked meals, reunions during visits, travel adventures, long calls, or a simple “How’s it going?”, these moments brought warmth, laughter, and kept me grounded.
Special thanks to Frances, Nidhi, Rachit, Shubham, and Sudarshan for making me feel welcome when I first moved to Chicago and easing my transition to life away from home. I have (almost :) always valued Shubham’s unfiltered opinions and sharp insights, and even more so, the close friendship we’ve shared over the years.  I will forever be grateful to Prachi and Tamaghna, whose conscious decision to stay in Chicago as my flatmates in early 2020 spared me from facing the ensuing uncertain times alone, and whose warmth, generosity, and sense of fun have brightened so many moments together. To my Chicago friends---Naren, Kshitij, Ahona, Shashank, Kavya, Anmol, Pushkar, Keziah, and Han---I’m glad we could spend meaningful time together, even during my shorter post-pandemic visits. I am deeply thankful for the friendships I have built with Ashlesha, Sid, Aditi, Pari, Akash, Kamal, Dewa, Purav, ADT, Agneya, and friends from my fellow STAB family, for standing by me through many years, for always being just a call away, and for being a constant reminder of all that life has to offer beyond work. My thanks also go to my Focusmate partners for the accountability and cheers that turned work hours into shared effort, especially during the pandemic years, and to ChatGPT, my ever-patient sounding board and brainstorming companion through the thesis-writing process.

I am grateful to my Mamas---Jay, Chirag, Gaurav, and Vijay---whose hard work, integrity, and generosity shaped my earliest ideas of success and gave me something to aspire to. Seeing them chart their own paths early in my life offered both an inspiration and a grounding sense of what I wanted to strive for. I also feel fortunate to have Maa and Dad as a second set of parents, whose care, encouragement, and belief in me have been a constant source of support. I owe a warm mention to my feline flatmates, Hubert, Mordecai, and now Pepper, for their unflagging commitment to supervising my work, warming my lap, and reminding me to take breaks.

To Riddhish, my constant and my catalyst, thank you for believing in me, always. 
To Mummy and Papa, thank you for everything.
Every milestone I have reached is as much yours as it is mine.
}

\dedicationstyle{\@empty}  
\dedication{
  {\itshape To Mummy and Papa.}
}

\committee{ %
Karen Livescu (Thesis Advisor)\par
Michael Auli \par
Allyson Ettinger \par
Kevin Gimpel 
}

\chair{Karen}

\showlistoftables

\begin{document}

\newpage
\section*{Bibliographic Note}

Parts of this thesis are based on prior peer-reviewed publications.
All the contents of
\chap~\ref{ch:slue}~\cite{shon2022slue, shon2023slue},
\chap~\ref{ch:ner-ext}~\cite{pasad2021use}, and
\chap~\ref{ch:slueperb}~\cite{arora2024evaluation}
are from prior published papers.
\chap~\ref{ch:analysis}~\cite{pasad2021layer, pasad2023comparative, pasad2023self},
\chap~\ref{ch:compare-tools}~\cite{pasad2021layer, pasad2023comparative} and 
\chap~\ref{ch:implications}~\cite{pasad2021layer, pasad2023comparative} have some parts being presented for the first time in this thesis. 

\newpage
\chapter{Introduction}
\label{ch:intro}


The last fifteen years have witnessed extraordinary progress in deep learning, from pioneering achievements in image classification~\cite{krizhevsky2012imagenet} to approaching human-like performance on speech recognition~\cite{xiong2016achieving} to the most recent widespread adoption of chatbots by hundreds of millions~\cite{chatgpt2024stats}. Through these advancements, carefully designed spectral representations of speech from decades ago~\cite{flanagan1972speech} have stood the test of time and have maintained their utility in successful speech technologies.
However, in the last few years, representations from speech foundation models (\sfms) have proven their effectiveness by almost ubiquitously replacing the spectrogram features~\cite{mohamed2022self, borgholt2022brief}.

\begin{figure}[ht]
\footnotesize
\centering
    \includegraphics[width=0.7\textwidth]{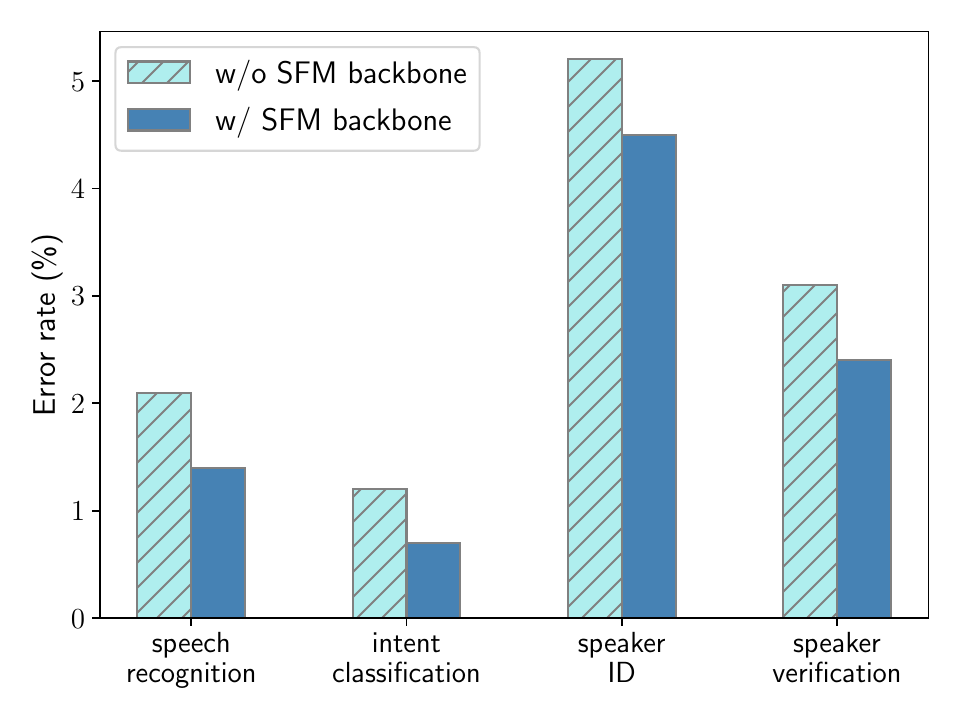}
    \caption{State-of-the-art for several speech processing tasks, with and without using a speech foundation model as a backbone, as reported in a survey by Mohamed et al.~\cite{mohamed2022self}. The respective error rates are on the y-axis: word error rate for speech recognition~\cite{zhang2020pushing}, miss rate for intent classification~\cite{chen2022unispeech}, equal error rate for speaker verification~\cite{wang2021fine}, and miss rate for speaker ID~\cite{chen2022wavlm}.
    }
    \label{fig:sota}
\end{figure}

The most common \sfms \ are self-supervised speech models optimized for a proxy task designed from unlabeled speech data~\cite{baevski2020wav2vec, ling2020decoar, chung2020generative}. Another class of \sfms \ is trained with some form of supervision, such as audio paired with images~\cite{peng2022fast} or videos of lip movements~\cite{shi2022learning} or text transcripts~\cite{radford2023robust, peng2023reproducing}.  
These pre-trained representations have been successfully incorporated into task-specific models spanning a variety of applications~\cite{mohamed2022self, feng2023superb}, including classification tasks such as speaker identification~\cite{chen2022wavlm}, intent classification~\cite{yang2021superb, chemudupati2023transferability}, and emotion recognition~\cite{feng2023peft}, speech-to-text tasks such as speech recognition~\cite{baevski2020wav2vec, hsu2021hubert} and speech translation~\cite{tsai2022superb}, as well as generative tasks such as voice conversion, speech enhancement, and speech synthesis~\cite{tsai2022superb, lakhotia2021generative, nguyen2023generative}. \fig~\ref{fig:sota} shows some example tasks for which task-specific models built on top of \sfms \ outperform previous state-of-the-art (SOTA) models. 

It is remarkable that \sfms \ seem to learn general-purpose representations, but these empirical successes alone provide limited insights. For instance, all four of the SoTA models presented in \fig~\ref{fig:sota} are contributed by independent research groups and use different \sfms \ and different adaptation strategies; this may suggest that different backbone models are specialized for specific tasks, but such a hypothesis cannot be validated by anything short of an extensive experimental study. Additionally, downstream task scores do not reveal the nature of generic speech properties learned during pre-training, nor do we learn how such knowledge is distributed across the layers of a deep network. Such insights can further our understanding of \sfms \ and help us navigate the ever-growing space of large pre-trained models~\cite{mohamed2022self}.



{\it To that end, this thesis advances our understanding of speech foundation models by addressing previously unanswered questions}: (i) Our lightweight analysis framework facilitates a quick discovery of properties encoded in \sfms' pre-trained representations and offers concrete takeaways for \sfms' adaptation to specific downstream tasks, (ii) Our spoken language understanding evaluation (SLUE) benchmark and the accompanying extensive evaluation provide insights into the efficacy of \sfms \ as a backbone for end-to-end language understanding models.

Our analysis framework studies \sfms' hidden representations using fast-to-compute intrinsic methods, such as training-free tasks and subspace analysis tools, to evaluate knowledge encoded in hidden representations (\chap~\ref{ch:analysis}).
While our study focuses mainly on acoustic and linguistic properties, the analysis framework is generic and easily scalable. For instance, our framework has inspired follow-up work studying speaker and language-specific characteristics~\cite{monteiro2023towards}, paralinguistics~\cite{li2023exploration}, similarities with brain prediction signals~\cite{chen2023self}, and also to study music representations~\cite{ragano2022learning} and text representations~\cite{xie2022hidden}. 


A stand-alone set of findings from neural analysis can suffer from confirmation bias~\cite{leavitt2020towards, jain2019attention, wiegreffe2019attention}. We mitigate such pitfalls with the help of baselines and by corroborating our findings with multiple evaluation methods (\chaps~\ref{ch:analysis} and \ref{ch:compare-tools}).
It is further encouraging to see that other subsequent studies confirm our findings via alternate analysis tools and tasks~\cite{sanabria2023analyzing,lin2023utility,shih2023speechclip,shi2023ml,ashihara2023speechglue,huang2023low,lin2022melhubert,lodagala2023pada,yang2023device,singh2023decoding,martin2023probing,bartley2023accidental,wang2023comparative,zhao2022improving,cho2023evidence,shi2023ml,ashihara2024unveiling,wiepertspeech,abdullahwave}.

We also present the implications of our task-agnostic findings for designing model adaptation strategies (\chap~\ref{ch:implications}). Our proposed adaptation strategy is more efficient in terms of either run time or performance and has been adopted in follow-up studies using newer \sfms~\cite{peng2022word, steinmetz2023transfer}.
Other follow-up work has proposed modeling decisions for improved pre-training or adaptation strategies motivated by our findings~\cite{zaiem2023fine,baevski2022data2vec,pasad2021layer,pasad2023comparative,chang2022distilhubert,hwang2022pseudo,berrebbi2022combining,zhu2021wav2vec,xue2023sshr,dorszewski2024convexity,chang2024exploring, Li2023parameterefficient}. 
The follow-up work either finds direct utility of our proposed methods~\cite{zhu2021wav2vec, xue2023sshr} and layer-specific findings~\cite{chang2024exploring, berrebbi2022combining, Li2023parameterefficient, hwang2022pseudo} or takes inspiration from our general observation of non-uniform distribution of knowledge across \sfm \ layers, by choosing to drop layers~\cite{zaiem2023fine, xue2023sshr, dorszewski2024convexity} during fine-tuning or use multiple layers (and not just the final layer, as is common otherwise) for distillation during pre-training~\cite{chang2022distilhubert, baevski2022data2vec}. 
A related study on text foundation models finds that layer-wise task specificity, measured using a similar framework as ours, can guide design choices for adapting text \sfm \ for text classification tasks~\cite{xie2022hidden}. 

Ultimately, the utility of an \sfm \ is quantified by its ability to address speech applications. 
To facilitate the evaluation of \sfms \ on complex spoken language understanding tasks, we contribute spoken named entity recognition (NER) and localization (NEL) to the open-source spoken language understanding evaluation (SLUE) benchmark (\chap~\ref{ch:slue}).
We evaluate \sfms \ as a part of both end-to-end (E2E) and cascaded (speech recognition followed by a text understanding model) language understanding models and find that, on leveraging external unannotated data, E2E NER models can surpass traditional cascaded approaches (\chap~\ref{ch:ner-ext}).
Further, we present an extensive evaluation of E2E \sfm-based language understanding models for SLUE tasks and find how the choice of \sfm \ and the adaptation strategy impacts the downstream NER and NEL performance (\chap~\ref{ch:slueperb}).
The SLUE benchmark has been extensively used for evaluating the semantic capability of novel pre-training~\cite{xia2023grass, peng2023study, wu2023wav2seq, xu2022introducing} and adaptation strategies~\cite{pasad2021use, he2023zero, arora2022token, peng2023comparative, arora2023integrating}.

Next, I discuss the contributions made by this thesis in more detail.



\section{Contributions}

\begin{enumerate}[leftmargin=*]
    \item {\bf Lightweight analysis of speech foundation models}\\ 
    We contribute a generic and lightweight analysis framework to study intermediate layers of \sfms. We evaluate learned representations using training-free tasks and analyze the representation subspace using canonical correlation analysis (CCA)~\cite{morcos2018insights} to measure acoustic and linguistic properties encoded across an \sfm's layers and representations of consecutive frames from a layer. Compared to popular parametric probing methods~\cite{ma2021probing,shah2021all}, CCA is computationally inexpensive and has a closed-form solution, making it quick to compute and scalable. More specifically, 
    \begin{enumerate}
        \item \sloppy We contribute the first such large-scale comparative analysis across several recently proposed \sfms, including self-supervised English speech models (wav2vec2.0 Base and Vox~\cite{baevski2020wav2vec}, HuBERT Base and Large~\cite{hsu2021hubert}, wavLM Base and Large~\cite{chen2022wavlm}, data2vec Base and Large~\cite{baevski2022data2vec}),  self-supervised multi-lingual speech models (XLSR-53~\cite{conneau2020unsupervised}), visually grounded speech models (FaST-VGS~\cite{peng2022fast}, FaST-VGS+~\cite{peng2022self}, AV-HuBERT~\cite{shi2022learning}, VG-HuBERT~\cite{peng2022word}), and supervised \sfms \ (Whisper Tiny, Base, Small, Medium, and Large~\cite{radford2023robust}).
        \item Our framework discovers properties related to acoustics, phonetics, and various word-level attributes, such as identity, pronunciation, syntax, and semantics. We also evaluate \sfms \ on training-free tasks, such as unsupervised word segmentation, word discrimination, and semantic utterance similarity, corroborating findings from our task-agnostic analysis.
        \item We compare multiple analysis tools and investigate how well layer-wise property trends can reliably guide the choice of layer for downstream tasks when using a frozen \sfm \ as a representation extractor. 
        Specifically, we perform layer-wise analysis of six \sfms \ using canonical correlation analysis~\cite{harold1936relations} and its variants~\cite{raghu2017svcca, morcos2018insights}, discrete mutual information, centered kernel alignment~\cite{kornblith2019similarity}, orthogonal Procrustess~\cite{everson1998orthogonal, schonemann1966generalized}, and linear classification. 
        \item  The analysis codebase is available at \href{https://github.com/ankitapasad/layerwise-analysis}{https://github.com/ankitapasad/layerwise-analysis}.
    \end{enumerate}   


    \item {\bf Implications for downstream task adaptation} \\
    We explore how our analytical findings can guide task-specific adaptation. 
    \begin{enumerate}
        \item When using a frozen \sfm \ as a feature extractor, we find that a single layer can match the performance of using all layers for various tasks; and our task-agnostic analytical findings can guide the decision of choosing the most effective layer.
        \item When fine-tuning an \sfm, specifically wav2vec2.0, for a downstream task model, we find that forgetting (re-initializing) uninformative layers, chosen based on our analysis results, before adaptation improves speech recognition for limited-data settings.
        \item When adapting \sfms \ using parameter-efficient fine-tuning, we find that placement of LoRA modules~\cite{hu2021lora} on only a handful of layers can match the performance of placing them on all layers. The optimal choice of layers does not have an obvious connection to our analysis findings, leaving this direction ripe for further exploration. 
    \end{enumerate}
    
    
    \item {\bf Spoken language understanding}\\
    We identify a lack of freely available natural spoken language understanding (SLU) datasets, consequently limiting the evaluation of \sfms \ for SLU tasks. So, we advance this line of work via the following:
    \begin{enumerate}
        \item We contribute spoken named entity recognition (NER) and named entity localization (NEL) tasks to the open-source spoken language understanding evaluation (SLUE) benchmark, which currently has six language understanding tasks, available on HuggingFace.\footnote{\href{https://huggingface.co/datasets/asapp/slue}{https://huggingface.co/datasets/asapp/slue}, \href{https://huggingface.co/datasets/asapp/slue-phase-2}{https://huggingface.co/datasets/asapp/slue-phase-2}.}
        \item We provide open-source SLU baselines using state-of-the-art foundation models for both popular approaches---cascaded (speech-to-text followed by text processing) and end-to-end (E2E)---to encourage further research in both directions.\footnote{ \href{https://github.com/asappresearch/slue-toolkit}{github.com/asappresearch/slue-toolkit}}
        \item We improve the baselines by leveraging external unannotated data, and we find that E2E NER outperforms the cascaded approach with comparably strong backbone models.\footnote{\href{https://github.com/asappresearch/spoken-ner}{github.com/asappresearch/spoken-ner}}
        \item We conduct a comprehensive evaluation of both self-supervised and supervised \sfms \ as E2E language understanding models using different adaptation strategies.\footnote{\href{https://github.com/espnet/espnet/pull/5685}{https://github.com/espnet/espnet/pull/5685}} 
    \end{enumerate}
\end{enumerate}

This thesis advances the research on speech foundation models (\sfms) by providing the tools and datasets needed to further our understanding of \sfms \ and enable the community to make informed design choices for future model development and adoption.

\section{Chapter guide}
The remaining chapters of this thesis are organized as follows:
\begin{itemize}
    \item \chap~\ref{ch:background} covers the relevant technical background on \sfms \ and analysis tools, discusses related prior work analyzing neural models, and provides a technical background on statistical analysis tools.
    \item \chap~\ref{ch:analysis} presents our analysis framework and findings from comparative layer-wise analysis of various \sfms \ using canonical correlation analysis and training-free tasks.
    \item \chap~\ref{ch:compare-tools} extends the analysis framework from the previous chapter to multiple statistical analysis tools and studies the correlation of findings with downstream task performance, eventually presenting a set of robust and reliable metrics.
    \item \chap~\ref{ch:implications} discusses the connection between our findings from the previous chapters and different adaptation strategies for \sfms \ and proposes a case study of an improved fine-tuning mechanism. 
    \item \chap~\ref{ch:slue} presents the Spoken Language Understanding Evaluation (SLUE) benchmark and \sfm-based baselines for end-to-end (E2E) and cascaded approaches to spoken named entity recognition (NER) and localization (NEL).
    \item \chap~\ref{ch:ner-ext} leverages external unannotated data to improve low-resource NER performance, with E2E models surpassing a cascaded approach. 
    \item \chap~\ref{ch:slueperb} presents an extensive evaluation of \sfms \ as a backbone for E2E NER and NEL models, comparing different \sfms \ and adaptation strategies.
    \item \chap~\ref{ch:conclusion} presents concluding remarks and discusses research directions worth exploring further. 
\end{itemize}

\newpage
\chapter{Background and Related Work}
\label{ch:background}

This thesis studies pre-trained \sfms \ by analyzing the properties encoded in their hidden representations and evaluating efficacy on various language understanding downstream tasks. In this chapter, we introduce \sfms \ (\sect~\ref{sec:background-sfm}), provide an overview of prior work on analysis and interpretation of neural models (\sect~\ref{sec:background-analysis}), and present a technical background on the analysis tools relevant for our work (\sect~\ref{sec:background-tools})


\section{Speech foundation models}
\label{sec:background-sfm}
Over the past decade, supervised speech models have provided impressive gains for applications with access to rich labeled data~\cite{hinton2012deep,lecun2015deep,bourlard2012connectionist}. Naturally, this progress is restrictive for languages and domains with limited labeled resources. To address this limitation, the concept of a {\it foundation model} has emerged, referring to a generic model trained on large-scale data that can be adapted to a wide range of downstream tasks, thus also encouraging re-usability and scalability~\cite{foundationmodels}. Although the term, ``foundation model" is relatively new~\cite{bommasani2021opportunities}, the past few years have seen a notable increase in the development of self-supervised speech models pre-trained on extensive unlabeled data~\cite{mohamed2022self}. 

Next, we will dive deeper into the pre-training (\sect~\ref{sec:background-sfm-pre}) and adaptation strategies (\sect~\ref{sec:background-sfm-adapt}) for speech foundation models while outlining the specific models and methods studied in this thesis.

\subsection{Pre-training}
\label{sec:background-sfm-pre}
A self-supervised speech model is trained to solve a pretext task, which involves optimizing for artificially designed ({\it input}, {\it output}) pairs derived from raw audio data. These modern-day self-supervised models share the motivation with much earlier research on unsupervised speech representation learning~\cite{wang2015unsupervised, Chung2016AudioWord2Vec, oord2018representation, milde2018unspeech, chung2018speech2vec, chung2019unsupervised, pascual2019learning, hsu2017unsupervised}, but they have demonstrated broader applicability across various tasks. For example, \fig~\ref{fig:sota} shows several quantitative results, and one such self-supervised model has also led to a competitive state-of-the-art for unsupervised speech recognition~\cite{baevski2021unsupervised}.

Building upon the success of self-supervised speech models~\cite{baevski2020wav2vec,ling2020decoar, liu2020mockingjay, hsu2021hubert, chen2022wavlm, baevski2022data2vec, BEST-RQ}, pretext tasks have been expanded to encompass multi-modal settings, such as visual grounding with images paired with spoken captions~\cite{peng2022fast, peng2022self, peng2022word} and lipreading datasets~\cite{shi2022learning, shi2022robust}, and textual grounding via limited supervision~\cite{bapna2021slam, sainath2023joist} or weak supervision~\cite{radford2023robust} as well as supervised settings~\cite{cheng2023mu}. Additionally, some of these monolingual models have been extended to support multiple languages~\cite{conneau2020unsupervised, babu2021xlsr, bapna2022mslam}. 
Given the diversity of pre-training techniques, we collectively refer to these pre-trained speech models as ``speech foundation models” (\sfms), in alignment with current terminology.

\sfms \ are generally pre-trained using one of three types of objective function\footnote{This categorization is offered by Mohamed et al. in their survey paper~\cite{mohamed2022self}}: generative~\cite{chung2018speech2vec, ravanelli2020multi, chung2020generative, ling2020decoar, liu2020mockingjay, chi2020audio}, contrastive~\cite{milde2018unspeech, oord2018representation, schneider2019wav2vec, baevski2019vq, baevski2020wav2vec, jiang2020speech, zhang2020pushing, chung2021w2v}, and predictive~\cite{hsu2021hubert, chen2022wavlm, baevski2022data2vec, BEST-RQ}. 
{\it Generative approaches} reconstruct the input data, typically using incomplete data as input; for instance, predicting future samples from the past (auto-regressive prediction) or original audio from a corrupted version (addition of noise or masking). 
In contrast to this reconstruction-based objective, {\it contrastive approaches} project the input to a learned representation space such that certain desirable attributes (e.g., spoken content) are preserved while being invariant to other properties of the input (e.g., speaker attributes, background noise). This organization of the representation space is obtained by the choice of ``contrasting" samples in the pre-training objective.
Lastly, the {\it predictive approaches} are trained with a prediction loss, where the discrete target label is obtained using a separate model or a previous iteration of the same model.  
With a small number of exceptions~\cite{wu2023wav2seq}, most self-supervised speech models are encoder-only models.
Encoder-decoder architectures are more common when pre-training with textual grounding, resulting in supervised \sfms~\cite{zhang2022speechut, sainath2023joist, radford2023robust, peng2023reproducing}. 
Most recently, ``speech language models" (speechLMs) with \sfm \ encoder and large language model decoder are being developed to perform as even better general-purpose foundation models that effectively tackle a much wider variety of tasks~\cite{peng2024survey, cui2024recent, arora2025landscape}.
\begin{table}[t]
\caption{Overview of all the \sfms \ that are analyzed in this thesis. The number of parameters for audio-visual models (\fastvgs, \fastvgsp, \vghubert, and \avhubert) denotes the parameter count for the audio branch. \whisper \ parameter counts represent the encoder module alone.}
\resizebox{\columnwidth}{!}{%
\begin{tabular}{l|ll|lrr|lll}
\hlineB{3}
\multicolumn{1}{c|}{\multirow{2}{*}{\begin{tabular}[c]{@{}c@{}}Speech\\ foundation\\ model\end{tabular}}} & \multicolumn{2}{c|}{Pretext task}   & \multicolumn{3}{c|}{Model architecture}  & \multicolumn{3}{c}{Pre-training data} \\
\multicolumn{1}{c|}{} & \multicolumn{1}{c}{\begin{tabular}[c]{@{}c@{}}Pre-training\\ objective\end{tabular}}   & \multicolumn{1}{c|}{\begin{tabular}[c]{@{}c@{}}Target\end{tabular}}  & \multicolumn{1}{c}{\begin{tabular}[c]{@{}c@{}}Local\\  encoder\end{tabular}} & \multicolumn{1}{c}{\begin{tabular}[c]{@{}c@{}}\# SA\\ layers\end{tabular}} & \multicolumn{1}{c|}{\begin{tabular}[c]{@{}c@{}}\#\\ params\end{tabular}} & \multicolumn{1}{c}{Datasets}   & \multicolumn{1}{c}{\# hours} & \multicolumn{1}{c}{\begin{tabular}[c]{@{}c@{}}Input\\ modality\end{tabular}}   \\ \hlineB{3}
\wavtovec-\baseM\tablefootnote{\label{fn:w2v2}\href{https://github.com/facebookresearch/fairseq/tree/main/examples/wav2vec}{https://github.com/facebookresearch/fairseq/tree/main/examples/wav2vec}} & \multirow{2}{*}{\begin{tabular}[c]{@{}l@{}}masked contrastive\\ discrimination\end{tabular}}   & \multirow{2}{*}{layer 0}   & \multirow{2}{*}{7 CNN}   & 12 & 95M  & LibriSpeech & 960 & \multirow{2}{*}{raw audio}  \\
\wavtovec-\largeM\textsuperscript{\ref{fn:w2v2}}   & & &  & 24 & 317M & LibriLight & 60k & \\ \hline
\hubert-\baseM\tablefootnote{\label{fn:hubert}\href{https://github.com/facebookresearch/fairseq/tree/main/examples/hubert}{https://github.com/facebookresearch/fairseq/tree/main/examples/hubert}}  & \multirow{2}{*}{\begin{tabular}[c]{@{}l@{}}iterative masked\\ prediction\end{tabular}} & \multirow{2}{*}{\begin{tabular}[c]{@{}l@{}}layer 9 from\\ last iteration\end{tabular}} & \multirow{2}{*}{7 CNN}   & 12 & 95M  & LibriSpeech & 960 & \multirow{2}{*}{raw audio} \\
\hubert-\largeM\textsuperscript{\ref{fn:hubert}} & & &  & 24 & 317M & LibriLight & 60k & \\ \hline
\multirow{2}{*}{\wavlm-\baseM\tablefootnote{\label{fn:wavlm}\href{https://github.com/microsoft/unilm/tree/master/wavlm}{https://github.com/microsoft/unilm/tree/master/wavlm}}}   & \multirow{2}{*}{\begin{tabular}[c]{@{}l@{}}masked prediction\\ with denoising\\ for overlapped\\ speech\end{tabular}}  & \multirow{4}{*}{\begin{tabular}[c]{@{}l@{}}layer 6 from\\ 1st-iteration\\ \hubert\end{tabular}} & \multirow{4}{*}{7 CNN}   & \multirow{2}{*}{12} & \multirow{2}{*}{95M}  & LibriSpeech & \multirow{2}{*}{960} & \multirow{4}{*}{raw audio} \\
\multirow{2}{*}{\wavlm-\largeM\textsuperscript{\ref{fn:wavlm}}}  & & &  & \multirow{2}{*}{24} & \multirow{2}{*}{317M} & \begin{tabular}[c]{@{}l@{}}LibriLight +\\ GigaSpeech + \\ VoxPopuli\end{tabular}   & \multirow{2}{*}{94k} & \\ \hline
\datatovec-\baseM\tablefootnote{\label{fn:data2vec}\href{https://github.com/facebookresearch/fairseq/tree/main/examples/data2vec}{https://github.com/facebookresearch/fairseq/tree/main/examples/data2vec}}  & \multirow{2}{*}{\begin{tabular}[c]{@{}l@{}}reconstruction \\loss\end{tabular}} & \multirow{2}{*}{\begin{tabular}[c]{@{}l@{}}average across\\ teacher layers\end{tabular}} & \multirow{2}{*}{7 CNN}   & 12 & 94M  & LibriSpeech & 960 & \multirow{2}{*}{raw audio} \\
\datatovec-\largeM\textsuperscript{\ref{fn:data2vec}} & & &  & 24 & 314M & LibriLight & 60k & \\ \hline
\fastvgs\tablefootnote{\label{fn:fastvgs}\href{https://github.com/jasonppy/FaST-VGS-Family}{https://github.com/jasonppy/FaST-VGS-Family}}   & \begin{tabular}[c]{@{}l@{}}cross-modal\\ contrastive loss\end{tabular} & n/a & 7 CNN & 8  & 110M & SpokenCOCO & 742 & \multirow{8}{*}{\begin{tabular}[c]{@{}l@{}}raw audio\\ paired with\\ images\end{tabular}}  \\
\fastvgsp\textsuperscript{\ref{fn:fastvgs}}  & \begin{tabular}[c]{@{}l@{}}cross-modal\\ contrastive loss + \\ masked contrastive\\ discrimination\end{tabular} & layer 0 & 7 CNN & 12 & 138M & \begin{tabular}[c]{@{}l@{}}SpokenCOCO +\\ LibriSpeech\end{tabular} & 1.7k    & \\
\vghubert\tablefootnote{\href{https://github.com/jasonppy/word-discovery}{https://github.com/jasonppy/word-discovery}}  & \begin{tabular}[c]{@{}l@{}}cross-modal\\ contrastive loss\end{tabular} & n/a & 7 CNN & 12 & 98M  & SpokenCOCO & 742 & \\ \hline
\begin{tabular}[c]{@{}l@{}}\avhubert\\ -\baseM\tablefootnote{\label{fn:avhubert}\href{https://github.com/facebookresearch/av_hubert}{https://github.com/facebookresearch/av\_hubert}}\end{tabular} & \multirow{2}{*}{\begin{tabular}[c]{@{}l@{}}iterative masked\\ prediction of\\ multi-modal units\end{tabular}} & \multirow{3}{*}{\begin{tabular}[c]{@{}l@{}}layer 12 from\\ last iteration\end{tabular}} & \multirow{3}{*}{1 linear} & 12 & 91M & \multirow{2}{*}{\begin{tabular}[c]{@{}l@{}}LRS3\\ lip-reading \\ corpus\end{tabular}} & \multirow{3}{*}{433} & \multirow{2}{*}{\begin{tabular}[c]{@{}l@{}}mel FBanks\\ paired with\\ images\end{tabular}} \\
\begin{tabular}[c]{@{}l@{}}\avhubert\\ -\largeM\textsuperscript{\ref{fn:avhubert}}\end{tabular} & & &  & 24 & 313M & & & \\ \hline
\xlsr\textsuperscript{\ref{fn:w2v2}}  & \begin{tabular}[c]{@{}l@{}}masked contrastive\\ discrimination\end{tabular} & layer 0 & 7 CNN & 24 & 317M & \begin{tabular}[c]{@{}l@{}}MLS +\\ CommonVoice +\\ BABEL\\ (53 languages)\end{tabular} & 56k & raw audio  \\   \hline 
\whisper\tablefootnote{\href{https://github.com/openai/whisper}{https://github.com/openai/whisper}}  & \begin{tabular}[c]{@{}l@{}}speech-to-text\\ prediction\end{tabular} & \begin{tabular}[c]{@{}l@{}}text\\ transcript,\\ translation\\ and\\language ID \end{tabular} & 2 CNN & \begin{tabular}[c]{@{}l@{}}4\\ 6\\ 12\\ 24\\ 32\end{tabular} & \begin{tabular}[c]{@{}r@{}}8M\\ 20M\\ 87M\\ 306M\\ 635M\end{tabular} & N/A & 680k & \begin{tabular}[c]{@{}l@{}}mel FBanks\\ paired with\\ text\end{tabular}  \\   \hlineB{3} 
\end{tabular}
}
\label{tab:sfms}
\end{table}

\subsubsection{Selected \sfms \ for analysis}
In this work, we present our analysis of {\it fifteen \sfms}, which differ in terms of (i) pre-training objective, (ii) pre-training data modality (using either just speech or speech paired with images or speech paired with text), (iii) pre-training data languages (English or multilingual), and (iv) model size. An overview of these \sfms \ is provided in Table~\ref{tab:sfms}. We use the publicly available pre-trained checkpoints for these models.

A typical \sfm \ initially processes raw audio (or filter banks) through convolutional layers (or linear projection). The resulting frame-level {\it local} features are further processed through self-attention layers. Incidentally, all the self-supervised \sfms \ we examine employ a masking-based pretext task, utilizing both left and right context to recover the masked segment (target).
For some of these \sfms, the pre-training target is derived from the local features and used in a contrastive setup (e.g., \wavtovec~\cite{baevski2020wav2vec}, and \xlsr~\cite{conneau2020unsupervised}). 
For another set of \sfms, one or more of the intermediate transformer layers are used as the target, either as a contextualized latent representation in a reconstruction setup (e.g., \datatovec~\cite{baevski2022data2vec} uses self-distillation) or as discrete cluster IDs in a predictive setup (e.g., \hubert~\cite{hsu2021hubert}, \wavlm~\cite{chen2022wavlm}, and \avhubert~\cite{shi2022robust} are all trained iteratively). \xlsr \ is a multi-lingual \sfm \ trained on spoken data from 53 languages, whereas the rest are English \sfms. \wavlm \ uses the cluster IDs from \hubert's intermediate layers and augments the input data to simulate noisy or overlapped speech.   

For the audio-visual \sfms \ such as \avhubert, \fastvgs~\cite{peng2022fast}, \fastvgsp, and \vghubert, we focus solely on the audio branch in our analysis of speech properties. The \avhubert \ model is trained on a lipreading dataset with a pre-training objective that involves multi-modal discrete units. The audio branch of \avhubert \ processes filter bank input through a single linear layer to obtain local features, whereas, for all other \sfms, local features are obtained by processing raw audio waveforms through a stack of CNN layers. The \fastvgs \ and \fastvgsp \ models are initialized from the pre-trained \wavtovec-\baseM \ model, with \fastvgs \ using 8 of the 12 transformer layers and \fastvgsp \ using all 12 layers while keeping the weights for the CNN module from \wavtovec-\baseM \ frozen. With additional CNN, self-attention, and cross-attention layers, the ``audio branch" is trained along with a visual branch using a cross-modal contrastive loss.
\fastvgsp \ extends \fastvgs \ with an additional masking-based contrastive loss for the transformer layers initialized from \wavtovec-\baseM. 

Lastly, \whisper~\cite{radford2023robust} is a multilingual encoder-decoder transformer-based \sfm \ trained with weak supervision on multiple tasks such as speech recognition, speech translation, and language identification. With the focus on increasing the scale and diversity of training data, \whisper \ models are trained on 680k hours of audio (sources are not disclosed), where 117k hours cover 96 non-English languages and X$\rightarrow$English translation data constitutes 125k hours of training data. 

\subsection{Adaptation of speech foundation models}
\label{sec:background-sfm-adapt}

\begin{figure*}[tbh]
 \centering
 \centerline{\includegraphics[width=0.95\linewidth]{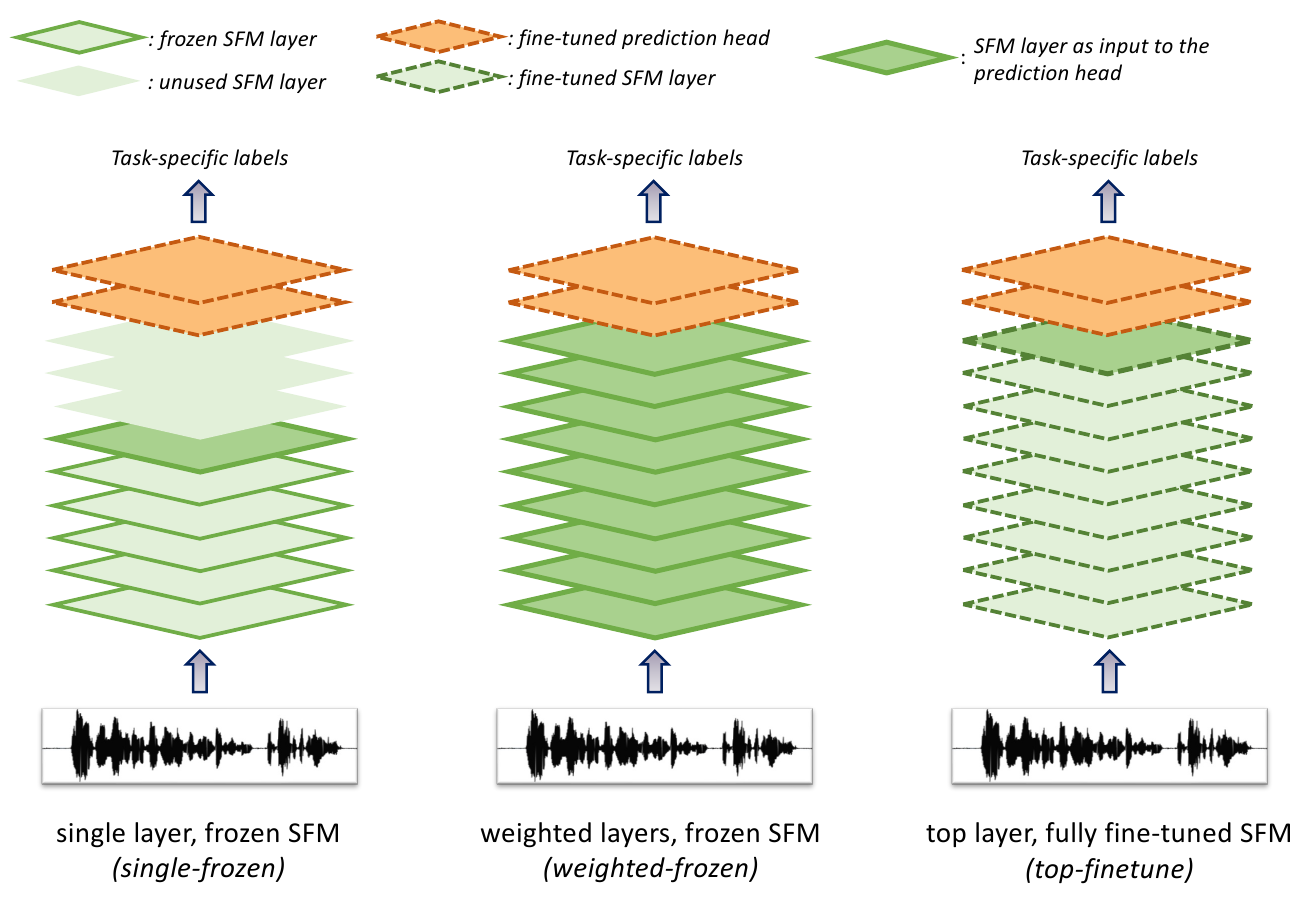}}
\caption{Illustration of some popular adaptation strategies.} 
 \label{fig:adapt-strat}
\end{figure*}

The effectiveness of an \sfm \ is ultimately determined by its performance on speech applications. To evaluate an \sfm \ on a downstream task, one needs to adapt the backbone \sfm \ to output classes/tokens specific to the task.
As the task-specific output space is typically different from the pre-trained \sfm, a prediction head is added to the \sfm \ and trained on the task-specific data. 
Some popular choices are illustrated in \fig~\ref{fig:adapt-strat} with the prediction head in \textcolor{dkorange}{orange} and \sfm \ self-attention layers in \textcolor{dkgreen}{green}.
The convolutional layers are not illustrated for simplicity, but retaining the frozen convolutional sub-module is now standard, except in a handful of cases studying speaker-related tasks that try an alternate approach~\cite{wang2021fine, chen2023chapter}.
Different outlines and shades in the figure indicate the choice of input to the prediction head and whether the pre-trained layers are dropped, retained as-is, or tuned with the supervised loss.

\sfl \ and \afl \ retain the pre-trained \sfm \ parameters as is, whereas \tff \ fine-tunes them on the supervised loss.
Some alternate approaches to combine frozen layer representations have been proposed~\cite{shih2024interface}, but \afl \ remains the most commonly used and widely tested approach for benchmarking across \sfms.
\tff \ is commonly used to obtain the best results on any specific task, and \sfms \ tuned with \tff \ are observed to provide competitive results with a very light prediction head and a much smaller amount of supervised data~\cite{baevski2020wav2vec, hsu2021hubert, chen2022wavlm}. But the \tff \ strategy is not easily scalable as these backbone models are huge (see \tab~\ref{tab:sfms}) and cumbersome to tune. 
Alternatively, \afl, popularized by the SUPERB benchmark~\cite{yang2021superb}, learns a weighted combination of representations from all layers. When compared to \tff, \afl \ offers a much faster evaluation and has been widely adopted by the community to compare various \sfms \ at scale.\footnote{\href{https://superbbenchmark.org/leaderboard}{https://superbbenchmark.org/leaderboard}} \sfl \ is a special case of \afl, and is commonly used in the analysis literature to perform parametric probing of intermediate representations with a lightweight prediction head~\cite{de2022probing, ma2021probing, yang2022autoregressive}.
We employ these adaptation protocols in various parts of this thesis (\chaps~\ref{ch:compare-tools}, \ref{ch:implications}, \ref{ch:slue}, \ref{ch:ner-ext}), and specifically in \chap~\ref{ch:slueperb}, we compare these approaches in the context of spoken language understanding.

\begin{figure*}[tbh]
 \centering
 \centerline{\includegraphics[width=0.7\linewidth]{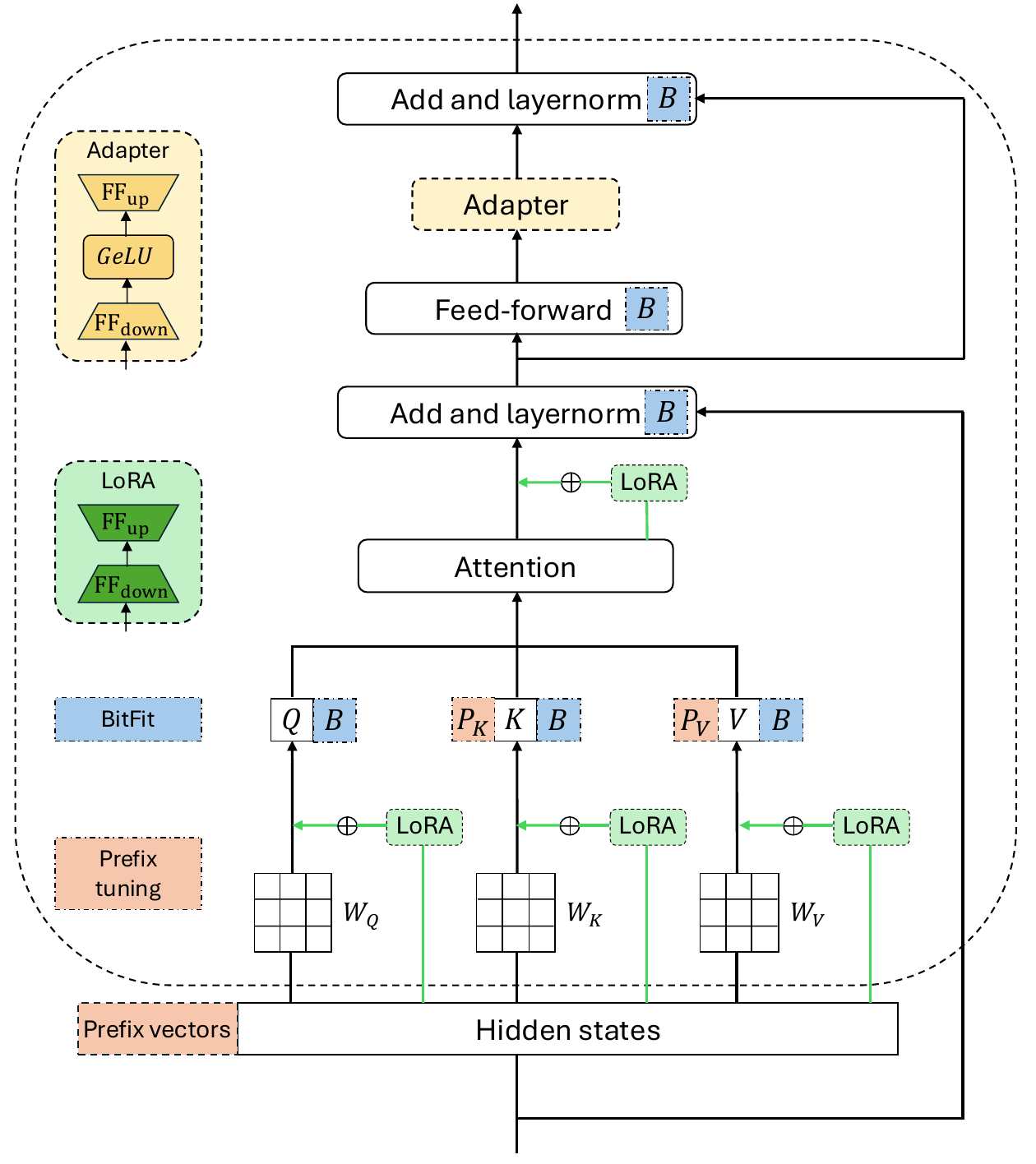}}
\caption{Illustration of some popular PEFT strategies.} 
 \label{fig:peft-strat}
\end{figure*}

While \tff \ has the most potential to offer the best performance on a given downstream task, it results in a separate large model for each downstream task, requiring significant computational resources and storage for each task-specific model. \afl, on the other hand, only learns a handful of parameters on top of the backbone \sfm. Still, it may not always deliver the best-performing model, especially when there is a mismatch between pre-training and target domains. Parameter-efficient fine-tuning (PEFT) strategies have emerged as a compelling alternative to full fine-tuning, addressing its inherent limitations in scalability and resource efficiency.

A PEFT strategy adapts a small number of parameters---relative to the size of the backbone \sfm---by either selecting a subset of existing parameters or introducing new ones, with the majority of the pre-trained \sfm \ kept frozen during task-specific training.
These parameters are typically introduced in one or more of the modules in the \sfm \ architecture, \fig~\ref{fig:peft-strat} demonstrates how a transformer layer is modified for some popular PEFT methods. {\it Adapters}, one of the earliest approaches, insert a small bottleneck module consisting of a down-projection (FF$_{down}$), a non-linearity (GeLU), and an up-projection (FF$_{up}$), with a skip-connection~\cite{houlsby2019parameter}. Low-rank adaptation ({\it LoRA}) learns a pair of rank-decomposition matrices that offset the pre-trained weight matrices and is typically added to the self-attention module~\cite{hu2021lora}. {\it BitFit} does not add additional parameters but instead tunes the bias term for each module~\cite{zaken2021bitfit}.
Motivated by prompting in large language models, {\it prefix-tuning} prepends trainable prefix vectors (``virtual tokens") to the input of transformer layers~\cite{li2021prefix}. 


Although many of these strategies were originally developed for large text foundation models~\cite{fu2023effectiveness, wang2024parameter}, PEFT-based adaptation has become relevant for \sfms, and is being widely explored for a variety of tasks, such as speech recognition~\cite{thomas2022efficient, vanderreydt2023parameter, kim2024convolution, huo2024adara, pratap2024scaling}, speech translation~\cite{chen2024parameter, ma2022cpt}, emotion recognition~\cite{lashkarashvili2024parameter, feng2023peft}, and audio captioning~\cite{kim2023prefix}. 
PEFT techniques have been effective in adapting an \sfm \ to a new language~\cite{wang2024learn, liu2024parameter, huang2023findadaptnet, song2024lora, xu2024towards} and new domains such as children's speech~\cite{liu2024sparsely, fan2024benchmarking, rolland2024shared}.
PEFT approaches are also being applied in speechLMs to effectively bridge speech and text modalities~\cite{baskar2024speech, hu2024wavllm, cappellazzo2024large, verdini2024connect}.
Some of this exploration also proposes modified PEFT strategies for \sfms \ and speech tasks~\cite{chen2024parameter, inoue2024elp, liu2024sparsely, vanderreydt2023parameter, rolland2024shared, kim2024convolution, huo2024adara, chen2023chapter}.
The ideal choice of a PEFT approach for a given task and an \sfm \ is still an open research question with multiple comprehensive studies comparing different PEFT modules and downstream tasks~\cite{inoue2024elp, suresh2024adapter, chang2023prompting, li2023evaluating, lin2024peft, chang2023speechprompt, otake2023parameter, chen2023exploring}. 

Typically, PEFT modules are introduced in all transformer layers, although some recent work has started exploring alternatives~\cite{thomas2022efficient, huang2023findadaptnet, chen2024parameter, vanderreydt2023parameter}. In \chap~\ref{ch:implications}, we dive deeper into the effect of placement of LoRA modules for spoken language understanding tasks. 

\section{Analysis of neural models}
\label{sec:background-analysis}
Analysis of neural models has been a topic of interest ever since neural approaches have had successful applications~\cite{elman1988learning, waibel1989phoneme, elman1991distributed, zhang2018visual, belinkov2019analysis}. These analysis studies have been helpful in understanding various aspects of a neural model, from the role of an individual neuron or a module in the network to the properties encoded in the intermediate activations to how a model makes a specific decision on input data. While we primarily discuss prior work on analyzing speech models, we will also note relevant methods from other domains (text, image, and brain activity).

\subsection{Categories of analysis methods}
{\bf Visualization tools} offer an intuitive way to interpret the inner workings of a neural network. Visualization of activation values for individual data points has been a common approach to interpret the role of a neuron~\cite{elman1988learning, waibel1989phoneme, bau2017network, krug2018neuron, krug2020gradient, wuand2024} or role of a specific neural module such as filters in convolutional neural networks (CNNs)~\cite{simonyan2013deep, zeiler2014visualizing, yosinski2015understanding, zhang2018visual}, gated components of recurrent neural networks (RNNs)~\cite{karpathy2015visualizing, wu2016investigating, strobelt2017lstmvis, tang2017memory, wang2017gate, silfverberg2021rnn}, cross-attention in encoder-decoder models~\cite{bahdanau2015neural, chan2015listen}, and self-attention in transformers~\cite{clark-etal-2019-bert, voita-etal-2019-analyzing}. 
Visualization tools developed using non-linear dimensionality reduction, such as t-SNE~\cite{van2008visualizing} and UMAP~\cite{McInnes2018}, have also been widely useful to study the structure of clusters formed by the high-dimensional space of hidden representations~\cite{mohamed2012understanding, elloumi2018analyzing, scharenborg2019representation, van2021analyzing, choi2022opening, de2022probing, sicherman2023analysing, liu2023dinosr, wang2023understanding}.

\sloppy
{\bf Neuron-level analysis} studies individual dimensions of the activation vectors to discover latent concepts in the representations and for interpreting the modeling mechanisms~\cite{qian-etal-2016-analyzing, bau2018identifying, dalvi2022discovering, templeton2024scaling, huben2024sparse}. Broadly, these approaches correlate the neuron behaviors with the input stimuli and design experiments to map the neuron activation to an explicit feature or a concept. Knowing the importance of individual neurons can provide fine-grained control of the system behavior and help with model editing, pruning, and efficient feature selection. Network dissection, motivated by neuroscience research~\cite{quiroga2005invariant}, is one example of a successful neuron-based approach. First introduced for vision models~\cite{bau2017network}, network dissection has been adopted for audio~\cite{wuand2024}. Only a handful of papers study the distribution of knowledge across individual neurons in \sfms~\cite{chowdhury2023end, lin2024property}.


Many studies {\bf evaluate hidden representations as a whole}, not distinguishing between individual neurons, to understand the properties encoded in intermediate layers. Our work utilizes this style of interpreting neural models. In the next three paragraphs we discuss three kinds of approaches designed to study hidden representations.  
The first two categories involve evaluation of hidden representations on a well-defined task, including both parametric and non-parametric approaches, also categorized as extrinsic and intrinsic evaluation, respectively, by the literature studying text embeddings~\cite{schnabel2015evaluation}.
The third category studies hidden representations by aligning the representation space with an external well-defined subspace.

Some widely used {\bf training-free, i.e., non-parametric, tasks} include semantic similarity of written or spoken words~\cite{faruqui2014community,nguyen2020zero}, semantic similarity of written or spoken sentence~\cite{conneau2018senteval,merkx2021semantic}, phonetic discriminability of representations (ABX task)~\cite{schatz2013evaluating, schatz2014evaluating}, and acoustic word discrimination~\cite{carlin2011rapid}. Intrinsic evaluation offers intuitive insights into the information encoded in the representations. The appeal of intrinsic evaluations is further enhanced by its fast and inexpensive computation, making it easy to scale.
The utility of these non-parametric evaluation tasks has been debated
\cite{faruqui2016problems, qiu2018revisiting}, but it's still found to be beneficial, when used with appropriate baselines, for studying speech representations and understanding the properties encoded by the representation spaces~\cite{pasad2021layer, pasad2023self, choi2024self}. ZeroSpeech challenges have also introduced a suite of tasks to analyze speech representations (discrete units, acoustic tokens, phone discovery, word discovery) learned by unsupervised speech models~\cite{dunbar2022self}.

{\bf Supervised parametric approaches} analyze hidden representations by training a simple neural model to solve a task specifically designed to evaluate certain fundamental characteristics.
For instance, Adi et al.~\cite{adi2016fine} designed low-level tasks related to length, content, and word order to study the information encoded in sentence embeddings.
This approach is commonly referred to as ``auxiliary prediction tasks’’~\cite{adi2016fine} or ``diagnostic classifiers’’~\cite{veldhoen2016diagnostic} or ``probing tasks’’~\cite{alain2016understanding, conneau-etal-2018-cram}.
For an extensive survey on analysis methods for neural language processing models, we direct the reader to Belinkov and Glass~\cite{belinkov2019analysis}, specifically table SM1.\footnote{\href{https://belinkov.com/nlp-analysis-methods/}{https://belinkov.com/nlp-analysis-methods/}}
Probing classifiers have been commonly used to study various properties encoded in supervised speech models, such as
speaker information~\cite{wang2017does, chowdhury2020does},
gender information~\cite{nagamine15_interspeech, wang2017does, chowdhury2020does},
speaking styles~\cite{elloumi2018analyzing},
channel~\cite{wang2017does, chowdhury2020does}, and
syntax-related linguistic content~\cite{triantafyllopoulos2022probing}. 
Audio-visual representations have been extensively analyzed for word-level knowledge~\cite{harwath-glass-2017-learning, chrupala2017representations}, testing the effectiveness of visual grounding in encoding sub-word units.

The third category studies {\bf subspaces of the hidden representations} using statistical tools such as dimensionality reduction or correlation with another well-defined representation space (either from an external variable or another model).
Many representation analysis studies evolved to quantify the similarity of neural representations from different neural networks~\cite{laakso2000content, li2015convergent, raghu2017svcca, morcos2018insights, wang2018towards, liang2020knowledge, conneau2018word}. Representation analysis was also extensively used in machine translation for studying the representation of bilingual or multilingual word pairs~\cite{artetxe2016learning, faruqui2014improving, conneau2018word, smith2017offline}.
Such approaches, unlike probing classifiers, are typically lightweight but may lack the interpretability of the evaluation score that the task-based approaches provide. Nonetheless, with the help of reasonable null hypotheses and observing relative scoring between different layers or models, subspace analysis has proved to be a valuable tool~\cite{ding2021grounding, kornblith2019similarity, raghu2017svcca, saphra2019understanding}.
When compared to intrinsic word similarity evaluation, subspace analysis is found to be more correlated with extrinsic task-based evaluation~\cite{tsvetkov2015evaluation}. Some of the tools and properties commonly explored to study neural models include canonical correlation analysis~\cite{tsvetkov2016correlation, voita2019bottom, pasad2021layer}, mutual information~\cite{hsu2021hubert, prasad2020accents}, centered kernel alignment~\cite{kornblith2019similarity, saphra2019understanding}, isotropy~\cite{ethayarajh2019contextual, mohamed2024orthogonality, chen2023isotropy}, and orthogonality~\cite{liu2023self, mohamed2024orthogonality}.


\subsection{Analyses of speech models}
In the context of neural speech models, {\bf phonetic properties} are the most explored. In 1988, Elman and Zipser~\cite{elman1988learning} trained a small phonetic labeling system (with a hidden dimension of six) for nine types of isolated syllable sounds and visualized the hidden unit response for individual data points to spot specificity to phones and articulatory contexts.
Interpreting phonetic knowledge encoded in speech neural networks has continued to be of major research interest in the community. A variety of techniques has been employed, such as visualizing the activation values~\cite{nagamine15_interspeech, rasanen2016analyzing, nagamine2017understanding}, comparison of discrete units with phoneme groups or individual phones, where the discrete units can either be learned~\cite{baevski2020wav2vec} or obtained from unsupervised clustering~\cite{nagamine15_interspeech, rasanen2016analyzing, nagamine2017understanding, hsu2021hubert, chang2023self}, using probing tasks to study intermediate representations~\cite{nagamine2017understanding, belinkov2017analyzing, de2022probing}, studying neuron responses to quantify specificity to phonemes~\cite{krug2018introspection, krug2018neuron}, or checking if phoneme boundaries are encoded in the learned representations, such as gates in LSTMs~\cite{wang2017gate}. Alishahi et al.~\cite{alishahi-etal-2017-encoding} use a combination of these techniques to investigate the encoded phonetic knowledge in audio-visual models. Phonetic features have also been studied in the context of EEG signals~\cite{khalighinejad2017dynamic} as well as to interpret the articulatory and phonotactic content~\cite{dunbar2015quantitative}.

As {\bf \sfms} have become more prevalent in recent years~\cite{mohamed2022self, yang2021superb, tsai2022superb, shi2023ml}, there has been growing work on interpreting their hidden representations.
To make sense of how pre-training helps \sfms, it is also reasonably common for the \sfm \ papers to have some analytical study of the pre-trained model. For instance, Chen et al.~\cite{chen2022wavlm} study speaker content in wavLM layers, and Baevski et al.~\cite{baevski2020wav2vec}, Hsu et al.~\cite{hsu2021hubert}, and Liu et al.~\cite{liu2023dinosr} study phonetic content in the learned codevectors and representations of wav2vec2.0, HuBERT, and DinoSR respectively. 
Generally, both probing~\cite{ma2021probing, de2022probing, yang2022autoregressive} and subspace analysis~\cite{pasad2021layer, pasad2023comparative, abdullah2023information, riera2023phone, wells22_interspeech} methods have found \sfms \ to be effective at encoding phonetic content.
Work on generative models based on \sfms \ also further confirms that some \sfms \ learn phone-like sub-word units~\cite{lakhotia2021generative, nguyen2023generative}.
Discrete units extracted from \sfms \ have also been evaluated as input speech features for various tasks~\cite{chang2024exploring, yang2024towards}.
Speaker identity content in \sfms \ has also been widely explored~\cite{fan2020exploring, van2021analyzing, chen2022wavlm, feng2022silence, de2022probing, liu2023self, riera2023phone, ashihara2024self}.
Researchers have also studied how phonetic content contrasts with speaker-related content in the representation space~\cite{riera2023phone, liu2023self, mohamed2024orthogonality, chen2023isotropy, lin2024property}.
\sfms \ have also been analyzed to study whether pre-trained representations encode knowledge related to language~\cite{de2022probing, chowdhury2023end}, para-linguistics~\cite{shah2021all, li2023exploration, saliba2024layer}, pronunciation and articulation~\cite{banno2023proficiency, cho2024self, kim2022automatic, ji2022predicting, bartelds2022neural} and suprasegmental and prosodic features~\cite{yang2022autoregressive, de2024layer, lin2023utility}. 
Other investigations of \sfms \ include the study of fine-grained linguistic aspects of phonological assimilation~\cite{pouw2024perception} and phonotactic biases in \sfms~\cite{kloots2024human}.
Acquisition of knowledge in \sfms \ has been compared with brain activity (fMRI and EEG signals)
\cite{chen2023self, li2023dissecting, vaidya2022self, millet2022toward}.


Analysis of neural models provides valuable insights into how training affects representations, enabling practitioners to make informed decisions regarding new models or adaptation strategies. For instance, visualization-based analysis has led to simplifications in ASR architectures~\cite{tang2017memory, miao2016simplifying}. CCA-based study of how representations evolve in a supervised model has motivated the design of efficient training regimes~\cite{raghu2017svcca}. Specific to \sfms, studying how the speaker and phonetic content are organized in the representation space has motivated a post-processing method to normalize speaker content~\cite {liu2023self, mohamed2024orthogonality, van2021analyzing}.
Insights into how linguistic knowledge is distributed across \sfm \ layers have motivated the choice of layers in the design of model adaptation~\cite{pasad2021layer,dorszewski2024convexity,zaiem2023fine, chang2024exploring,xue2023sshr} and model distillation~\cite{chang2022distilhubert,baevski2022data2vec,hwang2022pseudo} strategies.

\section{Background on analysis techniques}
\label{sec:background-tools}

The previous section (\sect~\ref{sec:background-analysis}) provided a broad overview of prior work analyzing neural models.
This section delves into an in-depth discussion of several statistical tools that are later used to study hidden representations of \sfms \ (\chaps~\ref{ch:analysis} and \ref{ch:compare-tools}). 

\subsection{Preliminaries}
\label{sec:tech-primer}

In the following subsections (Sections \ref{sec:tech-cca}-\ref{sec:tech-linear}), we explore tools for comparing two random vectors, $X$ and $Y$. In our context, $x_i$ represents a representation extracted from an \sfm. $x_i$ is paired with a corresponding property of interest, $y_i$, which may be a discrete class or a continuous-valued vector. The statistical tools discussed here measure various forms of similarity or dependence between $X$ and $Y$.

All metrics discussed here share invariance to orthogonal transformations, including permutations. This invariance is a desirable property when analyzing neural representations, where the ordering of features should not affect our conclusions.  
Some tools offer additional invariance properties, such as invariance to isotropic scaling or affine transformations.
Before evaluating all metrics, both $X$ and $Y$ are mean-normalized.


\subsection{Canonical correlation analysis}
\label{sec:tech-cca}

Canonical correlation analysis (CCA) is a statistical technique that measures the relationship between two continuous-valued random vectors as a scalar correlation score in the range $[0, 1]$~\cite{harold1936relations, hardoon2004canonical}. 
More concretely, CCA takes as input $n$ pairs of vectors $\{(x_1, y_1), ..., (x_n, y_n)\}$, sampled from the random vectors (or ``views") $X\in\mathbb{R}^{d_x}, Y\in\mathbb{R}^{d_y}$ and measures similarity as the maximum correlations between the linear projections of $X$ and $Y$.

CCA has a closed-form solution that can be stated iteratively by first defining directions ($a_1$, $b_1$) that yield the maximum correlation between projected variables $a_1^TX$ and $b_1^TY$. The subsequent directions ($a_k$, $b_k$) maximize the correlation between  $X$ and $Y$ subject to each new projection being uncorrelated with the previous ones in the same view. Mathematically,

\begin{align}
    a_1, b_1 &= \argmax_{a, b} \text{corr}(a^TX, b^TY)  \label{eq:cca1} \\
    a_k, b_k &= \argmax_{a, b} \text{corr}(a^TX, b^TY) 
        \quad \text{s.t. } a_k^TC_{xx}a_i = 0,\ b_k^TC_{yy}b_i = 0\ \forall i < k \notag \\
    &\qquad \qquad \qquad \qquad \qquad \qquad \quad \quad \: \text{and } \forall k \in [2, d], \text{ where } d = \min\{d_x, d_y\} \label{eq:cca2}
\end{align}


The optimal directions $a_i$ and $b_i$ can be obtained by solving the eigenvalue problem~\cite{harold1936relations}: 
\begin{align}
    C_{xx}^{-1} C_{xy} C_{yy}^{-1} C_{yx} a &= \lambda^2 a  \label{eq:cca-sol-1} \\
    b &= \frac{1}{\lambda}C_{yy}^{-1} C_{yx} a \label{eq:cca-sol-2}
\end{align}
To ensure the stability of the matrix inverse operations, two regularization parameters, $\epsilon_x$ and $\epsilon_y$, are added to the diagonal of the covariance matrices, $C_{xx}$ and $C_{yy}$, respectively. The CCA similarity between $X$ and $Y$ is then computed as the average of the top $d$ canonical correlations:
\begin{align}
    \text{CCA}(X, Y) = \frac{1}{d}\sum_{i=1}^d \rho_i = \frac{1}{d}\sum_{i=1}^d \text{corr}(a_i^TX, b_i^TY) \label{eq:vcca}
\end{align}

\eq~\ref{eq:vcca} represents standard or vanilla CCA, where all the correlated directions are given equal weight. However, some of these directions could result from spurious correlation of noise variables. A naive way to control this is by calculating CCA as the first (also the maximum) CCA coefficient~\cite{tsvetkov2016correlation}. Next, we describe two variants of standard CCA that were designed to be more robust to such spurious correlations.

{\bf Singular-value CCA (SVCCA)} was motivated by the idea that low-variance neurons from individual subspaces might contribute to noise~\cite{raghu2017svcca}. To achieve this, SVCCA performs singular value decomposition (SVD) on $X$ and $Y$, retaining the highest variance directions, before performing CCA. More precisely,
\begin{align}
    X &= U_x\Lambda_xV_x^\top, \ X' = U_x[:,:k_x]\Lambda_x[:k_x] \ \text{where} \ k_x = \min\{k: \sum_{i=1}^{k}\lambda_i/\sum_{i=1}^{d_x}\lambda_i \geq \tau_x\}, \ \lambda_i = \Lambda_x^2[i, i] \nonumber \\
    Y &= U_y\Lambda_yV_y^\top, \ Y' = U_y[:,:k_y]\Lambda_y[:k_y] \ \text{where} \ k_y = \min\{k: \sum_{i=1}^{k}\lambda_i/\sum_{i=1}^{d_y}\lambda_i \geq \tau_y\}, \ \lambda_i = \Lambda_y^2[i, i] \nonumber 
\end{align}
where $\tau_x$ and $\tau_y$ are thresholds for variance, typically set to $0.99$~\cite{raghu2017svcca}. Finally, SVCCA is computed as,
\begin{align}
\text{SVCCA}(X, Y) = \text{CCA}(X', Y') \label{eq:svcca}
\end{align}

Thus, in addition to $\epsilon_x$ and $\epsilon_y$, SVCCA has two additional hyperparameters, $\tau_x$ and $\tau_y$, that decide the number of SVD directions used to represent $X$ and $Y$.

{\bf Projection-weighted CCA (PWCCA)}, another CCA variant, challenges the assumption of vanilla CCA that all $d$ CCA directions are equally important.
PWCCA replaces mean computation in vanilla CCA (\eq~\ref{eq:vcca}) with a weighted average of CCA correlation coefficients, $\rho_i$~\cite{morcos2018insights}:
\begin{align}
    \text{PWCCA}(X, Y) = \sum_{i=1}^d \alpha_i\rho_i \quad \text{where }  \sum_{i=1}^d \alpha_i = 1 \label{eq:pwcca}
\end{align}

PWCCA assigns higher weights to ``more important" directions, where importance measures how well the CCA projections capture the original data structure. More precisely, once we have optimal CCA projections (\eqs~\ref{eq:cca1}-\ref{eq:cca2}), weights $\alpha_i$ are calculated as,
\begin{align}
    \alpha_i = \frac{\sum_{j=1}^d |\langle x_j, h_i \rangle|}{\sum_{i=1}^d \sum_{j=1}^d |\langle x_j, h_i \rangle|} \quad \text{where } [h_1,\ldots,h_d] = \text{orth}(A^TX)
\end{align}
$H$ represents the orthonormal basis of CCA projections $A^TX$. While these weights can be calculated using either $X$ or $Y$, the original implementation uses the view with fewer dimensions, which matches the number of optimal directions~\cite{morcos2018insights}.

{\bf In the context of representation analysis,} CCA similarity has been previously used to compare representations within and across neural network models~\cite{raghu2017svcca, bau2018identifying, kornblith2019similarity, voita2019bottom, kheir2024speech}. Some work has also employed CCA to compare neural representations with an external property such as brain activity~\cite{sussillo2015neural} and property-specific representations~\cite{saphra2019understanding}. 

Our analysis framework employs PWCCA to compare \sfm \ representations with various external properties (\chap~\ref{ch:analysis}).
While the literature has no consensus on the preferred CCA variant, we chose PWCCA because it provides a robust CCA solution without additional hyperparameters.
We also present a comparative study of different variants for a few representative analysis experiments from our framework (\chap~\ref{ch:compare-tools}).

\subsection{Centered kernel alignment} 
\label{sec:tech-cka}

Linear CKA evaluates the alignment between subspaces of the two views, $X$ and $Y$, by comparing the relative distances between points in each subspace, yielding a dissimilarity score in the range $[0, 1]$. Linear CKA was introduced as RV-coefficient by Robert and Escoufier~\cite{robert1976unifying} and re-introduced as linear CKA by Kornblith et al.~\cite{kornblith2019similarity}. The CKA distance is defined as,
\begin{align}
    CKA(X, Y) = \frac{\|Y^TX\|_F^2}{\|X^TX\|_F\|Y^TY\|_F} \text{\quad where}  \|\cdot\|_F \text{ is the Frobenius norm}
    \label{eq:cka}
\end{align}

{\bf In the context of representation analysis,} much like CCA, CKA has been used to compare neural representations between different models, such as BERT models~\cite{phang-etal-2021-fine}, speaker and speech models~\cite{ashihara2024self}, visually grounded text models~\cite{maniparambil2024vision}, and self-supervised speech models~\cite{chung2020similarity}. 

\subsection{Procrustes distance}
\label{sec:tech-op}
Orthogonal Procrustes (OP) is an optimal rotation, $R$, that aligns the two views, $X$ and $Y$, to have the minimum Euclidean distance possible~\cite{everson1998orthogonal, schonemann1966generalized}.
The minimum distance, also referred to as Procrustes distance, is used as a measure of the dissimilarity between the two views, and is defined as, 
\begin{align}
    \text{OP}(X, Y) &= \min_R\|Y-RX\|_F^2\text{\quad subject to }R^TR=I
\end{align}
This optimization has a closed-form solution~\cite{schonemann1966generalized}. Specifically, if the singular value decomposition of $XY^T$ is $XY^T = USV^T$, then the optimal rotation, $\hat{R}$, is given by
\begin{align}
    \hat{R} &= UV^T 
\end{align}
Plugging the optimal rotation matrix back into the original objective yields a distance formula that bypasses the need to compute $R$ explicitly:
\begin{align}
    \text{OP}(X, Y) &= \|X\|_F^2 + \|Y\|_F^2 - 2\|X^TY\|_* \text{\quad where } \|\cdot\|_* \text{ is the nuclear norm~\cite{schonemann1966generalized}}
    \label{eq:op}
\end{align}

The Procrustes distance defined in Equation~\ref{eq:op} is often preferred in practice because it avoids the need to explicitly solve for $R$. Procrustes distance is theoretically bounded in range $[0,2]$.

{\bf In the context of representation analysis,} Procrustes distance has been commonly used to align two subspaces, such as bilingual or multilingual word embeddings~\cite{conneau2018word, smith2017offline, artetxe2016learning}.

\subsection{Discrete mutual information}
\label{sec:tech-mi}
Methods described so far (CCA, linear CKA, Procrustes distance) are natural for relating continuous-valued vectors. Discrete mutual information (MI), in contrast, can measure dependency between representations and discrete features like phone or word classes.

The continuous-valued representations, $X$, are clustered, and each vector, $x_i$, is assigned a discrete cluster-ID, $\tilde{x}_i$. Once we obtain pairs of discretized representations and the corresponding discrete labels, $(\tilde{x}_i, y_i)$,
the MI score between the two distributions is calculated using the count-based probability estimates:
\begin{align}
    \text{MI}(\tilde{X}, Y) &= \sum_{i=1}^{|\tilde{X}|}\sum_{j=1}^{|Y|} p(\tilde{X}=i, Y=j)\text{log}\frac{p(\tilde{X}=i, Y=j)}{p(\tilde{X}=i)p(Y=j)}  \nonumber \\
    &= \sum_{i=1}^{|\tilde{X}|}\sum_{j=1}^{|Y|} \frac{|\tilde{X}_i\cap Y_j|}{N}\text{log}\frac{|\tilde{X}_i\cap Y_j|/N}{(|\tilde{X}_i|/N)(|Y_j|/N)} \nonumber \\
    &= \sum_{i=1}^{|\tilde{X}|}\sum_{j=1}^{|Y|} \frac{|\tilde{X}_i\cap Y_j|}{N}\text{log}\frac{N|\tilde{X}_i\cap Y_j|}{|\tilde{X}_i||Y_j|} 
    \label{eq:MI}
\end{align}
where\\
$|\tilde{X}|$ is the number of clusters formed by clustering continuous-valued representations, \\
$|Y|$ is the number classes of labels,\\
$|\tilde{X}_i|$ is the number of samples of $\tilde{X}$ in cluster $i$,\\
$|Y_j|$ is the number of samples of $Y$ in class $j$, and\\
$|\tilde{X}_i\cap Y_j|$ is the number of samples that are assigned to cluster $i$ and have label $j$.

Note that the number of clusters, $|\tilde{X}|$, is a user-defined hyperparameter. Discrete MI can, in principle, learn non-linear relationships between $X$ and $Y$.

{\bf In the context of representation analysis,} discrete MI can be employed in any setting where we have representations paired with discrete labels. Prior work has used discrete MI to compare with labels derived from internal sources, such as input text tokens~\cite{voita2019bottom}, as well as external sources, such as accent classes~\cite{prasad2020accents} or phone labels~\cite{hsu2021hubert}.




\subsection{Linear classification}
\label{sec:tech-linear}
Linear classification~\cite{bishop2006pattern}, like discrete MI, measures dependency between continuous-valued representations, $X\in\mathbb{R}^d$, and discrete labels, $Y \in \{1, \cdots, K\}$. 
It is optimized to learn an affine transformation of $X$ that generates a score for each class, with the goal of assigning the highest score to the correct class.

A common choice is to use the softmax cross-entropy loss, which interprets the output of the affine transformation as an unnormalized score for class $j$ for sample $i$:
\begin{align}
    \text{score}_j(x_i) = w_j^Tx_i + b_j \text{\quad} \forall i\in[1, N]\text{, }\forall j\in[1,K] 
\end{align}
where $N$ is the number of training samples, and $w_j \in \mathbb{R}^d$ and $b_j \in \mathbb{R}$ denote the respective weight vector and bias term for class $j$. The optimal affine transform parameters are obtained by minimizing the negative log-likelihood of the correct class over all samples in the training set:
\begin{align}
    \mathcal{L}(X, Y) &= -\sum_{i=1}^N \text{log} \Big(\frac{e^{\text{score}_{y_i}(x_i)}}{\sum_j{e^{\text{score}_j(x_i)}}}\Big) \text{ \quad where } y_i \text{ is the true class of } x_i
\end{align}

Linear classification has no closed-form solution and is trained via gradient descent. Its performance is evaluated using classification accuracy, which provides an intuitive measure of the relationship between the two views. For held-out samples $\tilde{x}\in\tilde{X}$, the predicted classes and accuracy are computed as:
\begin{align}
    \hat{y_i} &= \argmax_j (\text{score}_j(\tilde{x}_i)) \text{\quad} \forall i\in[1, |\tilde{X}|]\text{, }\forall j\in[1,K]  \\
    \text{Accuracy}(X, Y) &= \frac{1}{|\tilde{X}|}\sum_{i=1}^|\tilde{X}| \mathbbm{1}(y_i = \hat{y}_i)
\end{align}

{\bf In the context of representation analysis,} linear classification is a form of probing classifier, which is a widely used technique to intuitively interpret the information encoded in hidden representations~\cite{belinkov2019analysis}. The previous section (\sect~\ref{sec:background-analysis}) extensively covers related literature on the subject.

\subsection{Summary}
\label{sec:tech-summary}
\begin{table}[htbp]
\small
\centering
\caption{Summary of analysis tools discussed in \sect~\ref{sec:background-tools}.}
\label{tab:analysis-tools}
\begin{tabular}{l|lllr}
\hlineB{2}
\textbf{Analysis tool} &
\textbf{Hyperparameters}
& \textbf{\begin{tabular}[c]{@{}l@{}}Learnable\\  parameters\end{tabular}} & \textbf{$2^{nd}$ view} & \multicolumn{1}{l}{\textbf{Range}} \\[10pt] 
\hlineB{2}
CCA &
$\epsilon_x$, $\epsilon_y$
& \begin{tabular}[c]{@{}l@{}} $A\in\mathbb{R}^{d_x\times d}$ \\  $B\in\mathbb{R}^{d_y\times d}$ \end{tabular} & $y \in \mathbb{R}^{d_y}$ & $\in[0,1]$ \\[12pt] 
PWCCA & $\epsilon_x$, $\epsilon_y$ & \begin{tabular}[c]{@{}l@{}}$A\in\mathbb{R}^{d_x\times d}$ \\  $B\in\mathbb{R}^{d_y\times d}$\end{tabular} & $y \in \mathbb{R}^{d_y}$ & $\in[0,1]$ \\[12pt] 
SVCCA &
$\epsilon_x$, $\epsilon_y, \tau_x, \tau_y$
& \begin{tabular}[c]{@{}l@{}}$A\in\mathbb{R}^{d_x\times d}$ \\  $B\in\mathbb{R}^{d_y\times d}$\end{tabular} & $y \in \mathbb{R}^{d_y}$ & $\in[0,1]$ \\[12pt] 
CKA & None & None & $y \in \mathbb{R}^{d_y}$ & $\in[0,1]$ \\[12pt] 
\begin{tabular}[c]{@{}l@{}}Procrustes\\ distance\end{tabular} & None & None & $y \in \mathbb{R}^{d_y}$ & $\in[0,2]$ \\[12pt] 
\begin{tabular}[c]{@{}l@{}}Normalized\\ discrete MI\end{tabular} & \# of clusters & None & $y\in\{1,\ldots,d_y\}$ & $\in[0,1]$ \\[12pt] 
\begin{tabular}[c]{@{}l@{}}Linear classification\\ accuracy\end{tabular} & \begin{tabular}[c]{@{}l@{}}learning rate,\\ stopping criterion\end{tabular} & 
\begin{tabular}[c]{@{}l@{}}$W\in\mathbb{R}^{d_x\times d_y}$\\ $b\in\mathbb{R}^{d_y\times 1}$\end{tabular}
& $y\in\{1,\ldots,d_y\}$ & $\in[0,1]$
\\
\hlineB{2}
\end{tabular}%
\end{table}

\tab~\ref{tab:analysis-tools} summarizes key aspects of all the analysis techniques discussed in this section.  

Speed and simplicity are key requirements for representation analysis techniques, enabling their application across diverse experiments and models. Most methods discussed here offer closed-form solutions, with linear CKA and Procrustes distance being particularly efficient as they require no parameter optimization or hyperparameter tuning. While linear classification does require gradient descent, it remains straightforward, learning only a single linear transformation. This simplicity is crucial as complex analysis tools can introduce confounding effects, making it difficult to distinguish whether the findings stem from the model's representation or artifacts of the analysis method itself~\cite{hewitt2019designing, ravichander2020probing, belinkov2022probing}.

These methods also differ in their applicability to different types of variables. While CCA, linear CKA, and Procrustes distance naturally handle comparisons between continuous-valued representations, discrete MI and linear classification are designed for discrete variables. In our study, we convert discrete class labels to one-hot vectors, enabling CCA, linear CKA, and Procrustes distance to analyze discrete variables as well. More details about our analysis framework can be found in \chap~\ref{ch:analysis}.

Despite these differences in formulation and objectives, prior work has established theoretical connections between some of these methods---for example, CKA and CCA~\cite{kornblith2019similarity}, and CCA and Procrustes distance~\cite{lorbert2012kernel}. Empirical studies have also compared subsets of these tools: CKA has been shown to be a reliable measure of similarity in high-dimensional settings~\cite{kornblith2019similarity}, while Procrustes distance has been found to offer favorable statistical properties such as specificity and sensitivity~\cite{ding2021grounding}. However, prior work has largely focused on comparing internal representations within neural networks, rather than evaluating these tools in relation to external variables such as labels or predefined property vectors.  

To the best of our knowledge, no existing study has systematically compared all of these methods within a unified framework. In \chap~\ref{ch:compare-tools}, we conduct such a comparison to evaluate the behavior of these tools when analyzing \sfm \ representations.



\newpage
\chapter{Lightweight Analysis of Speech Foundation Models}
\label{ch:analysis}
In this chapter, we present a large-scale comparative study of speech foundation models.\footnote{The contents of this chapter are mainly from our prior published papers~\cite{pasad2021layer, pasad2023comparative, pasad2023self}. Experiments on acoustic word discrimination and word segmentation were contributed by Shane Settle and Chung-Ming Chien respectively~\cite{pasad2023self}.} We study \sfms \ varying in model size, pre-training modality, and pre-training objectives; for a detailed background and description of \sfms, refer to \sect~\ref{sec:background-sfm}.
Our study aims to understand how the representations evolve across \sfm \ layers and how the acoustic and linguistic knowledge is encoded across layers and frames. 
We also evaluate layer-wise \sfm \ representations on training-free tasks, giving us a more intuitive understanding of the encoded knowledge in a lightweight and easy-to-scale way.\footnote{\href{https://github.com/ankitapasad/layerwise-analysis}{https://github.com/ankitapasad/layerwise-analysis}}

\section{Analysis framework}
\begin{figure}[btp]
    \centering
    \includegraphics[width=0.65\textwidth]{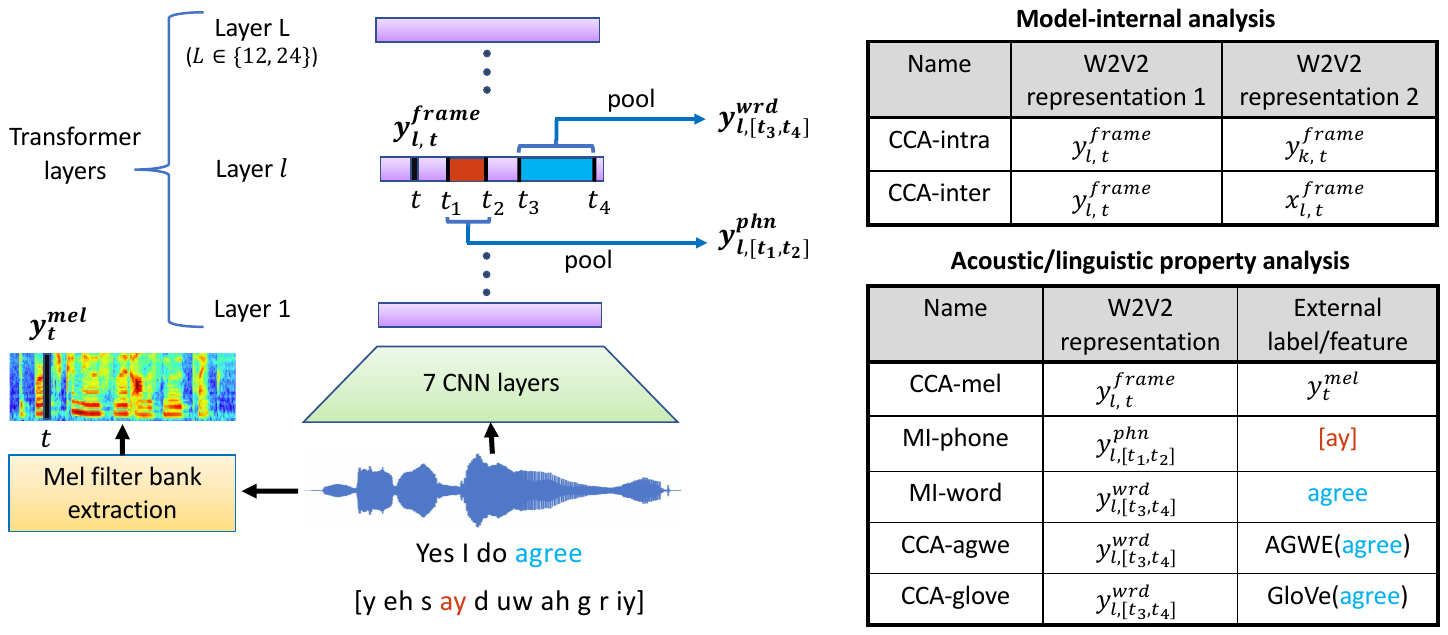}
    \caption{A typical \sfm \ architecture schematic when input is a raw waveform. The example input also shows corresponding span representations extracted at the frame, phone, and word levels.}
    \label{fig:schematic}
\end{figure}

As shown in \fig~\ref{fig:schematic}, a typical \sfm \ processes raw audio waveform through a few CNN layers followed by a stack of transformer layers~\cite{vaswani2017attention}. We'll refer to the CNN module as a feature extractor and the output of the CNN, i.e., input to the transformer module, as local features. Most of our analysis will study local features (layer $0$) and intermediate representations at the output of each transformer module, referred to as a single layer.

We extract frame-level and span-level representations from \sfm \ layers and evaluate the extracted representations using two categories of approaches.
\sect~\ref{sec:analysis-method-cca} describes our first approach using a variant of canonical correlation analysis (CCA)~\cite{harold1936relations}, projection-weighted CCA~\cite{morcos2018insights}; a detailed discussion on CCA is presented in \sect~\ref{sec:tech-cca}. Our second approach uses task-based evaluation where we choose tasks that require no specific training, such as acoustic word discrimination (\sect~\ref{sec:method-awd}), unsupervised word segmentation (\sect~\ref{sec:method-wseg}), and semantic similarity of spoken utterances (\sect~\ref{sec:method-sts}).

\subsection{CCA-based analysis}
\label{sec:analysis-method-cca}
As described in \sect~\ref{sec:tech-cca}, canonical correlation analysis (CCA) is a statistical technique that measures the relationship between two continuous-valued random vectors, where the similarity is calculated as the maximum correlation between their linear projections. CCA similarity has been previously used to compare representations within and across neural network models~\cite{raghu2017svcca, kornblith2019similarity, voita2019bottom}. In addition to providing a reasonable way to correlate random vectors from different subspaces, CCA optimization has a closed-form solution and is quick to compute. Thus, as an analysis tool, CCA provides a lightweight and scalable solution to analyze and understand the ever-growing space of \sfms.

\begin{table}[htb]
\centering
\small
\caption{List of linguistic feature vectors that are compared with the frame-level and span-level \sfm \ representations.}
\resizebox{\columnwidth}{!}{%
\begin{tabular}{lllr}
\hlineB{2}
Representation span & 
\begin{tabular}[c]{@{}l@{}}Linguistic property\end{tabular} & Attribute vector & Dimension \\\hlineB{2}
frame-level & \begin{tabular}[c]{@{}l@{}}spectral content\end{tabular}              & mel filter bank features & 80                                     \\[12pt]
phone-level & \begin{tabular}[c]{@{}l@{}}phone identity\end{tabular}              & one-hot embeddings & 39                                     \\[12pt]
word-level & \begin{tabular}[c]{@{}l@{}}word identity\end{tabular}              & one-hot embeddings & 500                                     \\[12pt]
word-level & \begin{tabular}[c]{@{}l@{}}word pronunciation\end{tabular} & acoustically grounded word embeddings~\cite{shi2021whole}
& 128                                                      \\[12pt]
word-level & \begin{tabular}[c]{@{}l@{}}part-of-speech tags\end{tabular}                                                            & attributes derived from PTB~\cite{tsvetkov2016correlation}
& 45                                                    \\[12pt]
word-level & semantic attributes                                           & 
attributes derived from SemCor~\cite{tsvetkov2015evaluation}
&  41                                                           
\\\hlineB{2}
\end{tabular}
}
\label{tab:cca-exp}
\end{table}
We use projection-weighted CCA (PWCCA) to measure the similarity between the model representations and various quantities of interest formulated as continuous-valued vectors. Next, we describe these feature vectors roughly in the increasing order of contextualization, starting from the localized frame-level \sfm \ representations and time-frequency domain spectrogram features to more complex attributes such as word-level syntactic and semantic embeddings.
\tab~\ref{tab:cca-exp} provides a comprehensive list of these linguistic attributes. 

{\bf \sfm \ representations  (\ccaintra):} We design an intra-model analysis experiment to study the effect of contextualization from self-attention layers. Specifically, we compare the frame-level representations from each self-attention layer with the corresponding local features at the input to the transformer module. The similarity value should reveal how much (or how little) a localized representation changes with depth, i.e., with access to surrounding context via self-attention.

\sloppy
{\bf Filterbank features (\ccafbank):} We compare frame-level representations with the popular mel filterbank features derived from the speech spectrogram. We extract 80-dimensional mel filter bank (FBank) features using a frame length of 25ms and a stride of 10ms. To make the \sfm \ representations comparable to the FBank features, we either compute an appropriate moving average of the extracted features (if the \sfm's stride is smaller) or downsample FBank features as
needed to ensure the same stride and receptive field for the comparison. Since most \sfms \ are trained with raw audio waveforms as input, the \ccafbank \ should reveal whether these models still learn a representation similar to the generic human-engineered FBank features.  

{\bf Phone identity (\ccaphone):} We compare an \sfm's representation of a phone segment with the corresponding phone identity. The discrete identity labels are converted to continuous-valued one-hot vectors for compatibility with CCA computation. \ccaphone \ should reveal whether the deep \sfms, trained without phonetic supervision, learn to encode phonetic content and, if so, how that information evolves across layers. 

{\bf Word identity (\ccaword):} We measure how well the \sfm \ representations encode the word-identifying information by comparing the word segment representations with word IDs. Similarly to \ccaphone, we convert the discrete word ID labels to continuous-valued one-hot vectors. In addition to studying the distribution of information across layers, we use \ccaword \ to also investigate how the word-identifying information is distributed across frames. We design the next three experiments to gain a fine-grained understanding of the type of word knowledge encoded in \sfm \ representations. 

{\bf Acoustically grounded word embeddings (\ccaagwe):} Prior work has proposed acoustically grounded word embeddings (AGWEs), which are vector representations of written words trained to reflect how those words sound~\cite{he2017multiview,settle2019acoustically,hu2020multilingual}. These are learned jointly with representations of spoken word segments, also known as acoustic word embeddings (AWEs). A contrastive learning objective is used to bring AGWEs and AWEs of the same word closer in a shared embedding space, while pushing apart embeddings of different words. This training encourages AGWEs to capture acoustic properties of words, so that words that sound similar (even if spelled differently) are embedded closer together.



We use AGWEs obtained from joint AWE+AGWE training~\cite{shi2021whole} on the LibriSpeech corpus and compare them with word segment representations from \sfms. We expect \ccaagwe \ to measure word-level pronunciation information encoded by the \sfms.

{\bf Syntactic features (\ccasyn):}
\label{sec:ptb}
Tsvetkov et al.~\cite{tsvetkov2016correlation} construct syntactic vectors from the Penn Treebank (PTB)~\cite{marcus1993building}. For each word, an empirical probability is calculated for each of the 45 part-of-speech (POS) tags based on the occurrence in the tagged corpus. This results in 45-dimensional syntactic vectors, and each vector sums to 1 (see \tab~\ref{tab:ptb}). \ccasyn \ trends should reveal if the word segment representations learn to encode any word-level syntactic properties.
\begin{table}[htbp]
\small
\centering
\caption{Examples of syntactic attribute vectors, constructed using PTB~\cite{tsvetkov2016correlation}. }
\begin{tabular}{llllll}
\textbf{WORD} & \textbf{PTB.NN} & \textbf{PTB.NNP} & \textbf{PTB.VB} & $\cdots$ & \textbf{PTB.JJ} \\\hlineB{2}
spring        & 0.94       & 0.04     & 0.02            &    $\cdots$          & 0.00            \\
fall          & 0.49        & 0.00    & 0.43            &     $\cdots$         & 0.00            \\
light         & 0.52        & 0.05    & 0.02            &   $\cdots$           & 0.41           
\end{tabular}
\label{tab:ptb}
\end{table}

PTB is sourced solely from news data, which differs from the typical pre-training data used for \sfms, i.e., audiobooks. We repeat this experiment with syntactic vectors derived from an alternate source to verify whether the domain impacts the CCA correlation trends. We use the tagged LinES corpus\footnote{\href{https://universaldependencies.org/}{https://universaldependencies.org/}}~\cite{ahrenberg2007lines}, which contains sentences from the literature domain. The PTB syntactic vectors are obtained from the original public source\footnote{\label{oracle}\href{https://github.com/ytsvetko/qvec/tree/master/oracles}{https://github.com/ytsvetko/qvec/tree/master/oracles}}, whereas for LinES we derive the vectors ourselves.

{\bf Semantic features (\ccasem):}
\label{sec:semcor-glove}
Tsvetkov et al.~\cite{tsvetkov2015evaluation} exploit word sense annotations in SemCor~\cite{miller1993semantic}, a WordNet-annotated version of the Brown Corpus. For each word, an empirical probability is calculated for each sense attribute (26 nouns and 15 verbs) based on the occurrence in the labeled corpus. This results in 41-dimensional semantic vectors, and each vector sums to 1 (see \tab~\ref{tab:semcor}). The resulting embedding space puts words with similar attributes closer together. For instance, the semantic vector of ``family" is most similar to other words that are predominantly defined with the {\it NN.GROUP} attribute: government, leaders, elite, platoon. This behavior differs from that of a more fine-grained distributional embedding space such as GloVe~\cite{pennington2014glove} where some of the nearest neighbors for ``family" are husband, father, mother, sister, and wife. The semantic vectors are obtained from their public source.\footnote{See footnote \ref{oracle}}
\begin{table}[htbp]
\centering
\small
\caption{Examples of semantic attribute vectors, constructed using SemCor~\cite{tsvetkov2015evaluation}.}
\begin{tabular}{llllll}
 {\bf WORD}            & {\bf NN.GROUP} & {\bf NN.ACT} & $\cdots$ & {\bf NN.ARTIFACT} & {\bf VB.CHANGE} \\\hlineB{2}
family       & 0.96        & 0.04 & $\cdots$           & 0.00  & 0.00      \\
mix          & 0.00          &  0.00     & $\cdots$         & 0.00  & 0.91    \\
headquarters & 0.10     & 0.00        & $\cdots$ & 0.90      & 0.00        \\
industry     & 0.79        & 0.21 & $\cdots$          & 0.00  & 0.00 \\     
hamper     & 0.00   & 0.00         & $\cdots$        & 0.33 & 0.67
\end{tabular}
\label{tab:semcor}
\end{table}

\subsection{Acoustic word discrimination} 
\label{sec:method-awd}
Acoustic word discrimination (AWD) is the task of determining whether or not a pair of acoustic waveform segments $({\bf X}_i, {\bf X}_j)$ correspond to the same word~\cite{carlin2011rapid}. The distance between ${\bf X}_i$ and ${\bf X}_j$ is computed, and the pair is predicted to be ``the same" (i.e., a match) if their distance falls below a threshold value and ``different" otherwise. AWD performance is reported as average precision, i.e., the area under the precision-recall curve generated by varying the threshold. Prior work uses this task to evaluate both segment-level acoustic word embeddings~\cite{levin2013fixed,kamper2016cnn_awe} and frame-level acoustic features~\cite{carlin2011rapid,jansen2013weak_topdown,kamper2015unsupervised}.

We use \sfms \ for AWD in two ways: (1) {\it pool-AWD}: compute cosine distance after mean-pooling the frame-level features, (2) {\it DTW-AWD}: compute dynamic time warping between segments to minimize cosine distance between their frame-level features. We specifically use the {\it DTW-AWD} experiment to tease out the information encoded in individual frame-level representations. 


\subsection{Word segmentation} 
\label{sec:method-wseg}
\begin{figure}[btp]
    \centering
    \includegraphics[width=0.65\linewidth]{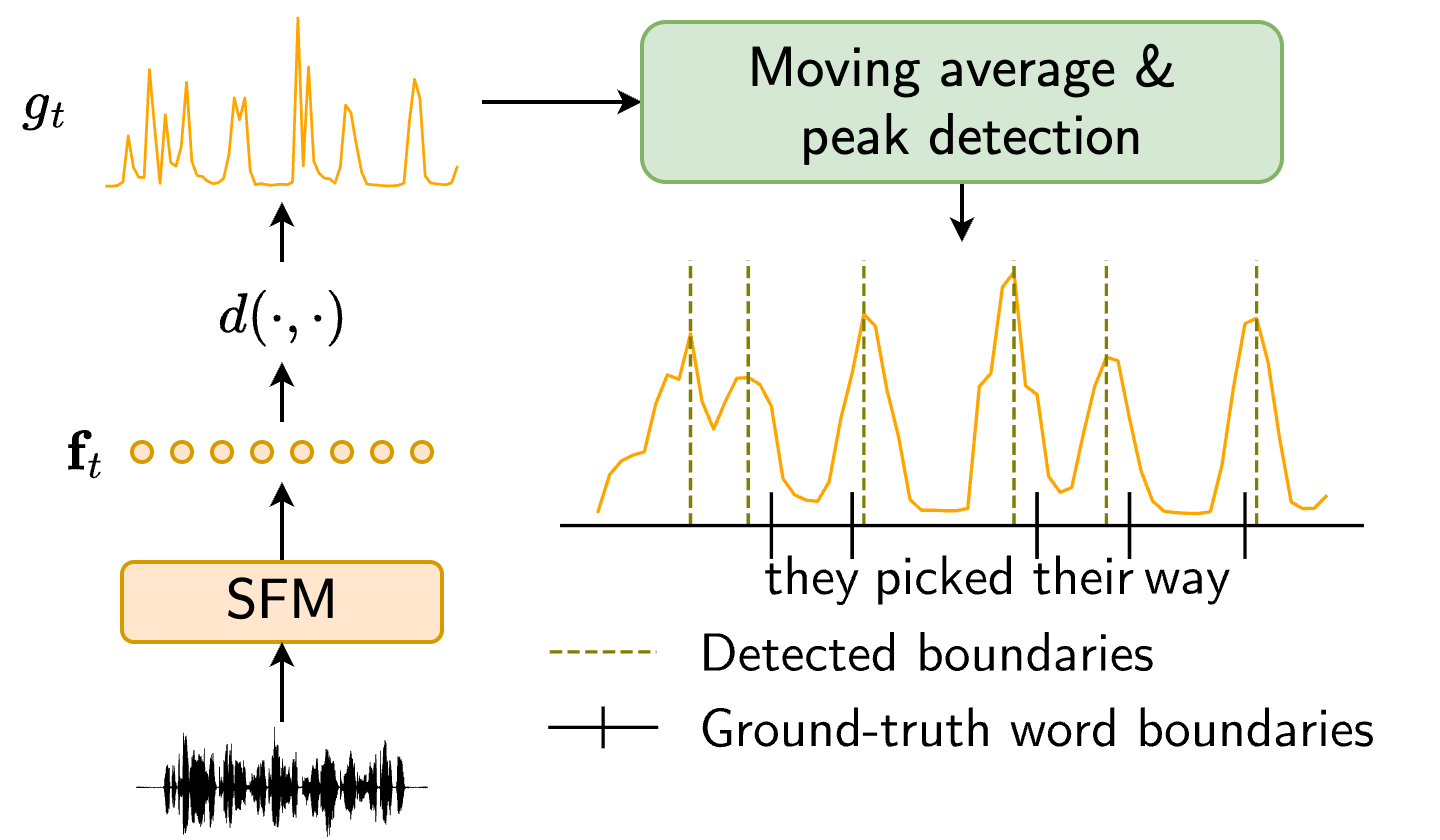}
    \caption{Our word segmentation algorithm.}
    \label{fig:wordseg-illustration}
\end{figure}

Word unit discovery and segmentation are common benchmark tasks that have also been used to study speech representations~\cite{dunbar2020zero, nguyen2022word, sanabria2021difficulty, algayres2022dp, bhati2021segmental, cuervo2022contrastive, ten2007computational}. Previous work examining the segmentation capabilities of \sfms \ rely on complex algorithms, post-processing self-attention weights of an audio-visual \sfm~\cite{peng2022word}, dynamic programming~\cite{kamper2022word}, or a two-stage prediction process with pseudo labeling~\cite{fuchs2023unsupervised}. This work asks how well \sfm \ representations can perform word segmentation ``intrinsically''. We design a straightforward training-free algorithm to identify the behavior of frame-level representations near word segment boundaries (see \fig~\ref{fig:wordseg-illustration}). 

Given a sentence comprising $T$ frames, first, we extract and normalize the frame-level features $\mathbf{f}_t$ from an \sfm \ layer for $1 \leq t \leq T$. Then we compute the dissimilarity between adjacent frames $d(\cdot, \cdot)$ to get $g_t = d(\mathbf{f}_{t+1}, \mathbf{f}_t)$, and smooth $g_t$ with a moving average. Finally, we use a peak detection algorithm to identify adjacent frames with higher dissimilarity than the surrounding frames. These detected word boundaries are evaluated using standard metrics: precision, recall, F1-score, and R-value. Peak detection algorithms have been commonly used for phoneme and word segmentation~\cite{kreuk2020self, bhati2021segmental, cuervo2022contrastive, rasanen2011blind}, but our approach does not rely on any explicit training.

\subsection{Sentence-level semantic similarity}
\label{sec:method-sts}
To evaluate the sentence-level semantics encoded by \sfms, we use {\it spoken STS}~\cite{merkx2021semantic}, which is a spoken (read) version of the popular semantic textual similarity (STS) database~\cite{conneau2018senteval}. STS has sentence pairs annotated with a semantic similarity judgment (see \tab~\ref{tab:sts}). For each utterance in a pair, we extract a sentence-level representation from an \sfm \ layer and calculate the predicted semantic similarity as the cosine similarity between these representations. We report Spearman's $\rho$ correlation between the annotated human judgments and the predicted similarity scores.
\begin{table*}[htbp]
\centering
\small
\caption{Example sentence pairs from the STS data.}
\resizebox{\columnwidth}{!}{%
\begin{tabular}{lc}
\hlineB{2}
\textbf{Sentence pairs}                                                                                                                               & \textbf{\begin{tabular}[c]{@{}c@{}}Human similarity judgement\\ (between 0 and 5)\end{tabular}} \\\hlineB{2}
\begin{tabular}[c]{@{}l@{}}police arrested 2 honduran drug trafficking suspects\\ bulgarian police arrest head of drug trafficking group\end{tabular} & 1.4                                                                                             \\[10pt]
\begin{tabular}[c]{@{}l@{}}cyprus bailout remarks alarm markets\\ why cyprus bailout is nothing more than usual euro nonsense\end{tabular}            & 2.6                                                                                             \\[10pt]
\begin{tabular}[c]{@{}l@{}}former alaska gov sarah palin is no long a commentator for fox news\\ fox news and sarah palin part ways\end{tabular}      & 4.4          \\\hlineB{2}                                                                                  
\end{tabular}
}
\label{tab:sts}
\end{table*}

\section{Experimentation details}
\label{sec:analysis-exp}
The pre-trained checkpoints for the \sfms \ we study are obtained from their publicly available sources. More details on individual \sfms \ are provided in \sect~\ref{sec:background-sfm}.

\subsection{CCA-based analysis}
\label{sec:analysis-exp-cca}
All the CCA experiments use utterances sampled from LibriSpeech~\cite{panayotov2015librispeech}. \tab~\ref{tab:data-specs} summarizes the number of samples used for each experiment. 
For frame-level evaluations, we sample 500 spoken utterances from the dev-clean split, yielding approximately 180k frames. For phone-level and word-level experiments, we extract around 7k segments each, with a roughly uniform distribution over 39 phones and 500 words, respectively. We found that varying the sample selection did not significantly affect results, so we fixed these counts for consistency. While most experiments use the development split, we use the training split for evaluations involving external embedding maps (\ccasyn \ and \ccasem). The training set offers broader coverage of vocabulary and spoken variants, enabling more representative sampling of the larger set of words covered by these external embedding maps.

\begin{table}[htbp]
\small
\begin{center}
\caption{Data subsets curated for our analysis. 
}
\begin{tabular}{l|llll}
{\bf Experiment}                               & {\bf \# labels}          & {\bf \# representation examples} & {\bf LibriSpeech split} & 
\\ \hline 
\begin{tabular}[c]{@{}l@{}}\ccaintra, \\ \ccafbank \end{tabular} & n/a            & 180k frames   & dev-clean     \\
\hline
\ccaphone                          & 39 phones         & 7k phone segments   & dev-clean      \\ 
\hline
\ccaword                          & 500 words         & 7k word segments   & dev-clean     \\ 
\ccasyn                          & 8.5k words         & 314k word segments   & train      \\ 
\ccasem                         & 4k words         & 167k word segments   & train      \\ 
\end{tabular}

\label{tab:data-specs}
\end{center}
\end{table}

For frame-level experiments we sample 500 spoken utterances from LibriSpeech dev-clean, which amount to roughly 180k frames. For phone-level and word-level experiments we use about 7k phone and word segments respectively with a roughly uniform distribution across 39 and 500 tokens respectively. We settle on these counts as we don't see a variance in the results with different sampling of the data. While we use development set for most our experiments, PTB and SemCor provide embeddings for a larger number of words, so we leverage the training data to have a representative distribution of the words. 

The sampled utterances are passed through a pre-trained \sfm \, and the desired span representations are extracted from within the context of the utterance representation at each layer. For span-level analysis on LibriSpeech, we use the ground-truth phone and word alignments generated by the Montreal forced aligner~\cite{mcauliffe2017montreal, lugosch2019speech}.

\fig~\ref{fig:schematic} provides an overview of different representations extracted from an \sfm \ for our analysis. Frame-level representations, $y_l^{frame}$, are extracted at each input frame, where the duration of a frame is decided by the resolution of the \sfm's feature extractor module, which is typically 25 milliseconds resulting from the convolutional neural network layers. Phone-level representations, $y_l^{phn}$, are obtained by averaging the frame representations of the central third of each phone segment, where the first and last third are discarded to reduce co-articulation effects. 

For most of our word-level analysis, $y_l^{wrd}$ is obtained by averaging the frame representations of all frames in a given word segment. For our experiments studying the distribution of the information within a word segment, we experiment with different word segment representations, using either a single frame or mean-pooling across a quarter of the contiguous frames. The single frame is sampled from one of five equidistant locations starting at the first frame. A quarter chunk of adjacent frames is extracted from one of the four quarters of the word segment.

To measure the effect of the model architecture's inductive bias, if any, on the observed trends, we include layer-wise trends for a randomly initialized model (\randinit) as a baseline. Two \randinit \ models are used with 7 CNN and 12 or 24 transformer layers.

For each CCA experiment, three sets of data points are sampled according to the statistics described in \tab~\ref{tab:data-specs}, and the mean score is reported. CCA on each set is evaluated using cross-validation. Specifically, a sample set is partitioned into ten splits, of which eight are used for training (learning the projection matrices $V$ and $W$), one for hyperparameter tuning ($\epsilon_x$ and $\epsilon_y$), and the last for testing. This process is repeated three times with a different train-dev-test split each time. This ensures that the correlation scores are robust against overfitting. 

\subsection{Acoustic word discrimination}
\label{sec:exp-awd}
We evaluate {\it pool-AWD} and {\it DTW-AWD} on ``clean" and ``other" partitions of the LibriSpeech~\cite{panayotov2015librispeech} development set. In all cases, the spoken word segments are $0.5$-$2$s in duration, and segments used for evaluation on LibriSpeech span the vocabulary of 5k words.


\subsection{Word segmentation}
\label{sec:exp-wseg}
We consider two dissimilarity measures between neighboring frames: Euclidean distance and cosine distance. We use a prominence-based algorithm~\cite{virtanen2020scipy} to detect peaks in the dissimilarity curve with a prominence value exceeding a specified threshold. For each layer in each \sfm, we grid search over the choice of distance metric, prominence value threshold, and moving-average window size. We choose the best combination based on F1-scores for word boundary detection on a randomly-sampled subset of the LibriSpeech~\cite{panayotov2015librispeech} dev-clean split ($\sim$2k utterances). We evaluate all layers on the  Buckeye~\cite{pitt2005the} validation set, and the best layer of each \sfm \ is evaluated on the Buckeye test set. 

\subsection{Sentence-level semantic similarity}
The natural speech recordings in {\it Spoken STS} constitute 5\% (638 sentence pairs) of the original STS corpus~\cite{merkx2021semantic}. Sentences in each pair are read by four speakers, and thus, each pair has 16 different speaker combinations. A spoken sentence is represented by mean-pooling across all frame-level representations from an \sfm \ layer. For \vghubert, we also extract the utterance-level {\it CLS} token representations. As in previous work~\cite{merkx2021semantic, zhu2022bootstrapping}, the predicted score for each sentence pair is the mean of cosine similarities between their representations for all speaker combinations.

\section{Results}
\label{sec:res-analysis}
We refer to the output of transformer layer $l$ as the representation at layer $l$ and the output of the CNN feature encoder (or linear layer for \avhubert) as layer 0 or ``local features". We report scores for \whisper-{\it small} and \whisper-{\it medium} in the main text as these match the encoder sizes for other \baseM \ and \largeM \ \sfms. We present results for all five sizes of \whisper \ in Appendix \sect~\ref{sec:appendix-whisper}.

We study frame-level representations (\sects~\ref{sec:res-cca-intra} and \ref{sec:res-cca-mel}), followed by an investigation of phone and word identities encoded in span-level representations and how these are distributed across layers (\sect~\ref{sec:res-cca-surface}) and across frames within the span (\sect~\ref{sec:res-cca-frame}). In the experiments described so far, we also include \randinit \ models and effectively conclude that the observed trends are not a function of the inductive bias of the model architecture. So, moving forward, we drop the \randinit \ models from our study, along with the multilingual model (\xlsr) as we probe span representations for linguistic attributes using English datasets via CCA (\sect~\ref{sec:res-cca-linguistic}) and task-based metrics (\sect~\ref{sec:res-wseg} and \sect~\ref{sec:res-sts}). For these experiments, we also drop \fastvgs, retaining \fastvgsp, an improvement over \fastvgs. Finally we study the effect of data domain for task-based evaluation metrics (\sect~\ref{sec:res-effect-of-domain}). 

\subsection{Evolution of representations across layers}
\label{sec:res-cca-intra}
\begin{figure*}[hbtp]
\begin{minipage}[b]{1.0\linewidth}
\small

 \centering
 \centerline{\includegraphics[width=0.95\linewidth, trim=0 310 0 0, clip]{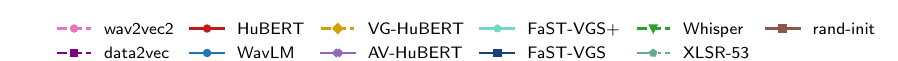}}
 \includegraphics[width=\linewidth, trim=0 252 0 0, clip]{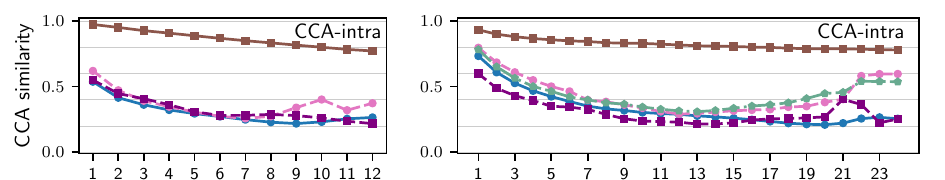}
\includegraphics[width=\linewidth, trim=0 235 0 0, clip]{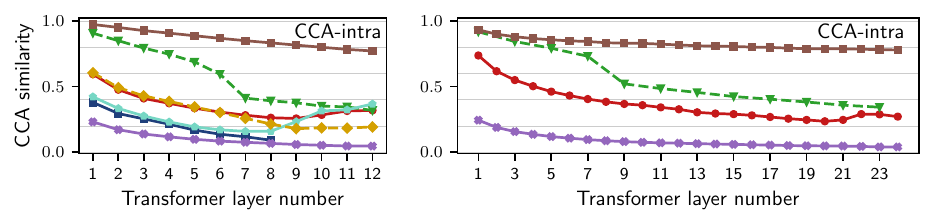}
\end{minipage}
\caption{CCA similarity between \sfm \ frame-level representations and local features, for \baseM \ (left) and \largeM \ (right) \sfms.}
\label{fig:res-cca-intra}
\end{figure*}
In \fig~\ref{fig:res-cca-intra}, we compare the transformer layer representations with the ``local features" at the input of the transformer module (layer $0$).
We find that \sfms \ with the same form of the pre-training objectives exhibit similar \ccaintra \ behavior irrespective of their pre-training data or model size. In a randomly initialized model, the representations change little across layers, providing evidence that our findings are indeed an effect of the pre-training and not an artifact of the architecture.

\sfms \ trained to recover the local features (\wavtovec, \xlsr, and \fastvgsp)
have a clear autoencoder-style trend, i.e., high similarity with the input for the initial and final layers and a drop in similarity for the middle layers. \sfms \ trained to recover an intermediate transformer layer (\wavlm \ and \hubert) also exhibit an increase in similarity for the final layers, but not as prominently as the previous category of \sfms.
The autoencoder-style trend is as expected since the \sfms \ discussed so far are trained to recover features from within the network, and it makes sense that using a deeper layer as a pseudo-ground-truth makes the final \sfm \ layers diverge more from the input than using a shallower layer.
Interestingly, when the pre-training objective combines the top few layers, including the topmost layer, as pseudo-ground-truth (\datatovec), we still see an autoencoder-style trend but with an eventual drop at the deepest layers. 
The autoencoder-style behavior has been observed before for text foundation models, such as BERT, where the initial drop and the following rise in similarity values are called the ``context-encoding" and the ``reconstruction" phases, respectively~\cite{voita2019bottom}.

When \sfms \ recover an external representation via textual (\whisper) or visual (\avhubert, \fastvgs, and \vghubert) supervision, the similarity with input consistently drops, diverging more and more as we go deeper into the network, possibly learning more from the external signal (text or images) and retaining less from the audio modality. Although \fastvgsp \ is also visually grounded and shares \fastvgs' pre-training objective, the final four layers are trained with the \wavtovec-style masked contrastive loss, which explains why the trend differs from other visually grounded \sfms. 

So, just like the final layers of a supervised model are most specific to the task it's trained for~\cite{syosinski2014transferable}, the final layers of a pre-trained \sfm \ are most specialized for its pre-training objective. 

\subsection{Frame-level representations: Spectral acoustic content}
\label{sec:res-cca-mel}
\begin{figure*}[hbtp]
\begin{minipage}[b]{1.0\linewidth}
\small

 \centering
 \centerline{\includegraphics[width=0.95\linewidth, trim=0 310 0 0, clip]{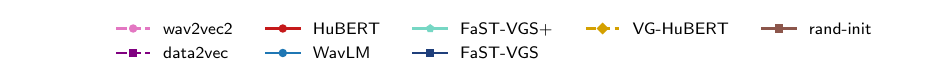}}
 \includegraphics[width=\linewidth, trim=0 252 0 0, clip]{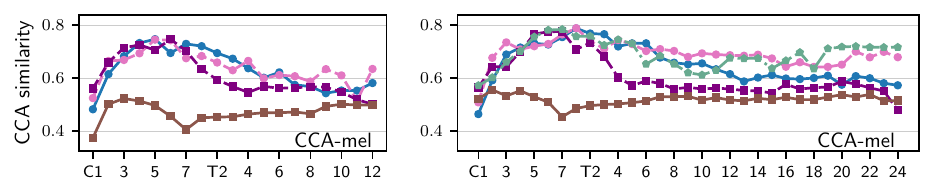}
\includegraphics[width=\linewidth, trim=0 235 0 0, clip]{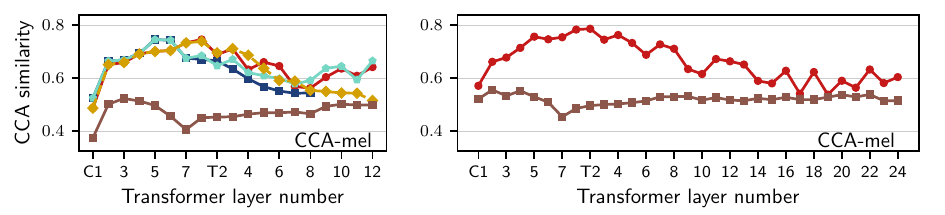}
\end{minipage}
\caption{Spectral content; CCA similarity between \sfm \ frame-level representations and FBank features, for \baseM \ (left) and \largeM \ (right) \sfms.}
\label{fig:res-cca-mel}
\end{figure*}
In \fig~\ref{fig:res-cca-mel}, we measure the correlation between pre-trained features from \sfms \ trained with raw audio inputs and the widely used spectrogram (mel filter bank) features. We drop \avhubert \ and \whisper \ from this analysis as these \sfms \ are trained with spectrogram features. 

For all \sfms \, the final CNN or the initial transformer layers are highly correlated with spectrogram features, suggesting a possible replacement of the CNN module with spectrogram features. Related work confirms our findings by proposing modifications to \wavtovec \ and \hubert~\cite{wu2022performance, lin2022melhubert, parcollet2023efficiency}.
The authors replace the CNN module with a much lighter-weight unit that processes filterbank features, and the modified \sfm \ is shown to achieve comparable performance to the original.
Moreover, our findings are consistent with a recent work using synthetic audio to analyze \wavtovec \ CNN layers~\cite{choi2022opening}. 

For the randomly initialized models, we see a non-trivial trend in the CNN layers, presumably because the CNN architecture has an inductive bias similar to the filtering mechanism of mel feature extraction. However, the correlation values are as low as the lowest scores for the \sfms. 

\subsection{Span representations: Phone and word identities}
\label{sec:res-cca-surface}
\begin{figure*}[hbt]
\begin{minipage}[b]{1.0\linewidth}
\small

 \centering
 \centerline{\includegraphics[width=0.95\linewidth, trim=0 310 0 0, clip]{images/analysis/legend-phone-word-intra.pdf}}
 \includegraphics[width=\linewidth, trim=0 252 0 0, clip]{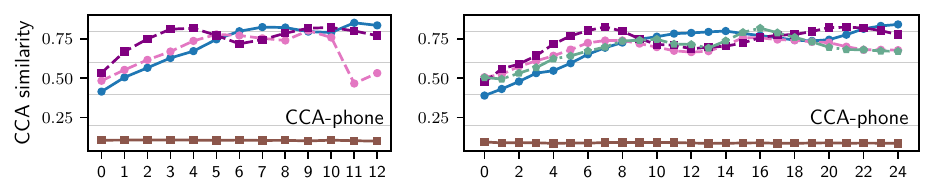}
\includegraphics[width=\linewidth, trim=0 235 0 0, clip]{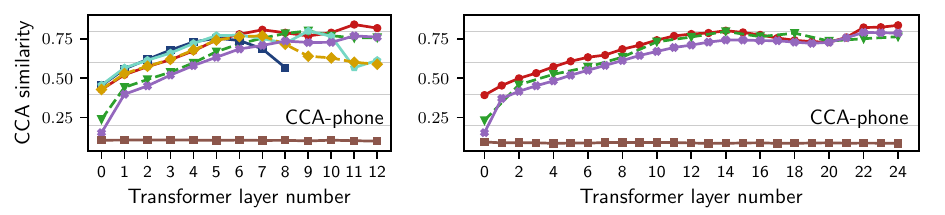}
\end{minipage}
\caption{Phonetic content; CCA similarity between \sfm \ phone segment representations and phone identity, for \baseM \ (left) and \largeM \ (right) \sfms.}
\label{fig:res-cca-phone}
\end{figure*}
\begin{figure*}[hbt]
\begin{minipage}[b]{1.0\linewidth}
\small

 \centering
 \centerline{\includegraphics[width=0.95\linewidth, trim=0 310 0 0, clip]{images/analysis/legend-phone-word-intra.pdf}}
 \includegraphics[width=\linewidth, trim=0 252 0 0, clip]{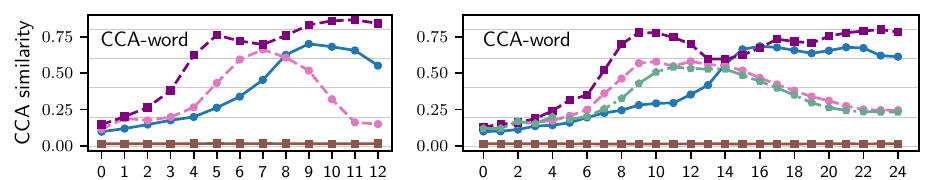}
\includegraphics[width=\linewidth, trim=0 235 0 0, clip]{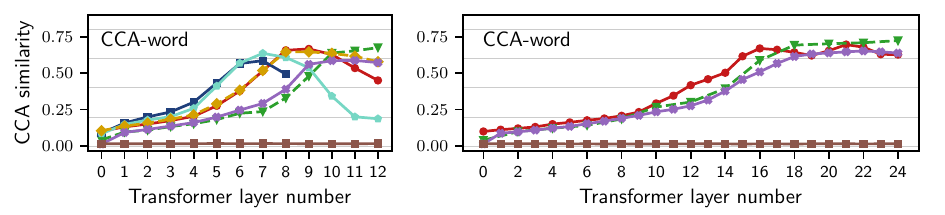}
\end{minipage}
\caption{Word-level content; CCA similarity between \sfm \ word segment representations and word identity, for \baseM \ (left) and \largeM \ (right) \sfms.}
\label{fig:res-cca-word}
\end{figure*}

\figs~\ref{fig:res-cca-phone} and \ref{fig:res-cca-word} show the layer-wise phonetic and word-level content. We note that the \sfms \ that have a strong autoencoder-style dynamic (\wavtovec, \xlsr, and \fastvgsp, as seen in \fig~\ref{fig:res-cca-intra}) tend to have a peak in both phonetic and word content in one or more of the intermediate layers, with a significant drop in the highest layers. These \sfms \ have the same masking-based contrastive objective that recovers the local features. 

In contrast, \sfms \ iteratively trained to predict discrete units learned in an intermediate layer (\hubert, \wavlm, \avhubert) encode these linguistic attributes towards higher layers than for the other (``autoencoder-like") models. Specifically, the phonetic content is best encoded in the final layers, suggesting that the intermediate discrete units, used as targets, may be akin to phone-like units. These phonetic-content trends are consistent with previous observations for \hubert \ using unsupervised clustering~\cite{hsu2021hubert}. Hsu \etal \ also report that the first iteration of \hubert \ trained to predict discrete units learned from mel cepstral features (as opposed to an intermediate layer) has a peak phonetic content at a lower layer. 

Such findings suggest that the trends may be more affected by the latent feature layer used in the pre-training objective rather than the form of the objective itself. In related work, analyses done by Chung et al.~\cite{chung2020similarity} suggest that self-supervised models are more affected by the training objective than the architecture; however, their work did not consider layer-wise trends or the multiple flavors of masked prediction objectives analyzed here.

We consistently see that \datatovec \ trends have two peaks. \datatovec's pre-training objective combines the top few layers in a student-teacher training setup. The location of the first peak coincides with the shallowest layer, which is a part of the pseudo-ground-truth, and the second peak is closer to the final layers, which are also the deepest in the pseudo-ground-truth. We also notice that the peaks for word content are located a couple of layers deeper than the ones for phonetic content.

Comparing \hubert \ to its visually grounded counterpart, \vghubert, we see a drastic drop in phonetic content in the final layers. This suggests that the cross-modal objective may not benefit from encoding phonetic information and modifies the initial \hubert \ parameters to ``forget" the originally encoded phonetic knowledge. Another visually grounded counterpart to \hubert, \avhubert, does not suffer a drop in the phonetic content. We hypothesize that the lip-reading corpus in \avhubert's pre-training could offer certain articulatory phonetic supervision, thus encouraging the final \avhubert \ layers to retain phonetic knowledge. Word content in both \vghubert \ and \avhubert \ has a lower drop in final layers than \hubert, suggesting that visual grounding (either via natural images or lip movements) leads to a better encoding of the word-related properties.

For \whisper, a textually supervised \sfm, the most phonetic content is encoded in an intermediate layer with a slight drop towards the end, whereas the word content monotonically improves as we move from the shallowest to deepest layers. \whisper \ is pre-trained with a speech recognition objective with sub-word tokens, and our findings suggest that retaining word knowledge is more relevant than phones for sub-word prediction.

\subsection{Word-identifying knowledge: Frame-wise distribution and accessibility}
\label{sec:res-cca-frame}
\begin{figure*}[btp]
\begin{minipage}[b]{1.0\linewidth}

\centering     
     \begin{subfigure}[b]{\linewidth}
         \centering
         \includegraphics[width=0.95\linewidth, trim=0 280 0 0, clip]{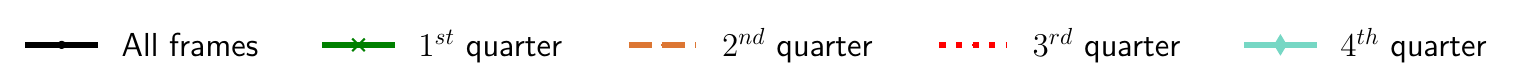}
        \includegraphics[width=\linewidth, trim=0 250 0 0, clip]{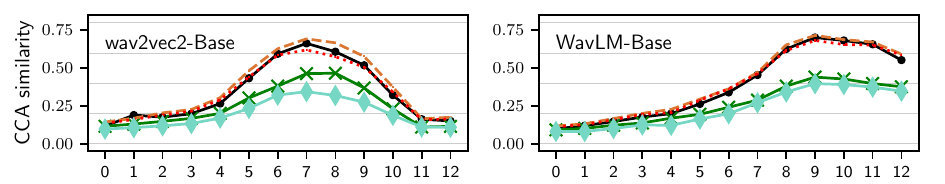}
        \includegraphics[width=\linewidth, trim=0 235 0 0, clip]{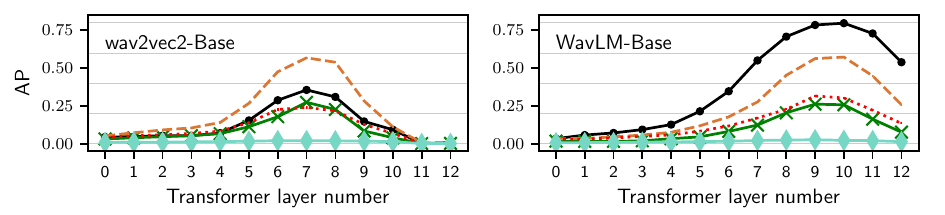}
        \caption{\ccaword \ (top) and {\it pool-AWD} (bottom) scores for \wavtovec-\baseM \ (left) and \wavlm-\baseM \ (right) \sfms \ when pooling over frames spanning quarter segments.}
         \label{fig:res-chunk}
     \end{subfigure}
 
     \begin{subfigure}[b]{\linewidth}
         \centering
         \includegraphics[width=0.95\linewidth, trim=0 320 0 0, clip]{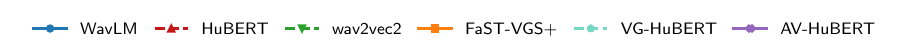}
            \includegraphics[width=\linewidth, trim=0 235 0 0, clip]{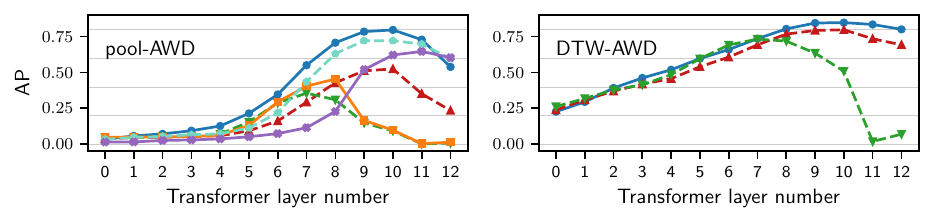}
         \caption{{\it pool-AWD} (left) and {\it DTW-AWD} (right) scores.}
         \label{fig:res-awd-pool-dtw}
     \end{subfigure}
\end{minipage}
\caption{Layer-wise \ccaword \ and word discrimination scores (average precision) with different variants of pooling and evaluation techniques.}
\label{fig:res-frame-loc}
\end{figure*}
We extend our \ccaword \ experiments to study how the word-identifying information is distributed within word boundaries. Specifically, we evaluate word segment representations obtained by pooling over frames spanning different quarters of the segment. Further, we assess these representations on AWD to investigate whether the word-identifying information, as evidenced by high CCA scores, is also easily accessible for task evaluation. We present the results in \fig~\ref{fig:res-frame-loc}.

From our \ccaword \ experiments on two candidate models, \wavtovec-\baseM \ and \wavlm-\baseM, we find that frames near the center of the word segment are most informative of the word identity (\fig~\ref{fig:res-chunk}). Specifically, the 2$^{nd}$ and 3$^{rd}$ quarter spans correlate as highly with the word identity as the mean-pooled representations. These findings are consistent across all \sfms \ we analyze, although all are not shown here.

On the other hand, when studying the chunked representations for {\it pool-AWD}, we see differences in trends for \wavtovec-\baseM \ and \wavlm-\baseM \ (\fig~\ref{fig:res-chunk}). While the frames in the second quarter provide the most informative AWD representation for \wavtovec-\baseM, using all frames is superior to using any subset of frames from \wavlm-\baseM. Additionally, for {\it pool-AWD} (\fig~\ref{fig:res-awd-pool-dtw} {\it left}), the best and worst performing models, \wavtovec-\baseM \ and \wavlm-\baseM, have a significant difference despite being similarly well-correlated with word ID vectors as per \ccaword \ (\fig~\ref{fig:res-cca-word} {\it left}). Since {\it pool-AWD} is a non-parametric approach, this possibly suggests that a more complex AWD model is needed to discover the word-identifying information that \ccaword's learned linear transform can recover from the same subset of frames. 

When we replace pooling with dynamic time warping in {\it DTW-AWD} (\fig~\ref{fig:res-awd-pool-dtw} {\it right}), we find that performance improves consistently, and the performance gap between these models is reduced. \wavtovec-\baseM \ sees the most improvement whereas \wavlm-\baseM \ sees the least change. Since DTW solves for an optimal frame-level alignment, this further corroborates our previous takeaway that some models (such as \wavtovec) distribute discriminative word information across frames in a way that is not easily extracted through pooling, indicating that more structured reasoning over the whole segment may be helpful. 


\subsection{Word segment representations: Pronunciation, syntax, and semantics}
\label{sec:res-cca-linguistic}
\begin{figure*}[hbt]
\begin{minipage}[b]{1.0\linewidth}
\small

 \centering
 \centerline{\includegraphics[width=0.95\linewidth, trim=0 310 0 0, clip]{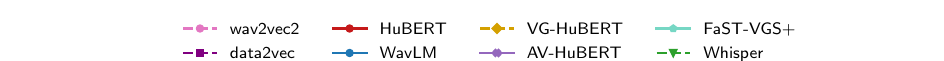}}
 \includegraphics[width=\linewidth, trim=0 252 0 0, clip]{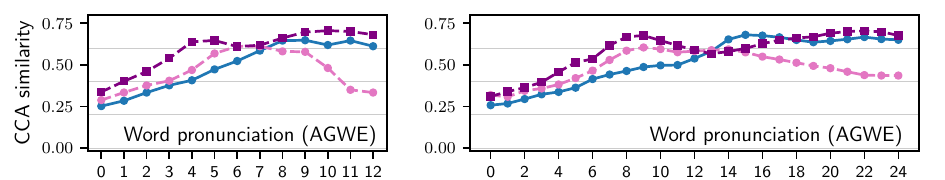}
\includegraphics[width=\linewidth, trim=0 235 0 0, clip]{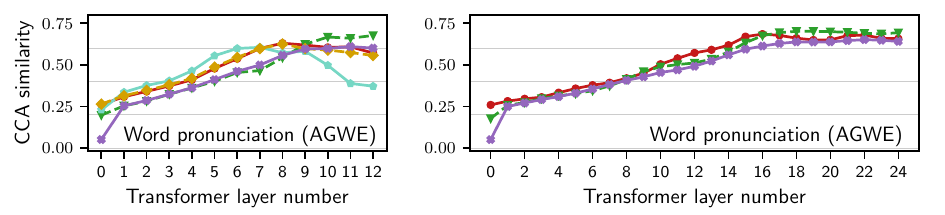}
\end{minipage}
\caption{Word pronunciation content; CCA similarity between \sfm \ word segment representations and AGWEs, for \baseM \ (left) and \largeM \ (right) \sfms.}
\label{fig:res-agwe}
\end{figure*}
\begin{figure*}[hbt]
\begin{minipage}[b]{1.0\linewidth}
\small

 \centering
 \centerline{\includegraphics[width=0.95\linewidth, trim=0 310 0 0, clip]{images/analysis/legend-syn-sem-agwe.pdf}}
 \includegraphics[width=\linewidth, trim=0 252 0 0, clip]{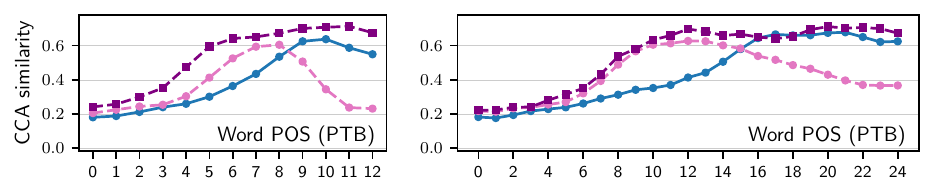}
\includegraphics[width=\linewidth, trim=0 235 0 0, clip]{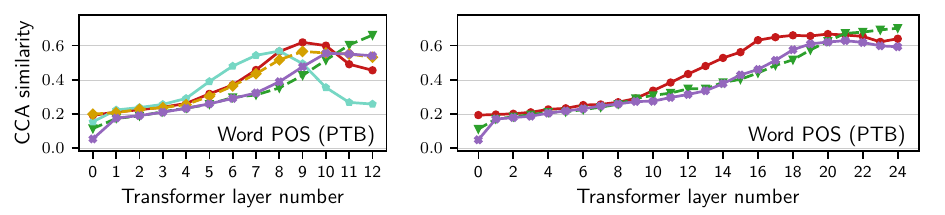}
\end{minipage}
\caption{Syntactic content; CCA similarity between \sfm \ word segment representations and POS attributes, for \baseM \ (left) and \largeM \ (right) \sfms.}
\label{fig:res-ptb}
\end{figure*}
\begin{figure*}[hbtp]
\begin{minipage}[b]{1.0\linewidth}
\small

 \centering
 \centerline{\includegraphics[width=0.95\linewidth, trim=0 310 0 0, clip]{images/analysis/legend-syn-sem-agwe.pdf}}
 \includegraphics[width=\linewidth, trim=0 252 0 0, clip]{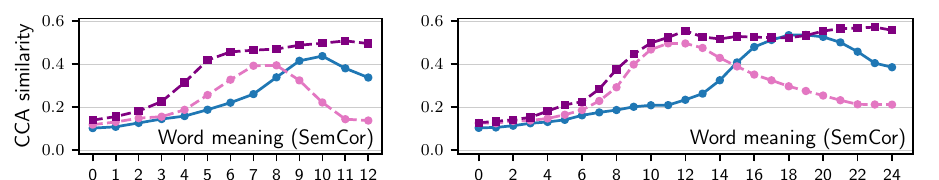}
\includegraphics[width=\linewidth, trim=0 235 0 0, clip]{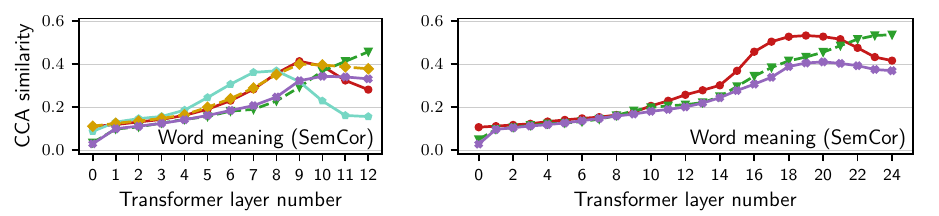}
\end{minipage}
\caption{Semantic content; CCA similarity between \sfm \ word segment representations and SemCor attributes, for \baseM \ (left) and \largeM \ (right) \sfms.}
\label{fig:res-semcor}
\end{figure*}

\figs~\ref{fig:res-agwe}-\ref{fig:res-semcor} shows the layer-wise trends for different word-level linguistic properties, namely the word pronunciation, syntactic, and semantic attributes. Similarly to \ccaphone \ and \ccaword \ trends, we see that the \sfms \ pre-trained to extract local features (\wavtovec \ and \fastvgsp) learn more meaningful features at a lower layer as compared to other models that are all pre-trained to recover discrete units from an intermediate layer. The audio-visual models (\avhubert \ and \vghubert) see no significant drop off even in the final layers. These models are optimized with an additional audio-visual objective, which suggests that meaningful linguistic content is retained better with visual grounding. Lastly, the deepest layers of supervised \whisper \ models consistently encode linguistic properties, better retaining knowledge that helps the decoder to perform accurate speech recognition.

For all \sfms, pronunciation content (\fig~\ref{fig:res-agwe}) is better correlated at lower layers than syntactic (\fig~\ref{fig:res-ptb}) and semantic properties (\fig~\ref{fig:res-semcor}). In \baseM \ models, the same set of intermediate layers is best correlated with syntactic and semantic attributes. This differs from some observations made for BERT, a pre-trained {\it text} model, where different linguistic features---POS, constituents, dependencies, entities, etc.---are encoded best at different layers~\cite{tenney2019bert}. This difference is possible since the speech pre-training objective is primarily local, with much of the model capacity (i.e., most layers) devoted to inferring local acoustic and lower-level phonetic features. Meanwhile, text models that start with higher-level segmented sub-word units can encode fine-grained linguistic properties in different layers.

In line with this reasoning, we observe that the \largeM \ models, with more than three times the parameters of \baseM \ models, have a more pronounced peak for semantic (\fig~\ref{fig:res-semcor}) than syntactic (\fig~\ref{fig:res-ptb}) content, which in turn has a broader plateau than the word pronunciation trends (\fig~\ref{fig:res-agwe}). Coincidentally, the peaks in the semantic trends coincide exactly with the layer that has a slight dip in \ccaword \ trends for the \largeM \ models---specifically layer 11 for \wavtovecl, and layer 19 for \wavlml \ and \hubertl. We hypothesize that since SemCor attributes are more fine-grained, the knowledge may be localized to fewer layers. In contrast, multiple layers surrounding these select few layers are equally good at encoding the comparatively lower-level attributes of syntax and word ID. 

\avhubert \ has the poorest correlation with syntactic and semantic properties, while its final layers have one of the best correlations with other phonetic and word-level properties. This suggests that the audio-visual objective from the lip-reading dataset induces much less meaning-related information in the representations than the speech-only masking-based objective. If the visual modality, lip motion in this case, provides the information needed to recover masked portions, then the model may not need to learn deeper semantics from speech signals.

We note that the syntactic trends are similar irrespective of whether the syntactic attributes are extracted from PTB or LinES corpus (not shown here). This suggests that the trends are not dependent on the domain of the data used to create syntactic attribute vectors. 

\begin{figure}[btp]
    \centering
    \includegraphics[width=0.65\linewidth]{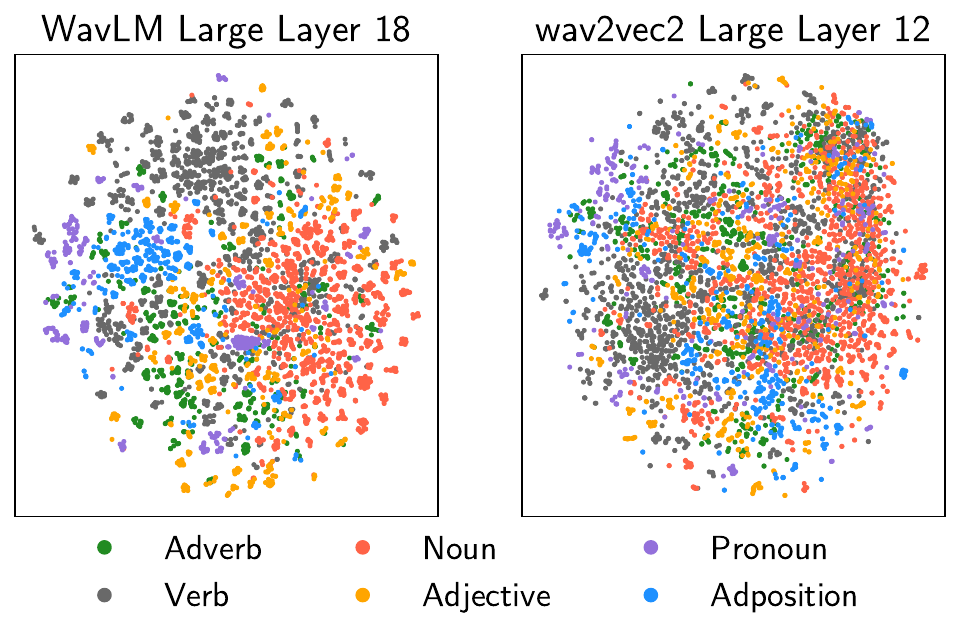}
    \caption{Visualization of the embedding spaces of the intermediate layers of \sfms. Each point represents one word sample. Only the 6 most common POS tags are shown.}
    \label{fig:tsne-pos}
\end{figure}

{\bf Qualitative analysis.} To qualitatively study the syntactic information encoded in \sfm \ representations, we visualize the mean-pooled word representations from the layers with high correlation with the PTB syntactic vectors (\fig~\ref{fig:res-ptb}). We sample $\sim$7k word instances across 500 distinct words and apply t-SNE to project the word representations to 2-dimensional space, as shown in \fig~\ref{fig:tsne-pos}. We find that, for \wavlm, word samples with the same POS tag (especially for verbs, nouns, and adpositions) are encoded into vectors close to each other. However, the representations of \wavtovec\ are not as well-separated. These visualizations further corroborate our findings from CCA trends (\fig~\ref{fig:res-ptb}), where \wavlm\ shows a higher correlation than \wavtovec.

\subsection{Unsupervised word segmentation}
\label{sec:res-wseg}
\begin{figure*}[bhtp]
 \centering
 \centerline{\includegraphics[width=0.95\linewidth, trim=0 320 0 0, clip]{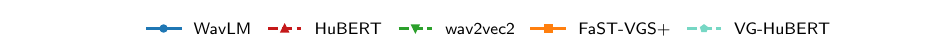}}
     \includegraphics[width=\linewidth, trim=0 235 0 0, clip]{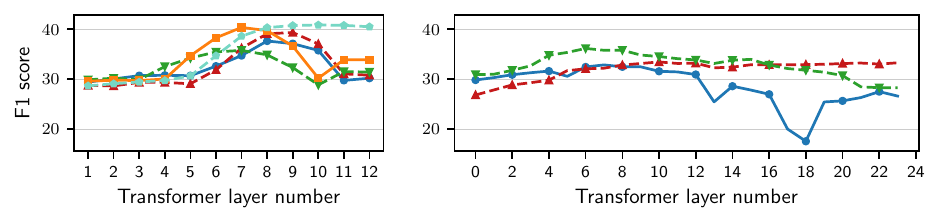}
\caption{F1-scores on the Buckeye validation set for unsupervised word segmentation using representations from \baseM \ (left) and \largeM \ (right) \sfms}. 
 \label{fig:wordseg}
\end{figure*}

\fig~\ref{fig:wordseg} shows the F1-scores of \sfms \ on the word segmentation task.
All of the models demonstrate non-trivial word segmentation capability. This suggests that \sfms \ can learn word boundary information and that the learned information can be easily extracted without the use of delicately designed algorithms.\footnote{We do not include \avhubert~in this experiment as its frame rate is 40 ms, which is larger than the maximum acceptable error of 20 ms on the Buckeye word segmentation task.} We also notice that visually-grounded models perform better than speech-only models, showing the potential of learning word boundary information better with the help of visual contexts. \baseM \ models, in general, have better performance than \largeM \ models, especially for \wavlm \ and \hubert.




\subsection{Sentence-level semantics}
\label{sec:res-sts}

\begin{figure*}[hbtp]
\begin{minipage}[b]{1.0\linewidth}
\small

 \centering
 \centerline{\includegraphics[width=0.95\linewidth, trim=0 310 0 0, clip]{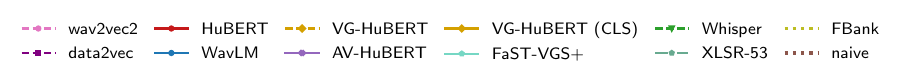}}
 \includegraphics[width=\linewidth, trim=0 252 0 0, clip]{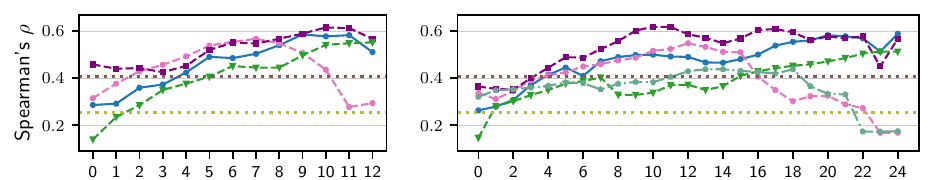}
\includegraphics[width=\linewidth, trim=0 235 0 0, clip]{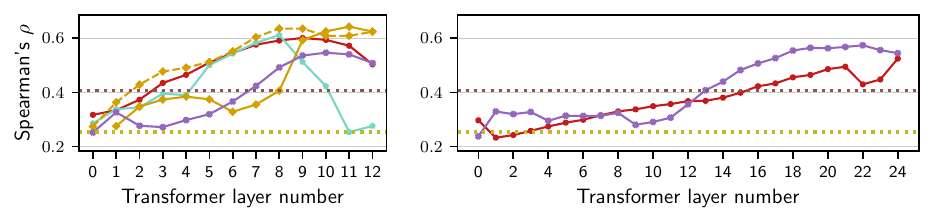}
\end{minipage}
\caption{Performance on spoken STS task using representations from \baseM \ (left) and \largeM \ (right) \sfms.}
\label{fig:sts}
\end{figure*}

\fig~\ref{fig:sts} shows the layer-wise performance on a spoken sentence similarity task.
Our acoustic baseline {\it FBank} represents sentences as mean-pooled filterbank features. The {\it naive} text baseline reports the fraction of word overlap in text transcripts between a pair of sentences and has a non-trivial correlation score of $0.4$. The best-performing layers outperform these baselines by at least 50\%, suggesting that the mean-pooled \sfm \ representations encode meaningful content beyond just the local acoustics and word identities.

The CLS token of \vghubert \ has the best correlation score of $0.64$ at layer 11, closely followed by layer 8 of \vghubert \ and \fastvgsp, both visually grounded models. Surprisingly, \whisper, a textually supervised \sfm \ does not outperform speech-only \sfms. This could be explained by one or both of the following reasons:
(i) \whisper's supervised training of its encoder-decoder architecture possibly encourages more semantic information to be encoded in the decoder layers, whereas our analysis here focuses on the encoder layers, and
(ii) the knowledge encoded in frame-level representations of the \whisper \ encoder may not be properly retained on mean-pooling; the cross-attention between \whisper \ encoder and decoder does not require nor encourage a uniform distribution of knowledge across all frames.

The speech-only \sfms \ we analyze outperform other \sfms \ previously evaluated on this task~\cite{merkx2021semantic, zhu2022bootstrapping}.\footnote{The comparison with \cite{merkx2021semantic} is based on Pearson's correlation, not reported here.} Zhu et al.~\cite{zhu2022bootstrapping} also report a text oracle baseline using self-supervised text embeddings ({\it SimCSE-unsup-RoBERTa}). {\it SimCSE-unsup-RoBERTa} has a correlation score of $0.77$, which, as expected is higher than our best correlation score for an \sfm \ ($0.64$), but is still far from a perfect score.

\subsection{Effect of domain on task-based evaluation}
\label{sec:res-effect-of-domain}
Prior work evaluating \sfms \ on downstream tasks has demonstrated how the relative ranking of \sfms \ may be influenced by the domain of an \sfm's pre-training data as well as the evaluation methodology~\cite{hsu2021robust, yang2021superb, tsai2022superb, zaiem2023speechIS}. For instance, similarly to our task-based experiments (\ref{fig:wordseg} and \ref{fig:sts}), the SUPERB benchmarks~\cite{yang2021superb, tsai2022superb}\footnote{\href{https://superbbenchmark.org/leaderboard}{https://superbbenchmark.org/leaderboard}} and Zaiem et al.\cite{zaiem2023speechIS} report instances where some \largeM \ \sfms \ under-perform their \baseM \ counterparts on downstream tasks.

Next, we discuss our takeaways related to the effect of (mis-)match between the domain of pre-training data and task data on some of our analysis experiments. Some parts of the results discussed here have already been presented before (\figs~\ref{fig:res-awd-pool-dtw} and \ref{fig:wordseg}), but we present these again for the ease of comparison across evaluation datasets.

\begin{figure*}[t]
\begin{minipage}[b]{1.0\linewidth}
\small

 \centering
 \centerline{\includegraphics[width=0.95\linewidth, trim=0 320 0 0, clip]{images/legend-scores.pdf}}
\end{minipage}
\begin{minipage}[b]{1.0\linewidth}

\centering
     
     \begin{subfigure}[b]{\linewidth}
         \centering
         \includegraphics[width=\linewidth, trim=0 235 0 0, clip]{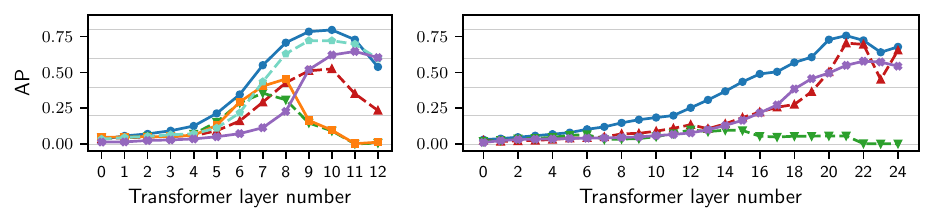}
         \caption{{\it pool-AWD scores on LibriSpeech dev-clean.}}
         \label{fig:res-awe-clean}
     \end{subfigure}

     \begin{subfigure}[b]{\linewidth}
         \centering
         \includegraphics[width=\linewidth, trim=0 235 0 0, clip]{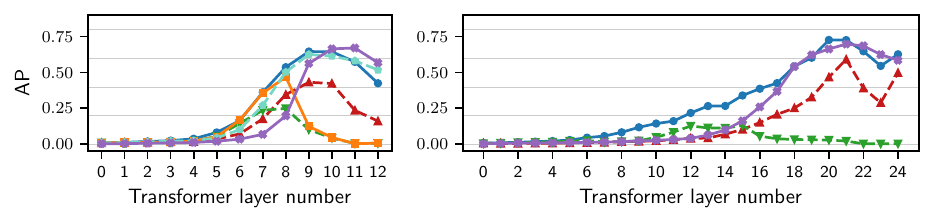}
         \caption{{\it pool-AWD scores on Switchboard dev.}}
         \label{fig:res-awe-swbd}
     \end{subfigure}
     
\end{minipage}
\caption{Evaluation of the word-identifying information in mean-pooled word segment representations from \baseM \ (left) and \largeM \ (right) S3Ms.}

  \label{fig:res-awd-domain}
\end{figure*}

\subsubsection{Acoustic word discrimination}
\label{sec:findings-domain-awd}
We evaluate {\it pool-AWD} on both LibriSpeech (\fig~\ref{fig:res-awe-clean}), a read speech domain, and Switchboard (\fig~\ref{fig:res-awe-swbd}), a conversational speech domain. We observe that the relative ranking of \sfms \ differs for the two settings. For instance, \avhubert \ has better performance on Switchboard, outperforming all \baseM \ models, whereas all other \sfms \ have higher scores on LibriSpeech. \wavlm-\largeM \ outperforms \wavlm-\baseM \ on Switchboard but the larger model under-performs on LibriSpeech. In both cases, the domain of pre-training data provides a potential explanation. Specifically, \avhubert \ models are pre-trained on TED videos~\cite{afouras2018lrs3} and \wavlm-\largeM \ is pre-trained on a mix of data~\cite{chen2021gigaspeech, wang2021voxpopuli} including orated speech and spontaneous speech, whereas all other \sfms \ are trained on read speech domains~\cite{panayotov2015librispeech, kahn2020libri, hsu2020text}. 

We note that some cross-model rankings are consistent across evaluation domains. For instance, \hubert \ and \wavlm, both pre-trained to predict discrete cluster IDs from intermediate layers, outperform \wavtovec, which is trained to recover local features. As seen for other task-based evaluation (\sects~\ref{sec:res-wseg},~\ref{sec:res-sts}), the visually grounded models, \fastvgsp \ and \vghubert, outperform the speech-only \baseM \ models, \wavtovec \ and \hubert, used to initialize them.

Additionally, we observe that the layer-wise trends for all \sfms \ are consistent across evaluation domains and follow a similar dependence on the pre-training objective as noted by our CCA-based results (\sects~\ref{sec:res-cca-surface} and \ref{sec:res-cca-linguistic}).

\begin{figure*}[tbh]
\begin{minipage}[b]{1.0\linewidth}
\small

 \centering
 \centerline{\includegraphics[width=0.95\linewidth, trim=0 320 0 0, clip]{images/legend-wseg.pdf}}
\end{minipage}
\begin{minipage}[b]{1.0\linewidth}

\centering
     \begin{subfigure}[b]{\linewidth}
         \centering
         \includegraphics[width=\linewidth, trim=0 235 0 0, clip]{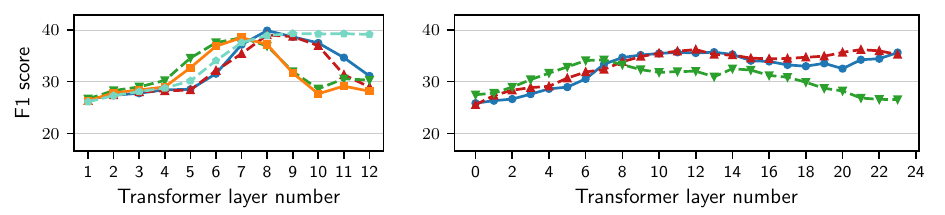}
         \caption{\it F1-scores on the LibriSpeech dev set}
         \label{fig:res-wseg-libri}
     \end{subfigure}
     \begin{subfigure}[b]{\linewidth}
         \centering
         \includegraphics[width=\linewidth, trim=0 235 0 0, clip]{images/wseg_fscore.pdf}
         \caption{\it F1-scores on the Buckeye validation set}
         \label{fig:res-wseg-buckeye}
     \end{subfigure}
\end{minipage}
\caption{{Unsupervised word segmentation using representations from \baseM \ (left) and \largeM \ (right) \sfms}. }
  \label{fig:res-wseg-domain}
\end{figure*}

\subsubsection{Word segmentation}
\label{sec:findings-domain-wseg}
We evaluate word segmentation on LibriSpeech (\fig~\ref{fig:res-wseg-libri}) and Buckeye (\fig~\ref{fig:res-wseg-buckeye}). Similarly to previous findings, we observe that the relative ranking of \sfms \ differs for the two settings. Specifically, \sfms \ pre-trained solely on LibriSpeech (\wavtovec-\baseM, \hubert-\baseM, \wavlm-\baseM) take a much larger hit in performance when evaluated on Buckeye, and the visually grounded models, on the other hand, have a slightly better performance on Buckeye than on LibriSpeech. 
Again, the layer-wise trends for most \sfms \ are invariant to the evaluation domain. \wavlm-\largeM \ does not follow this trend and more than half of the layers have a drastically poorer performance on Buckeye. We hypothesize that the hyperparameters (tuned on LibriSpeech, \sect~\ref{sec:exp-wseg}) transfer better for other \largeM \ models than for \wavlm-\largeM, due to domain mismatch, as discussed above for {\it pool-AWD} (\sect~\ref{sec:findings-domain-awd}).

\section{Summary}
This chapter discussed our findings from analysis of several \sfms, investigating how acoustic and linguistic knowledge is distributed across different layers and frames within spans. We find that the pre-training objective dictates both of these phenomena. Pre-trained representations from different \sfms \ require varying complexities of post-processing to retrieve the encoded knowledge. Additionally, \sfms \ trained to recover a more localized feature tend to have poorer representations at the final layer. \sfms \ that are trained to recover a more contextualized intermediate unit retain phonetic content at the final layers, but deeper linguistic information is still concentrated in the intermediate layers. We also find visually grounded models encode more linguistically meaningful information than their speech-only counterparts, and the textually supervised model encodes most linguistic content in the deepest layers. 

Our analyses have addressed several questions about \sfms' word-level representations, thereby providing a foundation to address more challenging questions.
For example, a natural next step is to ask how much (and where) phrase- and sentence-level properties, such as constituents, dependencies, and entities, are encoded. For some tasks, such as word segmentation, although our results with \sfms \ are stronger than prior work, they are far from solving the task.
Finally, we have noted (as have some prior studies) that larger models are not always better by all measures, raising the question of what the additional model capacity provides and whether there is a better way to train and utilize larger models.

Overall, our findings suggest that the choice of layers could be crucial to provide a richer representation for downstream tasks, and the optimal choice can vary from one \sfm \ to another. Based on this observation, we study the relationship between these task-agnostic trends and downstream task performance in \chap~\ref{ch:compare-tools}
and also discuss implications of our analysis for adaptation of \sfms \ in \chap~\ref{ch:implications}.

\newpage
\chapter{Comparative Study of Analysis Tools}
\label{ch:compare-tools}

In \chap~\ref{ch:analysis}, we learned that \pwcca \ offers a task-agnostic approach to study frame-level, phone-level, and word-level representations from \sfms.
\sect~\ref{sec:background-tools} provided an overview of several statistical tools for studying \sfm \ representations, with \pwcca \ being one of them. 
In this chapter, we'll examine various analysis tools in the context of the framework described in \chap~\ref{ch:analysis}. Specifically, we compare PWCCA and several other tools to understand their relative strengths and limitations, with the goal of informing future research on analyzing speech foundation models.\footnote{Some parts of \sect~\ref{sec:res-compare-tools-tasks} are from our prior published work~\cite{pasad2023comparative}. Bowen Shi helped set up task-specific evaluation of \sfms.}


\section{Methods}
\label{sec:method-compare-analysis-tools}
Our goal is to find a set of reliable and lightweight analysis tools that can be used to measure the knowledge encoded in different \sfm \ layers.
First, we evaluate segment-level (phone-level and word-level) \sfm \ representations using several 
task-agnostic analysis tools (\sect~\ref{sec:compare-tools-method-task-agnostic}); and then we compare these trends with layer-wise performance on specific tasks (\sect~\ref{sec:compare-tools-method-task-specific}).

\subsection{Layer-wise analysis of linguistic properties}
\label{sec:compare-tools-method-task-agnostic}
In \sect~\ref{sec:res-compare-tools-analysis}, we present the task-agnostic layer-wise trends and discuss how different analysis tools vary. We evaluate \sfm \ representations on a subset of the properties from our analysis framework (\sect~\ref{sec:analysis-method-cca}), specifically phonetic, word identity, and word-level semantic content. For evaluation metrics, we use canonical correlation analysis (\cca), singular-value CCA (\svcca), projection-weighted CCA (\pwcca), linear centered kernel alignment (\cka), and orthogonal Procrustes (\op). For phonetic and word-level content, where the second view has discrete labels, we also evaluate discrete MI (\mi) and linear classification accuracy (\linear). \tab~\ref{tab:analysis-tools} summarizes all the analysis tools we study in this chapter, along with corresponding parameters and hyperparameters.

When studying the layer-wise trends, we focus on three things. First, we verify that the scores are robust and not an artifact of a particular choice of data samples. Second, we visually compare the layer-wise trends from different analysis metrics. But this qualitative approach can soon become cumbersome as we compare twelve to eighteen layer-wise curves for each of the eighteen combinations of six \sfms \ and three properties being studied. So, to aid our assessment, we introduce the third approach that evaluates Pearson's and Spearman's correlation coefficient between layer-wise trends. This, in effect, provides a color-coded one-shot view of similarities and dissimilarities in various measures. 

\subsection{Transferability to downstream tasks}
\label{sec:compare-tools-method-task-specific}
In \sect~\ref{sec:res-compare-tools-tasks}, we study the relationship between task-agnostic trends and layer-wise performance of \sfms \ on downstream tasks. This comparison helps us establish whether an \sfm \ layer rich (or poor) in a specific property is also better (or ill-) suited for a related downstream task. We employ the \sfl \ approach described in \sect~\ref{sec:background-sfm-adapt} (\fig~\ref{fig:adapt-strat}) by training a prediction model on top of the layer-wise \sfm \ representations while keeping the \sfm \ frozen. 

We choose two tasks related to {\it token-level recognition}---automatic speech recognition (ASR) and phone recognition (PR) on LibriSpeech~\cite{panayotov2015librispeech}---and two tasks related to {\it semantic content}---intent classification (IC) on Fluent Speech Commands~\cite{lugosch2019speech} and scenario classification (SLURP) on the SLURP dataset~\cite{bastianelli2020slurp}. The relationship is measured as a correlation between layer-wise analysis scores and layer-wise task performance. We study individual scatter plots alongside a quantitative evaluation via Pearsons's and Spearman’s rank correlation coefficients. 

Pearson's correlation measures the linearity of the relationship, and Spearman's correlation can account for non-linear relationships by evaluating the monotonic alignment between the two, indicating whether the two approaches rank individual layers similarly. However, since Spearman's rank correlation ignores absolute values, it can overemphasize differences in ranking, even when the corresponding score values are very close. So, we present both correlation coefficients to provide a more comprehensive assessment. We discuss the individual cases in more depth when we present the scatter plots in \sect~\ref{sec:res-compare-tools-tasks}.

\section{Experimentation details}
We focus our study on \baseM \ and \largeM \ versions of \wavtovec, \hubert, and \datatovec. We choose these three \sfms \ as they vary in their pre-training objectives. See \tab~\ref{tab:sfms} for more details.

\subsection{Layer-wise analysis of linguistic properties}
As detailed in our analysis framework (\sect~\ref{sec:analysis-exp-cca}), we use data sampled from LibriSpeech to extract phone-level and word-level \sfm \ representations. \tab~\ref{tab:analysis-tools-data-specs} lists the number of samples used for each experiment. Prior to metric evaluation, the data is mean-centered along each feature dimension, and for \cka \ and \op, the matrices are further normalized to have a Frobenius norm of 1.



\begin{table}[htbp]
\small
\begin{center}
\caption{Data subsets sampled for analysis experiments.}
\resizebox{\textwidth}{!}{%
\begin{tabular}{l|l|l|rc|c}
\hlineB{2}
\multicolumn{1}{c|}{Granularity} & \multicolumn{1}{c|}{Property}                                  & 
\# distinct types
& \multicolumn{1}{c}{\# samples} &
LibriSpeech split &
Analysis tools
\\ \hlineB{2}
\multirow{2}{*}{phone}          & \multirow{2}{*}{phone ID}                                     & \multirow{2}{*}{39 phones}                                                      & 187k                           & train-clean            & \mi                                     \\
                                &                                                               &                                                                                 & 8k                             & dev-clean  & All tools                                                \\ \hline
\multirow{2}{*}{word}           & \multirow{2}{*}{word ID}                                      & \multirow{2}{*}{500 words}                                                      & 427k                           & train-clean       & \mi                                          \\
                                &                                                               &                                                                                 & 7k                             & dev-clean     & All tools                                           \\ \hline
word                            & \begin{tabular}[c]{@{}l@{}}semantic\\ attributes\end{tabular} & 4000 words                                                                      & 167k                           & train          & \begin{tabular}[c]{@{}l@{}}\svcca, \pwcca, \\ \cca, \cka, \op \end{tabular}                                         
\\
\hlineB{2}
\end{tabular}
}
\label{tab:analysis-tools-data-specs}
\end{center}
\end{table}

We use N-fold cross-validation for \pwcca, \cca, and linear classification, following the setup described in our analysis framework, \sect~\ref{sec:analysis-exp-cca}. For \pwcca \ and \cca \ we sweep $\epsilon_x$ and $\epsilon_y$ between three randomly chosen values from $\{10^{-5}, 10^{-6}, \dots, 10^{-10}\}$.\footnote{For some $(\epsilon_x, \epsilon_y)$ pairs the inverse problem becomes ill-defined and leads to \texttt{numpy.linalg.LinAlgError}; in that case, we sample a different pair of $(\epsilon_x, \epsilon_y)$ from the list.} For linear classification, we used the Adam optimizer~\cite{kingma2015adam} and sweep through learning rates $\in \{10^{-1}, 10^{-2}, 10^{-3}\}$.

CCA is solved using an eigenvalue problem where the resulting eigenvalues of a positive semi-definite matrix are correlation values of data projected in optimal directions (\eqs~\ref{eq:cca1}-\ref{eq:cca-sol-2}). We hypothesize that noise directions may be less correlated than signal directions, and we vary the number of CCA directions used for similarity calculation. Following \eq~\ref{eq:vcca}, we evaluate five variants of \vcca \ assuming $\rho_i$s are in descending order.
\begin{align}
    &\text{CCA}_\text{mean}(X, Y) = \frac{1}{d}\sum_{i=1}^d \rho_i \nonumber \\
    &\text{CCA}_\text{top-k}(X, Y) = \frac{1}{d_k}\sum_{i=1}^{d_k} \rho_i \ 
    \text{where,} \ d_k = \min\{l: \sum_{i=1}^{l}\rho_i/\sum_{i=1}^{d}\rho_i \geq k\} \ \forall k \in \{0.9, 0.7, 0.5\}\nonumber \\
    &\text{CCA}_\text{top-one}(X, Y) = \rho_1 \nonumber
\end{align}

Similarly, we evaluate four variants of SVCCA by varying the thresholds, $\tau_x = \tau_y \in \{0.99, 0.9, 0.7, 0.5\}$ that determine the number of retained singular vectors for the basis (\eq~\ref{eq:svcca}).

For discrete \mi, we cluster continuous-valued \sfm \ representations, using a mini-batch k-means algorithm,\footnote{\href{https://scikit-learn.org/stable/modules/generated/sklearn.cluster.MiniBatchKMeans.html}{https://scikit-learn.org/stable/modules/generated/sklearn.cluster.MiniBatchKMeans.html}} to obtain discrete cluster IDs. The k-means cluster centroids are trained on data sampled from the LibriSpeech train-clean subset (\tab~\ref{tab:analysis-tools-data-specs}). The data used for learning cluster centroids is curated to have roughly same number of examples for each label.\footnote{Similar trends are obtained when the chosen instances are uniformly sampled from the data instead.} \tab~\ref{tab:mi-exp-details} details hyperparameters used for our experiments.

\begin{table}[htbp]
\small
\begin{center}
\caption{Hyperparameters for k-means clustering in \mi \ experiments.}
\begin{tabular}{l|clcc}
\hlineB{2}
\multicolumn{1}{c|}{\bf Property} & \multicolumn{1}{c}{\bf \# Labels} & \multicolumn{1}{c}{\bf \# Clusters ($k$)} & \multicolumn{1}{c}{\bf Max iterations} & \multicolumn{1}{c}{\bf Batch size}
\\
\hlineB{2}
Phones & 39 & \phantom{spac}$k\in \begin{array}{l} \{39, 78, 150 \\ \phantom{\{}350, 500, 1000\} \end{array}$ & 500 & 1500 \\ 
Words & 500 & \phantom{spac}$k\in \begin{array}{l} \{500, 1000, 1500 \\ \phantom{\{}2500, 3500, 5000\} \end{array}$ & 500 & 4000
\\
\hlineB{2}
\end{tabular}

\label{tab:mi-exp-details}
\end{center}
\end{table}

\subsection{Transferability to downstream tasks}
\label{sec:compare-tools-exp-details-task}
We evaluate layer-wise \sfm \ representations on four supervised downstream tasks---intent classification (IC), scenario classification (SLURP), phone recognition (PR), and speech recognition (ASR).
\tab~\ref{tab:analysis-tools-task-datasets} provides the dataset details for each task.  
We closely follow the SUPERB benchmark and the corresponding code-base\footnote{\href{https://github.com/s3prl/s3prl}{https://github.com/s3prl/s3prl}} for designing and training the prediction heads~\cite{yang2021superb}.

\begin{table}[htbp]
\small
\begin{center}
\caption{Data subsets sampled for analysis experiments.}
\begin{tabular}{l|l|l|ccc}
\hlineB{2}
\multicolumn{1}{c|}{\multirow{2}{*}{Task}}  & \multicolumn{1}{c|}{\multirow{2}{*}{Dataset}} & \multicolumn{1}{c|}{\multirow{2}{*}{Output}} & \multicolumn{3}{c}{\# samples (\# hours)} \\
\multicolumn{1}{c|}{} & \multicolumn{1}{c|}{} & \multicolumn{1}{c|}{} & train & dev & test \\ \hlineB{2}
\begin{tabular}[c]{@{}l@{}}Intent\\ classification\end{tabular}   & \begin{tabular}[c]{@{}l@{}}Fluent Speech \\ Commands\end{tabular} & 24 classes & 23k (15h) & 3k (2h) & 4k (3h)  \\
\begin{tabular}[c]{@{}l@{}}Scenario\\ classification\end{tabular} & SLURP & 18 classes & 51k (85h) & 9k (7h) & 13k (10h)  \\
\begin{tabular}[c]{@{}l@{}}Phone\\ recognition\end{tabular} & LibriSpeech & 71 phones & \multirow{2}{*}{29k (100h)} & \multirow{2}{*}{3k (5h)} & \multirow{2}{*}{3k (5h)}  \\
\begin{tabular}[c]{@{}l@{}}Speech\\ recognition\end{tabular} & LibriSpeech & 28 characters & &    &  
\\
\hlineB{2}
\end{tabular}
\label{tab:analysis-tools-task-datasets}
\end{center}
\end{table}

For IC and SLURP, we obtain an utterance-level representation by mean-pooling across frame-level representations from an \sfm \ layer. The utterance-level representation is used as an input to a linear classifier trained with cross-entropy loss. For IC, three cross-entropy loss values, with a shared prediction head, are combined for action, object, and location~\cite{lugosch2019speech, yang2021superb}. IC and SLURP are evaluated using accuracy.

Phone recognition (PR) and speech recognition (ASR) are sequence-to-sequence tasks optimized with the connectionist temporal classification (CTC)~\cite{graves2006connectionist} objective, evaluated at the phone level and character level, respectively.
For PR, we use a frame-wise linear transformation, and for ASR, we train a 2-layer, 1024-dimensional bidirectional LSTM as a prediction head. PR and ASR are evaluated using phone error rate (PER) and word error rate (WER), respectively. 

All the models are trained with the Adam optimizer with a learning rate of $10^{-4}$~\cite{kingma2015adam}. The models are trained for 100k steps for PR and for 200k steps for IC, SLURP, and ASR. The best checkpoint is chosen based on validation set performance. For the \largeM \ \sfms, we evaluate every other layer to limit computation cost.

\section{Results}
\label{sec:res-compare-tools}
We discuss how different analysis tools compare for task-agnostic analysis of the encoded phonetic, word-level, and semantic content in \sect~\ref{sec:res-compare-tools-analysis}. In \sect~\ref{sec:res-compare-tools-tasks}, we compare the task-agnostic trends to layer-wise task-specific evaluation.

\subsection{Layer-wise analysis of linguistic properties}
\label{sec:res-compare-tools-analysis}

\begin{figure*}[btp]

\centering
    \includegraphics[width=\textwidth]{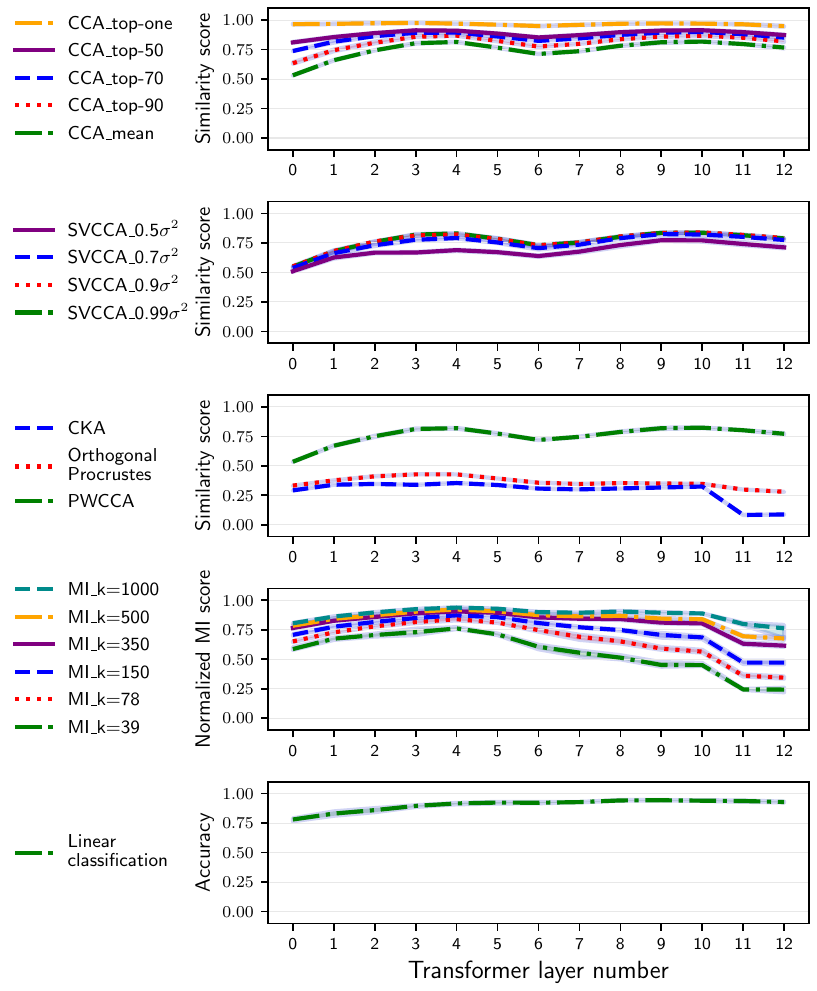}
    


     


\caption{Different tools comparing \sfm \ representations with phone identity for \datatovec-\baseM.}
\label{fig:res-data2vecb-phone-metric-cp}
\end{figure*}
\begin{figure*}[btp]
\centering
    \includegraphics[width=\textwidth]{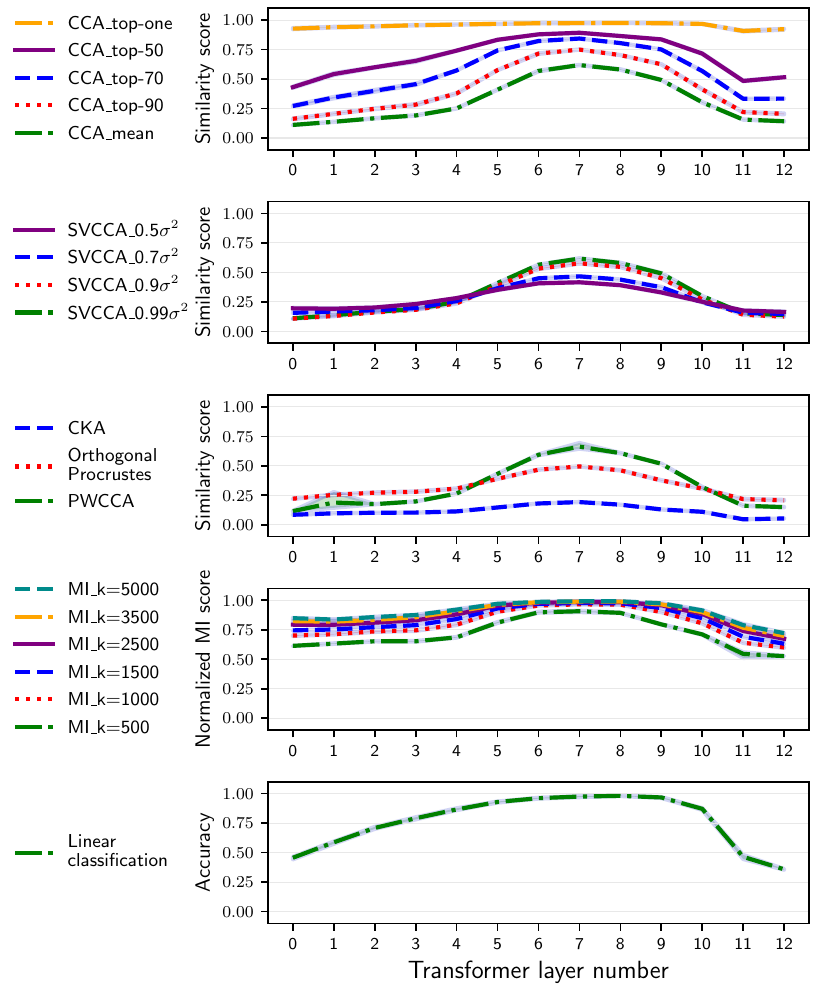}


     



\caption{Different tools comparing \sfm \ representations with word identity for \wavtovec-\baseM.}
\label{fig:res-wav2vecb-word-metric-cp}
\end{figure*}

\begin{figure*}[hbt]
\centering
    \includegraphics[width=\textwidth]{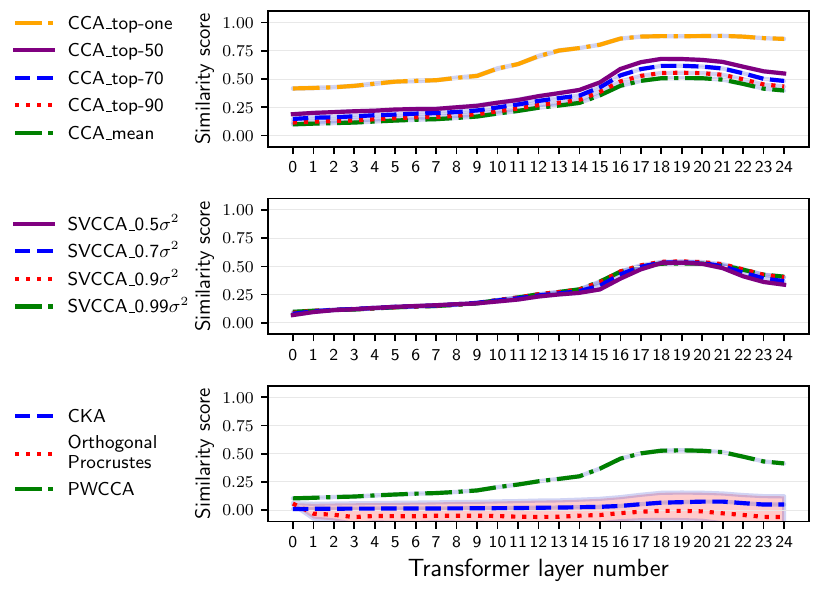}


     

\caption{Different tools comparing \sfm \ representations with semantic attributes for \hubert-\largeM.}
\label{fig:res-hubertl-sem-metric-cp}
\end{figure*}

In \figs~\ref{fig:res-data2vecb-phone-metric-cp}-\ref{fig:res-hubertl-sem-metric-cp}, we present layer-wise phonetic, word, and semantic content trends for \datatovec-\baseM, \wavtovec-\baseM, and \hubert-\largeM \ models, respectively.
Since \cka \ and \op \ are originally distance metrics, we report ($1-\text{CKA distance}$) and ($1-\frac{\text{Procrustes distance}}{2}$), respectively, to facilitate comparison with other metrics. For \mi, we report the score normalized by the entropy of the label distribution, so that it's in the range $[0, 1]$.
The gray shading around the lines indicates the variation in scores across different sample sets, reflecting the robustness of the results to data selection.

We note that some takeaways are very specific to the choice of \sfm \ and the property being studied. For instance, in \fig~\ref{fig:res-data2vecb-phone-metric-cp}, all \cca \ variants exhibit a two peak behavior for phonetic content in \datatovec-\baseM. This trend is more similar to \linear \ but differs from \mi, \op, and \cka, where the peak is at layer $4$, followed by a consistent drop. In \fig~\ref{fig:res-wav2vecb-word-metric-cp}, all metrics have similar trends for word-level content in \wavtovec-\baseM, with slight differences such as the dynamic range of \cka, and behavior of \linear \ around layers $9$ and $10$.
In \fig~\ref{fig:res-hubertl-sem-metric-cp}, all metrics, except \op, have similar trends for semantic content in \hubert-\largeM, but \op \ has a degenerate solution with near-zero scores for all layers.

\begin{figure}[ht]
\centering

\begin{subfigure}[b]{0.49\textwidth}
    \centering
    \includegraphics[width=\textwidth, trim=0 0 30 0, clip]{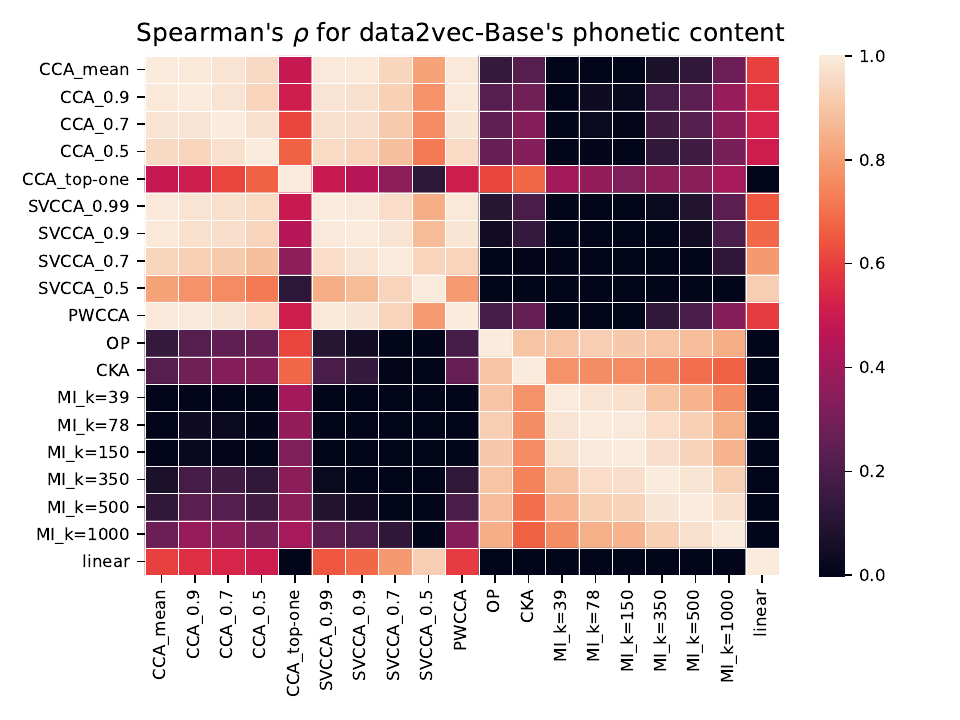}
\end{subfigure}
\hfill 
\begin{subfigure}[b]{0.49\textwidth}
    \centering
    \includegraphics[width=\textwidth, trim=0 0 30 0, clip]{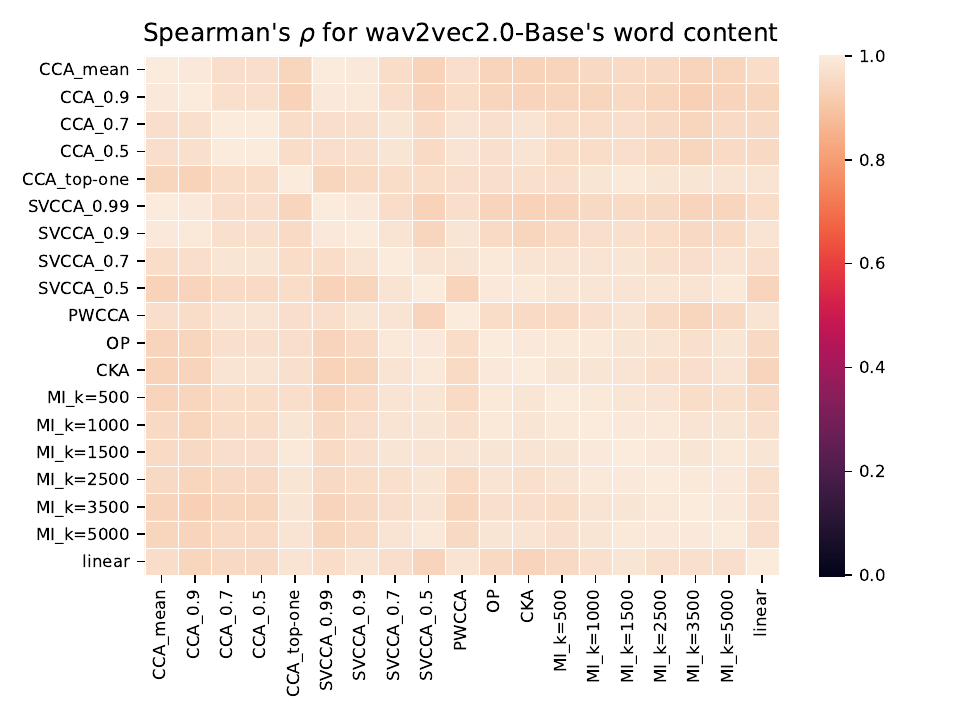}
\end{subfigure}

\begin{subfigure}[b]{0.49\textwidth}
    \centering
    \includegraphics[width=\textwidth, trim=0 0 30 0, clip]{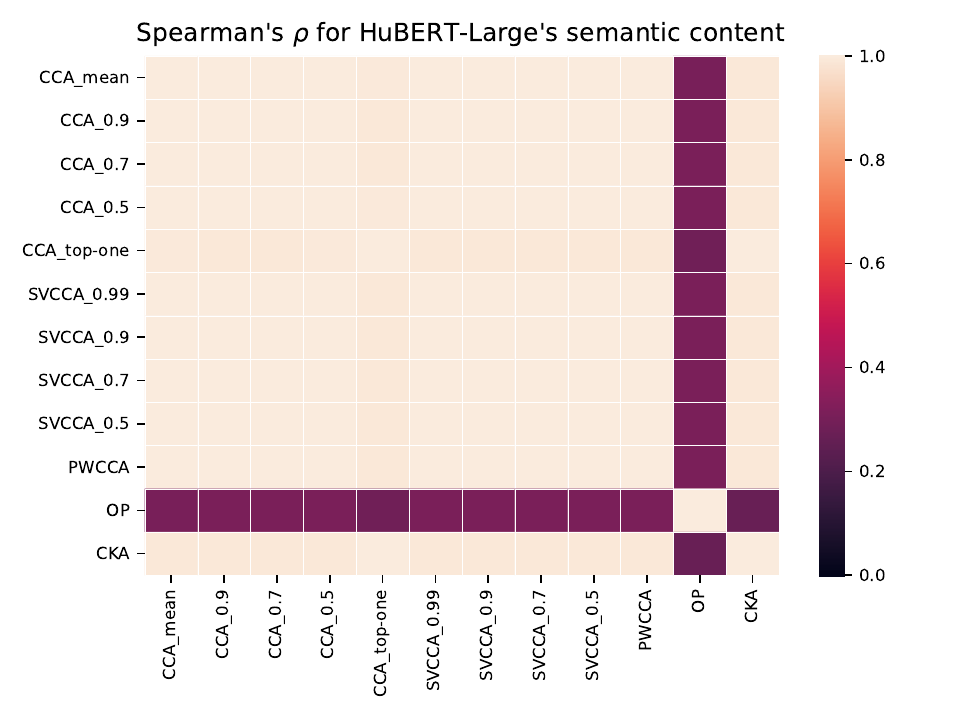}
\end{subfigure}
\hfill 
\begin{subfigure}[b]{0.49\textwidth}
    \centering
    \includegraphics[width=\textwidth, trim=0 0 30 0, clip]{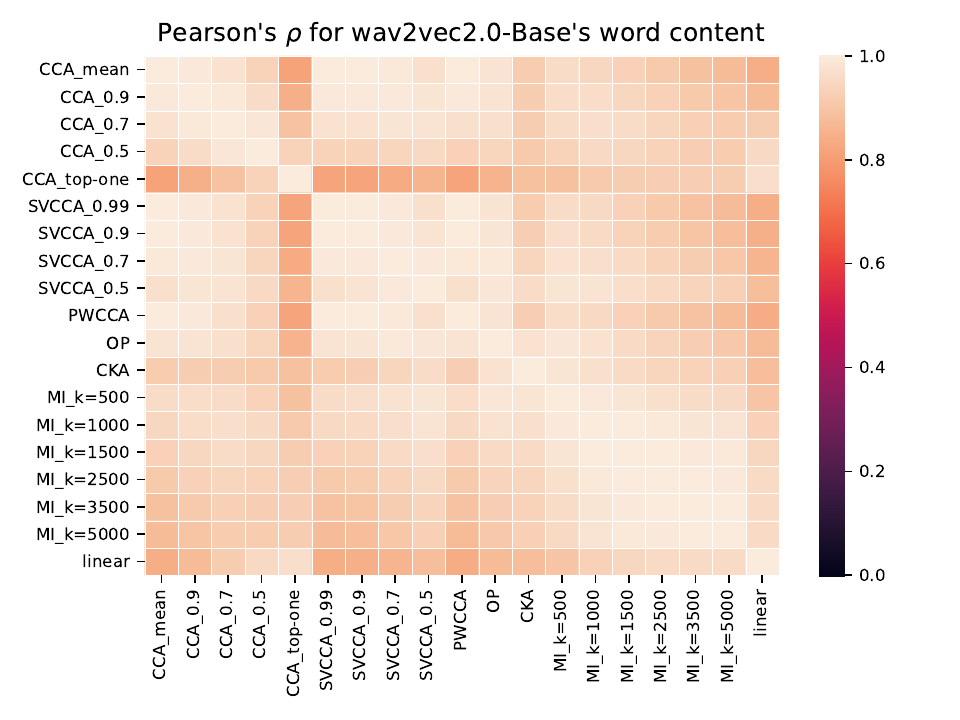}
\end{subfigure}

\caption{Correlation between different analysis tools.}
\label{fig:metric-corr}
\end{figure}

As discussed in Section~\ref{sec:method-compare-analysis-tools}, we also study the correlation between layer-wise trends. Each figure from \figs~\ref{fig:res-data2vecb-phone-metric-cp}-\ref{fig:res-hubertl-sem-metric-cp} can be represented with a single $18 \times 18$ (for phonetic and word content) or $12 \times 12$ (for semantic content) matrix of correlation values, as shown in \fig~\ref{fig:metric-corr}. Our findings from layer-wise trends are reflected here: 
(i) We see two distinct clusters of well-correlated metrics for \datatovec-\baseM's phonetic content, one with CCA variants and \linear, and another with \mi, \op, and \cka,
(ii) While all metrics have a high Spearman's $\rho$ rank-correlation for word-content in \wavtovec-\baseM,  we notice the slight differences in \cka, \linear, and \cca{\it -top-one} highlighted better with Pearson's correlation scores,
(iii) Lastly, all metrics, except \op, have well-correlated layer-wise semantic content trends for \hubert-\largeM.

Appendices~\ref{sec:appendix-metric-plots} and \ref{sec:appendix-metric-corr} present a complete set of plots of layer-wise trends and correlation values studying all the analysis tools for six \sfms \ studying phonetic, word-level, and semantic content. While some observations from comparing analysis tools are specific to the \sfm \ and the property being studied, we summarize some general findings below.

{\bf Are all analysis tools robust?}: Shaded regions around each curve in \figs~\ref{fig:res-data2vecb-phone-metric-cp}-\ref{fig:res-hubertl-sem-metric-cp} indicate how the scores vary when analyzing different subsets of the analysis data. Consistently small shaded areas suggest that all the metrics we study are robust and invariant to the choice of analysis data samples. Additionally, we perform cross-validation for \cca, \mi, and \linear \ by learning and evaluating the parameters (projection matrices for \cca \ and \linear, and cluster centroids for \mi) on different splits. We present results on the evaluation split here, which follow the same trends as results on the respective train splits (not shown here).

{\bf How different are CCA variants?}:
We study property content via \vcca \ with different number of \cca \ directions, \svcca \ with different thresholds on the variance, and \pwcca. 
All CCA variants, except  \vcca{\it -top-one} follow similar trends, as also indicated by the correlation maps in \fig~\ref{fig:metric-corr}.

For phonetic (\fig~\ref{fig:res-data2vecb-phone-metric-cp}) and word-level (\fig~\ref{fig:res-wav2vecb-word-metric-cp}) content, for all \sfms \ (\figs~\ref{fig:res-wav2vecb-phone-metric}-\ref{fig:res-data2vecl-word-metric}), \vcca{\it -top-one} values are consistently high across all layers, but as we add more CCA directions to the calculation of \vcca, we start to notice distinct trends across layers. 
Semantic content does not exhibit this behavior, \vcca{\it -top-one} follows the same trend as other CCA variants (\figs~\ref{fig:res-wav2vecb-sem-metric}-\ref{fig:res-data2vecl-sem-metric}).
The reason is that both phonetic and word-level content are represented as one-hot vectors, which makes it easy to have a highly correlated first projection.\footnote{See Appendix \sect~\ref{sec:appendix-metric-vcca} for an explanation.} So, \vcca{\it -top-one} is not a valid measure when comparing \sfm \ representations with a discrete property.

Since all valid CCA variants have similar trends and share the same insights, we recommend using \pwcca \ or {\it mean-}\vcca, both of which avoid the overhead of additional hyperparameters (number of directions for \vcca \ and \svcca) or pre-processing (SVD step of \svcca).

{\bf Effect of clustering on MI}: In \figs~\ref{fig:res-data2vecb-phone-metric-cp} and \ref{fig:res-wav2vecb-word-metric-cp} we see that \mi \ trends are typically invariant to the number of clusters used to discretize \sfm \ representations.
However, as the number of clusters increases, the absolute \mi \ scores saturate toward $1$, making it difficult to visually compare layer-wise trends due to reduced dynamic range.
This is not surprising, since in the limiting case where the number of clusters equals the number of data points---i.e., each data point forms its own cluster---the normalized discrete \mi \ score would be 1, irrespective of the choice of representations.

{\bf How does CKA compare to the rest?}: \cka \ trends are often dissimilar to either \cca \ or \mi \ or both. Additionally, the dynamic range of \cka \ scores is consistently low, with the highest \cka \ score not exceeding $0.4$ for phonetic and word-level content and $0.2$ for semantic content (\cka \ score can be at most $1$). This could be related to the lower sensitivity of \cka, as reported by Ding et al.~\cite{ding2021grounding} after comparing \cka, \pwcca, and \op. So, using \cka \ as a stand-alone analysis tool can make it difficult to comment if the trends mean anything when the scores are consistently low.

{\bf Is \op \ reliable?}:
Orthogonal Procrustes consistently yields degenerate solutions when evaluating semantic content in \sfms \ (\figs~\ref{fig:res-hubertl-sem-metric-cp}, \ref{fig:res-wav2vecb-sem-metric}-\ref{fig:res-data2vecl-sem-metric}), with scores\footnote{The plots present $1-\frac{\text{Procrustes distance}}{2}$.} near zero across all layers and occasional negative values. These negative scores correspond to Procrustes distance exceeding $2$---its theoretical upper bound, as discussed in \sect~\ref{sec:tech-op}---indicating numerical instability or representational mismatch. This implies that \op \ is not directly suitable for comparing \sfm \ representations with \sem \ attributes.

While further exploration of this failure mode is left for future work, the observed inconsistency raises concerns about \op{}'s reliability for studying \sfm{} representations, despite its alignment with other metrics at the phonetic and word levels.

{\bf Linear classification}: We find that linear classification tends to produce high scores across a slightly broader range of layers, including layers where other metrics show a decline in the trend. This difference is most prominent in the \datatovec \ models, as also evidenced by a lower correlation of linear classification with other metrics (\fig~\ref{fig:metric-corr}). For example, \datatovec-\baseM \ shows consistently high \linear-\phone \ scores in mid-to-late layers (\fig~\ref{fig:res-data2vecb-phone-metric-cp}), and a similar pattern is observed in the word-level setting across both \datatovec \ models (Appendix \figs~\ref{fig:res-data2vecl-word-metric}, \ref{fig:res-data2vecb-word-metric}).

These differences reflect the fact that, although multiple methods involve learning a linear transformation, they optimize different objectives: \cca \ and \op \ assess structural alignment between spaces, while linear classification is trained to extract task-discriminative information. As discussed in the previous chapter, \sect~\ref{sec:res-cca-frame}, \datatovec \ representations tend to make task-relevant information more directly accessible, particularly for word-level discrimination, which may further explain the high linear classification scores in these models.

Taken together, these findings highlight how different analysis methods, despite sharing a common parametric form---a linear transformation---can yield diverging insights depending on their objectives and sensitivity to task-relevant structure.



\subsubsection{Runtime analysis of metrics}
Table~\ref{tab:analysis-tools-time} reports the runtime for a single evaluation run of each analysis metric on a CPU and an NVIDIA L40S GPU.
The runtime for \mi \ is proportional to the number of clusters chosen for k-means clustering; we report durations for the smallest number of clusters from our experiments, i.e., $k=39$ for phone ID and $k=500$ for word ID. 
While we do not report explicit runtimes for \cca \ and \svcca, \cca \ is expected to have a runtime similar to that of \pwcca, and \svcca \ would require added processing time for the SVD computation.


\begin{table}[htb]
\small
\centering
\caption{Time required, in milliseconds, for a single run of different analysis metrics on layer 1 of \wavtovec-\baseM.}
\label{tab:analysis-tools-time}
\begin{tabular}{l|rr|rr|cr}
\hlineB{2}
\multicolumn{1}{c|}{\multirow{2}{*}{\textbf{Analysis tool}}} & \multicolumn{2}{c|}{\textbf{phone ID}} & \multicolumn{2}{c|}{\textbf{word ID}}  & \multicolumn{2}{c}{\textbf{semantic attributes}}  \\
\multicolumn{1}{c|}{}  & \multicolumn{1}{c}{\textbf{CPU}} & \multicolumn{1}{c|}{\textbf{GPU}} & \multicolumn{1}{c}{\textbf{CPU}} & \multicolumn{1}{c|}{\textbf{GPU}} & \textbf{CPU} & \multicolumn{1}{c}{\textbf{GPU}} 
\\ \hlineB{2}
PWCCA & 1,443  & 659  & 4,272  & 1,927  & \multicolumn{1}{r}{13,467} & 6,932  \\[8pt]
CKA & 339  & 115  & 566  & 214  & \multicolumn{1}{r}{7,656}  & 2,527  \\[8pt]
OP  & 96 & 77 & 311  & 159  & \multicolumn{1}{r}{2,413}  & 819  \\[8pt]
discrete MI & 6,613  & 1,910  & 129,358  & 37,145 & \multicolumn{2}{c}{N/A} \\[8pt]
\begin{tabular}[c]{@{}l@{}}Linear\\ classification\end{tabular} & 10,892 & 1,797  &  85,198  & 1,657  & \multicolumn{2}{c}{N/A}  
\\ \hlineB{2}
\end{tabular}
\end{table}

The reported durations do not account for robustness assessments, such as evaluating across different random subsets of data (relevant for all tools), multiple clustering initializations (for \mi), repeated runs needed for N-fold cross-validation (for \pwcca \ and \linear), or hyperparameter tuning (\pwcca \ and \linear).
The number of runs required for robust results varies by method and application.

For instance, while we perform cross-validation for CCA analysis, prior work commonly uses it simply as a scoring algorithm, i.e., using the same set of samples to learn projections and to evaluate the correlation scores. 
While cross-validation can guard against overfitting and spurious correlations, our experience shows that \pwcca \ trends\footnote{layer-wise trends and not necessarily the absolute values} remain consistent with and without cross-validation, as well as across different hyperparameter choices ($\epsilon_x, \epsilon_y$). This suggests that the computational overhead of cross-validation may be unnecessary.

In contrast, hyperparameter tuning is essential for linear classification, which benefits from extensive search over factors such as learning rate, optimizer choice, and stopping criteria. Given this variability, we report single-run durations, allowing users to estimate total computational costs based on their specific choices and use cases.

Finally, our runtime analysis does not address the data efficiency of these algorithms. While some methods may perform well with fewer samples, this investigation is beyond our current scope. When choosing an appropriate analysis method, users should consider these runtime measurements in conjunction with the data specifications provided in Table~\ref{tab:analysis-tools-data-specs}.

\subsection{Transferability to downstream tasks}
\label{sec:res-compare-tools-tasks}

We study the correlation between task-agnostic analysis scores from the previous section and the task-specific scores for the tasks described in \sect~\ref{sec:compare-tools-exp-details-task}. Based on the discussion in the previous section, we study the following metrics: \pwcca, \cka, \mi, and \linear \ for phonetic and word content, and \pwcca \ and \cka \ for semantic content. We select \mi \ with $78$ and $500$ clusters for phonetic and word content, respectively.
We drop all other \cca \ variants as they mostly follow very similar trends to \pwcca, and \pwcca \ has the fewest additional hyperparameters. We also drop \op \ as it leads to a degenerate solution for semantic content. 

\begin{figure}[htb]
    \includegraphics[width=\textwidth]{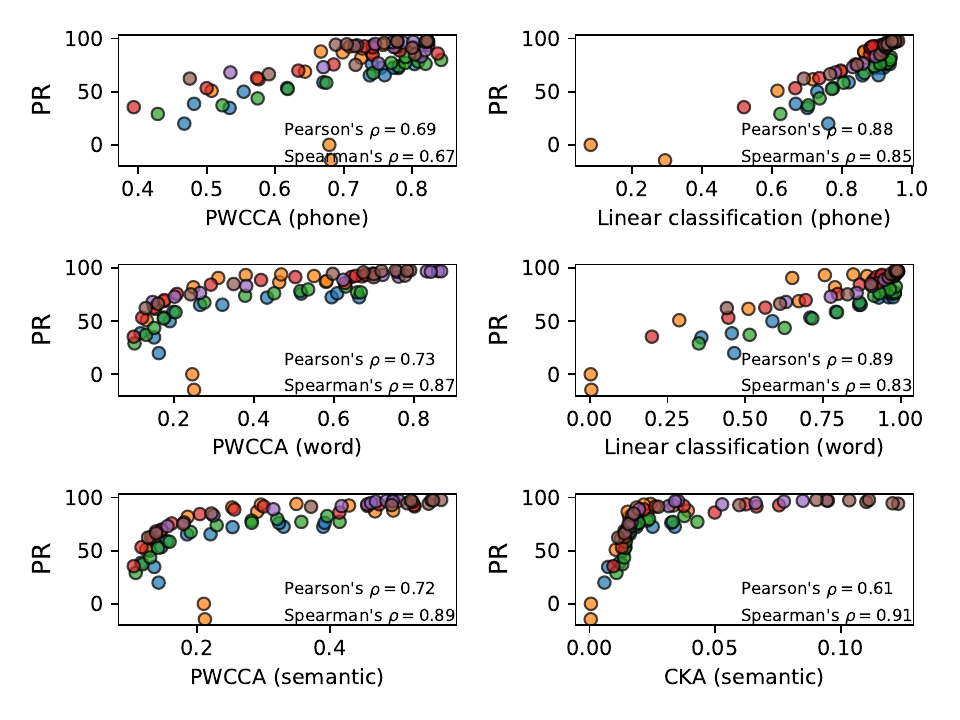}
    \caption{Scatter plots comparing PR performance with task-agnostic layer-wise trends.
    PR is measured as $100 - $error\_rate (in \%), \pwcca \ and \cka \ shown as similarity scores, and linear classification as classification accuracy.} 
    \label{fig:phone-all}
\end{figure}

We find that \pwcca \ and \linear \ are always better correlated with tasks than \cka \ and \mi, so we report the former two in the main text. We also notice that, unsurprisingly, phonetic content consistently correlates poorly with IC and SLURP tasks, so we don't present those in the main text. Lastly, we study the transferability of scores across all \sfms, i.e., we evaluate correlation of task-agnostic and task-specific scores across all layers of all \sfms. This way we can test the utility of analysis metrics to compare layers from different \sfms. We report all the scatter plots, including plots for individual \sfms, in Appendix~\ref{sec:appendix-scatter-plots}.

\begin{figure}[htb]
    \includegraphics[width=\textwidth]{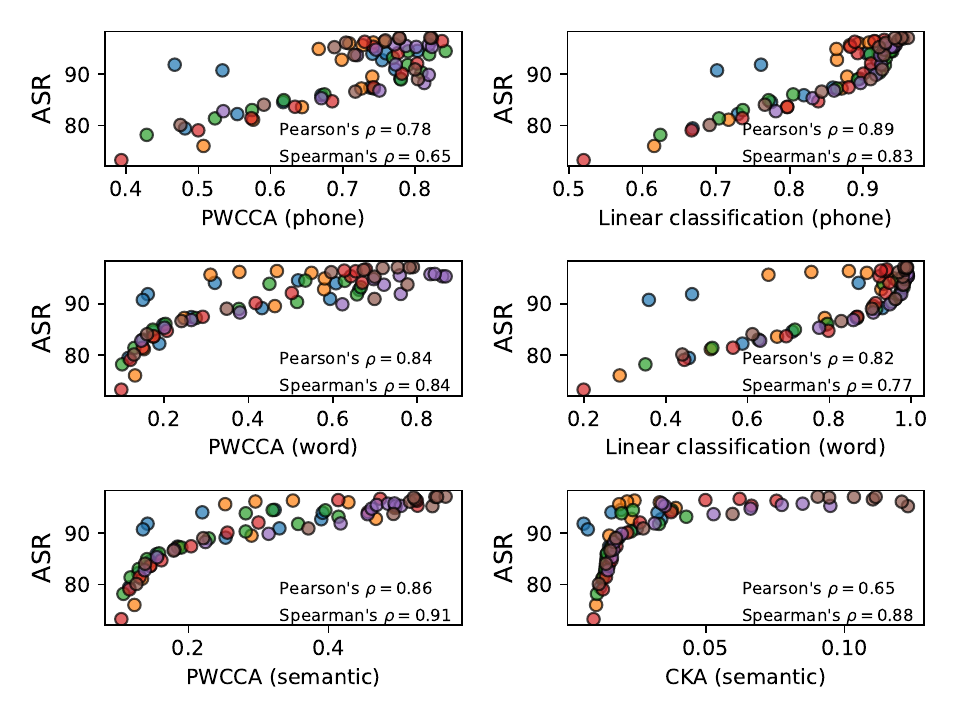}
    \caption{Scatter plots comparing ASR performance with task-agnostic layer-wise trends.
    ASR is measured as $100 - $error\_rate (in \%), \pwcca \ and \cka \ shown as similarity scores, and linear classification as classification accuracy.} 
    \label{fig:asr-all}
\end{figure}

\begin{figure}[htb]
    \includegraphics[width=\textwidth, trim=0 0 0 125, clip]{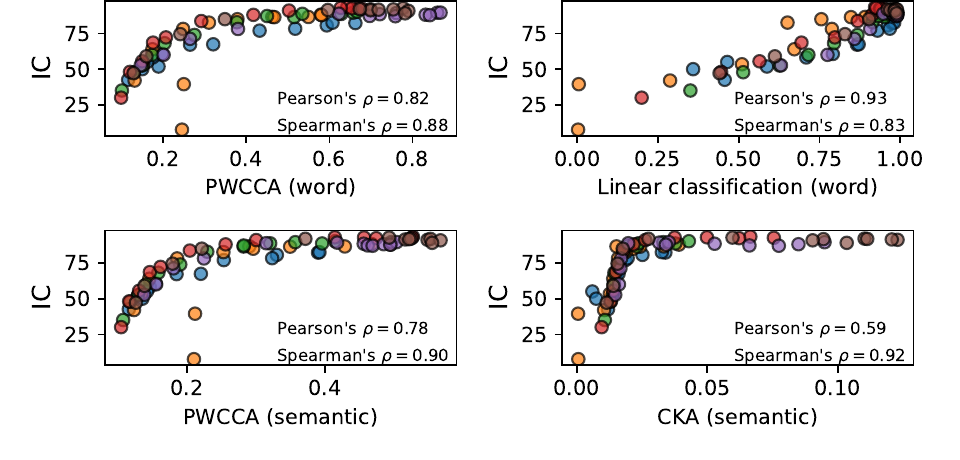}
    \caption{Scatter plots comparing IC performance with task-agnostic layer-wise trends.
    IC is measured as accuracy (in \%), \pwcca \ and \cka \ shown as similarity scores, and linear classification as classification accuracy.} 
    \label{fig:fsc-all}
\end{figure}

\begin{figure}[htb]
    \includegraphics[width=\textwidth, trim=0 0 0 125, clip]{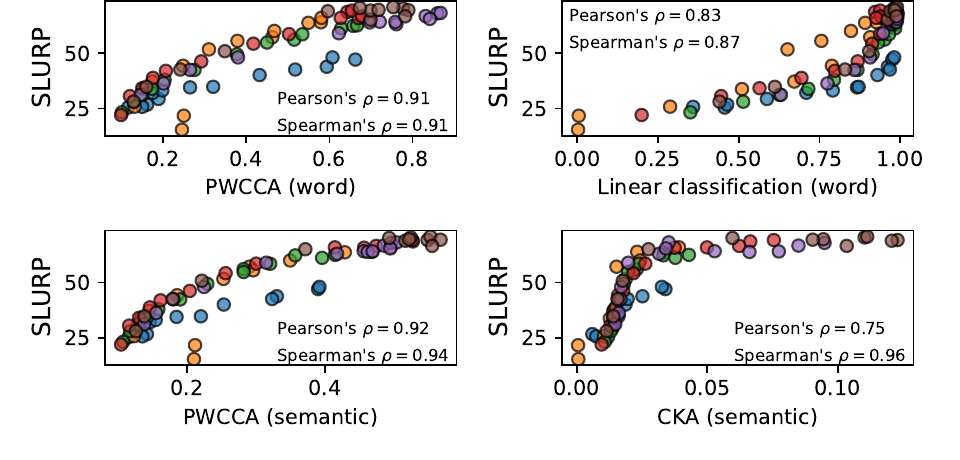}
    \caption{Scatter plots comparing SLURP performance with task-agnostic layer-wise trends.
    SLURP is measured as accuracy (in \%), \pwcca \ and \cka \ shown as similarity scores, and linear classification as classification accuracy.} 
    \label{fig:slurp-all}
\end{figure}

\figs~\ref{fig:phone-all}-\ref{fig:slurp-all} present scatter plots comparing task-agnostic analysis scores from the previous section, for all six \sfms, and task-specific performance for phone recognition (PR) (\fig~\ref{fig:phone-all}), automatic speech recognition (ASR) (\fig~\ref{fig:asr-all}), intent classification (IC) (\fig~\ref{fig:fsc-all}), and scenario classification (SLURP) (\fig~\ref{fig:slurp-all}). We report $100 - $error\_rate for PR and ASR, and for IC and SLURP, we report accuracy. A well-correlated scatter plot indicates that a task-agnostic metric can provide implications for task-based adaptation of an \sfm \ layer representation. 

Most of the scatter plots presented here have reasonable behavior, where we typically observe that (i) layers with a higher property content have a better task performance, (ii) layers with a low property content have poorer task performance, (iii) there is no isolated cluster of points on the bottom right (high content and low task score) or the top left (low content and high task score) corner. However, as discussed below, there are non-trivial differences in how different metrics behave.

All scatter plots with \cka \ (\figs~\ref{fig:phone-all}-\ref{fig:slurp-all}) have a reverse L-shape, where layers with a high task score vary considerably in their \cka \ similarity scores. So, Pearson's correlation is consistently low. Although Spearman's correlation is consistently high, choosing a threshold for a ``high" score will be tricky. This could be connected to \cka's lower dynamic range, also discussed previously. As an exception, \cka \ has well-correlated trends just for \hubert-\baseM, specifically \cka-\phone \ for PR (\fig~\ref{fig:appendix-phone-data2vec_small}) and ASR (\fig~\ref{fig:appendix-asr-hubert_small}), and \cka-\word \ for IC (\fig~\ref{fig:appendix-fsc-hubert_small}) and SLURP (\fig~\ref{fig:appendix-slurp_scenario-hubert_small}). 

All scatter plots with \linear \ (\figs~\ref{fig:phone-all}-\ref{fig:slurp-all}) consistently have a reasonable Pearson's correlation. However, we also notice a cluster of points in the top right corner of all these plots. These points correspond to layers from different \sfms, with equally high linear \ classification accuracy but different task performance. For most individual \sfms, \linear \ offers a high correlation, especially for PR and IC downstream performance (example \figs~\ref{fig:appendix-phone-wav2vec_small}, \ref{fig:appendix-phone-hubert_small}, \ref{fig:appendix-phone-data2vec_small}, \ref{fig:appendix-fsc-wav2vec_vox}, \ref{fig:appendix-fsc-hubert_large}).

Unlike \linear \ and \cka, all \pwcca \ scatter plots have a gradual gradient of increase in performance and analysis scores, and also present a uniform distribution of points. This implies that the \pwcca \ scores offer a reliable comparison, not just for layers within a single \sfm, but also layers from different \sfms.

We notice that \pwcca-\phone \ is an exception to this trend; for both PR (\fig~\ref{fig:phone-all}) and ASR (\fig~\ref{fig:asr-all}) it is not as well-correlated as word-level metrics are. On a closer inspection of individual \sfms \ (\figs~\ref{fig:appendix-phone-wav2vec_small}-\ref{fig:appendix-phone-data2vec_large} for PR and \figs~\ref{fig:appendix-asr-wav2vec_small}-\ref{fig:appendix-asr-data2vec_large} for ASR), we find that \datatovec \ models (\figs~\ref{fig:appendix-phone-data2vec_small}, \ref{fig:appendix-phone-data2vec_large}, \ref{fig:appendix-asr-data2vec_small}, \ref{fig:appendix-asr-data2vec_large}) are key contributors to the poor correlation. For \wavtovec \ and \hubert \ models, as also noted by prior work~\cite{hsu2021hubert, chang2023self}, we see a strong correlation of phonetic content with PR performance. We hypothesize that \datatovec \ may not localize the phonetic content in the frame-level representations as well as \wavtovec \ and \hubert, which are trained on quantized and discretized representation respectively.


For ASR, we remove two outlier points from \wavtovec-\voxM \ (refer to Appendix \fig~\ref{fig:appendix-asr-wav2vec_vox} for the entire plot) to make the trends more straightforward to interpret. Specifically, layers 22 and 24 of \wavtovec-\voxM \ do not converge on ASR despite having a decent \cca \ score of 0.6. So, although such analyses often indicate good correlation with downstream performance, it is essential to remember that task performance depends not only on the representation but also on modeling and optimization issues.


\section{Summary}
In this chapter, we analyzed six \sfms, varying in their pre-training objectives and model sizes, to compare canonical correlation analysis and its variants, centered kernel alignment, orthogonal Procrustes, discrete MI, and linear classification to measure phonetic, word-level, and semantic content. We find that all these metrics are stable across sampling variation, but exhibit some differences in layer-wise trends. Our study indicates that \pwcca \ (or most other \cca \ variants) is most clearly correlated with task performance based on the linguistic properties contained in \sfm \ layers, and also offers a reliable way to compare across layers from different \sfms.

\linear \ scores are well-correlated with task performance when studying a single \sfm, but when studying multiple \sfms, layers with similarly high values could lead to differences in task scores. \cka \ fails to provide a distinct scale of scores, thus making it harder to decide the threshold for a ``high" similarity value. Although \op \ is more lightweight than \cca \ and may seem like a good candidate, it leads to degenerate solutions when studying semantic attributes, which raises concerns about its scalability to a broader range of experiments within a comprehensive analysis framework. \mi \ scores do not offer a reliable correlation with task performance and we suspect that the discretization step distorts the trends.
Based on our findings, we recommend \pwcca \ as the most reliable measure. \pwcca \ has the flexibility to compare with continuous-valued as well as discrete-valued (by converting to one-hot) features and is also lightweight (\tab~\ref{tab:analysis-tools-time}).

We would like to caution the reader that the scale of this analysis, although broad, is still limited: We study four tasks with a single choice of prediction head for each task; and even within the scope of our analysis we notice some unexplained phenomena such as \datatovec's near perfect \linear-\word \ accuracy for most layers that is not captured by any other analysis metric.

\chapter{Implications for Task-Specific Adaptation}
\label{ch:implications}
Pre-trained representations from \sfms \ have provided a new standard for solving speech applications~\cite{mohamed2022self}, and a variety of \sfm-specific adaptation strategies have started to emerge \cite{tsai2022superb, chen2023chapter}.
Some prior work has also observed that a subset of \sfm \ layers can offer similar performance as using all layers~\cite{huang2023findadaptnet, chiulearnable2024, shi2024mmm}.
But, despite the wide adoption of \sfms, the community still lacks a standardized adaptation strategy or a principled approach to designing one. 


In this chapter, we seek to leverage insights from our task-agnostic layer-wise analysis (\chap~\ref{ch:analysis}) to motivate design decisions for adaptation strategies.\footnote{Some parts of this chapter (\sects~\ref{sec:adapt-afl} and \ref{sec:adapt-tff}) are from our prior published papers~\cite{pasad2021layer, pasad2023comparative}. Ju-Chieh Chou ran speech recognition experiments with the re-init adaptation strategy.} 
\sect~\ref{sec:background-sfm-adapt} covers the adaptation strategies that we will use in this chapter, specifically, \sfl, \afl, \tff, and parameter-efficient fine-tuning (PEFT) approaches.

\section{Implications for \afl}
\label{sec:adapt-afl}




\begin{figure}[htb]
  \begin{minipage}[c]{\textwidth}
 \centerline{\includegraphics[width=0.75\textwidth, trim=0 305 0 10, clip]{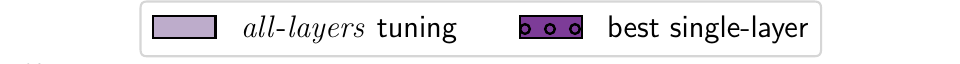}}
 \centerline{\includegraphics[width=0.75\textwidth]{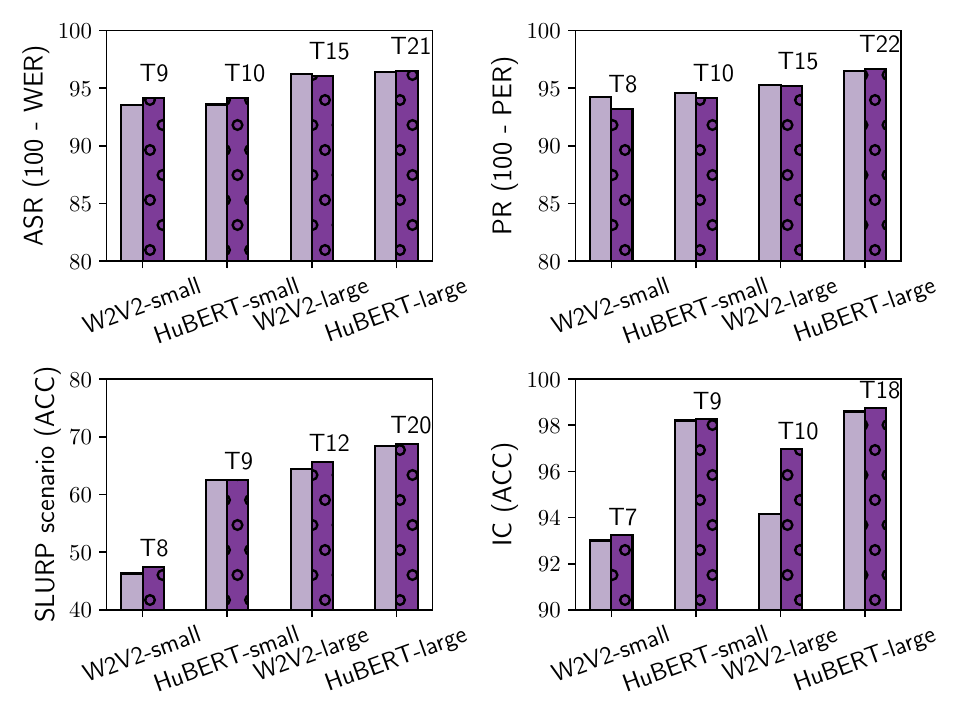}}
  \end{minipage}
  \begin{minipage}[c]{\textwidth}
\caption{Task performance and best layer for all tasks.}
  \label{fig:task-scores}
  \end{minipage}
\end{figure}
In \sect~\ref{sec:res-compare-tools-tasks} we presented the correlation of task-agnostic analysis metrics with task-specific performance.
Here we compare the performance of \afl \ with \sfl; specifically, we investigate whether a single layer can match the performance of using a weighted combination of all layers; and if so, is there a way to choose that layer based on our task-agnostic analysis. 

\fig~\ref{fig:task-scores} shows the scores from \afl \ and the best \sfl \ performance, along with the best layer for the latter. We not only find that the best \sfl \ performance is at least as good as \afl \ performance for most tasks but also that the best-performing layer is always lower than at least the top two layers and is close to the layers observed to have the most phonetic and word-level content as measured by CCA (see \figs~\ref{fig:res-cca-phone} and~\ref{fig:res-cca-word}). Specifically, for all semantic tasks, the best-performing layer is one of the top 3 in terms of \ccaword, and for ASR and PR, the best-performing layer is one of the top 6 in terms of both \ccaphone \ and \ccaword. Thus, our CCA-based analysis framework can effectively help narrow down the choice of layers relevant to a downstream task, thus reducing the compute requirements for downstream task adaptation compared to using all layers in \afl.


Alternatively to the presented approach, it is natural to ask whether the layer weights learned in \afl \ experiments are a good indicator of usefulness for downstream tasks. The learned layer weights in the \afl \ strategy and the \sfl \ performance have a mean rank correlation of 0.66 across all ten (task, model) pairs. In contrast, the mean rank correlation between \ccaword  \ and task performance is 0.90, even though \ccaword \ is a generic measure while layer weights are task-specific. This further strengthens our proposed analysis-driven design, which is also more lightweight than training a \afl \ model for each layer to obtain layer importance.


\section{Implications for \tff}
\label{sec:adapt-tff}
\begin{figure}[htbp]
\centering
    \includegraphics[width=0.75\textwidth, trim=0 160 0 0, clip]{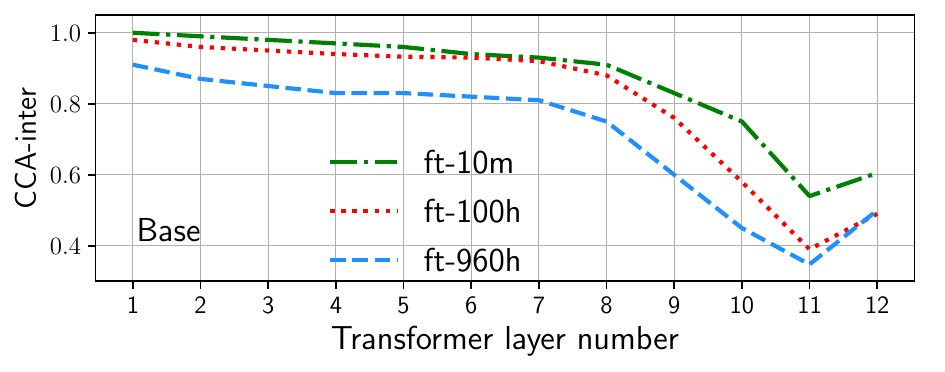}
    \caption{CCA similarity between each layer of \wavtovec-\baseM \ before and after \tff \ on 960 hours of LibriSpeech for ASR.}
  \label{fig:res-cca-inter}
\end{figure}
We know from our previous analysis that the linguistic content that should be helpful for ASR is less well represented in the final few layers of \wavtovec \ (\ccaphone \ in \fig~\ref{fig:res-cca-phone} and \ccaword \ in \fig~\ref{fig:res-cca-word}). We also study the effect of \tff \ on \wavtovec-\baseM \ by computing the CCA similarity between frame-level representations from the corresponding layers of the pre-trained and fine-tuned models ({\it CCA-inter}).\footnote{ASR fine-tuned \wavtovec-\baseM \ is obtained from the fairseq public repo.} \fig~\ref{fig:res-cca-inter} presents the {\it CCA-inter} scores, and as expected, we see that the last few layers of \wavtovec-\baseM \ change the most during \tff.

Based on these observations, we hypothesize that some final layers do not provide a good initialization for the task. To test this hypothesis, we modify the standard \tff \ approach by re-initializing the top layer(s) before performing \tff. We find that re-initializing the final 1-3 layers indeed outperforms the standard approach of initializing all layers from the pre-trained \sfm \ (\tab~\ref{tab:w2v2-asr}), with significant improvements when fine-tuning on the 10-minute training set and minor improvements for larger training sets.

\begin{table}[tbh]
\small
\centering
\caption{WER (\%) for the modified fine-tuning protocol for the \wavtovec-\baseM \ model, using the best value of $n$ based on dev-clean performance. {\it A\textrightarrow{}B} indicates that standard fine-tuning produces WER {\it A}, and the proposed protocol produces WER {\it B}.}
\label{tab:w2v2-asr}
    
\begin{tabular}{cc|cc}
\multirow{2}{*}{\textbf{train set}} & \multirow{2}{*}{\textbf{$n$}} & \multicolumn{2}{c}{\textbf{standard \textrightarrow{} re-init }12-$n$ \textbf{layers}} \\ \cline{3-4}
                                    &                             & \textbf{test-clean}      & \textbf{test-other}     \\ \hline
10m                                 & 9                           & 49.0 \textrightarrow{} 44.1               & 56.7 \textrightarrow{} 51.8              \\
1h                                  & 11                          & 20.3 \textrightarrow{} 19.8               & 29.8 \textrightarrow{} 29.3              \\
10h                                 & 11                          & 11.3 \textrightarrow{} 10.9               & 20.6 \textrightarrow{} 19.4             
\end{tabular}
\end{table}

\section{Implications for PEFT}

There are a variety of approaches to adapt an \sfm, and most recently, parameter-efficient fine-tuning (PEFT) approaches are becoming commonplace~\cite{chen2023exploring, chang2023prompting, li2023evaluating}.
These task-specific models use the frozen parameters of a pre-trained \sfm \ and rely on only a handful (compared to the size of an \sfm) of additional parameters specifically adapted for the task.
We provide background on various PEFT approaches in \sect~\ref{sec:background-sfm-adapt}.

We aim to extend our analytical tools to understand how these adaptation approaches interact with the pre-trained representations and further reduce the parameter size of PEFT algorithms. Specifically, we experiment with the placement of LoRA modules at only a subset of layers. Fewer LoRA modules imply fewer trainable parameters, and if the optimal LoRA placement involves only the deepest layers, the backward pass need not pass through the complete model, thus reducing training time. 

We build on top of Chen et al.'s public codebase~\cite{chen2023exploring}.\footnote{Original codebase:\href{https://github.com/virginiakm1988/s3adapter}{https://github.com/virginiakm1988/s3adapter}\\Updated codebase maintained at: \href{https://github.com/ankitapasad/s3adapter/tree/wip}{https://github.com/ankitapasad/s3adapter/tree/wip}}

\subsection{Experimental details}
We train \hubert-\baseM \ with LoRA modules for dialog act classification (DAC)~\cite{shon2023slue} and SLURP scenario classification~\cite{bastianelli2020slurp} tasks on 7 and 85 hours of training data respectively. For both DAC and SLURP we use a linear prediction head with 18 classes.\footnote{Coincidentally, both DAC and SLURP have the same number of classes.}

All LoRA experiments use a hidden dimension of 8, added to the key and query projection matrices of the self-attention layer. We tried higher dimensions up to 64 but did not see much relative improvement in performance. We use Adam optimizer with a linear warmup schedule. We tune the learning rate between $0.1$, $0.01$, $0.001$, and $0.0001$ and warm up steps between $500$, $1000$, and $3000$. A learning rate of $0.001$ consistently works the best, but we don't notice a significant impact with different numbers of warm up steps. The prediction head has $13.8k$ parameters and each LoRA module adds $24.6k$ parameters.



We tune the learning rate and warmup steps for \tff \ and \afl \ experiments as well. We find that $0.001$ and $0.00005$ are the best learning rates for \afl \ and \tff \ respectively.

We experiment with different numbers of LoRA layers and report the scores for the best placement for each configuration. We experiment with every possible combination of layers for up to 4 layers for DAC; i.e., 12 possible placements when placing a single LoRA layer, 66 possible placements when placing two LoRA layers, and so on. Lastly, we experiment with LoRA placement for 5 and 6 layers, where we either choose {\it top-k} layers or a combination of the bottommost and topmost layers. 

As SLURP experiments take longer to converge (it has much more fine-tuning data), we don't experiment with all possible combinations of layers. We choose a set of best combinations that work for DAC.

\subsection{Results}
\label{sec:res-implications-peft}
\tab~\ref{tab:lora-res-dev} presents results on the development sets of the DAC and SLURP scenario tasks.\footnote{For completeness, corresponding test set results are in Appendix \tab~\ref{tab:lora-res-test}. We report results on the development set in the main text, as they more directly reflect the effect of layer placement.}
For both of these tasks, we see that the fine-tuned model (\tff) significantly outperforms the frozen model (\afl). When we place LoRA on all layers, DAC performance matches that of \tff. For SLURP, applying LoRA on all layers improves over \afl \ but still falls short of \tff.

\begin{table}[htb]
\centering
\small
\caption{Comparison of DAC dev macro F1 and SLURP scenario dev accuracy across different LoRA configurations. As both tasks, DAC and SLURP, have the same number of classes, the prediction head adds the same number of trainable parameters.}
\label{tab:lora-res-dev}
\resizebox{\columnwidth}{!}{%
\begin{tabular}{l|l|c|l|c|l|r}
\hlineB{2}
\multicolumn{2}{c|}{} & \multicolumn{2}{c|}{\textbf{DAC}} & \multicolumn{2}{c|}{\textbf{SLURP scenario}} & \multirow{2}{*}{\begin{tabular}{c} \textbf{Trainable} \\ \textbf{params} \\ \textbf{(M)} \end{tabular}} \\
\cline{3-6}
\multicolumn{2}{c|}{\multirow{-2}{*}{\textbf{Method}}} & \textbf{macro F1} & \begin{tabular}{c} \textbf{LoRA} \\ \textbf{placement} \end{tabular} &
\textbf{Accuracy} & \begin{tabular}{c} \textbf{LoRA} \\ \textbf{placement} \end{tabular} & 
\\
\hline
\multicolumn{2}{c|}{\afl}
  & \cellgradsg{68.0} & \multicolumn{1}{c|}{N/A} & \cellgradsh{62.7} & \multicolumn{1}{c|}{N/A} & 0.01 \\
\multicolumn{2}{c|}{\tff} & \cellgradsg{74.8} & \multicolumn{1}{c|}{N/A} & \cellgradsh{88.6} & \multicolumn{1}{c|}{N/A} & 90.18 \\
\hline
\multirow{6}{*}{\shortstack[l]{\# LoRA\\layers}}
  & 1   & \cellgradsg{71.5} & 12 & \cellgradsh{69.1} & 12 & 0.04 \\
  & 2   & \cellgradsg{73.6} & 11,12 & \cellgradsh{71.7} & 11,12 & 0.06\\
  & 3   & \cellgradsg{74.6} & 1,3,10 & \cellgradsh{72.5} & 10,11,12 & 0.09 \\
  & 4   & \cellgradsg{75.7} & 1,7,11,12 & \cellgradsh{76.8} & 2,8,11,12 & 0.11 \\
  & 4+  & \cellgradsg{74.9} & 1,2,3,10,11,12 & \cellgradsh{77.8} & 1,2,3,10,11,12 & 0.16 \\
  & ALL & \cellgradsg{75.5} & ALL & \cellgradsh{79.0} & ALL & 0.31 \\
\hlineB{2}
\end{tabular}
}
\end{table}

The goal of our experimentation is to see if an optimal placement of LoRA modules exists. For DAC, we notice that placing LoRA modules at just four layers matches the \tff \ performance.
And interestingly, the optimal placement combines the bottommost (layer $1$) and top layers (layers $7,11,12$). For SLURP, the performance improves with more LoRA modules, but even with six LoRA modules, it does not match having LoRA at all layers. Similarly to our observations on DAC, the best-performing combination of layers includes the bottommost and topmost layers. 

We hypothesize that LoRA modules at the shallowest layers are relevant for domain adaptation, since the downstream task datasets differ from \hubert-\baseM's pre-training data. LoRA modules at the topmost layers must be necessary to close the gap between pre-training and the fine-tuning task.

\subsection{Analysis and discussion}
\begin{figure}[btp]



\centering
\centerline{\includegraphics[width=\linewidth]{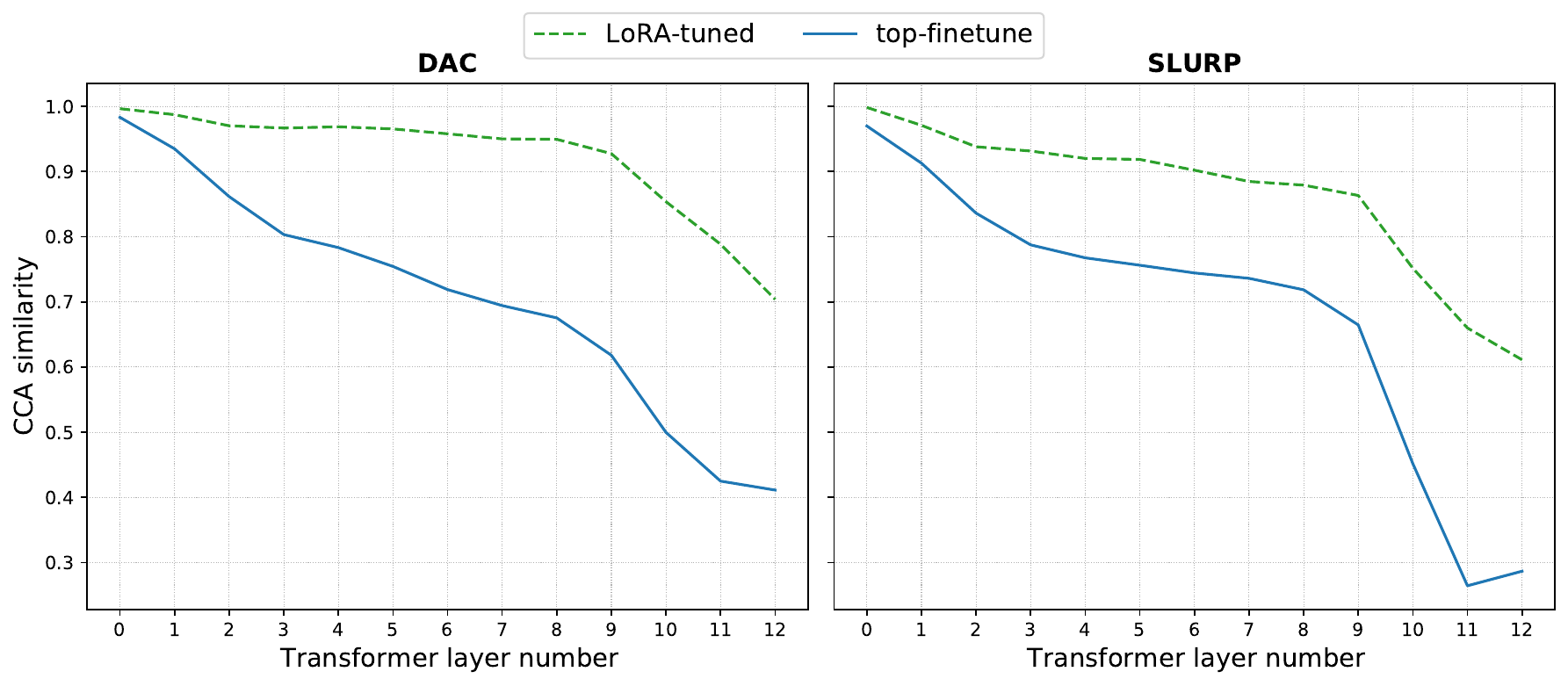}}
\caption{CCA-inter comparing \hubert-\baseM \ with its LoRA-tuned and fully fine-tuned counterparts. For the LoRA experiment, LoRA modules are placed on all layers.}
\label{fig:res-cca-inter-lora}
\end{figure}
We see that specific placements perform better than other layer combinations of LoRA modules. Next, we investigate whether our CCA-inter analysis is informative of the placement, as it was for \tff \ in \sect~\ref{sec:adapt-tff}.

\fig~\ref{fig:res-cca-inter-lora} shows the CCA similarity between \hubert-\baseM \ representations before and after LoRA-tuning, when LoRA modules are placed at all layers. We notice that CCA similarity is pretty high ($>0.9$) for layers $0-9$, and then it drastically reduces for the final three layers. This suggests that LoRA modules placed on the final few layers bring about the most modification to the representations of the \hubert-\baseM \ model. Contrary to our observations, the CCA-inter plot does not suggest a larger impact of LoRA modules for the bottom-most layers. Our empirical experiments suggest that optimal placements include a combination of bottommost and topmost layers.

In the same \fig~\ref{fig:res-cca-inter-lora}, we also plot CCA similarity between \hubert-\baseM \ and a \tff \ model for the respective tasks. The plots have a larger \% drop (slope) in similarity for layers $0-2$ and $9-12$ than for other intermediate layers. This possibly indicates a relatively larger modification of representations at those layers for the downstream task adaptation, and these layers also match the optimal LoRA placement layers. However, we lack an explanation for why we don't see a similar pattern for the LoRA-tuned model. 

In conclusion, this line of work remains open-ended. Prior work has also noticed that fewer LoRA modules can match the performance of placing LoRA at all layers, and at times also match the performance of \tff~\cite{thomas2022efficient, huang2023findadaptnet}, but we still lack a lightweight principled approach to choosing those layers. While we observe a trend that a combination of the shallowest and deepest layers works the best, future work needs to evaluate more \sfms \ and more tasks varying in task definition and data domains.


\section{Summary}
In this chapter, we draw on the analytical insights from our task-agnostic CCA-based study (\sect~\ref{ch:analysis}) to derive implications for task-specific adaptation of \sfms. The \afl \ approach for adapting \sfms \ has been widely adopted since its introduction for the SUPERB benchmark~\cite{tsai2022superb}. In our study, we noticed that a single layer (\sfl) can match the performance of \afl, and our CCA-based analysis can help effectively narrow down the search space.

Based on our layer-wise analysis of \wavtovec, we propose a re-initialization strategy for full-finetuning (\tff) of \wavtovec \ for speech recognition. We see improved performance in limited labeled data settings. We also employ CCA-based analysis to draw insights into the placement of LoRA modules for parameter-efficient fine-tuning (PEFT). While we don't find a direct correlation with our task-agnostic insights, our experiments suggest that an optimal placement of LoRA modules does seem to follow a pattern that combines the shallowest and deepest layers.  

Researchers have explored other lightweight approaches for model transferability based on latent space measurement and likelihood estimation, but unlike our motivation~\cite{chen2023estimate}, these approaches use task-specific labeled data to derive conclusions. In general, the choice of the best layer based on our analysis framework is more scalable than popular task-specific probing approaches for layer-wise analysis~\cite{shah2021all, feng2022silence, ji2022predicting}, and our analysis techniques are easily extensible to additional properties besides the ones we have studied here (e.g., speaker, prosody, syntax). 


\newpage
\chapter{Spoken Language Understanding Benchmark: Named Entity Recognition and Localization}
\label{ch:slue}
The effectiveness of a speech foundation model (SFM) is ultimately determined by its ability to address speech applications. Most commonly, the assessment of new \sfms \ is restricted to speech recognition on LibriSpeech~\cite{panayotov2015librispeech}. But this provides a very narrow view of the utility of learned features. This constraint is partly due to the absence of a comprehensive benchmark encompassing various tasks. Although recent efforts, such as SUPERB and its successors~\cite{yang2021superb, tsai2022superb}, have made the much-needed contribution of larger-scale benchmarks and tasks, there remains a shortage of datasets that can effectively quantify the linguistic understanding (SLU) capabilities of \sfms.




In order to fill this gap, we introduced the Spoken Language Understanding Evaluation (SLUE) benchmark in collaboration with multiple research groups. SLUE currently comprises six English SLU tasks~\cite{shon2022slue, shon2023slue}---sentiment analysis, named entity recognition, named entity localization, dialog act classification, summarization, and spoken question answering. \tab~\ref{tab:slue-datasets} provides an overview of the task definitions and the corresponding datasets. Several expert transcribers and annotators were hired from a third-party vendor and an in-house annotation team from ASAPP. All annotated datasets are freely accessible on HuggingFace,\footnote{\href{https://huggingface.co/datasets/asapp/slue}{https://huggingface.co/datasets/asapp/slue}, \href{https://huggingface.co/datasets/asapp/slue-phase-2}{https://huggingface.co/datasets/asapp/slue-phase-2}.}
accompanied by data processing and training guidelines for the published baseline models.\footnote{\href{https://github.com/asappresearch/slue-toolkit/}{https://github.com/asappresearch/slue-toolkit/}}

The remainder of this chapter will focus on the two tasks contributed by our group: named entity recognition and localization.\footnote{The contents of this chapter are from our prior published papers~\cite{shon2022slue, shon2023slue}.}

\begin{table*}[ht]
  \centering
  \caption{Overview of the datasets and tasks in the SLUE benchmark~\cite{shon2022slue,shon2023slue}.}
  \resizebox{\linewidth}{!}{
  \begin{tabular}{llccccc}
    \toprule
    \multirow{2}{*}{Dataset} & \multirow{2}{*}{Tasks} & \multirow{2}{*}{Speaking Style} & \multirow{2}{*}{Output} &\multicolumn{3}{c}{Size (utterances / hours)}\\\cmidrule(lr){5-7}
    & & & & Train & Dev & Test\\
    \midrule
    SLUE-VoxCeleb & SA & Conversational & sentiment class & \hphantom{0}5,777 / \hphantom{00}12.8 & \hphantom{0,}955 / \hphantom{0}2.1 & 4,052 / \hphantom{0}9.0\\
    \multirow{2}{*}{SLUE-VoxPopuli} & NER & Orated Speech & (entity phrase, entity tag) pairs & \hphantom{0}5,000 / \hphantom{00}14.5 & 1,753 / \hphantom{0}5.0 & 1,842 / \hphantom{0}4.9\\
    & NEL & Orated Speech & named entity time-stamps & N/A & 1,750 / \hphantom{0}5.0 & 1,838 / \hphantom{0}4.9\\
    SLUE-HVB & DAC & Scripted conversations & dialogue act classes & 11,344 / \hphantom{00}6.8 & 1,690 / \hphantom{0}1.0 & 6,121 / \hphantom{0}3.6\\
    SLUE-TED & SUMM & Orated Speech & text summary & \hphantom{0}3,384 / 664.0 & \hphantom{0,}425 / 81.0 & \hphantom{0,}424 / 84.0\\
    SLUE-SQA-5 & QA & Read Speech & answer time-stamps & 46,186 / 244.0 & 1,939 / 21.2 & 2,382 / 25.8\\
    \bottomrule
  \end{tabular}
  }
  \label{tab:slue-datasets}
\end{table*}

\section{Spoken named entity recognition and localization}
\begin{figure}[htbp]
    \centering
    \includegraphics[width=0.55\textwidth]{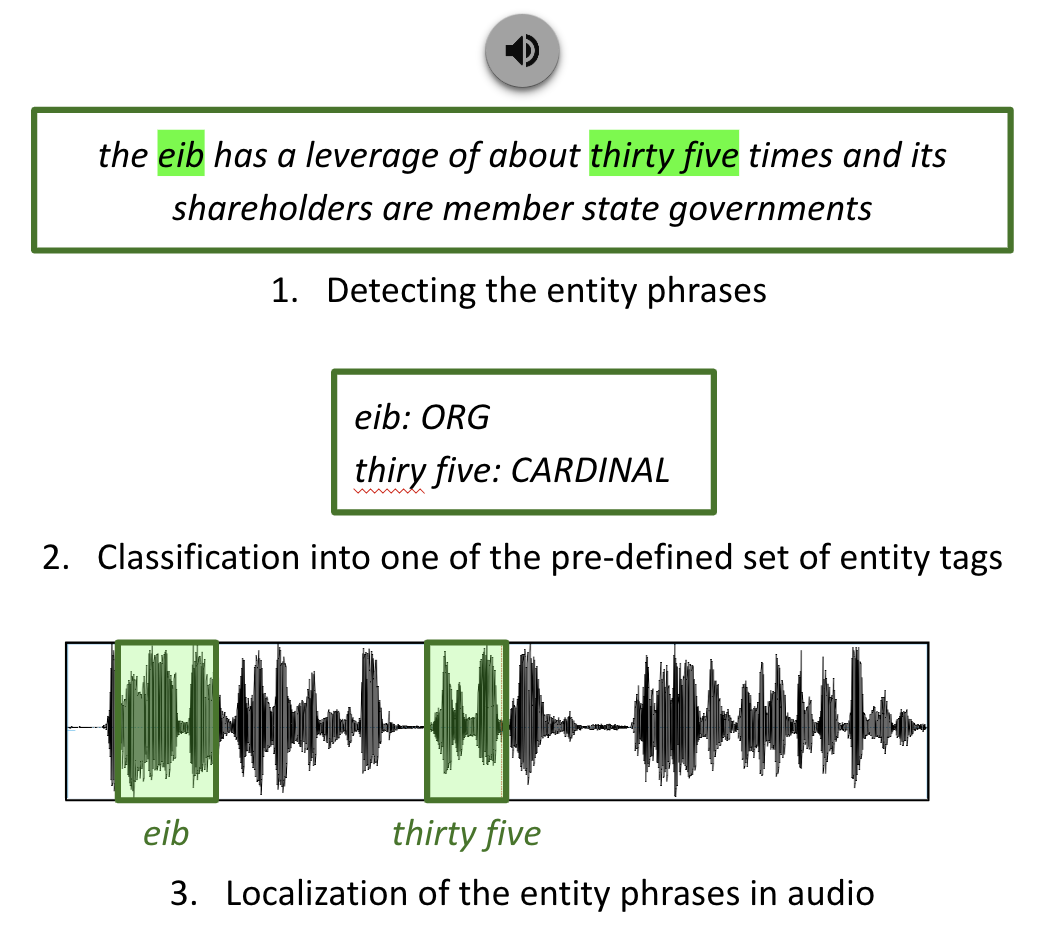}
    \caption{Named entity recognition (steps 1 and 2) and named entity localization (step 3).
}
  \label{fig:ner-nel}
\end{figure}
{\it Named entities} are specific words or phrases that represent objects, people, locations, dates, organizations, or other structured information. Named entity recognition (NER) is the task of identifying and categorizing these named entities into predefined classes, such as ``Person", ``Organization", ``Location", etc., to extract and structure information from spoken or written language (see \fig~\ref{fig:ner-nel} {\it steps 1 and 2}). In the context of spoken language, it may be relevant to query the start and end time stamps of these named entity phrases, and this constitutes the task of named entity localization (NEL) (see \fig~\ref{fig:ner-nel} {\it step 3}). NER and NEL are vital components of information extraction systems and play a crucial role in downstream tasks, such as coreference resolution~\cite{durrett2014joint} and de-identification~\cite{cohn2019audio}. 

We use F1 scores to evaluate the performance of our NER and NEL systems. NEL performance can be assessed either at the frame-level ({\it frame-F1}) or word-level ({\it word-F1}) by measuring the overlap between predicted and the ground-truth entity spans~\cite{cohn2019audio}. The {\it word-F1} score incorporates a hyperparameter, $\rho \in [0,1]$, which represents the required fraction of overlap between a ground-truth word segment and the predicted region for the ground-truth word to count as detected; $\rho=1$ necessitates a perfect match to be considered as a true positive. NEL evaluation focuses solely on the time stamps and does not consider the entity tags or the entity phrases.

In contrast, NER performance is evaluated on the predicted named entity phrase and their corresponding tags using a micro-averaged F1 score~\cite{ghannay2018end, shon2022slue}. Additionally, we report the {\it label-F1} metric, which only considers the tag predictions and allows for some tolerance towards misspellings or segmentation errors in the decoded text.

\section{Dataset}
We use audio recordings from VoxPopuli~\cite{wang2021voxpopuli}  and annotate them for named entities. VoxPopuli is a large multilingual speech corpus consisting of European Parliament event recordings with audio, transcripts, and utterance timestamps from the official Parliament website. By the nature of the source, the spoken data includes abundant named entities, making it an ideal choice for our use case. We use the English subset of the data and retain the canonical splits provided in the official repository for the {\it dev} and the {\it test} sets.\footnote{\href{https://github.com/facebookresearch/voxpopuli}{https://github.com/facebookresearch/voxpopuli}} For the {\it fine-tune} set, we sample about 15 hours of data from the official train set. More details about the dataset split can be found in \tab~\ref{tab:slue-datasets}. 

\begin{table}[htb]
\centering
\small
\caption{SLUE-VoxPopuli NER label statistics.}
\begin{tabular}{c|l|r|r|r}
\hlineB{2}
\multicolumn{1}{c|}{\multirow{2}{*}{\begin{tabular}[c]{@{}c@{}}Combined\\ label\end{tabular}}} & 
\multicolumn{1}{c|}{\multirow{2}{*}{\begin{tabular}[c]{@{}c@{}}Raw label\\ (ontonotes5)\end{tabular}}} & 
\multicolumn{3}{c}{\# of NER phrases} \\
& & Fine-tune & Dev & Test \\\hline
PLACE & GPE, LOC & 2012 & 642 & 731 \\[3pt]
QUANT & \begin{tabular}[l]{@{}l@{}}CARDINAL, MONEY,\\ ORDINAL, PERCENT,\\ QUANTITY\end{tabular} & 923 & 327 & 246 \\[0.5cm]
ORG & ORG & 864 & 259 & 273 \\[3pt]
WHEN & DATE, TIME & 762 & 260 & 186 \\[3pt]
NORP & NORP & 647 & 220 & 348 \\[3pt]
PERSON & PERSON & 272 & 51 & 81 \\[3pt]
LAW & LAW & 250 & 60 & 96 \\[3pt]
\hline
\hlineB{2}
\end{tabular}%
\label{tab:ner-labels}
\end{table}

\subsection{Named entity recognition}
\label{sec:slue-ner-data}
For annotation of named entities, we follow the OntoNotes Release 5.0~\cite{hovy2006ontonotes} guidelines and entity labels. The label-wise counts in the annotated data are reported in \tab~\ref{tab:ner-labels}. As the domain of OntoNotes 5 is slightly different from VoxPopuli, for evaluation, we combine similar categories and discard the rare ones, resulting in 7 categories. Both combined and raw labels are included in the dataset. See Appendix \tab~\ref{tab:voxpopuli-stats-detail} for the distribution statistics of raw labels.

We hired four annotators, and all annotation was done on text transcripts. We obtain a second pass of annotations for the test set to estimate human performance. The second pass achieved a micro-averaged F1 score of 0.79 when evaluated against the first pass. The disagreement between the two passes can be classified as a mismatch in detecting the entity phrase (missed/over/partial detection) or a mismatch in the label when they agree on the entity phrase (mislabel). We see that 88\% of these disagreements were detection errors. On a closer look at the data, we notice specific recurring systematic differences in the two passes, leading to most of these errors. We decided to use the original annotations for all further evaluation. See Appendix \fig~\ref{fig:ner-annotation-pie} for a fine-grained comparison between two annotation passes.

\subsection{Named entity localization}
\label{sec:slue-nel-data}
We extend SLUE-VoxPopuli to NEL by adding word-level time stamps in the dev and test sets. We use the Montreal Forced Aligner (MFA)~\cite{mcauliffe2017montreal} to obtain word-level time stamps, using MFA's public English acoustic model~\cite{mfa-acoustic-model}.
We manually verify the MFA-produced entity alignments for 188 utterances (20\% of the utterances with entity tags) in the dev set and conclude that the MFA output provides a reliable ground truth. Refer to Appendix \sect~\ref{sec:slue-appendix-annotation-nel} for discussion of our findings from manual verification or MFA alignments. 

We do not publish NEL annotations for the {\it fine-tune} set as we focus on re-purposing NER models for NEL, which we believe is a more realistic use-case, as is also common for the speech-to-text forced alignment models, such as MFA, to be trained without ground-truth alignments.

\section{Baselines}
\label{sec:slue-baselines}
\begin{figure}[ht]
  \centering
  \includegraphics[width=\linewidth]{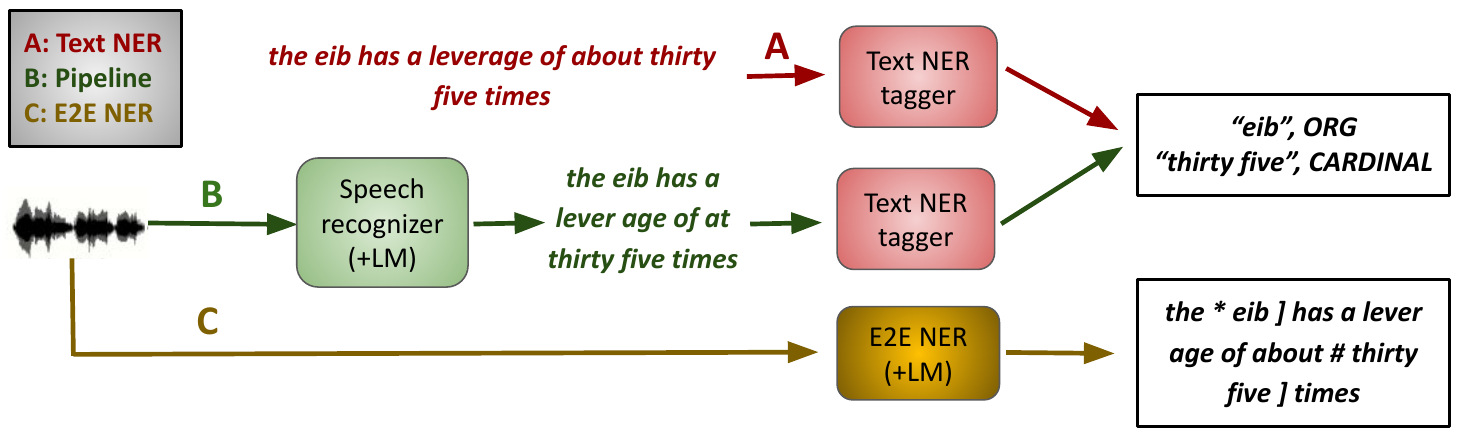}
  \caption{High-level summary of approaches typically used to solve spoken and textual NER tasks. Optional LM decoding is applied in ASR and E2E-NER models.} 
  \label{fig:baseline-systems}
\end{figure}

Most SLU tasks are typically tackled using one of the two types of approaches: (i) Pipeline or cascaded, and (ii) End-to-end (E2E). As shown in Fig.~\ref{fig:baseline-systems}, a pipeline approach decodes speech to text using a speech recognizer and then passes the decoded text through a text understanding module, text NER in this case. An E2E system directly maps the input speech to the output task labels. Each approach has its own set of advantages and shortcomings. Pipeline systems can enjoy the individual advances from both the speech and the text models, whereas combining two modules increases inference time, and propagation of ASR errors can have unexpected detrimental effects on the text NER module performance. On the other hand, E2E models directly optimize a task-specific objective and tend to have faster inference. However, such models typically require a large amount of task-specific labeled data to perform well. This can be seen from previous papers on E2E NER~\citep{yadav2020end, ghannay2018end}, where using at least 100 hours of labeled data is typical.

We present both pipeline and E2E baselines for NER and NEL. We also include a {\it pipeline-oracle}, i.e., a model that uses human transcription as input. It serves as a top line for the {\it pipeline} approach in the absence of recognition errors.

\begin{table}[hp!]\centering
\small
\caption{Modes used in our experiments.}
\label{tab:slue_baseline_models_sizes}
\begin{tabular}{llr}\toprule
Type &model name & parameter size \\\midrule
\multirow{3}{*}{Speech model} & \wavtovec-\baseM & 95M \\
& \hubert-\baseM & 95M \\
& \wavtovec-\largeM & 317M\\
\midrule
Text model &DeBERTa &139M \\ \midrule
\multirow{35}{*}{\shortstack{off-the-shelf\\[4pt]ASR model}} &Whisper-en &71M \\
&QuartzNet15x5Base-En &18M \\
&stt\_en\_citrinet\_1024 &143M \\
&stt\_en\_citrinet\_1024\_gamma\_0\_25 &141M \\
&stt\_en\_citrinet\_256 &10M \\
&stt\_en\_citrinet\_256\_gamma\_0\_25 &9M \\
&stt\_en\_citrinet\_512 &36M \\
&stt\_en\_citrinet\_512\_gamma\_0\_25 &36M \\
&stt\_en\_conformer\_ctc\_large &121M \\
&stt\_en\_conformer\_ctc\_large\_ls &121M \\
&stt\_en\_conformer\_ctc\_medium &30M \\
&stt\_en\_conformer\_ctc\_medium\_ls &30M \\
&stt\_en\_conformer\_ctc\_small &13M \\
&stt\_en\_conformer\_ctc\_small\_ls &12M \\
&stt\_en\_conformer\_ctc\_xlarge &635M \\
&stt\_en\_conformer\_transducer\_large &120M \\
&stt\_en\_conformer\_transducer\_large\_ls &120M \\
&stt\_en\_conformer\_transducer\_medium &32M \\
&stt\_en\_conformer\_transducer\_small &14M \\
&stt\_en\_conformer\_transducer\_xlarge &644M \\
&stt\_en\_conformer\_transducer\_xxlarge &998M \\
&stt\_en\_contextnet\_1024 &144M \\
&stt\_en\_contextnet\_1024\_mls &144M \\
&stt\_en\_contextnet\_256 &14M \\
&stt\_en\_contextnet\_256\_mls &14M \\
&stt\_en\_contextnet\_512 &40M \\
&stt\_en\_contextnet\_512\_mls &40M \\
&stt\_en\_jasper10x5dr &332M \\
&stt\_en\_quartznet15x5 &18M \\
&stt\_en\_squeezeformer\_ctc\_large\_ls &236M \\
&stt\_en\_squeezeformer\_ctc\_medium\_large\_ls &125M \\
&stt\_en\_squeezeformer\_ctc\_medium\_ls &77M \\
&stt\_en\_squeezeformer\_ctc\_small\_ls &18M \\
&stt\_en\_squeezeformer\_ctc\_small\_medium\_ls &28M \\
&stt\_en\_squeezeformer\_ctc\_xsmall\_ls &9M \\
\bottomrule
\end{tabular}
\end{table}

Our E2E and pipeline baselines use pre-trained \sfms \ from the official Fairseq repository.\footnote{\href{https://github.com/pytorch/fairseq}{https://github.com/pytorch/fairseq}}
In addition to using \sfm-based ASR models, we evaluate pipeline baselines with off-the-shelf ASR models from  Whisper\footnote{\href{https://github.com/openai/whisper}{https://github.com/openai/whisper}}~\cite{radford2023robust} and NeMo\footnote{\href{https://docs.nvidia.com/nemo-framework/user-guide/latest/nemotoolkit/asr/models.html}{https://docs.nvidia.com/nemo-framework/user-guide/latest/nemotoolkit/asr/models.html}}~\cite{kuchaiev2019nemo} collections.
\tab~\ref{tab:slue_baseline_models_sizes} lists all the backbone models used as part of the NER and NEL baselines in this chapter.

\subsection{Named entity recognition}
\label{sec:slue-baselines-ner}
For the {\it pipeline-oracle}, we fine-tune a pre-trained DeBERTa-L~\cite{he2020deberta} after adding a linear layer on top of the final hidden-state output. We use the HuggingFace's transformers toolkit~\cite{wolf2020transformers} to fine-tune it for NER with a token-level classification loss. 

For the {\it pipeline} experiments, we first train an ASR model. We add a linear layer on top of a pre-trained \sfm \ (\wavtovec~\cite{baevski2020wav2vec} or \hubert~\cite{hsu2021hubert}) and fine-tune all the parameters, except the CNN module, with a character-level CTC objective~\cite{graves2006connectionist}. Optionally, we decode with a trigram language model (LM) trained on the TED-LIUM 3 LM corpus~\cite{hernandez2018ted} and the decoding parameters optimized on dev set performance (beam size 500, LM weight 2, and word insertion penalty -1).\footnote{We found this LM to (slightly) outperform bigram and 4-gram models, as well as LMs trained on LibriSpeech.} We use the DeBERTa-based {\it pipeline-oracle} as the text understanding module to perform NER inference on the decoded text.  

The {\it E2E NER} baseline models are trained similarly to the ASR models with the only difference of the target character sequence, which is delimited by special tokens corresponding to entity labels~\cite{ghannay2018end, yadav2020end}. For example, the phrases ``irish" and ``eu" are tagged as NORP (\textcolor{blue}{\$}) and GPE (\textcolor{blue}{\%}) respectively in `\textit{`the \textcolor{blue}{\$ irish ]} system works within a legal and regulatory policy directive framework dictated by the \textcolor{blue}{\% eu ]}}". The vocabulary includes 19 special characters, 18 for each entity tag (\tab~\ref{tab:ner-labels}), inserted at the beginning of an entity phrase, and one to denote the end of the entity phrase. Optionally, we decode with a 4-gram language model trained on the SLUE-VoxPopuli fine-tune set and the decoding parameters optimized on dev set performance (beam size 500, LM weight 2, and word insertion penalty 1).

\subsection{Named entity localization}
\label{sec:slue-baselines-nel}
Since NEL does not have a dedicated {\it fine-tune} set, no dedicated model is trained for NEL. This is intentional since NER and NEL are related tasks, and a realistic use case would require a single model that performs both tasks. For NEL inference, we use the baseline NER models described above. 
Since the {\it pipeline-oracle} assumes access to the ground-truth transcripts, the force-aligned timestamps (the same ones that we use as ground-truth) are used with the output of the {\it pipeline-oracle} NER model.


\begin{figure}[ht]
  \centering
  \includegraphics[width=0.75\linewidth]{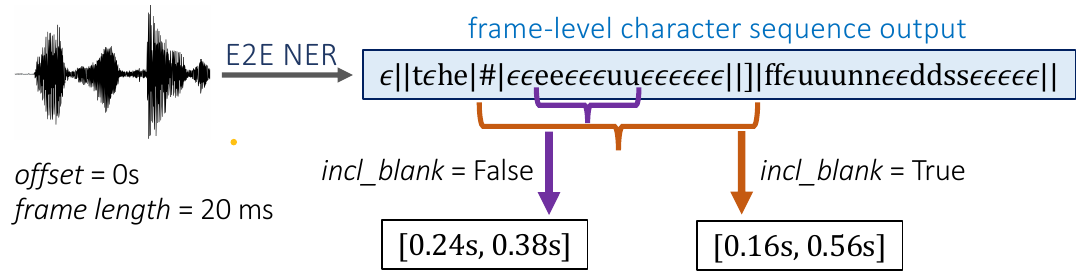}
  \caption{Example inference for an E2E NEL model using a CTC recognizer. The transcript is ``the eu funds''. `\#' and `]' are the start and end labels of an ORG entity.} 
  \label{fig:method-nel}
\end{figure}

The baseline CTC-based {\it E2E} and {\it pipeline NER} models produce a posterior probability matrix, $\mathcal{E} \in \mathbb{R}^{T\times V}$, consisting of the posterior of each character in the vocabulary of size $V$ for each of the $T$ frames in the input audio. For ASR (within {\it pipeline}), the character vocabulary consists of the English alphabet, a word separator token ``$\vert$'', and a blank token ``$\epsilon$''. For the {\it E2E} model, the vocabulary also includes special characters for the start and end of an entity phrase. We obtain a frame-level character sequence output via greedy decoding on $\mathcal{E}$. The time stamps corresponding to ``$\vert$'' tokens in the output character sequence provide word-level start and end boundaries. The NeMo-ASR models are either trained at a character or subword sequences, and the time stamps are obtained from the posterior probability vector of each token, similar to SFM-based CTC ASR models.

As the ASR and NER models are not trained with an explicit alignment signal, the word boundary tokens may not be a reliable indicator of the actual time stamps, and we introduce two hyperparameters as a heuristic fix for possible misalignments: {\it offset} is a fixed duration by which we shift the time stamp predictions, and {\it incl\_blank} $\in \{0,1\}$ denotes whether any trailing $\epsilon$ tokens are considered a part of the predicted entity segment. These parameters are tuned on the dev set, see Appendix \tab~\ref{tab:nel_bestparams} for the optimal values of {\it offset} and {\it incl\_blank} we found for each model.

For {\it pipeline NEL}, the predicted text from ASR is passed to a {\it oracle-pipeline} NER model, and the time stamps for detected entities are extracted from the ASR's $\mathcal{E}$. For the {\it E2E NEL}, the time stamps corresponding to the entity start and end special characters are extracted directly from its $\mathcal{E}$. An example is presented in \fig~\ref{fig:method-nel}. 

\section{Results}
\label{sec:slue-results}
Next, we report baseline results for NER and NEL tasks. As previously stated, we use the same NER baseline models for NEL inference. All three results reported here are on SLUE-VoxPopuli test set; corresponding dev set results are reported in Appendix \sect~\ref{sec:slue-appendix-results}.

\subsection{Named entity recognition}
\begin{table}[htbp]
\centering
\small
\caption{Named entity recognition performance on test set.}
\begin{tabular}{lcc|cc}
\hlineB{2}
\multicolumn{1}{c}{Speech model} & LM & Text model & F1 (\%) & label-F1 (\%)  \\
\hlineB{2} 
\textbf{Pipeline-oracle:}  & & & &  \\
\hspace{3mm} \multirow{1}{*}{ N/A (GT Text)} & \multirow{1}{*}{ N/A} & DeBERTa-L  & 81.4 & 85.7 \\
\hlineB{2}
\textbf{Pipeline approaches:}  & & & &  \\
\hspace{2mm} \wavtovec-\baseM & - & DeBERTa-L & 49.5 & 74.2 \\
\hspace{2mm} \wavtovec-\largeM & - & DeBERTa-L & 57.8 & 78.8 \\
\hspace{2mm} \wavtovec-\baseM & \cmark & DeBERTa-L & 68.0 & 79.8 \\
\hspace{2mm} \wavtovec-\largeM & \cmark & DeBERTa-L & 69.6 & 82.2 \\
\hlineB{2}
\textbf{E2E approaches:}  & & & &  \\
\hspace{2mm} \wavtovec-\baseM & - & \multirow{6}{*}{\normalsize N/A} & 50.2 & 64.0 \\
\hspace{2mm} \hubert-\baseM & - &  & 49.8 & 62.9 \\
\hspace{2mm} \wavtovec-\largeM & - &  & 50.9 & 64.7 \\
\hspace{2mm} \wavtovec-\baseM & \cmark &  & 63.4 & 71.7 \\
\hspace{2mm} \hubert-\baseM & \cmark &  & 61.9 & 70.3 \\
\hspace{2mm} \wavtovec-\largeM & \cmark &  & 64.8 & 73.3 \\
\hlineB{2}
\end{tabular}
\label{tab:ner-baseline}
\end{table}

Baseline results for NER are reported in \tab~\ref{tab:ner-baseline}. 
The best checkpoints are chosen based on the word error rate of NER-annotated sentences in the dev set. 
We make the following observations:
\begin{itemize}[leftmargin=*,noitemsep,nolistsep]
    \item There is significant room for improvement for both Pipeline and E2E models, even while leveraging state-of-the-art pre-trained models.
    \item Decoding with an n-gram language model provides consistent and significant improvements. 
    \item Improvements from larger speech models are less evident when using LM decoding; that is, using a small amount (5k utterances) of unlabeled text is as beneficial as leveraging 60 times more unlabeled audio data with the current methods.
\end{itemize}
The last point may suggest that the pre-trained speech models do not learn significant semantic information, so even a small amount of additional semantic knowledge (in the form of language models here) should help immensely. 

\begin{figure*}[htb]\centering
    \begin{subfigure}[b]{0.45\textwidth}
         \includegraphics[width=\textwidth]{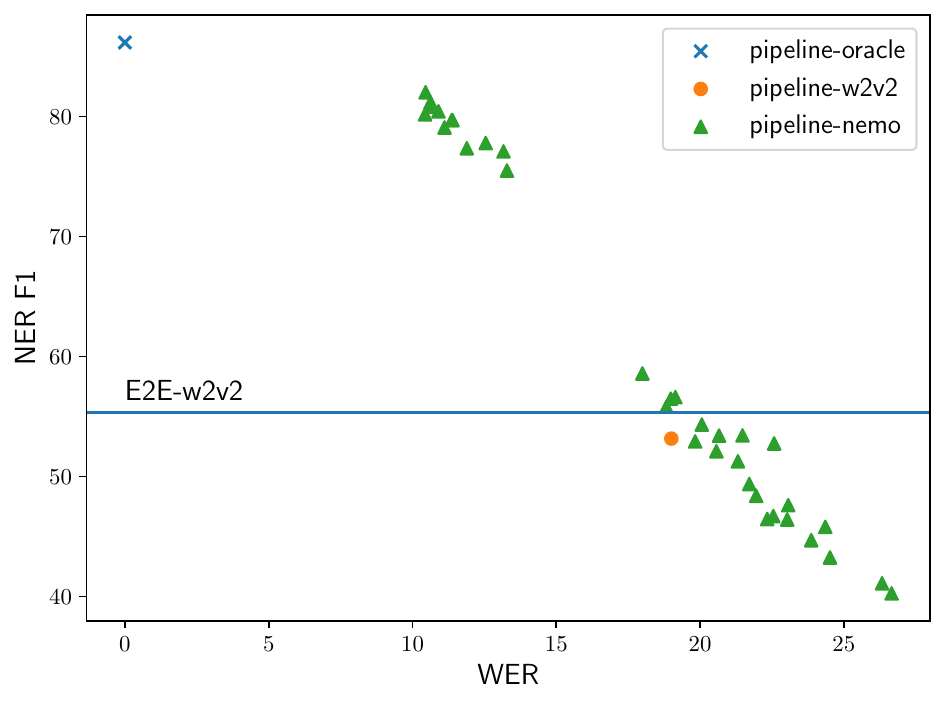}
         \caption{WER and F1 score on NER dev set}
         \label{fig:f1-wer-scatter-ner}
     \end{subfigure}
    \begin{subfigure}[b]{0.45\textwidth}
         \includegraphics[width=\textwidth]{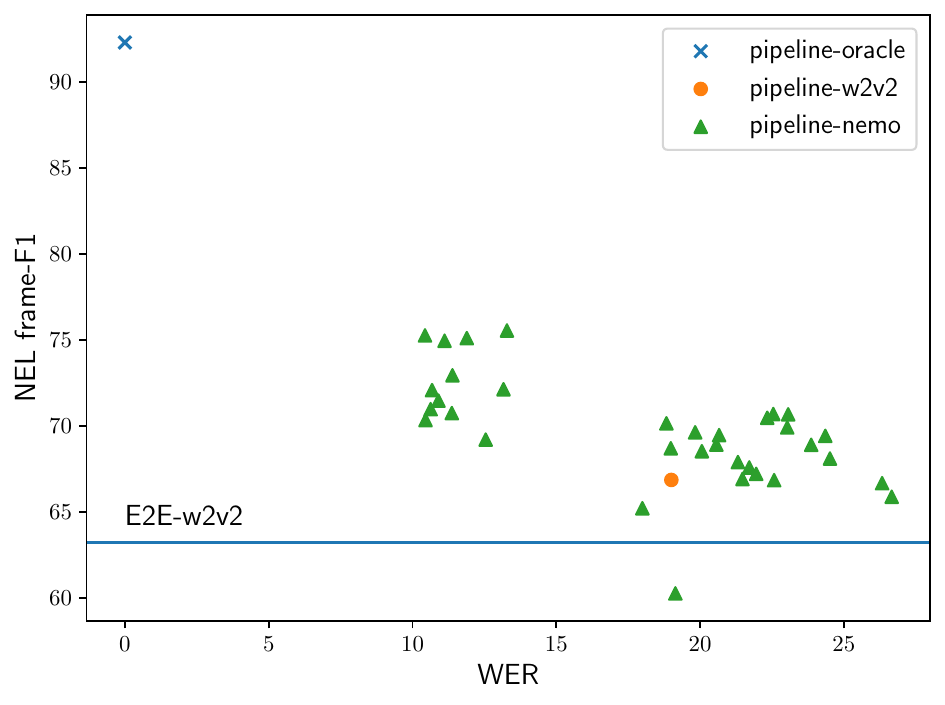}
         \caption{WER and F1 score on NEL dev set}
         \label{fig:f1-wer-scatter-nel}
     \end{subfigure}
    \caption{Impact of WER on text understanding performance.}
    \label{fig:f1-wer-scatter}
\end{figure*}

Figure~\ref{fig:f1-wer-scatter-ner} shows a scatter plot of NER and WER scores for various NeMo-based pipeline models. We see a consistent improvement in the NER performance with increasing ASR quality. Interestingly, {\it E2E} outperforms {\it pipeline} approach when the word error rate degrades above 17\%. 

\subsection{Named entity localization}

\begin{table}[htb]\centering
\small
\caption{NEL task baseline performance on test set. The \wavtovec-\baseM \ models are fine-tuned on slue-voxpopuli data.*the best NeMo model based on NEL frame-f1 score on dev is ``stt\_en\_conformer\_ctc\_small"}
\begin{tabular}{lcc|c|c}\toprule
\multirow{2}{*}{System} &\multirow{2}{*}{\makecell{Speech\\model}} & \multirow{2}{*}{\makecell{Text\\model}} & \multirow{2}{*}{frame-F1} & \multicolumn{1}{c}{word-F1} \\
& & &  & ($\rho$=0.8)  \\\midrule
pipeline-oracle & x & DeBERTa & 89.0 & 90.0  \\
pipeline-w2v2 & \wavtovec-\baseM & DeBERTa & 65.2 & 72.0  \\
E2E-w2v2 & \wavtovec-\baseM & x & 56.3 & 59.6  \\
pipeline-nemo & best model* & DeBERTa & 74.1 & 81.4 \\
\bottomrule
\end{tabular}
\label{tab:nel-baseline}
\end{table}

Baseline results for NEL are reported in \tab~\ref{tab:nel-baseline}. We report word-F1 with $\rho=0.8$ here, see Appendix \tab~\ref{tab:nel_baseline_all} for results with different tolerance ($\rho$) values.

Although {\it pipeline-w2v2} and {\it E2E-w2v2} baselines have somewhat similar frame-F1, these approaches have complementary strengths (see Appendix \sect~\ref{sec:slue-appendix-error-analysis}). We also find that the off-the-shelf NeMo ASR model ({\it pipeline-nemo}) outperforms the dataset-specific ASR model ({\it pipeline-w2v2}).

Figure~\ref{fig:f1-wer-scatter-nel} shows a scatter plot of NEL and WER scores for various NeMo-based pipeline models. Although models with the lowest WER have the best frame-F1, the overall correlation is not high. The NeMo models have different training objectives and model architectures, and we note that within each model class, the ASR and NEL metrics are much better correlated (see Appendix \fig~\ref{fig:nel_nemo_corr}). This suggests that, unlike NER, where a better speech recognizer always boosts pipeline-NER performance, for NEL, model architecture and/or training objectives also play a significant role in alignment quality.

\section{Related work}
The community needs stable benchmarks to perform standardized comparisons of novel architectures and design choices. SLUE was motivated by the lack of realistic and publicly available datasets for spoken language understanding tasks. 

\textbf{SUPERB}~\cite{yang2021superb} aggregates several existing speech tasks to evaluate frozen speech backbones and soon became a commonplace for comparing \sfms. It mainly focuses on low-level tasks based on acoustics, paralinguistics, and speaker information, but also contains two SLU tasks---intent classification (from Fluent Speech Commands~\cite{lugosch2019speech}) and slot filling (from SNIPS~\cite{coucke2018snips}). However, the former is an easy task where many models have close to 100\% accuracy, and the latter uses synthesized rather than natural speech. 
\textbf{SLURP}~\cite{bastianelli2020slurp} is a spoken version of written conversations between humans and personal robot assistants~\cite{liu2019benchmarking}.
It includes three SLU tasks---scenario prediction, action prediction, and entity prediction labeled for short speech commands.
\textbf{ASR-GLUE}~\cite{feng2021asrglue} and \textbf{SpeechGLUE}~\cite{ashihara2023speechglue} are spoken and synthesized versions of the well-known GLUE benchmark~\cite{wang2018glue} respectively. While ASR-GLUE has natural speech, it only provides a test set, so the researchers must rely on other datasets for training.
\textbf{Timers and Such}~\cite{Lugosch2021TimersAS} is a dataset of speech commands involving numbers designed for intent classification and slot filling, good for limited use cases.
\textbf{Spoken SQuAD}~\cite{lee2018spoken} and \textbf{Spoken CoQA}~\cite{you2022end} are synthesized speech versions of the text SQuAD~\cite{Rajpurkar2016SQuAD1Q} and CoQA~\cite{reddy2019coqa} datasets.
\textbf{NMSQA}~\cite{lin2022dual} is a multi-speaker spoken QA dataset whose test set contains natural speech, but the train and validation sets are synthesized. 
Other well-known SLU datasets include \textbf{ATIS}~\cite{hemphill1990atis} and \textbf{Switchboard NXT}~\cite{calhoun2010nxt}, which have been used for tasks like intent and dialog act classification, but the data is available under license constraints. 
Wu et al.~\cite{wu2020harpervalleybank} published an open-sourced speech dataset; however, the dialog act labels are not manually annotated but predicted using a commercial API. While NEL is a reasonably new task, a similar task, audio de-identification (audio de-ID), has been introduced with annotations for conversational data from {\bf Switchboard} and {\bf Fisher}~\cite{cohn2019audio, baril2022named},  but these datasets are not free. Audio de-ID is a special case of NEL where the entities of interest are related to personal identifiers. 

Unlike the predecessors that are either limited in public availability~\cite{hemphill1990atis, calhoun2010nxt, sanabria2018how2, attention-fusion} or lack task complexity~\cite{yang2021superb, feng2021asrglue} or have synthetic datasets~\cite{lee2018spoken, lin2022dual, you2022end}, the datasets in the SLUE benchmark constitute free-to-use public datasets with either orated, conversational, or read speech. While our work focuses on English speech, the community has contributed language understanding datasets for other languages~\cite{tomashenko2019recent,evain2021lebenchmark}.

Since the recent introduction of the SUPERB benchmark~\cite{yang2021superb}, researchers have started contributing their pre-trained models to the SUPERB leaderboard,\footnote{\href{https://superbbenchmark.org/leaderboard}{https://superbbenchmark.org/leaderboard}} thus providing a broader view that could help us understand the relative metrics/demerits of various \sfms. We hope to achieve the same with SLUE~\cite{shon2022slue, shon2023slue} for complex language understanding tasks. Towards that goal, SLUE datasets have not only been employed by the community to test new \sfms~\cite{peng2023structured, arora2024evaluation, chou2023toward, wu2023wav2seq, peng2023comparative}, but have also been incorporated into broader benchmarks such as dynamic SUPERB~\cite{huang2024dynamic}.

\section{Summary}
This chapter describes a part of our open-source and publicly available spoken language understanding evaluation (SLUE) benchmark. We focus on the named entity recognition and localization tasks in limited labeled data settings and describe their dataset collection, annotation, and performance of \sfm-based baselines. We extensively study both {\it E2E} and {\it pipeline} approaches using the state-of-the-art speech and text foundation models as backbones. We observe that {\it pipeline} systems outperform {\it E2E} baselines overall, but their performance degrades significantly as the quality of speech-to-text transcription worsens.
In comparison, {\it oracle-pipeline} outperforms both {\it pipeline} and {\it E2E} baselines, highlighting the scope of improvement on these datasets.


\chapter{On the Use of External Data for Spoken Named Entity Recognition}
\label{ch:ner-ext}

In the previous chapter, we compared several approaches for named entity recognition (NER) (\tab~\ref{tab:ner-baseline}).
The end-to-end (E2E) model uses a strong speech foundation model (\sfm) as a backbone but exhibits the poorest performance. Leveraging text knowledge, either by decoding with an external language model or by cascading a pre-trained text foundation model during training (cascaded approach), improves the spoken NER performance over the E2E baseline. 
Finally, directly using ground-truth text tokens with a text foundation model outperforms both E2E and cascaded approaches, underscoring the value of rich linguistic information.

A pre-trained text foundation model offers linguistically meaningful representations~\cite{tenney2019bert} that are not as effectively learned by the \sfms~\cite{choi2024self}. This knowledge gap leads to a more pronounced performance gap in our low-labeled data setting, where the limited NER data may be insufficient for the models to learn the relevant linguistic knowledge from scratch.

\begin{figure}[t]
    \centering
    \hspace{-.15in}
    \includegraphics[width=0.65\linewidth, trim=0 65 0 80]{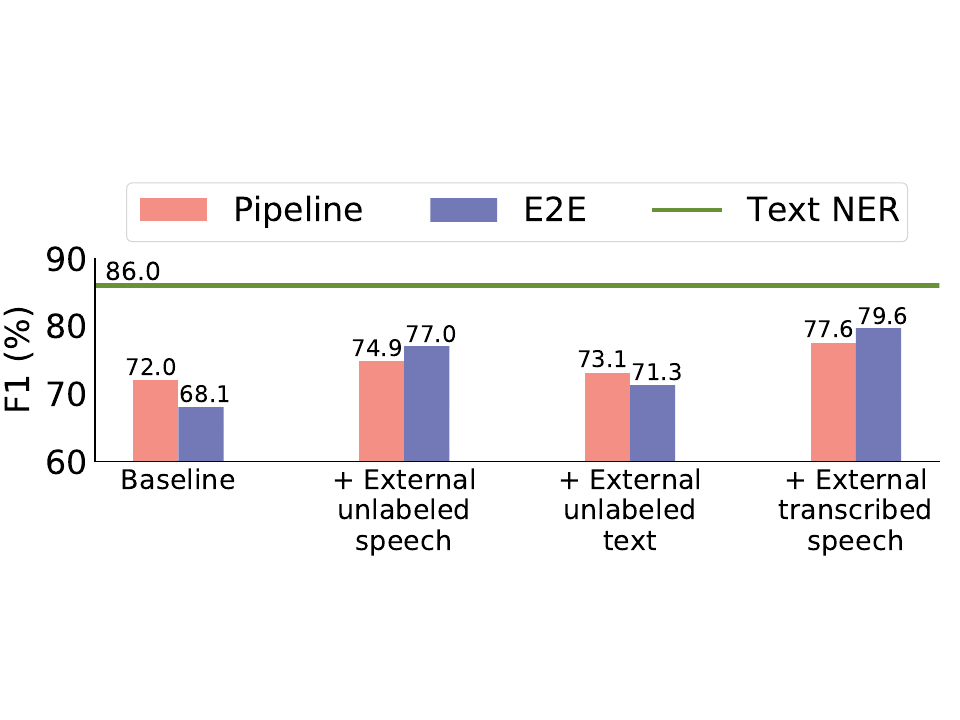}
    \caption{Improvements in spoken NER with 100 hours of external data of different types. ``Pipeline'' refers to approaches consisting of speech recognition followed by a text NER model; ``E2E'' refers to approaches that directly map from speech to NER-tagged text. The ``Baseline'' and ``Text NER'' numbers are from previously established baselines~\cite{shon2022slue}.}
    \label{fig:ner-ext-summary}
\end{figure}

In this chapter,\footnote{The contents of this chapter are from our prior published paper~\cite{pasad2021use}.} we aim to close the knowledge gap between different approaches by employing modeling tools such as knowledge distillation and self-training.
We continue to work with the low labeled data setting (15 hours of training data), but tackle a realistic scenario where additional unannotated data is available.
We show that narrowing the knowledge gap indeed results in a reduced performance gap between modeling paradigms (\fig~\ref{fig:ner-ext-summary}). Notably, our proposed end-to-end model outperforms the cascaded baseline, achieving state-of-the-art performance on SLUE-VoxPopuli.

\section{Methods}
\label{sec:ner-ext-approach}

Similarly to the previous chapter (\chap~\ref{ch:slue}), we work on improving and evaluating both cascaded and E2E approaches. The performance is evaluated using {\it micro-averaged F1} scores on an unordered list of tuples of named entity phrases and tag pairs predicted for each sentence. An entity prediction is considered correct if both the identified entity phrase and the detected entity tag are correct.

\subsection{Utilizing external data}
\label{sec:ext-data}
Next, we describe our approaches that use data external to the task-specific labeled data to improve both the pipeline and the E2E models for spoken NER. We consider four types of external data: (i) unlabeled speech ({\it Un-Sp}), (ii) unlabeled text ({\it Un-Txt}), (iii) transcribed speech ({\it Sp-Txt}), and (iv) text-based NER data. 

\begin{figure}[htp]
    \centering
    \hspace{-.15in}
    \includegraphics[width=0.8\linewidth]{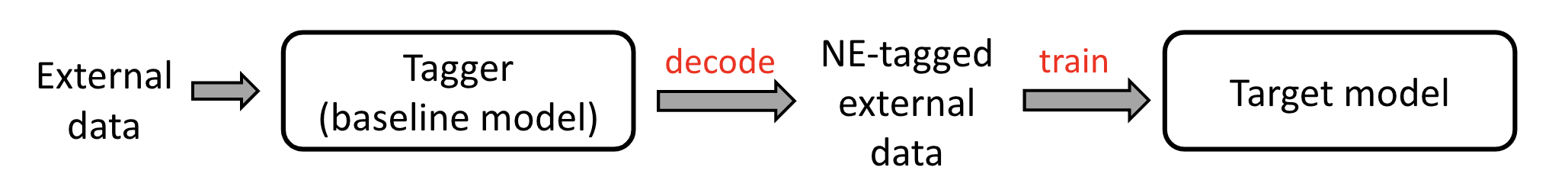}
    \caption{Illustration of how pseudo-labeled data is generated from external unannotated data.}
    \label{fig:ext-data-system}
\end{figure}

As shown in \fig~\ref{fig:ext-data-system}, a majority of techniques we consider involve labeling the external data with a {\it tagger model} (typically one of the baseline models) to produce {\it pseudo-labels}. The {\it target model} is then trained on these generated pseudo-labels along with the original labeled NER data. \tabs~\ref{tab:approach-ppl} and \ref{tab:approach-e2e} present a detailed list of all methods we consider for improving pipeline and E2E models respectively. The methods we include use the first three kinds of external data listed above.  The fourth kind, external text-based NER data, is used in experiments attempting to improve the text NER model; since it does not succeed (Sec.~\ref{sec:ext-text-ner}), this data source is not explored further for the pipeline and E2E models.

\begin{table}[ht]
\centering
\small
\caption{Methods for using external data for pipeline models.
The method for external transcribed data ({\it Sp-Txt}) is based on transfer learning and thus there is no {\it tagger model}. More details are provided in \sect~\ref{sec:ext-data}.}
\resizebox{\columnwidth}{!}{%
\begin{tabular}{lllll}
\toprule
{\bf External data type} & {\bf Method} & {\bf Tagger model} & {\bf Target model} & {\bf LM for decoding}\\
\midrule
Un-Sp              & SelfTrain-ASR    &  ASR & ASR  & tedlium 3-gram         \\ 
Un-Txt             & SelfTrain-txtNER & text NER &  text NER   & tedlium 3-gram                              \\ 
Sp-Txt             & Pre-ASR          & n/a & ASR    & tedlium 3-gram                                 \\ 
\bottomrule
\end{tabular}
}
\label{tab:approach-ppl}
\end{table}

\begin{table*}[ht]
\centering
\caption{Methods for using external data for E2E models. More details are provided in \sect~\ref{sec:ext-data}.}
\small
\resizebox{\columnwidth}{!}{%
\begin{tabular}{lllll}
\toprule
{\bf External data type}      & {\bf Method}           & {\bf Tagger model}                                                               & {\bf Target model} & {\bf LM for decoding} \\
\midrule
\multirow{2}{*}{Un-Sp}  & SelfTrain-E2E    & E2E-NER                                                                      & E2E-NER      & pLabel 4-gram   \\
                        & Distill-Pipeline & \begin{tabular}[l]{@{}l@{}}Pipeline-NER\\ (after SelfTrain-ASR)\end{tabular} & E2E-NER      & pLabel 4-gram   \\
\midrule
Un-Txt                  & Distill-txtNER-lm   & text NER                                                                     & n/a          & pLabel 4-gram   \\
\midrule
\multirow{2}{*}{Sp-Txt} & Distill-txtNER      & text NER                                                                     & E2E-NER      & pLabel 4-gram   \\
                        & Pre-ASR          & n/a                                                                          & n/a          & ftune 4-gram    \\ 
\bottomrule
\end{tabular}}
\label{tab:approach-e2e}
\end{table*}

When the tagger model is the same as the target model, this is a well-established process called self-training~\citep{scudder1965probability,yarowsky1995unsupervised,riloff1996automatically,xu2020iterative,xu2021self}.
When the tagger and target models are different, we refer to it as knowledge distillation~\citep{hinton2015distilling}, where the information is being distilled from the tagger model to the target model. This approach enables the target model to learn from the better-performing tagger model via pseudo-labels. Among the baseline models, the pipeline performs better than E2E approaches, presumably since the former uses a strong pre-trained text representations. So, for instance, distilling from the pipeline (tagger model) into the E2E model (target model) is expected to boost the performance of the E2E model.

The LMs used for decoding in different approaches are also mentioned in \tabs~\ref{tab:approach-ppl} and~\ref{tab:approach-e2e}. All the ASR experiments use language models trained on the TED-LIUM 3 LM corpus~\citep{hernandez2018ted} as in \chap~\ref{ch:slue}. The language model used in baseline E2E NER experiments is trained on the 15-hour fine-tune set ({\it ftune 4-gram}). All the E2E experiments leveraging external data are trained on external data pseudo-labeled with named entity tags. These pseudo-labels also provide additional annotated data for LM training. These are referred to as {\it plabel 4-gram} (for ``pseudo-label 4-gram") in \tab~\ref{tab:approach-e2e}. 

Next, we elaborate on \tabs~\ref{tab:approach-ppl} and \ref{tab:approach-e2e}, and describe modeling approaches for each type of external data.

\textbf{Unlabeled speech: } The unlabeled speech is used to improve the ASR module of the pipeline approach via self-training ({\it SelfTrain-ASR}).

For improving the E2E model, the improved pipeline can be used as the tagger model, followed by training the E2E model on the generated pseudo-labels ({\it Distill-Pipeline}). Alternatively, the unlabeled audio can be directly used to improve the E2E model via self-training ({\it SelfTrain-E2E}).

\textbf{Unlabeled text: } The text NER module in the pipeline approach is improved by self-training using the unlabeled text data ({\it SelfTrain-txtNER}). The E2E model uses the pseudo labels generated from the text NER baseline module on the unlabeled text to update the LM used for decoding ({\it Distill-txtNER-lm}).

\textbf{Transcribed speech: }The pipeline approach is improved by using the additional transcribed speech data to improve the ASR module ({\it Pre-ASR}). The E2E model uses this updated ASR as an initialization in a typical transfer learning setup.  Alternatively, for paired speech text data, the pseudo-labels generated from the text NER model can be paired with audio and used for training the E2E model, thus distilling information from a stronger text NER model into it ({\it Distill-txtNER}).

\textbf{Text NER data: } In addition to improving the pipeline and E2E models using the approaches mentioned above, we also look for any possible improvements in the text NER model by leveraging a larger external annotated text NER corpus. The DeBERTa base model is first fine-tuned on the larger external corpus, and then further fine-tuned on the in-domain labeled data. The first fine-tuning step is expected to help avoid shortcomings in performance due to the limited size of the in-domain labeled data. 

This approach is limited by the availability of external datasets with the same annotation scheme as the in-domain corpus. We use the OntoNotes5.0~\citep{pradhan2013towards} corpus, whose labeling scheme inspired that of VoxPopuli. See \tab~\ref{tab:datastats} for more information on OntoNotes5.0.

\section{Experimental setup}
\label{sec:ner-ext-exp}
\subsection{Dataset}
\begin{table}[ht]
\centering
\small
\caption{Data statistics. The ``ext-" prefix denotes external datasets. The external data doesn't have named entity annotations, except for OntoNotes 5.0.}
\begin{tabular}{l|rrr}
\toprule
\textbf{Data split}   & \textbf{\# utt} & \begin{tabular}[c]{@{}c@{}} \textbf{Duration}\\\textbf{(hours)} \end{tabular} & \begin{tabular}[c]{@{}c@{}} \textbf{\# entity}\\\textbf{phrases} \end{tabular} \\
\midrule
fine-tune            & 5k              & 15    & 5820                  \\
dev                   & 1.7k            & 5    & 1862                 \\
test                  & 1.8k            & 5    & 2006      \\
\midrule
ext-100h &  350k            & 101         & \multirow{2}{*}{N/A}            \\
ext-500h & 177k            & 508                     \\
\begin{tabular}[l]{@{}l@{}}{ext-NER}\\{(ontonotes-}\\{train)}\end{tabular} & 66.6k & N/A & 81.8k\\

\bottomrule
\end{tabular}
\label{tab:datastats}
\end{table}
For our experiments with external in-domain data, we use uniformly sampled 100-hour and 500-hour subsets of the remainder of the VoxPopuli train set. The statistics for these splits are reported in \tab~\ref{tab:datastats}.

Following the same format as the baseline models in \chap~\ref{ch:slue}, we train on the finer label set (18 entity tags) and evaluate on the combined version (7 entity tags).


\subsection{Models}
We use the fairseq library~\citep{ott2019fairseq} to fine-tune wav2vec 2.0 models for the E2E NER and ASR tasks. We fine-tune the model for 80k (160k) updates on 100 (500) hours of pseudo-labeled data. It takes 20 (40) hours (wall clock time) to fine-tune on 100 (500) hours of data using 8 TITAN RTX GPUs. We use HuggingFace's transformers toolkit~\citep{wolf2020transformers} for training the text NER model on pseudo-labels. Detailed config files are provided in our codebase.\footnote{\href{https://github.com/asappresearch/spoken-ner}{https://github.com/asappresearch/spoken-ner}}

For baselines that do not use pre-trained representations, we utilize the DeepSpeech2 (DS2) toolkit\footnote{\href{https://github.com/SeanNaren/deepspeech.pytorch}{https://github.com/SeanNaren/deepspeech.pytorch}}~\citep{amodei2016deep}. DS2 first converts audio files into spectrogram features. The model processes the spectrogram features through two 2-D convolutional layers followed by five bidirectional 2048-dim LSTM layers and a softmax layer. The softmax layer outputs the probabilities for a sequence of characters. The model has 26M parameters and is trained with SpecAugment data augmentation~\citep{park2019specaugment} and a character-level CTC objective.

\section{Results}
\label{sec:results}
\subsection{Baseline models}

\begin{table}[h]
\centering
\small
\caption{Dev set \% f-score performance of baseline models. All models here are trained on the 15-hour fine-tune set. The pre-trained speech and text models are mentioned wherever used or applicable. The last three rows are from previously established baselines~\cite{shon2022slue}.}
\begin{tabular}{l|cc|c}
\toprule
\multirow{2}{*}{\begin{tabular}[l]{@{}l@{}}{\bf NER}\\{\bf system}\end{tabular}} & \multicolumn{2}{c|}{\bf Pretrained model} & \multirow{2}{*}{\bf F1} \\
 & {\bf Speech} & {\bf Text} & \\
\midrule
Pipeline & \xmark & DeBERTa-B & 52.4 \\
E2E & \xmark & \xmark & 51.8       \\
\midrule
Pipeline & W2V2-B & DeBERTa-B & 72.0      \\
E2E & W2V2-B & \xmark & 68.1  \\
\midrule
Text NER & \xmark & DeBERTa-B & 86.0      \\
\bottomrule
\end{tabular}
\label{tab:ner-ext-baseline}
\end{table}
Results from all the baseline models are reported in \tab~\ref{tab:ner-ext-baseline}. The models here are trained on the 15-hour fine-tune set. We see that self-supervised pre-training gives a significant performance boost over no pre-training. The text NER model (which uses ground-truth transcripts) is far better than the pipeline method, which is better than the E2E model.

\subsection{Leveraging external data}
\begin{table}[h]
\centering
\small
\caption{Dev set \% f-score performance of the pipeline models. Note the baseline Pipeline (72) and text NER (86.0) performances without using any additional data from \tab~\ref{tab:ner-ext-baseline}.}
\begin{tabular}{l|l|cc}
\toprule
{\bf Ext. data} & {\bf Method}           & \multicolumn{1}{c}{{\bf 100h}} & \multicolumn{1}{c}{{\bf 500h}} \\
\midrule
Un-Sp              & SelfTrain-ASR    &       73.8                   &       74.4                   \\
Un-Txt             & SelfTrain-txtNER &         72.3                 &         70.8                 \\
Sp-Txt             & Pre-ASR          &       75.6                   &       77.7                  \\
\bottomrule
\end{tabular}
\label{tab:ner-ext-results-ppl}
\end{table}
\begin{table}[htbp]
\centering
\caption{Dev set \% f-score performance of the E2E models. Note the baseline E2E (68.1) and text NER (86.0) performances without using any additional data from \tab~\ref{tab:ner-ext-baseline}.}
\begin{tabular}{l|l|cc}
\toprule
{\bf Ext. data}      & {\bf Method}           & \multicolumn{1}{c}{{\bf 100h}} & \multicolumn{1}{c}{{\bf 500h}} \\
\midrule
\multirow{2}{*}{Un-Sp}  & SelfTrain-E2E    &          70.6                &       72.1                   \\
                        & Distill-Pipeline &          76.5                &    77.5                      \\
\midrule
Un-Txt                  & Distill-txtNER-lm      &         71.0                 &         71.7                 \\
\midrule
\multirow{2}{*}{Sp-Txt} & Distill-txtNER      &       79.2                   &          82.2               \\
                        & Pre-ASR          &         70.7                 &       73.2                  \\
\bottomrule
\end{tabular}
\label{tab:ner-ext-results-e2e}
\end{table}
We report F1 scores on the dev set using different pipeline and E2E approaches in Tables~\ref{tab:ner-ext-results-ppl} and \ref{tab:ner-ext-results-e2e}, respectively. \fig~\ref{fig:ner-ext-summary} presents key results when using each external data type for both E2E and pipeline models. The key findings are: \\
(i) Using external data reduces the gap between spoken NER baselines and text NER. \\
(ii) With access to either unlabeled speech or transcribed speech, E2E models outperform pipeline models, whereas, for the baselines, the opposite holds. \\
(iii) Using unlabeled text gives the smallest boost among the three types of external data, and the pipeline approach performs better in that setting.

We evaluated the test set scores only for the best-performing model in each category. A summary of test set results is presented in Appendix~\ref{sec:appendix-ner-ext-test-set}. The results follow the same trend as on the dev set.

\subsubsection{External text NER data}
\label{sec:ext-text-ner}
We try to improve the text NER model by using the OntoNotes5.0 NER corpus~\citep{pradhan2013towards}. Fine-tuning DeBERTa-base on OntoNotes5.0 produces an F1 of 60\% on the VoxPopuli dev set. Fine-tuning it further on VoxPopuli gives F1 86\% on the dev set. Since we do not see any boost over the existing vanilla approach (86\%, see \tab~\ref{tab:ner-ext-baseline}), we retain the original text NER model using only in-domain data and do not perform further experiments using the OntoNotes-finetuned model. 

\section{Discussion and analysis}
It is not surprising that the models perform poorly when not leveraging a pre-trained \sfm \ (\tab~\ref{tab:ner-ext-baseline}).
The limited labeled data is not enough for the baseline E2E approach, but the pipeline model can leverage a strong text representation model, which gives it an edge. 

Next, we discuss the improvements seen in both E2E and pipeline approaches and analyze the similarities and differences in errors made by different approaches.


\subsection{Improved E2E results}
When using external data with the E2E model, the best performing methods use either (a) external unlabeled speech ({\it Distill-Pipeline}) or (b) transcribed speech ({\it Distill-txtNER}) (\tab~\ref{tab:ner-ext-results-e2e}). The tagger models have a stronger semantic component than the E2E baseline in both of these scenarios because of their strong text NER module. The same cannot be said for the other competing approaches for these external data categories, {\it SelfTrain-NER} and {\it Pre-ASR}, which provide much lower improvements. {\it SelfTrain-NER} distills information from the LM into the model layers, but the n-gram LM is much less powerful than the transformer-based text NER module used in {\it Distill-Pipeline}. The {\it Pre-ASR} approach has no means to improve the semantic component in the updated model.

In the presence of unlabeled text data, the modification comes from a better 4-gram LM trained on pseudo-labels. Note that the baseline E2E model parameters do not change, unlike when using the other two types of external data. This can explain why this approach only has a small improvement over the baseline.

\subsection{Improved pipeline results}
The baseline pipeline model already takes advantage of the text NER module, which leaves little room for improvement in the semantic understanding component. In \tab~\ref{tab:ner-ext-results-ppl} we see that, using unlabeled text data to improve the text NER module ({\it SelfTrain-txtNER}) gives a small boost of 0.4\%. For comparison, note that the improvement from using unlabeled speech is 2.5\% over baseline. So, the hope with pipeline models is for the external data to improve the speech-to-text conversion, which can then help reduce error propagation between the independent pipeline modules.

\subsection{Amount of external data}
Almost all experiments produce a larger improvement when using 500 hours of external data than 100 hours. Only {\it SelfTrain-txtNER} has a reverse trend (see \tab~\ref{tab:ner-ext-results-ppl}). The higher amount of external data naturally increases the fraction of noisy data that the target model is trained on, and that may lead to a poorer model. We hypothesize that methods for balancing between the effects of manually annotated and pseudo-labeled examples could help tackle this issue~\citep{park2020improved}. However, we leave an in-depth investigation of this phenomenon to future work.

\subsection{Error analysis}
\label{sec:ner-ext-analysis}
An accurate spoken NER model needs to correctly identify both the entity phrase(s) and the entity label(s). So far, our evaluation uses the F1 score, which penalizes every detection that is wrong in either of these two attributes. 
A prerequisite of this evaluation is that the named entity phrase should be correctly spelled and present in the predicted text.
Here, we evaluate two more metrics that are focused mainly on the textual correctness of the output---word error rate ({\it WER}) and named entity accuracy ({\it NE accuracy}), and also look at the breakdown of F1 into precision and recall.
With the help of these metrics and fine-grained error categories, we study whether the errors in our spoken NER systems stem from erroneous text prediction (for example, a misspelled entity phrase could lead to its missed detection) or a lack of semantic understanding (i.e. incorrect identification of phrase and/or the entity tag, despite the phrase being present in the text output). 

%

{\it WER} is the word-level Levenshtein distance between the ground-truth text and the decoded text generated by the model. {\it NE accuracy} is the proportion of entity phrases correctly decoded in the speech-to-text conversion. An entity phrase is considered accurate only if all the words in the phrase are correctly decoded in the right order. For analysis, we choose the best-performing models within each category.

\begin{figure}[h]
\begin{minipage}[b]{1.0\linewidth}
\small
 \centering
 \centerline{\includegraphics[width=0.6\linewidth, trim=0 105 0 125, clip]{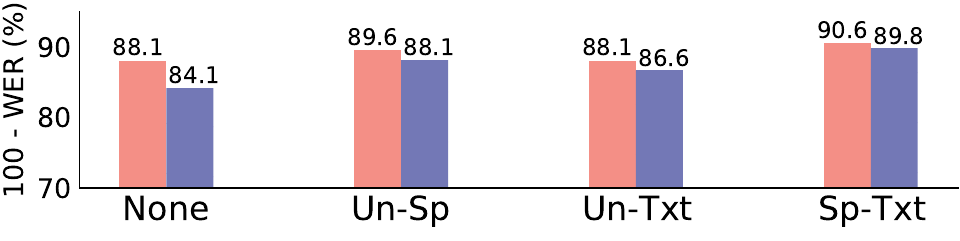}}
\end{minipage}
\begin{minipage}[b]{1.0\linewidth}

\footnotesize
 \centering
  \centerline{\includegraphics[width=0.6\linewidth, trim=0 85 0 95, clip]{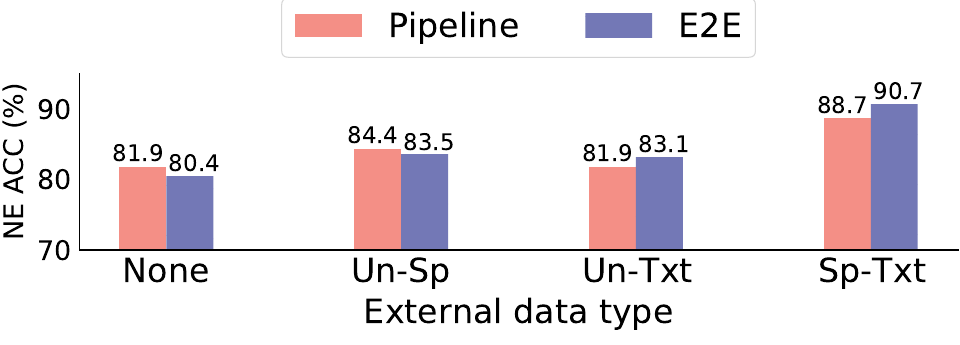}}
\end{minipage}

\caption{{$100-$WER (\%) and NE accuracy (\%) values on the dev set for the best-performing models in each category with access to 100 hours of external data.}}
  \label{fig:wer-ne-acc}
\end{figure}
\begin{figure}[htbp]
\begin{minipage}[b]{1.0\linewidth}
\small
 \centering
 \centerline{\includegraphics[width=0.6\linewidth, trim=0 75 0 100, clip]{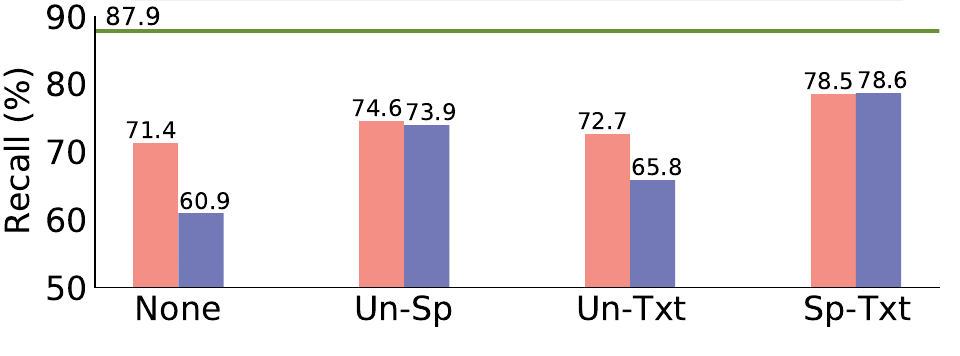}}
\end{minipage}
\begin{minipage}[b]{1.0\linewidth}

\footnotesize
 \centering
  \centerline{\includegraphics[width=0.6\linewidth, trim=0 65 0 75, clip]{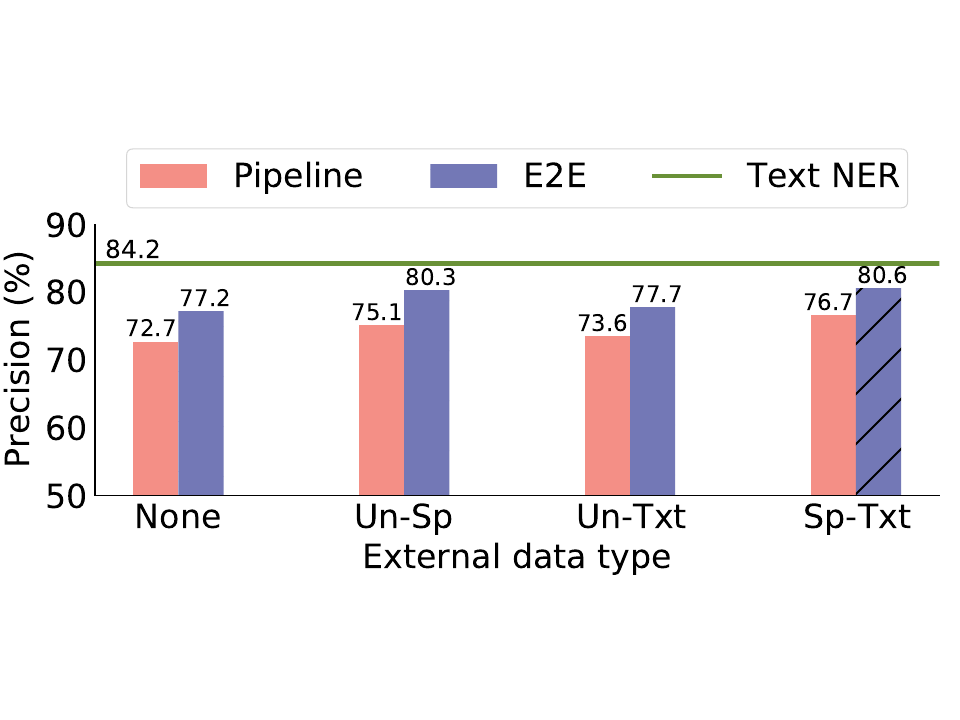}}
\end{minipage}

\caption{{Recall and precision on the dev set for the best-performing models in each category with access to 100 hours of external data.}}
  \label{fig:prec-recall}
\end{figure}

\fig~\ref{fig:wer-ne-acc} presents the NE accuracy and WER. We strip off the tag-specific special character tokens when evaluating WER for the E2E NER models. Note that we report $100 - $ WER so that higher is better in both plots. We observe that the ASR used in pipeline models typically performs better than the speech-to-text conversion of E2E models, even when the former has a poorer F1 (\fig~\ref{fig:ner-ext-summary}). This may lead us to hypothesize that the E2E model recognizes NE words better while doing worse for other words. However, this hypothesis is not supported by the {\it NE-ACC} results (\fig~\ref{fig:wer-ne-acc}).

Next, we look at the breakdown of F1 into precision and recall (\fig~\ref{fig:prec-recall}). We see that pipeline models have worse precision, thus suggesting that these suffer from a higher false-positive rate than the E2E models. This explains why {\it NE-ACC} is not predictive of F1; the former can inform us about errors due to false negatives, but not false positives.

\subsubsection{Error categories}
\begin{figure*}[t]
    \centering
    \hspace{-.15in}\includegraphics[width=\linewidth]{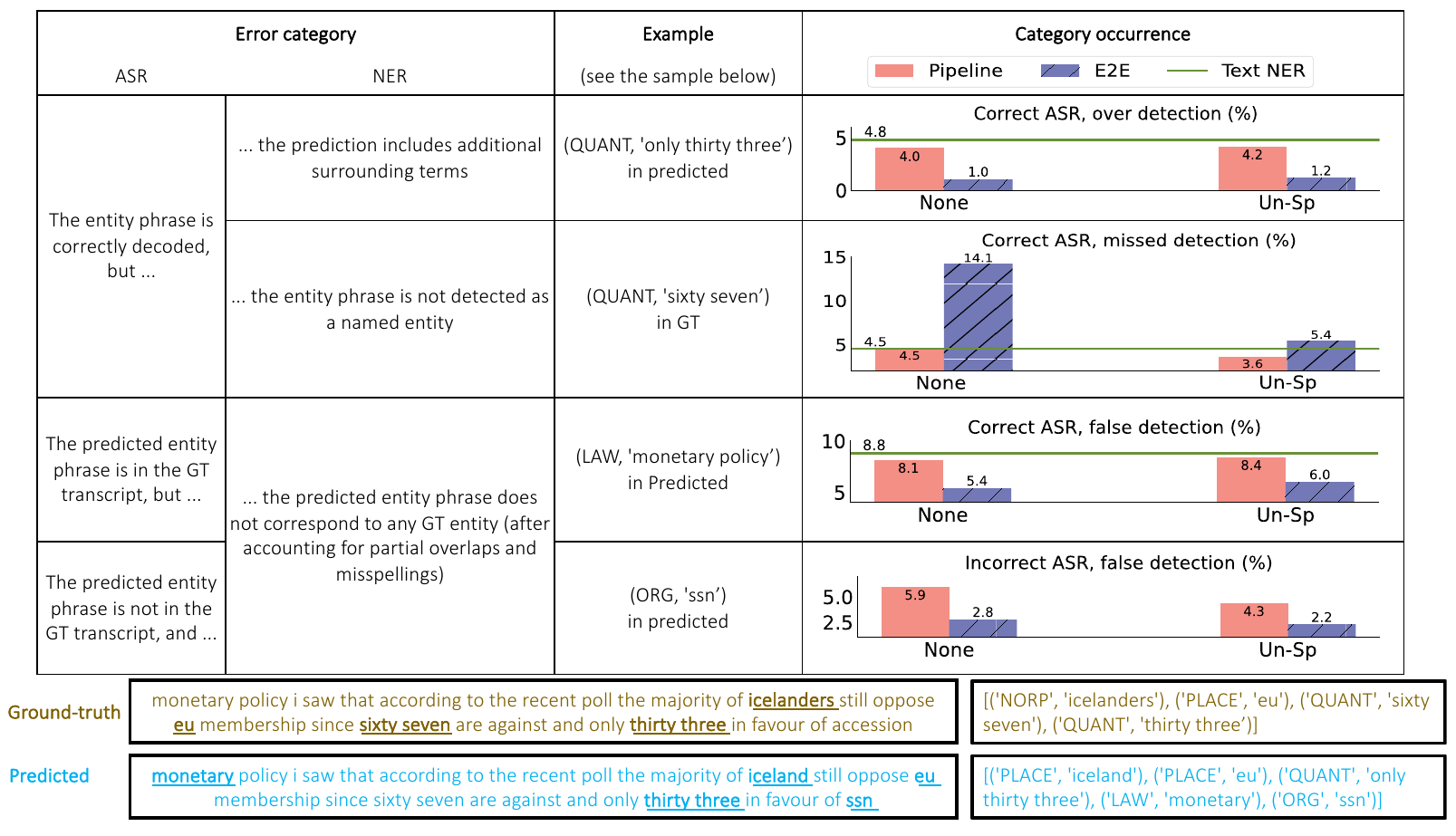}
    \caption{NER error category distribution on the dev set. The category-specific error rates in the plots are normalized by the total number of ground-truth (GT) entities.
    The examples here are artificially created from the same ground-truth example for ease of presentation. Actual examples of these categories are presented in Appendix~\ref{sec:appendix-ner-ext-error-analysis}.
    }
    \label{fig:error-cat}
\end{figure*}
For a more detailed understanding of model behavior, we categorize the NER errors into an exhaustive list of types (details in Appendix~\ref{sec:appendix-ner-ext-error-analysis}). We focus on four major categories showing noteworthy differences between pipeline and E2E approaches. We provide this analysis for the baselines, {\it Distill-Pipeline}, and {\it SelfTrain-ASR} models using external unlabeled speech data. The trends and observations presented here are consistent with the other two external data types. 

The major error categories, along with examples, are presented in \fig~\ref{fig:error-cat}. We observe that:\\
(i) False detections are 1.5 times more common in pipeline models than in E2E models, as expected based on the lower precision for the former. This happens even when the falsely detected text is not a speech-to-text conversion error.\\
(ii) Over-detections are 3.5 to 4 times more common in the pipeline models even when the entity phrase is decoded correctly.\\
(iii) Missed detections for the E2E {\it Distill-Pipeline} model are drastically reduced compared to the E2E baseline. Missed detections refer to cases where the entity phrases are correctly transcribed but are not labeled as named entities.  The improvement here therefore suggests that {\it Distill-Pipeline} improves the understanding capability of the E2E model, in addition to its speech-to-text capability. Also, note that the pipeline model does not enjoy the same benefit from unlabeled speech since this only involves self-training (instead of knowledge distillation from a much richer model for E2E).

Overall, the pipeline models suffer disproportionately from false positives. This seems to stem from the text NER model, which has even higher over-detection and false detection rates than the pipeline baseline models (\fig~\ref{fig:error-cat}). The reasons behind this difference between E2E and pipeline models need further investigation.

\section{Related work}
{\it Self-training}~\citep{scudder1965probability,yarowsky1995unsupervised,riloff1996automatically} is a popular approach to improve supervised models when some additional unannotated data is available. Self-training has been observed to improve ASR~\citep{parthasarathi2019lessons, xu2021self} and is also complementary to pre-training~\citep{xu2021self}. To the best of our knowledge, this is the first work to introduce it to spoken NER while also studying its effects on both E2E and pipeline approaches. 

{\it Knowledge distillation} is widely used in model compression research. In this approach, some intermediate output from a teacher model is used to train a smaller student model~\citep{hinton2015distilling}. In the context of our work, the teacher and student networks are two different approaches to solving NER tasks, and the latter is trained on the final output tags of the former.

{\it Transfer learning} has been widely employed for SLU tasks~\citep{lugosch2019speech, jia2020large}, including E2E spoken NER~\citep{ghannay2018end, caubriere2020we}. 
Automatic speech recognition (ASR) is a typically chosen task for pre-training a model before fine-tuning it for SLU. This choice is facilitated by the wider availability of transcribed speech than SLU annotations. Specifically for NER, ASR pre-training is expected to help since the accuracy of decoded texts can directly affect the final NER predictions.

\section{Summary}
In this chapter, we study the potential benefits of using a variety of external data types: (a) unlabeled speech, (b) unlabeled text, (c) speech with transcripts, and (d) text-based NER data.
We design E2E and pipeline approaches to leverage external data and observe an improvement from leveraging every type of external data. Our analysis also quantifies the pros and cons of the pipeline (speech recognition followed by text NER) and end-to-end (E2E) approaches. The best-performing model when using external data is an E2E approach. This is one of the few results in the literature thus far showing better performance for E2E over pipeline methods that use state-of-the-art modules for spoken language understanding.
Our specific contributions include: \\
(i) Overall, we obtain F1 improvements of up to 16\% for the E2E model and 6\% for the pipeline model over previously published baselines, setting a new state of the art for NER on this dataset (\fig~\ref{fig:ner-ext-summary}). \\
(ii) We benchmark the advantage of self-supervised representations (SSR) over a baseline that uses standard spectrogram features. SSR gives relative improvements of 36\%/31\% for pipeline/E2E models, respectively. To our knowledge, prior work has not directly measured this improvement over competitive baselines tuned for the task. \\
(iii) We establish that E2E models outperform pipeline approaches on this task, given access to external data, while the baseline models without the external data have the opposite relationship. \\
(iv) We provide a detailed analysis of model behavior, including differences in error types between pipeline and E2E approaches and the reasoning for the superiority of E2E over pipeline models when using external data but not in the baseline setting.


This work also leaves some interesting research questions for future work. For example, we see minor improvements between 100h and 500h of external data (see \tab~\ref{tab:ner-ext-results-ppl} and \ref{tab:ner-ext-results-e2e}), which suggests the question: What is the smallest amount of external data needed to obtain significant improvements in NER performance? Additionally, one preliminary experiment with external, out-of-domain text NER data (OntoNotes 5.0) fails to improve the text NER performance, suggesting the challenges of dealing with out-of-domain datasets. The scenario where we have access to out-of-domain external data is common but challenging, and warrants further study.

Overall, we hope that our work provides guiding principles for researchers working on SLU tasks in similar low-resource domains when some form of external data is abundant. 

\chapter{Extensive Evaluation of Speech Foundation Models on Spoken Named Entity Tasks}
\label{ch:slueperb}

In the previous chapters we introduced the spoken language understanding (SLU) benchmark tasks of named entity recognition (NER) and localization (NEL) along with \sfm-based baselines for both end-to-end (E2E) and cascaded approaches (\chap~\ref{ch:slue}). We also saw how \sfms \ can offer an effective backbone for E2E models, surpassing traditional cascaded approaches, by leveraging external unannotated data (\chap~\ref{ch:ner-ext}). 

So far, our study of \sfm-based models for SLU was limited to two different \sfms \ (\wavtovec \ and \hubert) adapted in one specific way that adds a linear prediction layer to an \sfm \ and adapts all \sfm \ transformer layers on the task-specific data (\tff \ introduced in \sect~\ref{sec:background-sfm-adapt}). While this is one of the most popular paradigms, in reality, we are also limited in our understanding of the comparative utility, if any, of several other \sfms \ and different ways to adapt them.

In this chapter we seek to fill that gap by performing an extensive evaluation of multiple self-supervised and supervised \sfms \ using three adaptation strategies varying in their training and inference costs: (i) \emph{frozen} \sfms \ with a \emph{lightweight} prediction head, (ii) \emph{frozen} \sfms \ with a \emph{complex} prediction head, and (iii) \emph{fine-tuned} \sfms \ with a \emph{lightweight} prediction head. We continue to focus on named entity recognition (NER) and localization (NEL) tasks.\footnote{The contents of this chapter are from our prior published paper~\cite{arora2024evaluation}. Siddhant Arora helped set up experiments on ESPnet and contributed results using a complex prediction head.}


\section{Method}
\label{sec:slueperb-method}
\begin{table*}[t]
\small
  \centering
  \caption{Summary of the \emph{encoder} of self-supervised and supervised pre-trained \sfms \ used in this work.}
  \resizebox{\linewidth}{!}
  {
  \begin{tabular}{lllrll}
    \toprule
    \multicolumn{1}{c}{Type} & 
    \multicolumn{1}{c}{SFM} &
    \multicolumn{1}{c}{Architecture} &
    \begin{tabular}[c]{@{}c@{}}Model\\ size\end{tabular} & \begin{tabular}[c]{@{}c@{}}Pre-training\\ data (hours)\end{tabular} &\begin{tabular}[c]{@{}c@{}}Pre-training\\ objective\end{tabular}\\
    \midrule
    \multirow{6}{*}{\begin{tabular}[l]{@{}l@{}}Self-\\supervised\end{tabular}}&
    \wavtovec-\largeM &
    \begin{tabular}[l]{@{}l@{}}7 CNN\\ 24 Transformer\end{tabular} &
    317.4M & LibriLight 60k & contrastive\\
    &
    \hubert-\largeM &
    \begin{tabular}[l]{@{}l@{}}7 CNN\\ 24 Transformer\end{tabular} & 316.6M & LibriLight (60k) & \begin{tabular}[l]{@{}l@{}}masked\\ prediction\end{tabular}\\
    &
    \wavlm-\largeM &
    \begin{tabular}[l]{@{}l@{}}7 CNN\\ 24 Transformer\end{tabular} & 315.5M & \begin{tabular}[l]{@{}l@{}}LibriLight (60k)\\ GigaSpeech (10k)\\ VoxPopuli (24k)\end{tabular} & \begin{tabular}[l]{@{}l@{}}masked\\ prediction\\ and de-noising\end{tabular}\\ \midrule
    \multirow{5}{*}{Supervised} &
    \whisper-\mediumM &
    \begin{tabular}[l]{@{}l@{}}2 CNN\\ 24 Transformer\end{tabular} & 315.7M & Web data (680k) & \multirow{3}{*}{\begin{tabular}[l]{@{}l@{}}Speech to text\\ transcription\\ and translation\end{tabular}}\\
    & \owsm & \begin{tabular}[l]{@{}l@{}}7 CNN\\ 18 Branchformer\end{tabular} & 560.8M & \begin{tabular}[l]{@{}l@{}}Mix of open-\\source data (180k)\end{tabular} & \\
    & {\it Pre-trained SLU} & \begin{tabular}[l]{@{}l@{}}2 CNN\\ 12 Conformer\end{tabular} & 83.2M & SLURP (58) & \begin{tabular}[l]{@{}l@{}}CTC with\\ entity tokens\end{tabular}
    \\
    \bottomrule
  \end{tabular}
  }
  \label{tab:slueperb-sfm-details}
\end{table*}
We evaluate \sfms \ listed in \tab~\ref{tab:slueperb-sfm-details} on named entity recognition (NER) and localization (NEL) tasks. An in-depth overview of \sfms \ is covered in \sect~\ref{sec:background-sfm}.

The self-supervised \sfms, \wavtovec~\cite{baevski2020wav2vec}, \hubert~\cite{hsu2021hubert}, and \wavlm~\cite{chen2022wavlm}, are encoder-only models, whereas \whisper~\cite{radford2023robust}, \owsm~\cite{peng2024owsm}, and {\it pre-trained SLU}~\cite{arora2022espnet} are supervised encoder-decoder models. While \whisper \ and \owsm \ are general-purpose multitask \sfms, {\it pre-trained SLU} model, trained on the SLURP data~\cite{bastianelli2020slurp}, is a supervised backbone that is trained on the task of entity recognition, similar to the evaluation tasks of NER and NEL. Thus, we compare the effect of leveraging large general-purpose \sfms \ with a much smaller model, both in terms of model size and pre-training data, trained on the external data for the task of interest.

In this work, we only use the encoder module of the supervised \sfms, similarly to prior work studying supervised \sfms \ for other downstream tasks~\cite{gong2023whisper}. This approach enables a standardized comparison with the self-supervised \sfms, which are encoder-only models.
 
\subsection{Evaluation protocols}
Here, we discuss three approaches for leveraging \sfms \ for NER and NEL. We employ a learnable weighted sum of hidden (transformer/conformer/branchformer) layer representations for all three approaches. The training objective, inference, and evaluation setup for NER and NEL are as described previously in \sect~\ref{sec:slue-baselines}.

{\bf Lightweight prediction head} is a \afl \ strategy (\sect~\ref{sec:background-sfm-adapt}) with a lightweight prediction head. This strategy has a comparatively low training and inference cost and is often used for a quick evaluation of \sfms~\cite{yang2021superb}. 

{\bf Fine-tuned representations} is a {\it weighted-finetune} strategy that uses the same prediction head as the previous strategy, and the \sfm \ parameters are fine-tuned along with the prediction head. Fine-tuning the backbone \sfm \ has been the most popular adaptation approach in the literature, especially when the final downstream task performance is of interest~\cite{baevski2020wav2vec, hsu2021hubert, chen2022wavlm, shon2022slue, ott-etal-2019-fairseq}.  
However, this approach significantly increases the computation cost during fine-tuning, which might make it challenging to use in scenarios with a limited computation budget.

{\bf Complex prediction head} is a \afl \ strategy with a complex, encoder-decoder-style prediction head. This strategy offers a middle ground with a comparatively low training cost but suffers from a much higher decoding cost due to autoregressive decoding. The inclusion of this strategy is also motivated by prior work~\cite{zaiem2023speechIS} demonstrating how a change in prediction head can lead to different takeaways when comparing different \sfms. 

\section{Experiments}
\label{sec:slueperb-exp-details}

\sloppy
All our experiments are conducted using the ESPnet toolkit.\footnote{\href{https://github.com/espnet/espnet}{https://github.com/espnet/espnet}}
We apply SpecAugment~\cite{park2019specaugment}, dropout~\cite{srivastava2014dropout} and label smoothing~\cite{muller2019does} techniques. The models are trained using NVIDIA A40 (40GB) GPUs. 
\tab~\ref{tab:slueperb-hyperparameters} shows training and inference hyperparameters for our hyperparameter search. We perform extensive tuning of training parameters, particularly warmup and learning rate, and the final values are chosen based on the performance on the validation set. The final model is an average of the ten best-performing checkpoints saved through the training process. The final fine-tuning configurations of all our models are made public.\footnote{\href{https://github.com/ankitapasad/espnet/tree/SLUE_clean/egs2/slue-voxpopuli/slu1/conf/tuning}{https://github.com/ankitapasad/espnet/tree/SLUE\_clean/egs2/slue-vox Populi/slu1/conf/tuning}}

\begin{table}[t]
\small
  \centering
  \caption{List of training and inference hyperparameters, along with the search values used for tuning, where applicable.}
 {
  \begin{tabular}{lr}
    \toprule
    Hyperparameter & Value \\
    \midrule
    Convolution Subsampling & [1/2x, 1/4x]\\
    Dropout Rate & [0, 0.1, 0.2] \\
    LR schedule & [inv. sqrt., exp. lr.]\\
    Max learning rate & [1e-1, 1e-2, 5e-3, 1e-3, 4e-4, 1e-4, 1e-5, 1e-6] \\
    Warmup steps &  [2500, 5000, 10000] \\
    Number of epochs & [30, 50, 70] \\
    Adam eps  &  1e-8\\
    Adam betas  & (0.9, 0.999)\\
    Weight decay & [1e-5, 1e-6, 1e-7]\\
    \midrule
    Beam Size & [1, 2, 10]\\
    Length Penalty & [0, 0.1]\\
    CTC weight & [0.0, 0.3] \\
    \bottomrule
  \end{tabular}
  }
  \label{tab:slueperb-hyperparameters}
\end{table}

Both {\it lightweight prediction head} and {\it fine-tuned representations} use a 2-layer conformer~\cite{Gulati2020ConformerCT} prediction head. {\it Complex prediction head} has a 12-layer conformer encoder and 6-layer transformer decoder. In line with prior work, CNN layers are kept frozen, and the intermediate CNN representations are not used to form an input to the prediction head. 
 
\section{Results and discussion}
\label{sec:slueperb-results}

\begin{table*}[t]
\small
  \centering
  \caption{Performance of various \sfms \ and adaptation strategies on the test set of SLUE-VoxPopuli for NER, ASR, and NEL tasks; darker shades correspond to better scores and lighter shades correspond to poorer scores. The suffix {\it -L} and {\it -M} for \sfms \ indicate \largeM \ and \mediumM \ sizes respectively. {\it Size} indicates the number of trainable parameters in millions. Dev set results are in Appendix \tab~\ref{tab:slueperb-dev-results}.}
\begin{tabular}{cc|r|cccc}
\toprule
\multirow{2}{*}{\shortstack{Adaptation \\ strategy}} & \multirow{2}{*}{SFM} & \multirow{2}{*}{\centering Size (M)} & \multicolumn{2}{c}{NER} & ASR & NEL \\
\cmidrule(r){4-5}\cmidrule(r){6-6}\cmidrule(r){7-7}
 &  & & label F1 $\uparrow$ & F1 $\uparrow$ & WER $\downarrow$ & frame F1 $\uparrow$ \\
\midrule
\multirow{6}{*}{\shortstack{{\it Frozen} \sfm \\ with a \\ {\it lightweight} \\ prediction head}} & HuBERT-L & 6.5 & \cellgradsc{76.5} & \cellgradsd{59.3} & \cellgradse{14.2} & \cellgradsf{67.7} \\
& wav2vec2.0-L & 6.5 & \cellgradsc{73.6} & \cellgradsd{57.5} & \cellgradse{16.0} & \cellgradsf{64.1} \\
 & WavLM-L & 6.5 & \cellgradsc{80.6} & \cellgradsd{64.5} & \cellgradse{10.4} & \cellgradsf{72.0} \\
& Whisper-M & 9.1 & \cellgradsc{79.6} & \cellgradsd{63.1} & \cellgradse{12.5} & \cellgradsf{71.8} \\
& OWSM 3.1 & 9.1 & \cellgradsc{78.4} & \cellgradsd{61.7} & \cellgradse{12.8} & \cellgradsf{70.5} \\
& Pre-trained SLU & 9.1 & \cellcolor{verylightgreen}{60.8} &  \cellcolor{verylightgreen}{45.5} &  \cellcolor{verylightgreen}{39.1} &  \cellcolor{verylightgreen}{47.8} \\
\midrule
\multirow{6}{*}{\shortstack{{\it Frozen} \sfm \\ with a \\ {\it complex} \\ prediction head}} & HuBERT-L & 32.4 & \cellgradsc{78.5} & \cellgradsd{63.1} & \cellgradse{13.0} & \cellgradsf{69.8} \\
&  wav2vec2.0-L & 32.4 & \cellgradsc{78.2} & \cellgradsd{63.7} & \cellgradse{14.0} & \cellgradsf{71.2} \\
& WavLM-L & 32.4 & \cellgradsc{82.7} & \cellgradsd{69.7} & \cellgradse{10.1} & \cellgradsf{72.6} \\
& Whisper-M & 32.4 & \cellgradsc{79.2} & \cellgradsd{64.1} & \cellgradse{13.2} & \cellgradsf{70.1} \\
& OWSM 3.1 & 35.0 & \cellgradsc{79.6} & \cellgradsd{66.0} & \cellgradse{12.6} & \cellgradsf{68.6} \\
& Pre-trained SLU & 34.9 & \cellcolor{verylightgreen}{68.7} & \cellcolor{verylightgreen}{54.8} & \cellcolor{verylightgreen}{28.5} & \cellcolor{verylightgreen}{54.4} \\
\midrule
\multirow{6}{*}{\shortstack{{\it Fine-tuned} \sfm \\ with a \\ {\it lightweight} \\ prediction head}} & HuBERT-L & 318.9 & \cellgradsc{78.8} & \cellgradsd{62.6} & \cellgradse{12.0} & \cellgradsf{69.4} \\
&  wav2vec2.0-L & 319.7 & \cellgradsc{78.2} & \cellgradsd{62.9} & \cellgradse{11.7} & \cellgradsf{68.6} \\
& WavLM-L & 317.8 & \cellgradsc{82.5} & \cellgradsd{66.3} & \cellgradse{9.7} & \cellgradsf{71.7} \\
& Whisper-M & 314.8 & \cellgradsc{76.9} & \cellgradsd{59.8} & \cellgradse{18.2} & \cellgradsf{56.6} \\
& OWSM 3.1 & 569.9 & \cellgradsc{78.5} & \cellgradsd{61.5} & \cellgradse{14.3} & \cellgradsf{65.1} \\
& Pre-trained SLU & 92.3 & \cellcolor{verylightgreen}{60.8} & \cellcolor{verylightgreen}{47.6} & \cellcolor{verylightgreen}{37.1} & \cellcolor{verylightgreen}{49.1} \\
\bottomrule
\end{tabular}
\label{tab:slueperb-test-results}
\end{table*}

In \tab~\ref{tab:slueperb-test-results} we see the performance of various \sfms \ and adaptation straregies for NER, ASR, and NEL tasks. Speech recognition performance is evaluated on the output of the NER model after stripping the entity-specific tokens. 

\subsection{Findings}
We notice that {\bf pre-trained SLU consistently performs the worst} with a large margin. {\it Pre-trained SLU} offers a realistic setup with a small backbone model trained on a moderate amount of task-specific data (\tab~\ref{tab:slueperb-sfm-details}). It remains to be seen whether more complicated strategies, such as multi-stage training or self-distillation, are better able to leverage task-specific data. Our experiments establish that it is straightforward to effectively employ bigger general-purpose \sfms, even the ones pre-trained without any supervision. However, the same approaches don't necessarily apply to smaller-scale task-specific supervised pre-trained backbones. 

On the other end of the trend, {\bf WavLM always performs the best} for all tasks on SLUE-VoxPopuli, even better than supervised \sfms. The superior performance of \wavlm \ as compared to its SSL counterparts (\wavtovec \ and \hubert) has been observed in the literature~\cite{yang2021superb}, but we hypothesize that having VoxPopuli as a part of \wavlm's pre-training data must provide an added advantage when evaluating on SLUE-VoxPopuli. Further supporting our hypothesis, we note that \wavlm's superiority does not necessarily extend to other SLU tasks with no domain overlap with the pre-training data~\cite{arora2024evaluation}.

For the remainder of the discussion, we will focus on \sfms \ other than {\it pre-trained SLU} and \wavlm. Interestingly, we see {\bf no universal trend between supervised and self-supervised SFMs}.
For the {\it frozen}-{\it lightweight} evaluation on all tasks, supervised \sfms, (\whisper \ followed by \owsm) outperform self-supervised \sfms \ (\hubert \ followed by \wavtovec). The performance gap between supervised and self-supervised \sfms \ reduces as the prediction head becomes more complex, but the trend is no longer consistent across tasks. For instance, on the NEL task, \wavtovec \ has the poorest performance with a {\it frozen}-{\it lightweight} strategy, but it is the best performing \sfm \ with the {\it frozen}-{\it complex} strategy.

We notice {\bf a general degradation of performance with fine-tuned SFMs as compared to frozen SFMs with a complex prediction head}, despite pretty extensive hyperparameter tuning (\tab~\ref{tab:slueperb-hyperparameters}). For instance, while \whisper \ is a top-performing \sfm \ (ranking second or a close third) for all experiments using frozen representations, fine-tuning \whisper \ leads to a significant performance degradation, making it rank last.
This suggests that, while fine-tuning \sfms \ allows a much higher modeling capacity, it is not necessarily straightforward to attain a good checkpoint with increased model capacity, especially when working with a small fine-tuning data (less than 20 hours). Most surprisingly, {\bf fine-tuned supervised SFMs perform the poorest on ASR across all SFMs and adaptation strategies}.

\subsection{Takeaways}
\label{sec:slueperb-takeaways}
Overall, our findings indicate that \wavlm \ provides the best backbone for ASR, NER, and NEL tasks on SLUE-VoxPopuli.
However, since \wavlm's pre-training data overlaps with SLUE-VoxPopuli, it remains unclear whether its superior performance generalizes beyond these specific downstream tasks and dataset.

When considering the performance of other \sfms, we observe that no single adaptation strategy or pre-training approach (supervised vs. self-supervised) consistently outperforms across all tasks; this conclusion is further supported by an extension of this study across a broader range of tasks~\cite{arora2024evaluation}. Despite this variability across methods, we can still extract meaningful patterns and guidance.

Next we highlight a few key takeaways from our findings that can  guide practitioners in selecting and implementing backbone \sfms \ for downstream tasks.

\subsubsection{Effect of backbone \sfm}
We also observe that general-purpose \sfm \ backbones are much more effective than using a smaller, task-specific backbone ({\it pre-trained SLU} in our experiments).

While our findings indicate no clear choice between self-supervised and supervised \sfms, it is important to note that our experimental setup does not leverage the pre-trained decoder of the supervised backbone. Supervised \sfms \ may encode semantic content in their decoder layers, which we discard by using just the pre-trained encoder. We anticipate that SLU tasks could benefit from integrating the pre-trained decoder of supervised \sfms\space, although we leave this exploration to future work.

Lastly, \wavlm's consistent lead stresses the importance of pre-training data; data domain (mis-)match between pre-training and task-specific data could have a significant (dis)advantage. We have also observed this phenomenon in our task-specific findings in the previous chapter studying individual \sfm \ layers (\sect~\ref{sec:res-effect-of-domain}).

\subsubsection{Effect of adaptation strategy}
The differences in performance of \sfm \ backbones are most pronounced in the {\it frozen}-{\it lightweight} strategy and least in the {\it frozen}-{\it complex} strategy.
Thus, the choice of frozen \sfm \ backbone is possibly less significant as we add more learnable additional parameters.

Both {\it frozen}-{\it complex} and {\it finetune}-{\it lightweight} strategies generally improve upon the {\it frozen}-{\it lightweight} strategy. While {\it frozen}-{\it complex} generally outperforms {\it finetune}-{\it lightweight}, it is important to note that the use of complex prediction heads leads to a substantial increase in inference time ($>$ 2.5x for all tasks). Thus, employing a complex prediction head is, in general, better when inference speed is not a bottleneck. Conversely, if latency is a concern, fine-tuned representations with a lightweight prediction head serve as a good option, enhancing performance without compromising inference time.

\section{Summary}
This is the end of the three-chapter focus on the applicability of \sfms \ for spoken language understanding tasks.
Collectively, \chaps~\ref{ch:slue},~\ref{ch:ner-ext}, and~\ref{ch:slueperb} study various ways to employ backbone \sfms \ for named entity recognition (NER) and localization (NEL) tasks.
\chap~\ref{ch:slue} introduces the tasks and establishes cascaded and end-to-end (E2E) baselines. These baselines use state-of-art speech and text foundation models trained with the {\it finetune}-{\it lightweight} strategy.
\chap~\ref{ch:ner-ext} presents effective ways to leverage external unannotated data to improve both cascaded and E2E baselines, with the E2E approach eventually outperforming the cascaded counterpart.  

Promising results from \sfm-based E2E approaches motivate the current chapter, where we focus solely on E2E solutions. Moreover, we extend our study to different adaptation strategies as well as a larger variety of backbone \sfms. 
Our study highlights trends and guidelines for practitioners (\sect~\ref{sec:slueperb-takeaways}) and also answers previously unresolved questions, such as whether supervised \sfm \ backbones are better than self-supervised ones (not necessarily and not by much) or whether fine-tuning \sfm \ is always the best approach (not quite; a complex prediction head with much fewer training parameters performs better). 

Note that the previous chapters use an external language model for decoding, which provides a significant boost in performance (see \tab~\ref{tab:ner-baseline}). In this chapter, we choose not to use an external language model to focus our study on the impact of the choice of \sfm \ backbones and adaptation strategies without any confounding factors. So, it is not straightforward to compare the results presented in the three chapters. 





\newpage
\chapter{Closing Remarks}
\label{ch:conclusion}

Even before supervised deep learning models became mainstream, researchers have been investigating ways to utilize unlabeled audio data in speech technology~\cite{kemp1999unsupervised, lamel2002lightly}.
So, the advent of successful self-supervised speech models, trained only on unlabeled speech utterances, marked a revolutionary shift in the landscape~\cite{mohamed2022self}.
Pre-trained speech foundation models (\sfms)---including self-supervised and supervised backbone models---can be used directly as representation extractors for downstream tasks~\cite{yang2021superb}, as initialization for task-specific models~\cite{baevski2020wav2vec, shon2022slue, shon2023slue}, or as components of more general-purpose backbones, such as multilingual~\cite{pratap2024scaling, zhang2023google}, multimodal~\cite{hu2024wavllm, tang2023salmonn, wang2023slm}, or generative~\cite{rubenstein2023audiopalm, yang2023uniaudio, borsos2023audiolm} foundation models.

This thesis complements the empirical success stories by providing a principled way to interpret the knowledge acquired by \sfms \ during pre-training. 
The lightweight nature of our analysis framework enables us to present a comparative study across a diverse set of \sfms \ (\chap~\ref{ch:analysis}).
We also test the ability of various statistical tools to provide a reliable way to evaluate the knowledge encoded in \sfm \ representations (\chap~\ref{ch:compare-tools}).
While we focus on acoustic and linguistic properties, the analysis framework is generic and easily scalable. For instance, the tools we use have been extended to the study of more properties in \sfms, such as speaker and language-specific characteristics~\cite{monteiro2023towards}, paralinguistics~\cite{li2023exploration}, and similarities with brain prediction signals~\cite{chen2023self}. 
Our CCA-based framework analyzes representations as a whole, but future work can study the learned CCA projection matrices to interpret the role of individual neurons. This direction can contribute to the otherwise limited research on neuron-level analysis of \sfm \ representations~\cite{chowdhury2023end, lin2024property}.

Our extensive study indicates that the intermediate layers of self-supervised \sfms \ encode the most linguistic content, where the location of the most meaningful layer(s) varies based on the pre-training objective. In contrast, visual or textual supervision during pre-training encourages the final \sfm \ layers to encode more linguistically meaningful representations. 
Such analytical findings enable us to make informed decisions when adapting \sfms \ for downstream tasks (\chaps~\ref{ch:compare-tools} and \ref{ch:implications}).
In \chap~\ref{ch:implications}, we successfully connect our findings to adaptation strategies that use \sfms \ as frozen extractors or when fine-tuned on task-specific data.
However, more work is needed to fully understand how different adaptation strategies interact with pre-trained \sfms.
For instance, our experiments show that the optimal placement of LoRA modules---determined through extensive empirical evaluation---does not align with the distribution of encoded knowledge across layers, as measured by the our analysis metrics.
As the backbone models become larger, such parameter-efficient modules are more common. So, future work studying how parameter efficient modules interact with the pre-trained \sfms \ could be invaluable in developing principled approaches for model use.

This thesis also contributes to the empirical study of \sfms \ by facilitating their evaluation on spoken language understanding (SLU) tasks, specifically named entity recognition and localization (\chap~\ref{ch:slue}). While \sfms \ are far from solving spoken language understanding, they have brought us significantly closer to oracle models that operate on ground-truth transcripts (\chaps~\ref{ch:ner-ext} and \ref{ch:slueperb}). Our extensive evaluation of \sfms \ on SLU tasks suggests that frozen representations with a complex prediction head (lower training and higher inference budgets) perform better than fine-tuning the representations with a lightweight prediction head (higher training and lower inference budgets). 
Choosing between these adaptation strategies, like selecting the optimal placement of LoRA modules, required extensive experimentation. 
These outcomes could not be predicted from off-the-shelf analyses of \sfms \ alone. This highlights a gap in our current understanding of how fine-tuning interacts with pre-trained \sfm \ parameters.
We propose some CCA-based analysis in this thesis, but it is limited to comparing representations of pre-trained and fine-tuned \sfms. Future work can complement this representation analysis with the study of how learned weights change during fine-tuning or PEFT.

For task-based evaluations, we consistently find an \sfm's downstream performance depends on both the encoded knowledge in the pre-trained model and the domain shift between pre-training and evaluation data (\chaps~\ref{ch:analysis} and \ref{ch:slueperb}). This effect of domain shift is very common in transfer learning, and prior work has also highlighted this in the context of \sfms~\cite{zaiem2023speechIS, hsu2021robust}. Future studies can benefit from developing insights into how an \sfm's pre-training data can influence downstream task performance, such as recent work studying the impact of data bias for SUPERB tasks~\cite{meng2022don}.
Developing these insights about the encoded knowledge, the interaction with adaptation modules, and the effect of pre-training data can collectively aid users in choosing an optimal \sfm \ and the corresponding adaptation strategy for a downstream task without extensive experimentation. 

All the analysis and evaluation presented in this thesis use either encoder-only \sfms \ or the encoder part of encoder-decoder \sfms \ (in the case of Whisper).
Especially for supervised \sfms \ and generative models, analysis of decoders could provide useful insights into the relative role of the encoder and decoder, as previously investigated for machine translation models~\cite{belinkov-etal-2017-neural}.
Although the analytical tools we develop for encoder-only \sfms \ cannot be directly extended to decoders because of their auto-regressive nature, one can adopt the same tools using approaches like teacher-forcing and aggregating decoder representations spanning predicted phones/words.
Understanding the knowledge encoded in the decoder and potentially incorporating the decoder for downstream task adaptation could especially benefit tasks that involve generation, such as spoken question answering or speech summarization.


Another promising avenue for future research includes studying the training dynamics of \sfms. While this has been extensively studied for text foundation models~\cite{saphra-thesis}, work on \sfms \ is limited~\cite{chung2020similarity}.
Using a fixed number of gradient updates during pre-training is common, but checkpoints saved at different training steps may be better suited for specific tasks and use cases. Considering the concrete connection between CCA scores and task performance (\chap~\ref{ch:compare-tools}), a promising future direction could be to explore the use of CCA scores for early stopping. 

Building new \sfms \ is still an active area of research. \sfms \ are being studied as a backbone for under-explored, yet challenging, downstream speech applications such as recognizing disordered~\cite{wiepertspeech} or children's speech~\cite{fan2024benchmarking}. A section of the community working on spoken language models is primarily using \sfms \ as acoustic encoders while leveraging large language models for reasoning capabilities~\cite{tang2023salmonn, gong2023joint}. Many speech synthesis models use \sfms \ to obtain meaningful discrete representations of speech~\cite{rubenstein2023audiopalm, yang2023uniaudio, borsos2023audiolm}. 
The optimal way to incorporate \sfms \ in any of these models continues to be an active area of research that is often tackled via extensive experimentation~\cite{verdini2024connect, arora2024evaluation, peng2024probing}, which, although reliable, is not a scalable approach.
It also remains uncertain whether the future of our field lies in developing a single, large general-purpose \sfm \ or maintaining an inventory of many specialized, ``smaller" \sfms \ to select from. Regardless of the path taken, robust analytical tools will be essential for guiding users in making informed decisions about the hows, whys, and which-es of \sfms.






\bibliographystyle{plain}
\bibliography{refs-clean,refs-arxiv,refs-unpolished}

\newpage
\phantomsection  
\addcontentsline{toc}{chapter}{Appendices} 

\begin{appendices}
\linespread{1.25}\selectfont
\normalfont  
\normalsize  
\addtocontents{toc}{\protect\setcounter{tocdepth}{-1}}
\chapter{Lightweight analysis of \sfms}
\section{Results on Whisper}
\label{sec:appendix-whisper}
\begin{figure}[htbp]
\centering
    \includegraphics[width=0.7\textwidth]{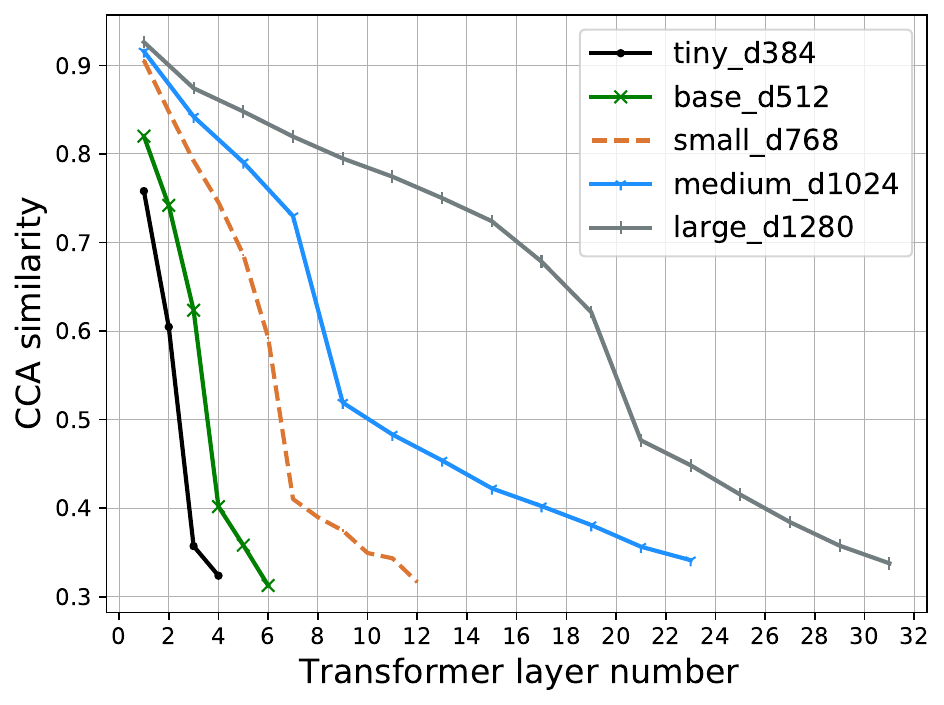}
    \caption{CCA similarity between \whisper \ frame-level representations and local features. Refer to \sect~\ref{sec:res-cca-intra} in the main text for a detailed discussion.}
    \label{fig:res-intra-whisper}
\end{figure}

\begin{figure}[htbp]
\centering
    \includegraphics[width=0.7\textwidth]{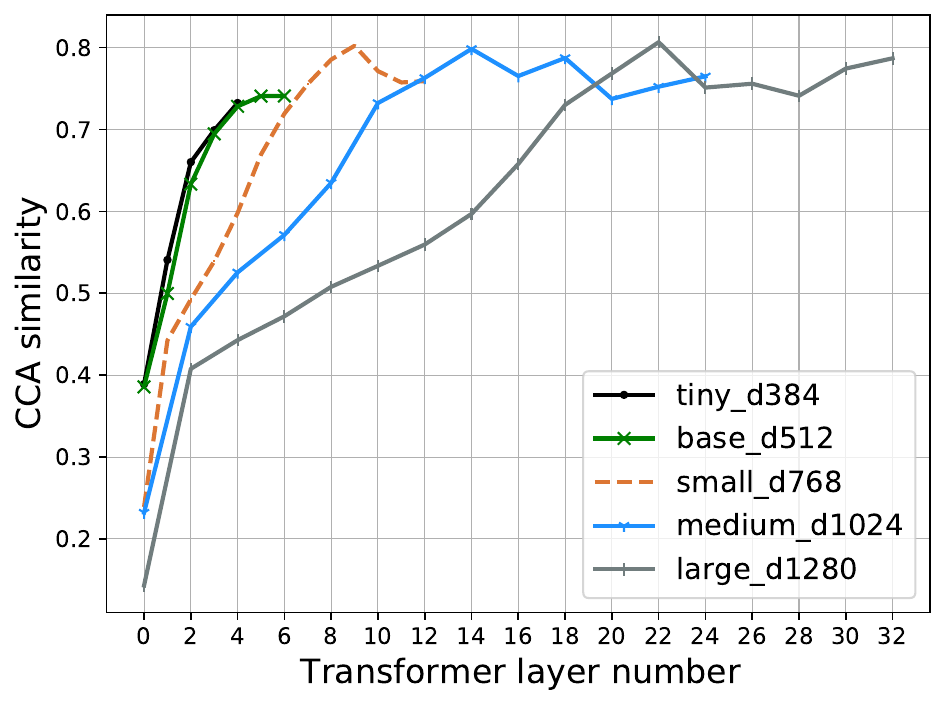}
    \caption{Phonetic content; CCA similarity between \whisper \ phone segment representations and phone identity. Refer to \sect~\ref{sec:res-cca-surface} in the main text for a detailed discussion.}
    \label{fig:res-phone-whisper}
\end{figure}

\begin{figure}[htbp]
\centering
    \includegraphics[width=0.7\textwidth]{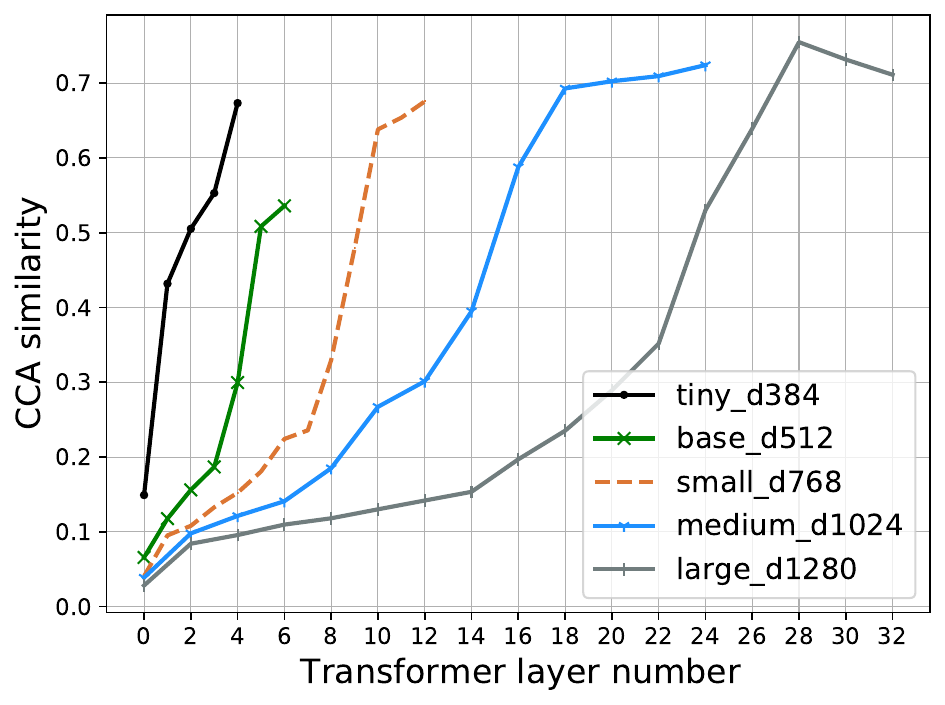}
    \caption{Word-level content; CCA similarity between \whisper \ word segment representations and word identity. Refer to \sect~\ref{sec:res-cca-surface} in the main text for a detailed discussion.}
    \label{fig:res-word-whisper}
\end{figure}

\begin{figure}[htbp]
\centering
    \includegraphics[width=0.7\textwidth]{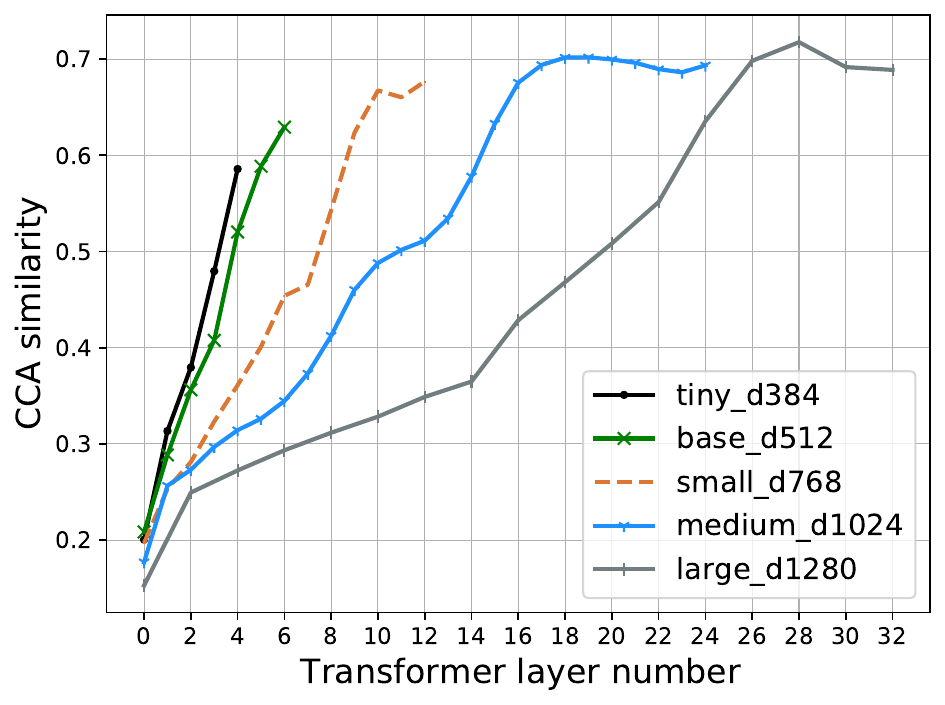}
    \caption{Word pronunciation content; CCA similarity between \whisper \ word segment representations and AGWEs. Refer to \sect~\ref{sec:res-cca-linguistic} in the main text for a detailed discussion.}
    \label{fig:res-agwe-whisper}
\end{figure}

\begin{figure}[htbp]
\centering
    \includegraphics[width=0.7\textwidth]{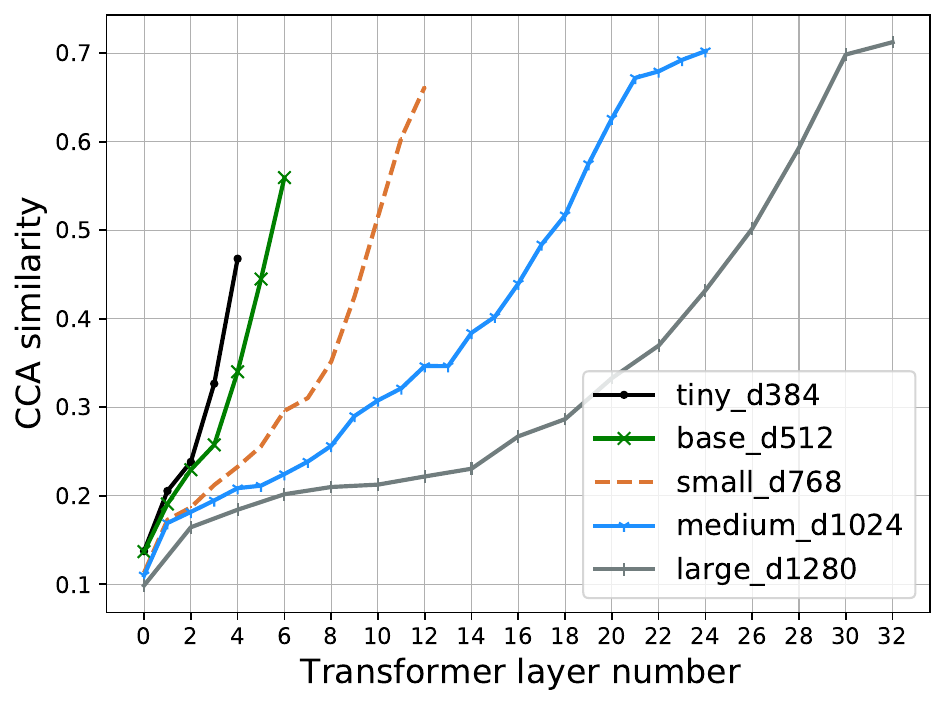}
    \caption{Syntactic content; CCA similarity between \whisper \ word segment representations and POS attributes. Refer to \sect~\ref{sec:res-cca-linguistic} in the main text for a detailed discussion.}
    \label{fig:res-syn-whisper}
\end{figure}

\begin{figure}[htbp]
\centering
    \includegraphics[width=0.7\textwidth]{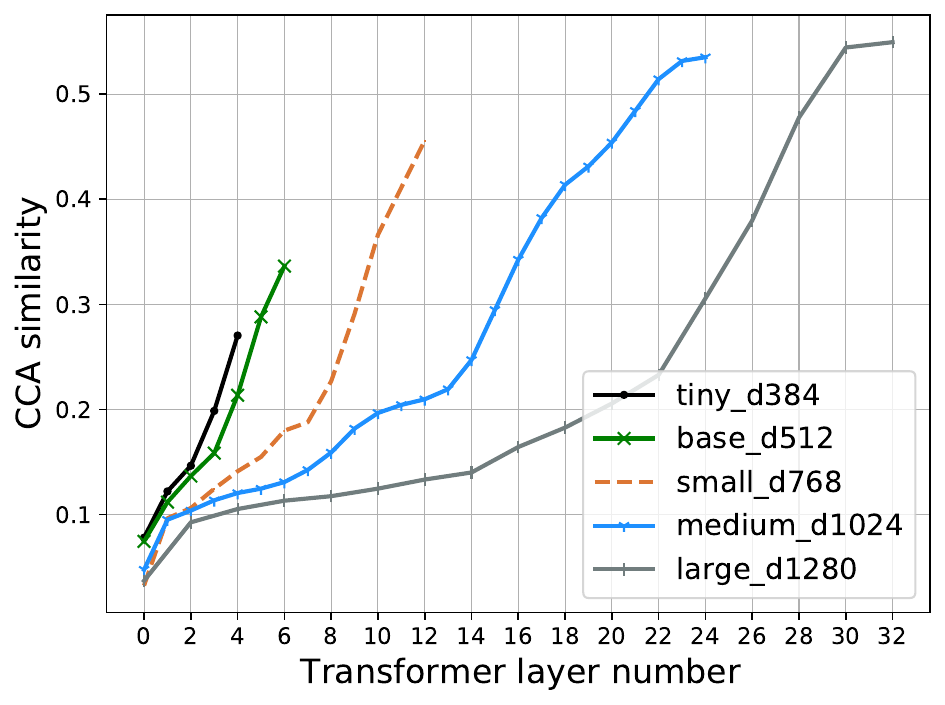}
    \caption{Semantic content; CCA similarity between \sfm \ word segment representations and SemCor attributes. Refer to \sect~\ref{sec:res-cca-linguistic} in the main text for a detailed discussion.}
    \label{fig:res-sem-whisper}
\end{figure}

\begin{figure}[htbp]
\centering
    \includegraphics[width=0.7\textwidth]{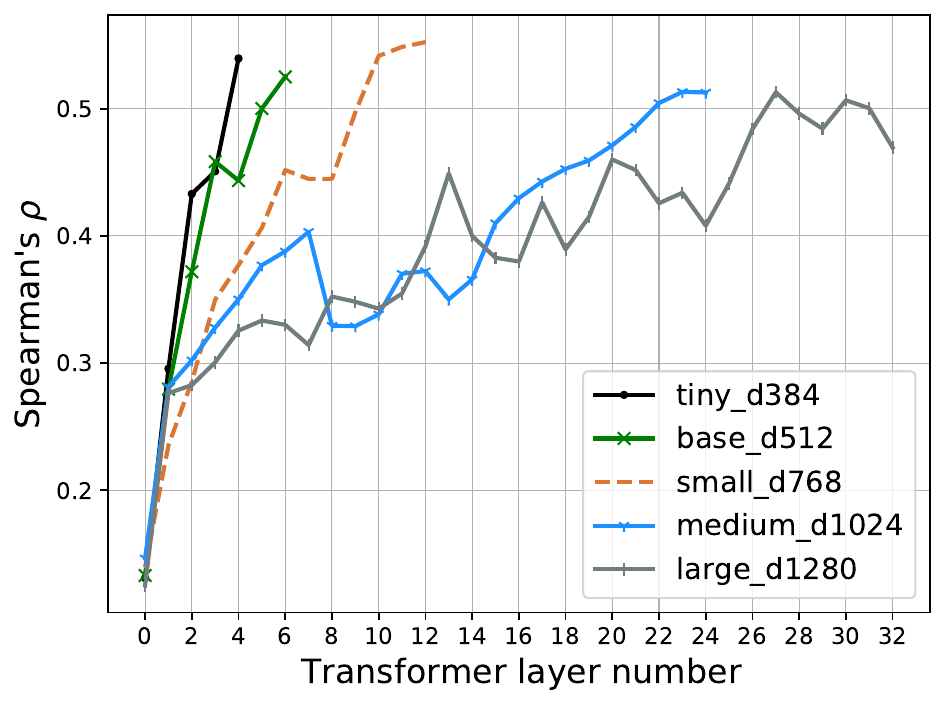}
    \caption{ Performance on spoken STS task using representations from \whisper. Refer to \sect~\ref{sec:res-sts} in the main text for a detailed discussion.}
    \label{fig:res-sts-whisper}
\end{figure}

Analysis and task-based evaluation for all five sizes of \whisper \ models presented in \figs~\ref{fig:res-intra-whisper}-\ref{fig:res-sts-whisper}. The hidden dimensions for different \whisper \ model sizes are mentioned in the legends. 

Refer to \sect~\ref{sec:res-analysis} for the corresponding discussion in the main text.

\chapter{Comparative study of analysis tools}
Results presented below complement the results and discussion in \chap~\ref{ch:compare-tools}.

\section{\vcca{\it -top-one} for discrete labels}
\label{sec:appendix-metric-vcca}
In \sect~\ref{sec:res-compare-tools-analysis} we note how \vcca{\it -top-one} scores are consistently high across all layers for phonetic and word-level content.
In both these instances we are comparing \sfm \ representations to one-hot vectors.
We argue that in such cases, the CCA optimization problem to find the first direction (\eq~\ref{eq:cca1}) is at least as simple as finding a projection that differentiates one phone (or word) from all other phones (or words), which is not informative of whether the learned representations encode knowledge that helps us distinguish between all phones (or words).

Let's revisit CCA Equation~\ref{eq:cca1} for an intuitive understanding of our hypothesis. Let's say $X\in\mathbb{R}^{d_x\times N}$ are \sfm \ representations and $Y\in\mathbb{R}^{d_y\times N}$ are one-hot vectors. Now, if projection $b\in\mathbb{R}^{d_y}$ is a one-hot vector, where $b_k=1$, then the resulting $b^TY\in\mathbb{R}^N$ will be a multi-hot vector with ones for samples that correspond to class $k$ and zeros for all other samples. So, a very high correlation can be achieved between $a^TX$ and $b^TY$ by a projection $a\in\mathbb{R}^{d_x}$ that maps $x_i$s corresponding to class $k$ to a higher value than samples corresponding to all other classes.

This differentiates class $k$ from all other classes and does not inform whether the $x_i$s encode knowledge that can help differentiate all classes. So, studying the trend of \vcca{\it -top-one} can be misleading.

\section{Layer-wise trends}
\label{sec:appendix-metric-plots}
\begin{figure*}[btp]

\centering
    \includegraphics[width=\textwidth]{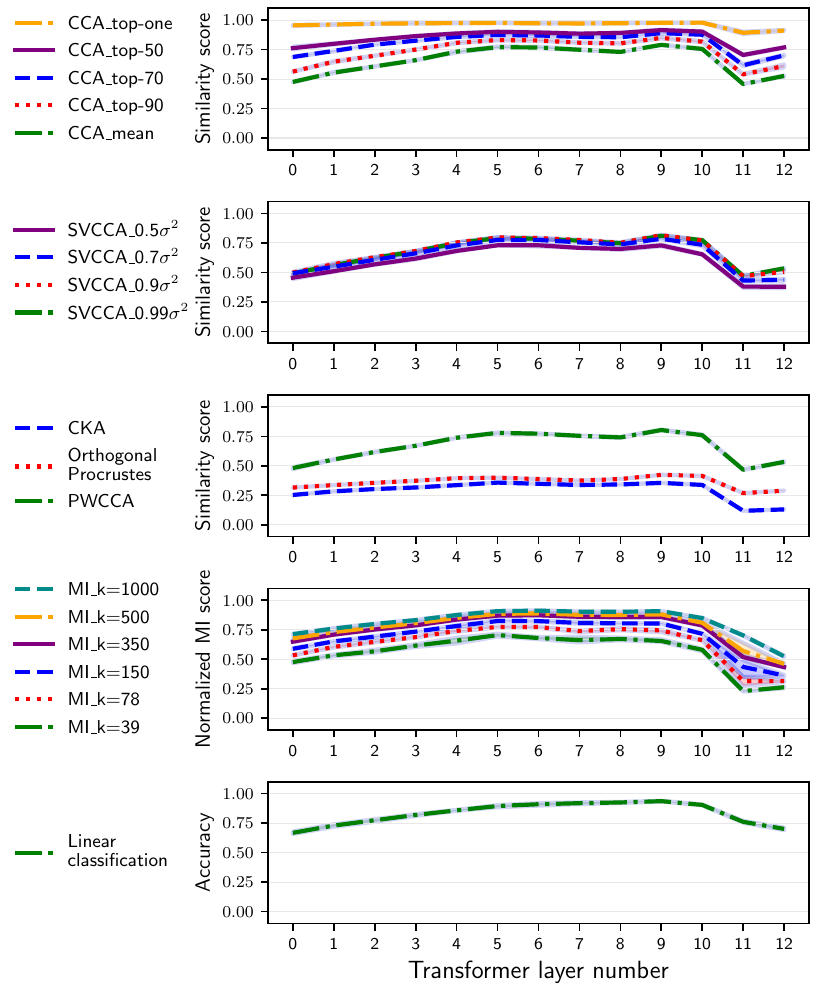}


     


\caption{Different tools comparing \sfm \ representations with phone identity for \wavtovec-\baseM.}
\label{fig:res-wav2vecb-phone-metric}
\end{figure*}

\begin{figure*}[btp]
\centering
    \includegraphics[width=\textwidth]{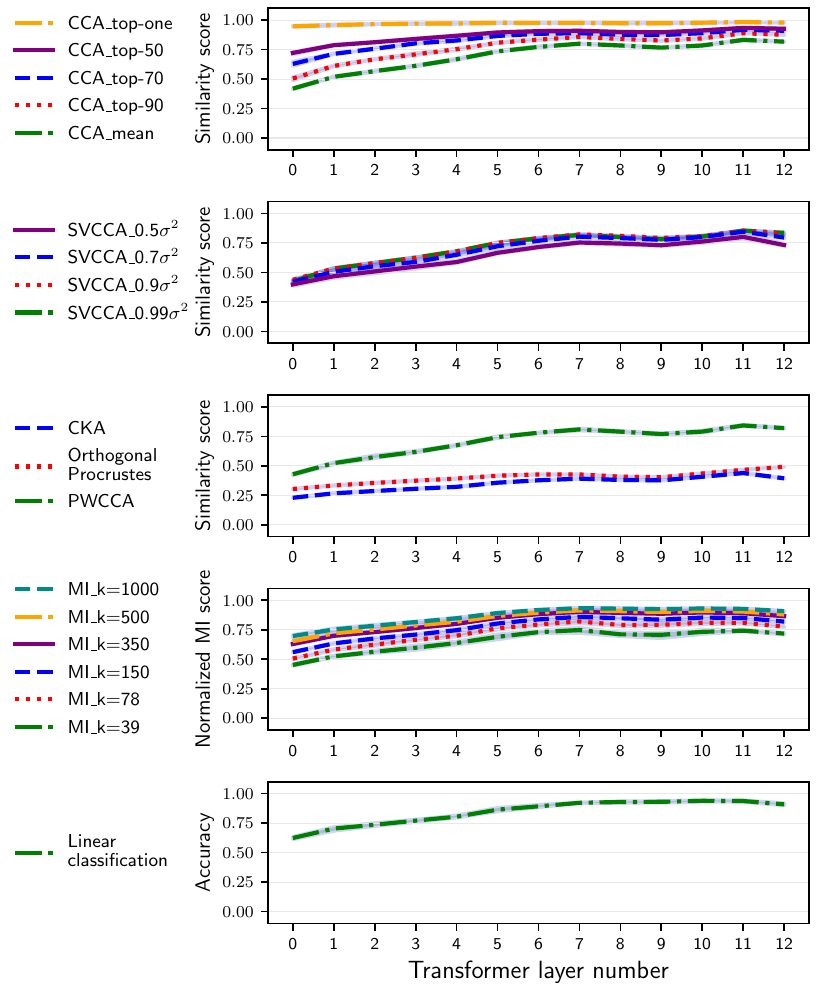}


     

     
\caption{Different tools comparing \sfm \ representations with phone identity for \hubert-\baseM.}
\label{fig:res-hubertb-phone-metric}
\end{figure*}

\begin{figure*}[btp]

\centering
    \includegraphics[width=\textwidth]{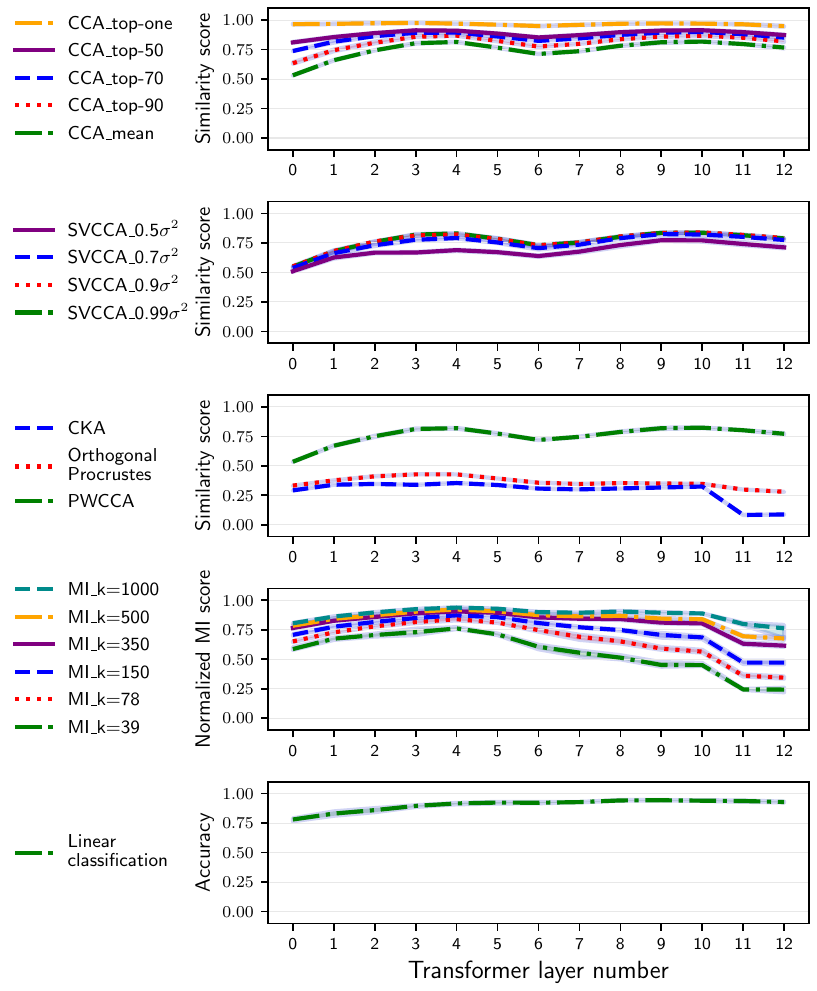}



     


\caption{Different tools comparing \sfm \ representations with phone identity for \datatovec-\baseM.}
\label{fig:res-data2vecb-phone-metric}
\end{figure*}

\begin{figure*}[btp]
\centering
    \includegraphics[width=\textwidth]{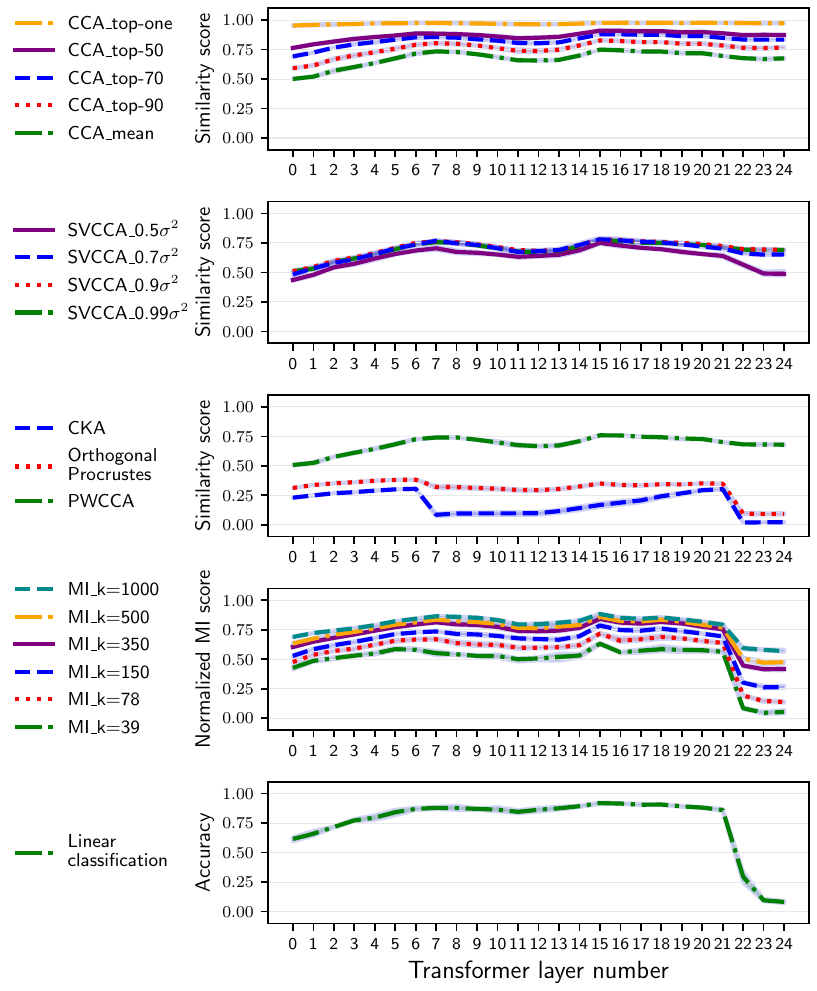}


     


\caption{Different tools comparing \sfm \ representations with phone identity for \wavtovec-\largeM.}
\label{fig:res-wav2vecl-phone-metric}
\end{figure*}

\begin{figure*}[btp]
\centering
    \includegraphics[width=\textwidth]{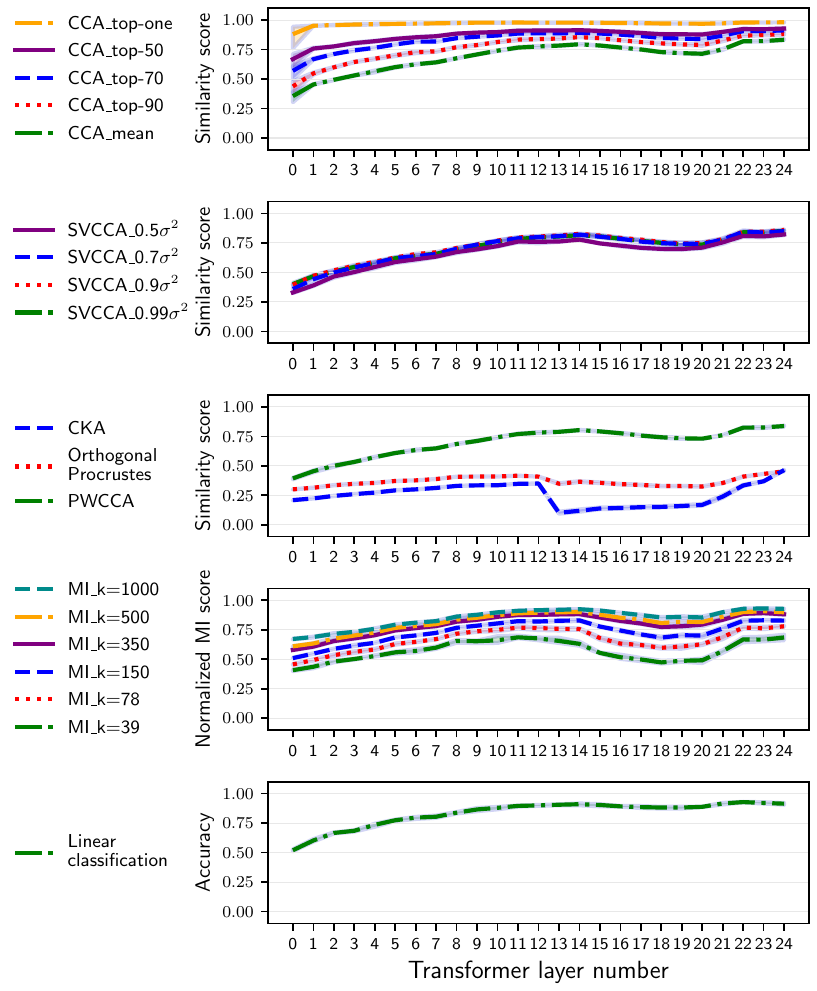}


     


\caption{Different tools comparing \sfm \ representations with phone identity for \hubert-\largeM.}
\label{fig:res-hubertl-phone-metric}
\end{figure*}

\begin{figure*}[btp]
\centering
    \includegraphics[width=\textwidth]{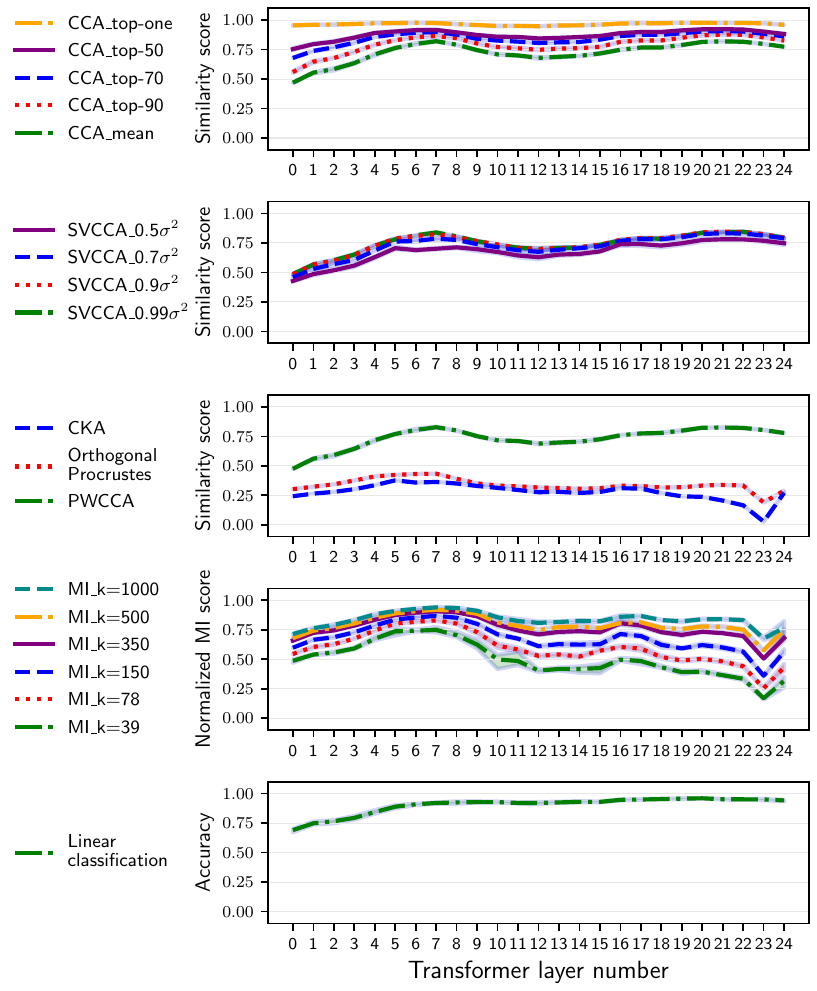}


     


     
\caption{Different tools comparing \sfm \ representations with phone identity for \datatovec-\largeM.}
\label{fig:res-data2vecl-phone-metric}
\end{figure*}

\begin{figure*}[btp]

\centering
    \includegraphics[width=\textwidth]{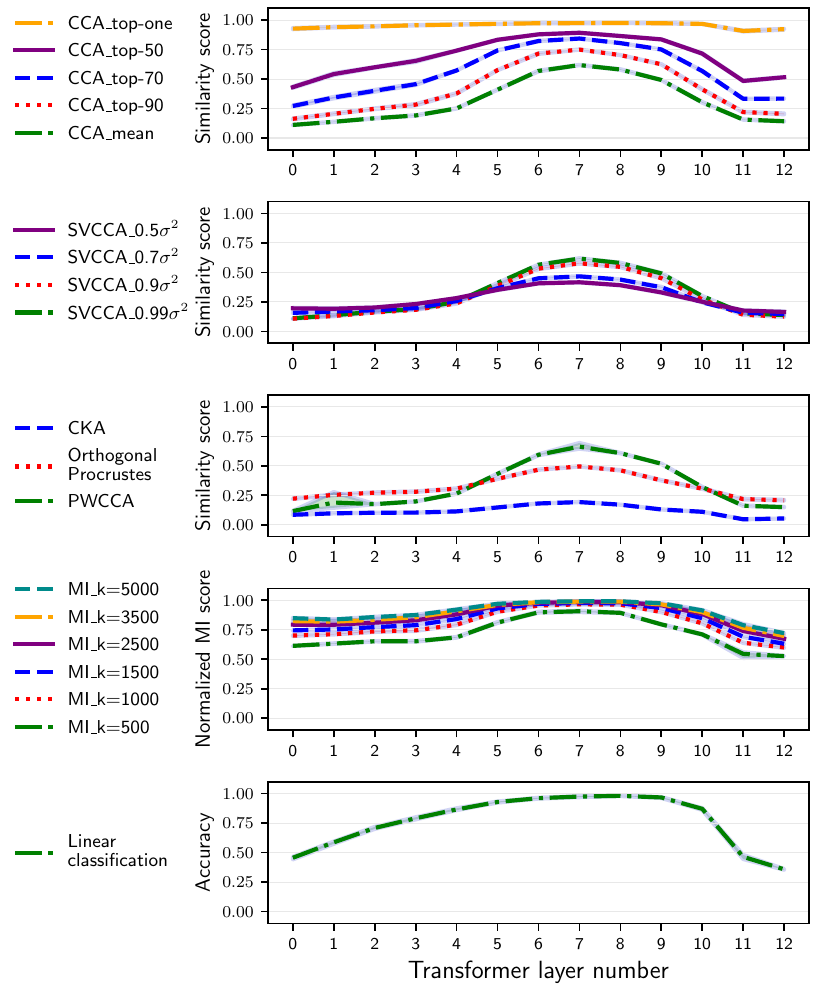}
    


     



\caption{Different tools comparing \sfm \ representations with word identity for \wavtovec-\baseM.}
\label{fig:res-wav2vecb-word-metric}
\end{figure*}

\begin{figure*}[btp]
\centering
    \includegraphics[width=\textwidth]{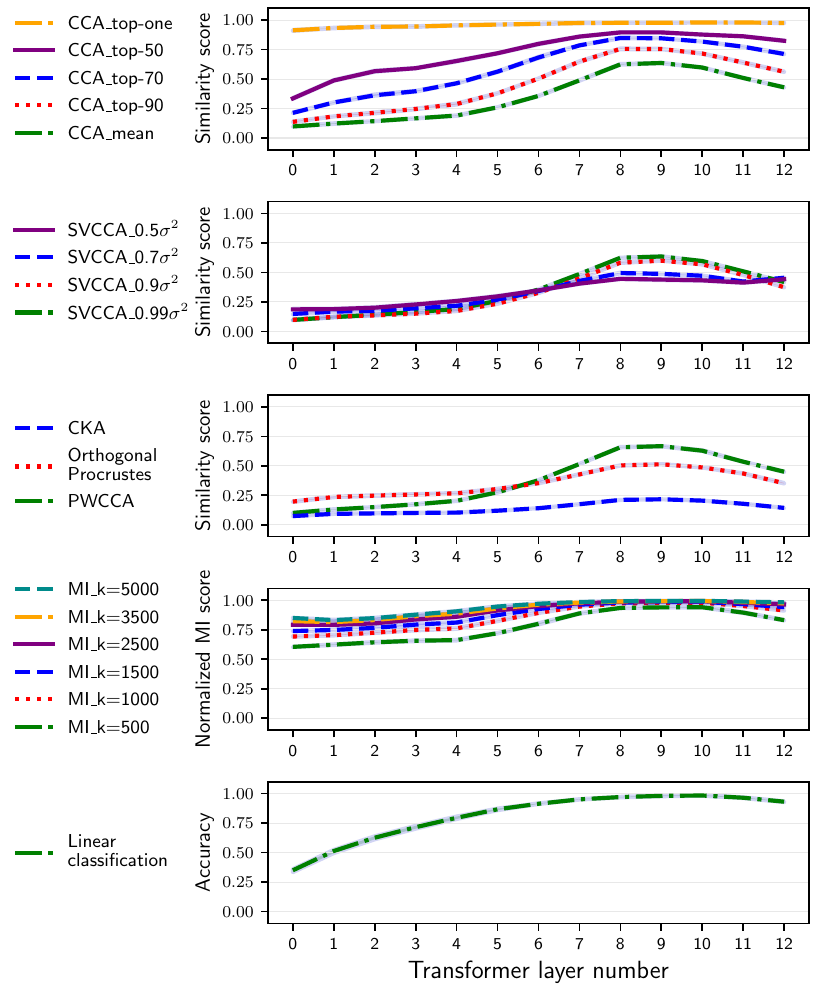}


     


     
\caption{Different tools comparing \sfm \ representations with word identity for \hubert-\baseM.}
\label{fig:res-hubertb-word-metric}
\end{figure*}

\begin{figure*}[btp]

\centering
    \includegraphics[width=\textwidth]{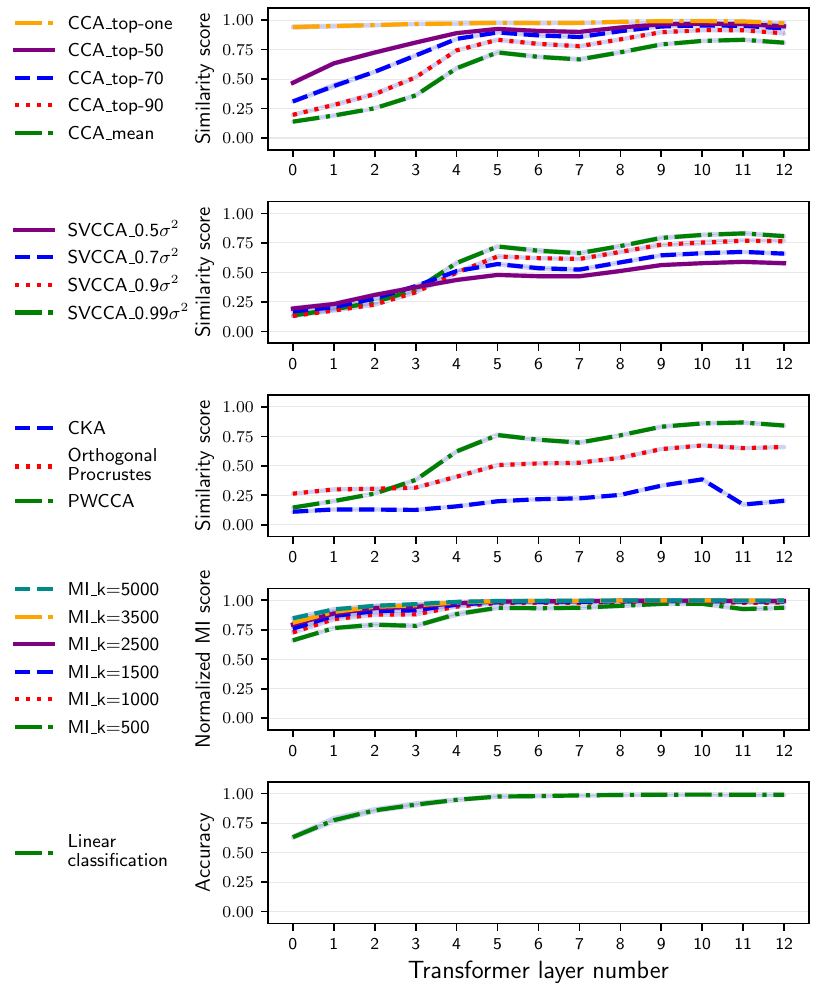}
    


     


\caption{Different tools comparing \sfm \ representations with word identity for \datatovec-\baseM.}
\label{fig:res-data2vecb-word-metric}
\end{figure*}

\begin{figure*}[btp]

\centering
    \includegraphics[width=\textwidth]{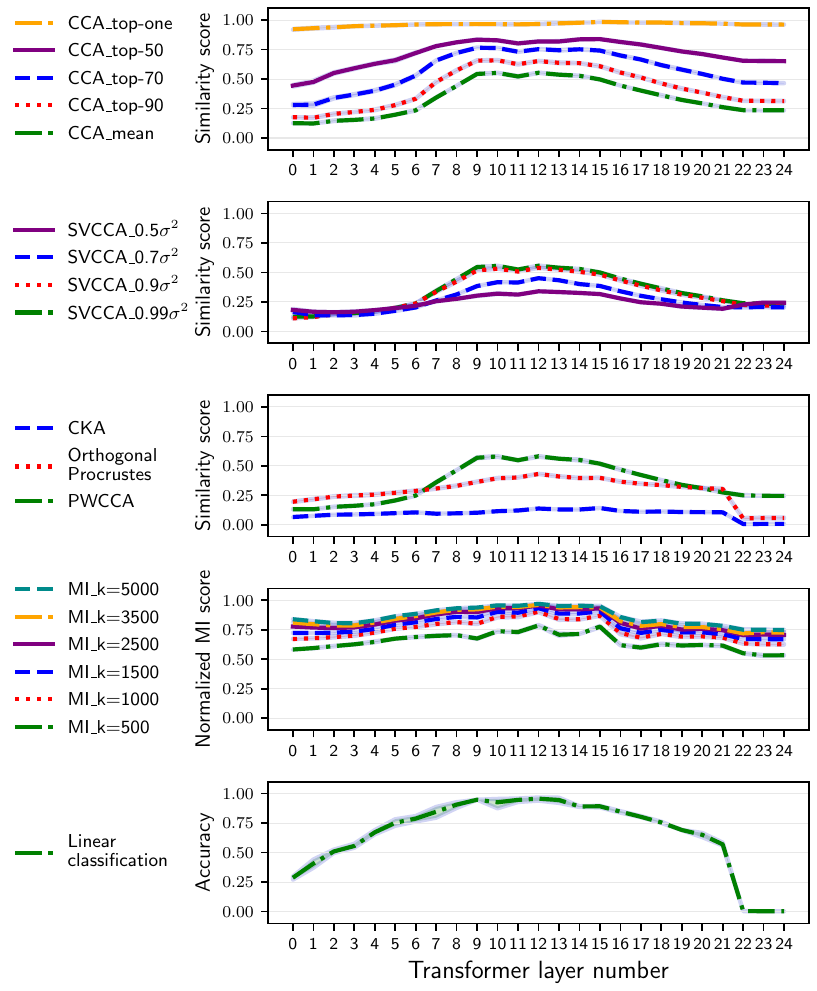}


     


\caption{Different tools comparing \sfm \ representations with word identity for \wavtovec-\largeM.}
\label{fig:res-wav2vecl-word-metric}
\end{figure*}

\begin{figure*}[btp]
\centering
    \includegraphics[width=\textwidth]{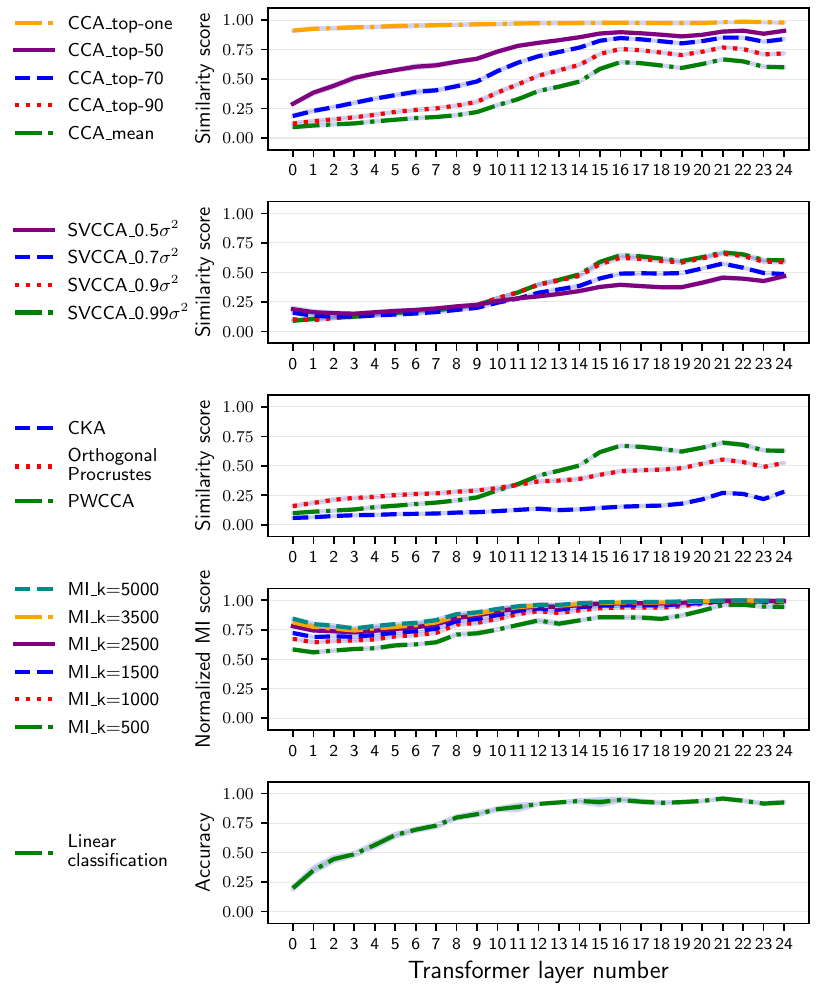}


     


\caption{Different tools comparing \sfm \ representations with word identity for \hubert-\largeM.}
\label{fig:res-hubertl-word-metric}
\end{figure*}

\begin{figure*}[btp]

\centering
    \includegraphics[width=\textwidth]{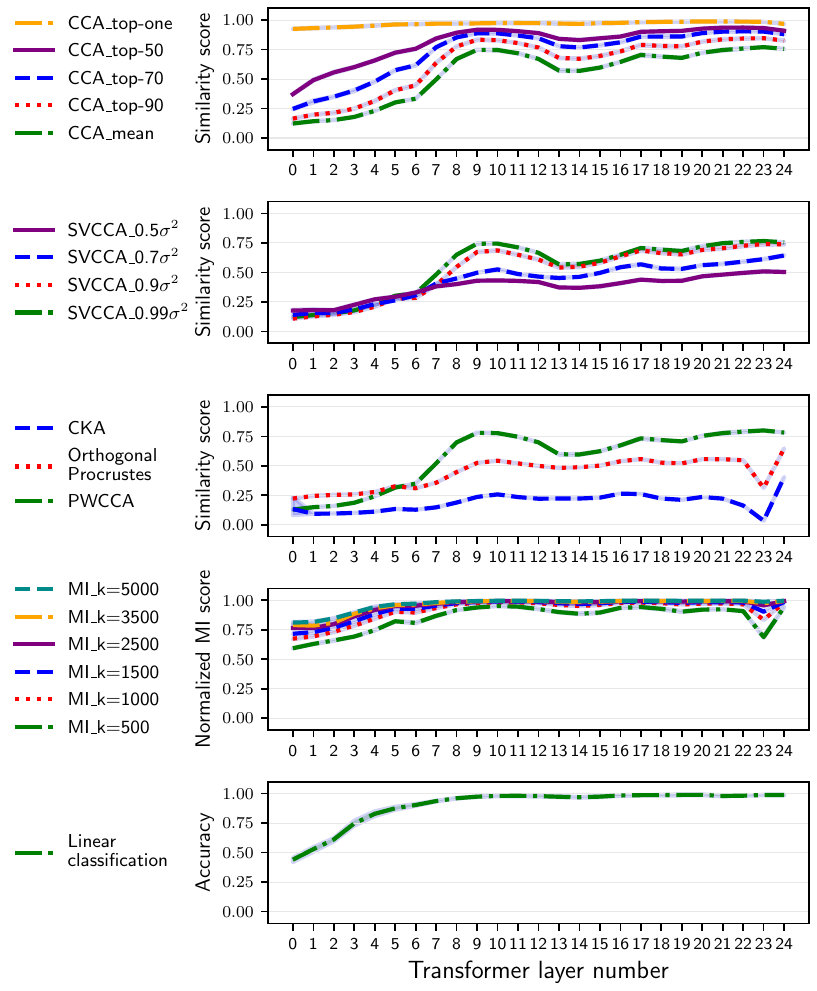}
    


     


\caption{Different tools comparing \sfm \ representations with word identity for \datatovec-\largeM.}
\label{fig:res-data2vecl-word-metric}
\end{figure*}

\begin{figure*}[btp]
\centering
    \includegraphics[width=\textwidth]{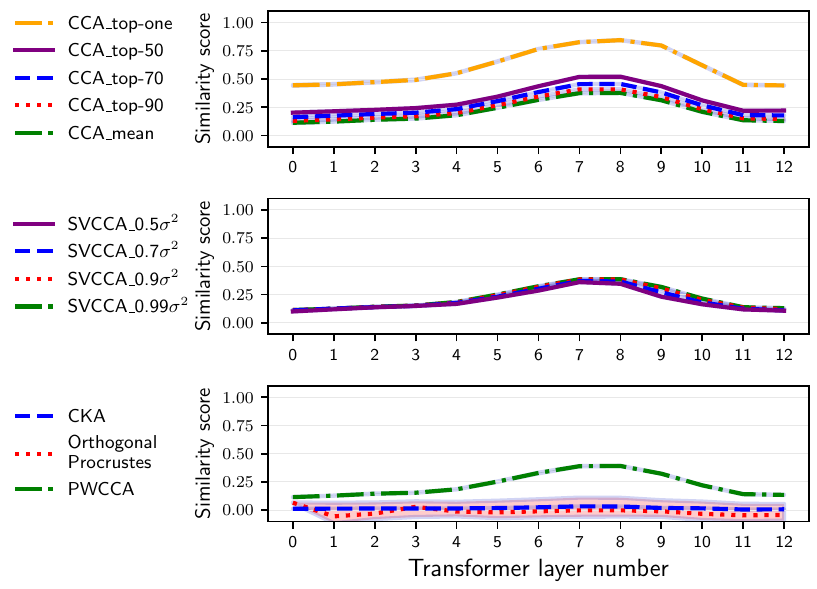}


     

\caption{Different tools comparing \sfm \ representations with semantic attributes for \wavtovec-\baseM.}
\label{fig:res-wav2vecb-sem-metric}
\end{figure*}

\begin{figure*}[btp]
\centering
    \includegraphics[width=\textwidth]{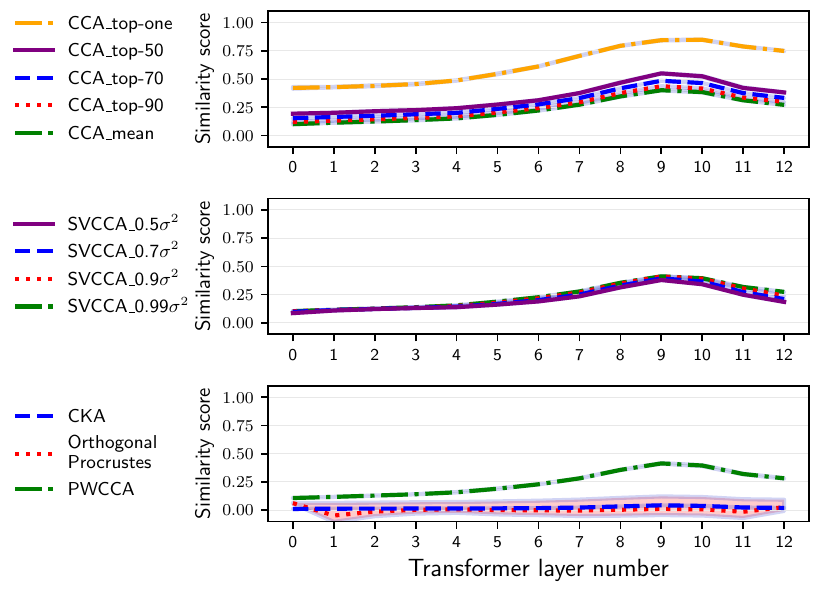}


     

\caption{Different tools comparing \sfm \ representations with semantic attributes for \hubert-\baseM.}
\label{fig:res-hubertb-sem-metric}
\end{figure*}

\begin{figure*}[btp]

\centering
    \includegraphics[width=\textwidth]{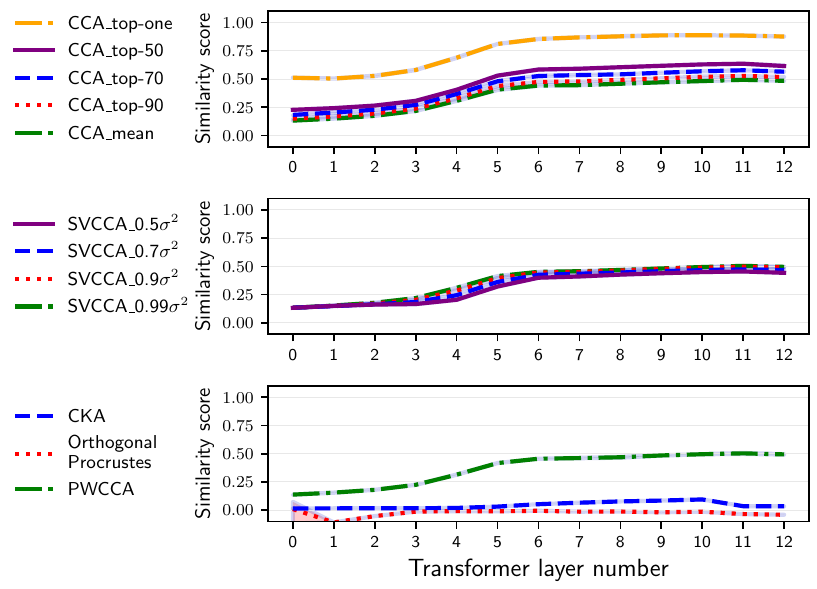}
    


     

\caption{Different tools comparing \sfm \ representations with semantic attributes for \datatovec-\baseM.}
\label{fig:res-data2vecb-sem-metric}
\end{figure*}

\begin{figure*}[btp]
\centering
    \includegraphics[width=\textwidth]{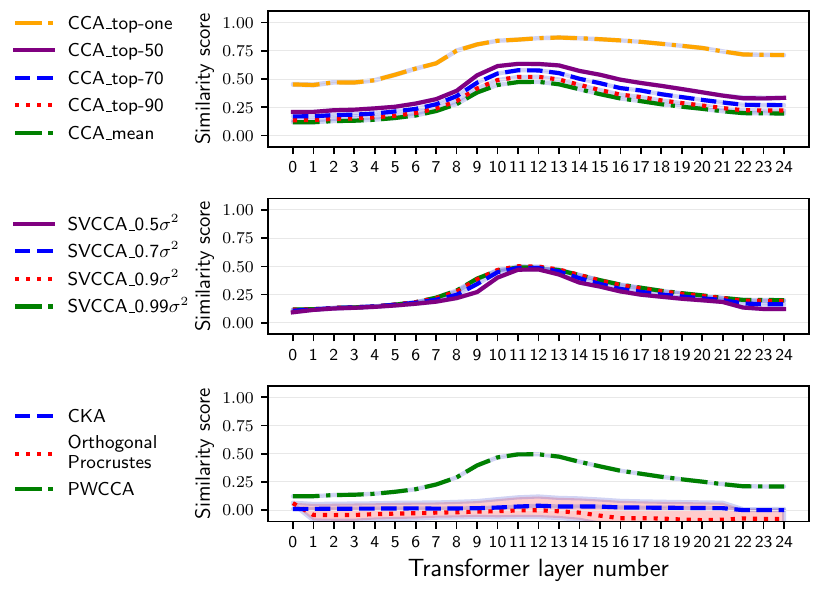}


     

\caption{Different tools comparing \sfm \ representations with semantic attributes for \wavtovec-\largeM.}
\label{fig:res-wav2vecl-sem-metric}
\end{figure*}

\begin{figure*}[btp]
\centering
    \includegraphics[width=\textwidth]{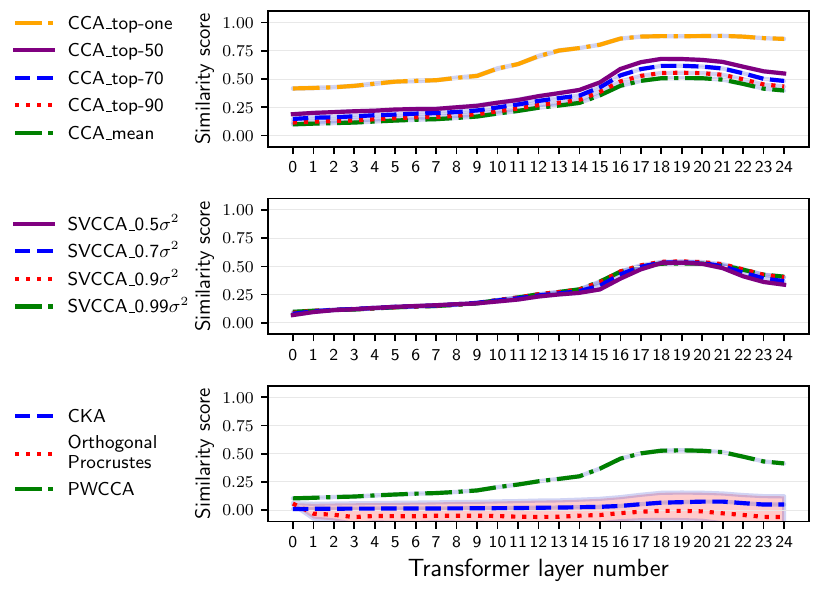}


     

\caption{Different tools comparing \sfm \ representations with semantic attributes for \hubert-\largeM.}
\label{fig:res-hubertl-sem-metric}
\end{figure*}

\begin{figure*}[btp]

\centering
    \includegraphics[width=\textwidth]{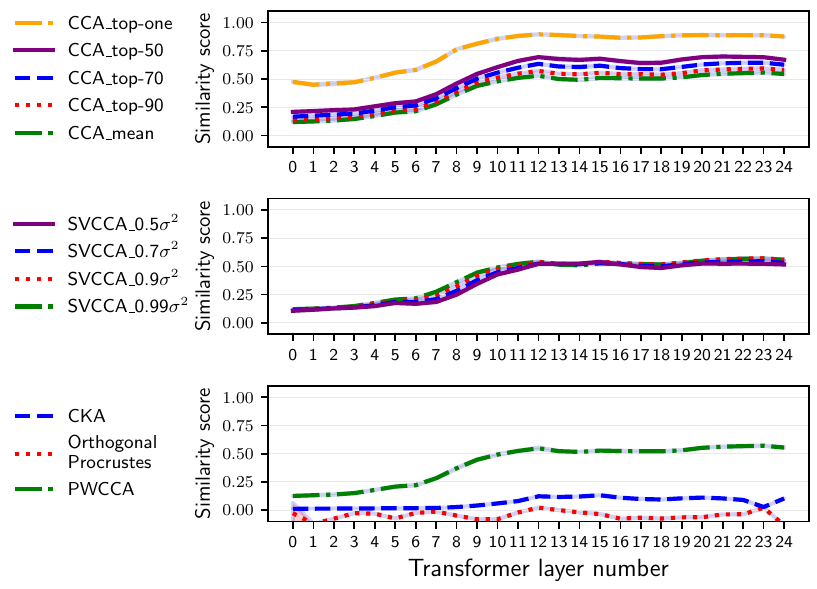}
    


     

\caption{Different tools comparing \sfm \ representations with semantic attributes for \datatovec-\largeM.}
\label{fig:res-data2vecl-sem-metric}
\end{figure*}

Figures~\ref{fig:res-wav2vecb-phone-metric}-\ref{fig:res-data2vecl-sem-metric} below are layer-wise trends for \sfms \ and several analysis metrics discussed in \chap~\ref{ch:compare-tools}. These are additional results for discussion presented in \sect~\ref{sec:res-compare-tools-analysis}.


\section{Comparing different metrics: Correlation scores}
\label{sec:appendix-metric-corr}
\begin{figure}[ht]
\centering

\begin{subfigure}[b]{0.49\textwidth}
    \centering
    \includegraphics[width=\textwidth, trim=0 0 30 0, clip]{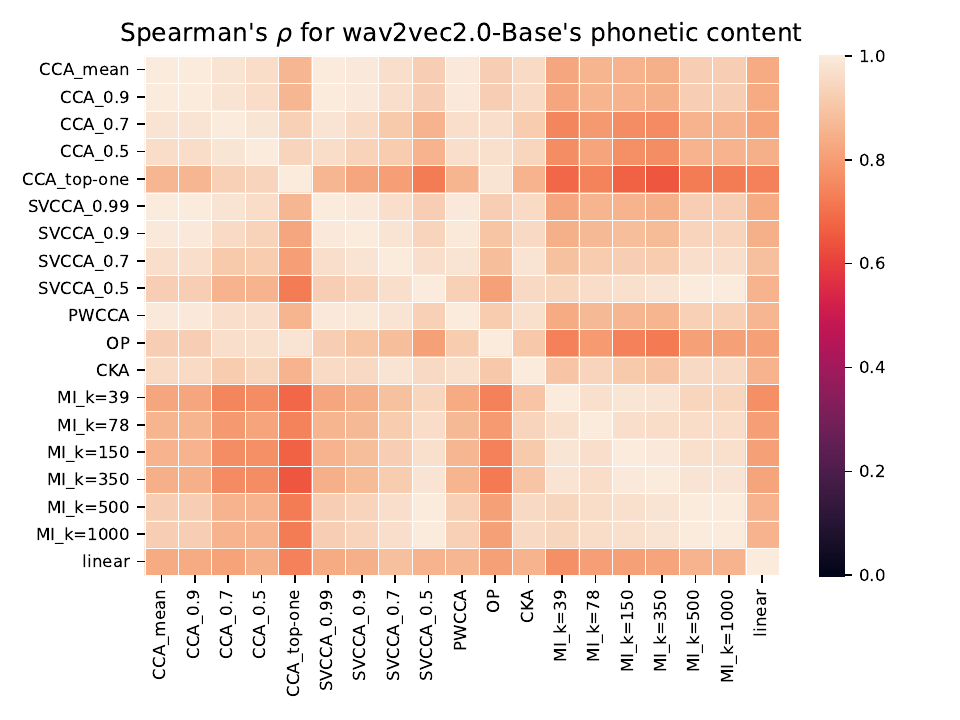}
    \label{fig:wav2vecb-phone-spearman}
\end{subfigure}
\hfill 
\begin{subfigure}[b]{0.49\textwidth}
    \centering
    \includegraphics[width=\textwidth, trim=0 0 30 0, clip]{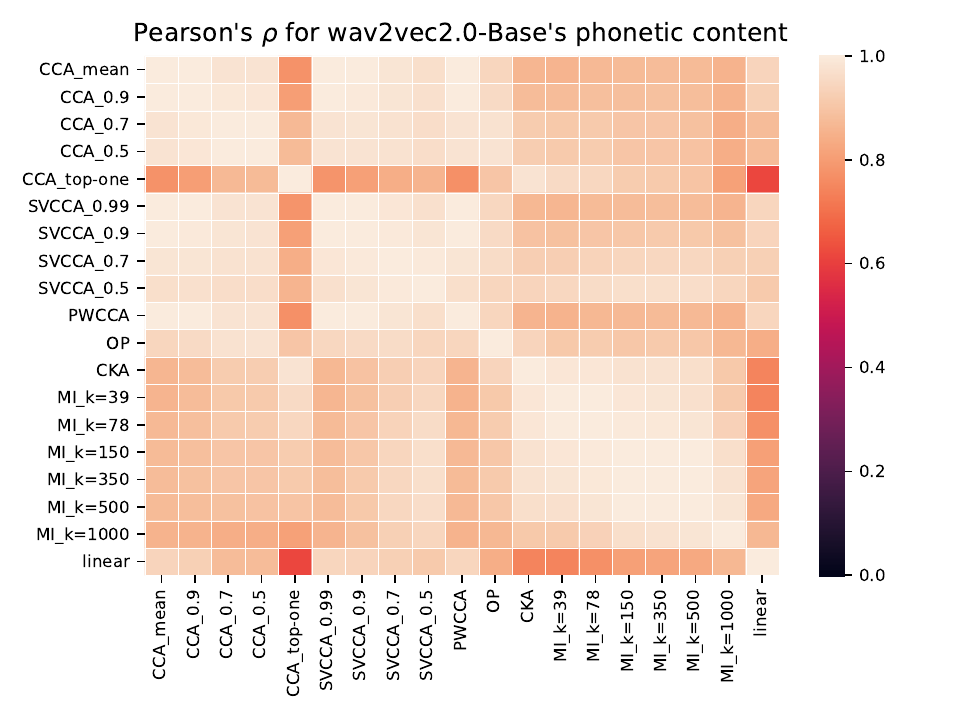}
    \label{fig:wav2vecb-phone-pearson}
\end{subfigure}

\begin{subfigure}[b]{0.49\textwidth}
    \centering
    \includegraphics[width=\textwidth, trim=0 0 30 0, clip]{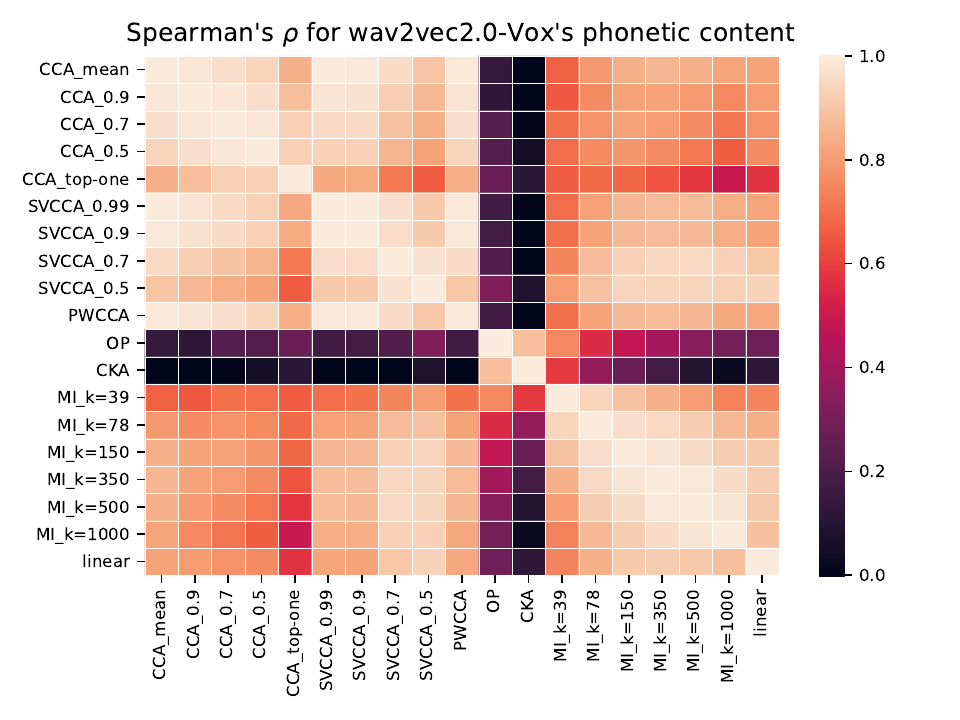}
    \label{fig:wav2vecl-phone-spearman}
\end{subfigure}
\hfill 
\begin{subfigure}[b]{0.49\textwidth}
    \centering
    \includegraphics[width=\textwidth, trim=0 0 30 0, clip]{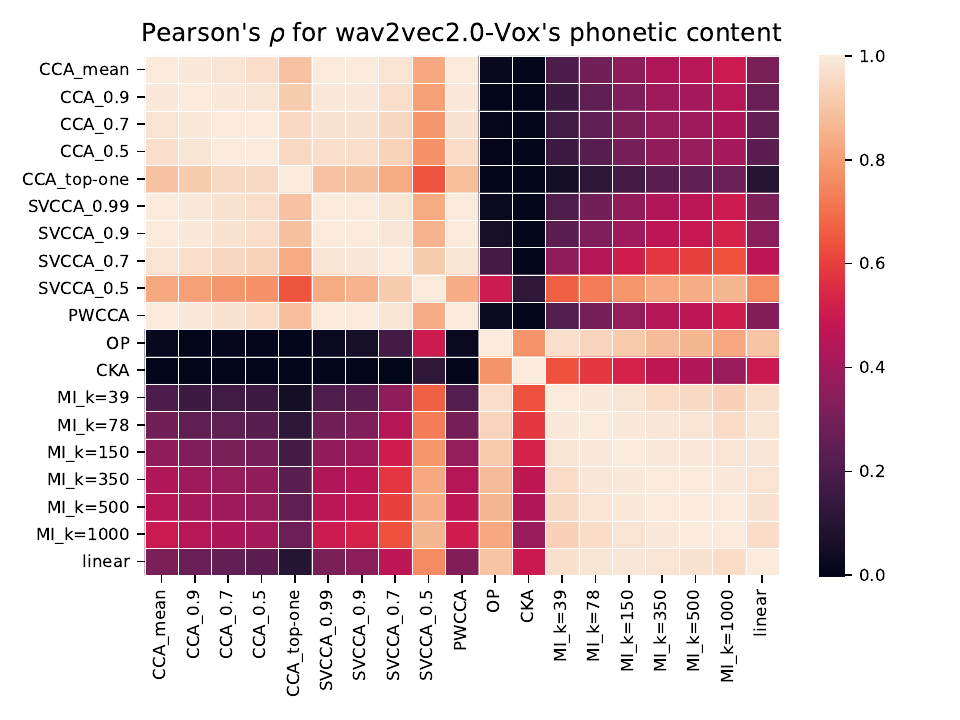}
    \label{fig:wav2vecl-phone-pearson}
\end{subfigure}

\caption{Correlation between different analysis tools for phonetic content in \wavtovec \ models.}
\label{fig:metric-corr-w2v2-phone}
\end{figure}

\begin{figure}[ht]
\centering

\begin{subfigure}[b]{0.49\textwidth}
    \centering
    \includegraphics[width=\textwidth, trim=0 0 30 0, clip]{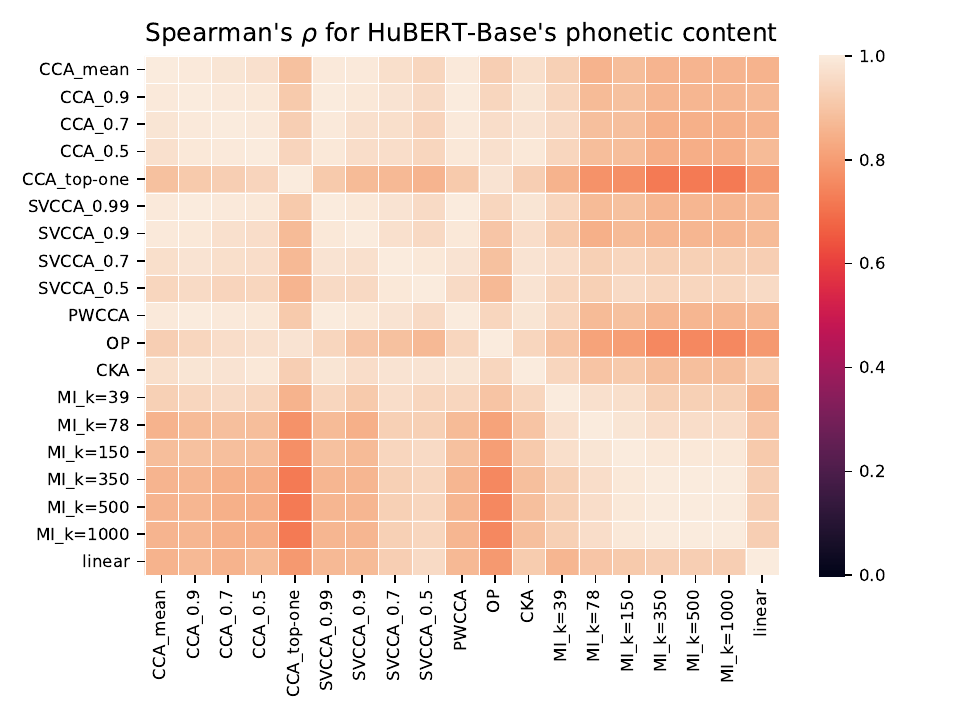}
    \label{fig:hubertb-phone-spearman}
\end{subfigure}
\hfill 
\begin{subfigure}[b]{0.49\textwidth}
    \centering
    \includegraphics[width=\textwidth, trim=0 0 30 0, clip]{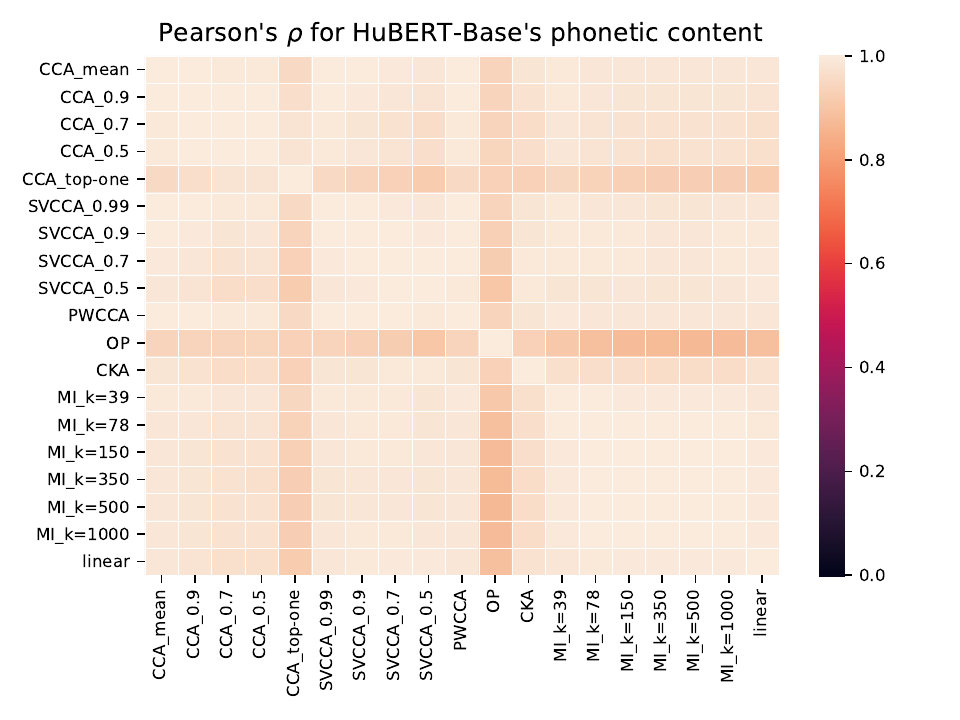}
    \label{fig:hubertb-phone-pearson}
\end{subfigure}

\begin{subfigure}[b]{0.49\textwidth}
    \centering
    \includegraphics[width=\textwidth, trim=0 0 30 0, clip]{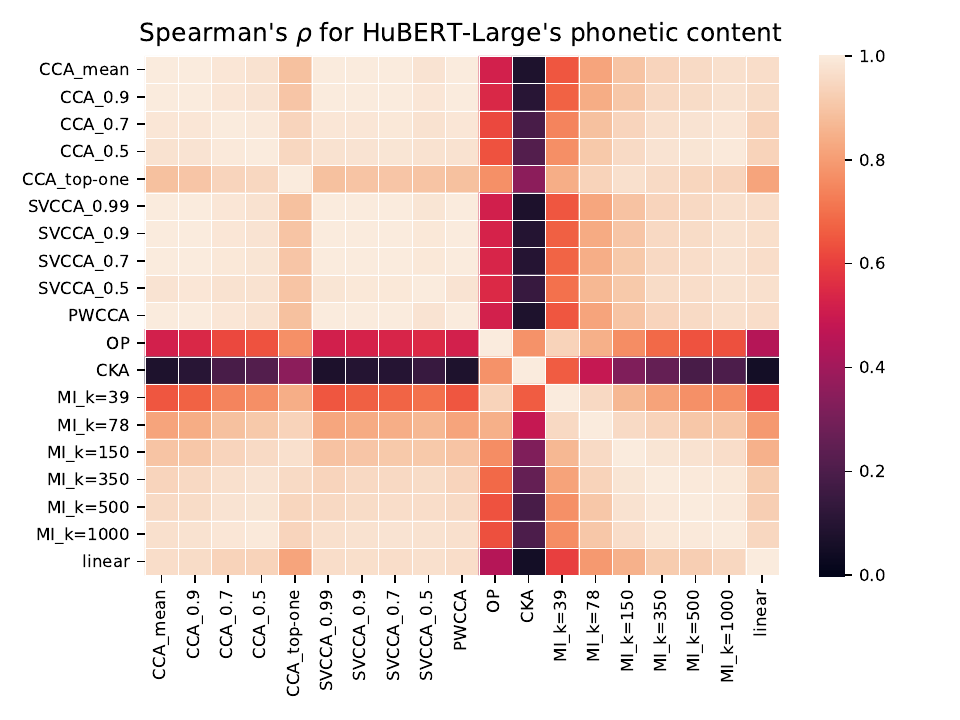}
    \label{fig:hubertl-phone-spearman}
\end{subfigure}
\hfill 
\begin{subfigure}[b]{0.49\textwidth}
    \centering
    \includegraphics[width=\textwidth, trim=0 0 30 0, clip]{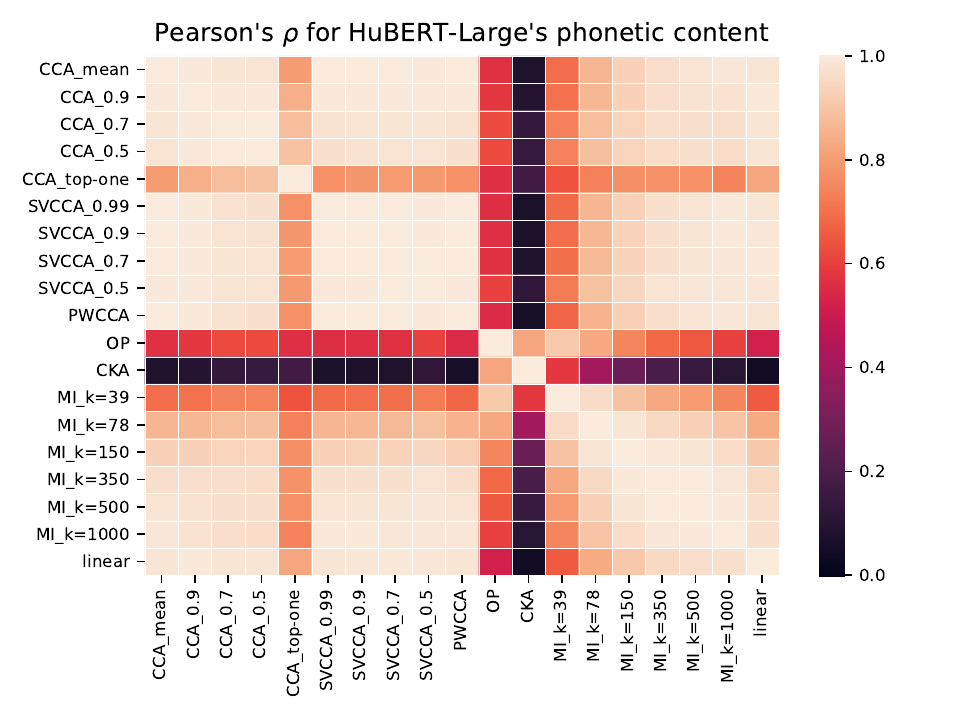}
    \label{fig:hubertl-phone-pearson}
\end{subfigure}

\caption{Correlation between different analysis tools for phonetic content in \hubert \ models.}
\label{fig:metric-corr-hubert-phone}
\end{figure}

\begin{figure}[ht]
\centering

\begin{subfigure}[b]{0.49\textwidth}
    \centering
    \includegraphics[width=\textwidth, trim=0 0 30 0, clip]{images/metric_analysis/data2vec_small_phone_spearman.pdf}
    \label{fig:data2vecb-phone-spearman}
\end{subfigure}
\hfill 
\begin{subfigure}[b]{0.49\textwidth}
    \centering
    \includegraphics[width=\textwidth, trim=0 0 30 0, clip]{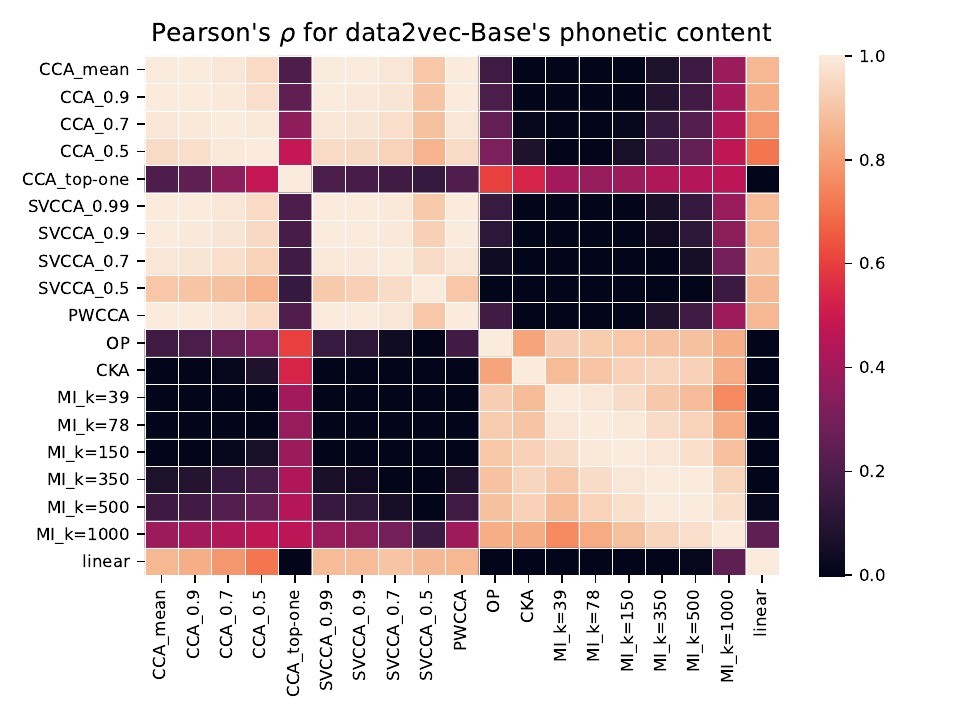}
    \label{fig:data2vecb-phone-pearson}
\end{subfigure}

\begin{subfigure}[b]{0.49\textwidth}
    \centering
    \includegraphics[width=\textwidth, trim=0 0 30 0, clip]{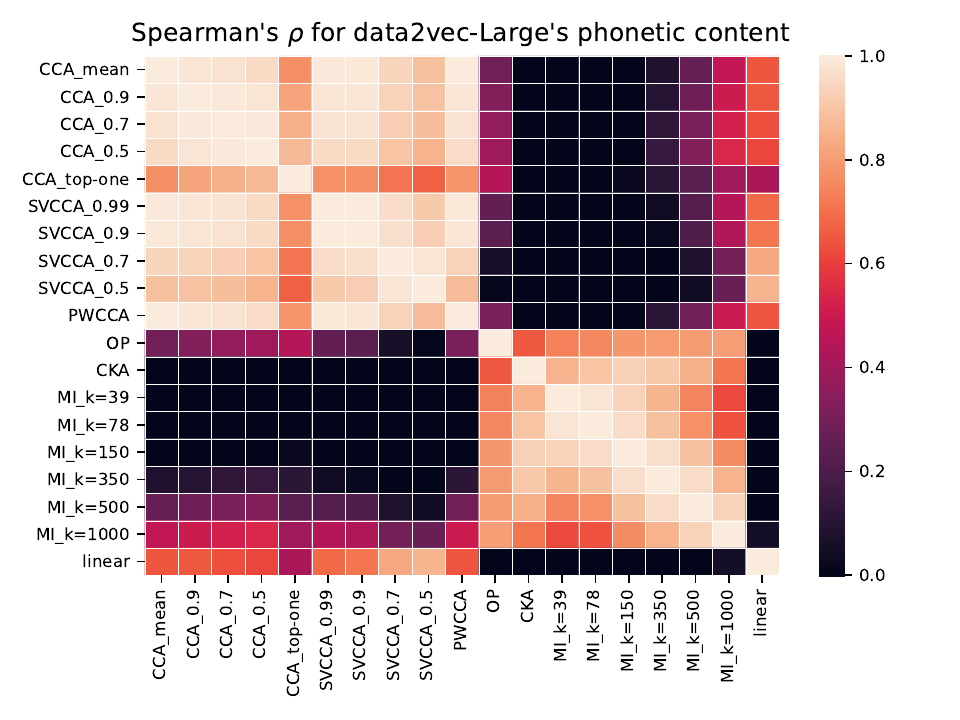}
    \label{fig:data2vecl-phone-spearman}
\end{subfigure}
\hfill 
\begin{subfigure}[b]{0.49\textwidth}
    \centering
    \includegraphics[width=\textwidth, trim=0 0 30 0, clip]{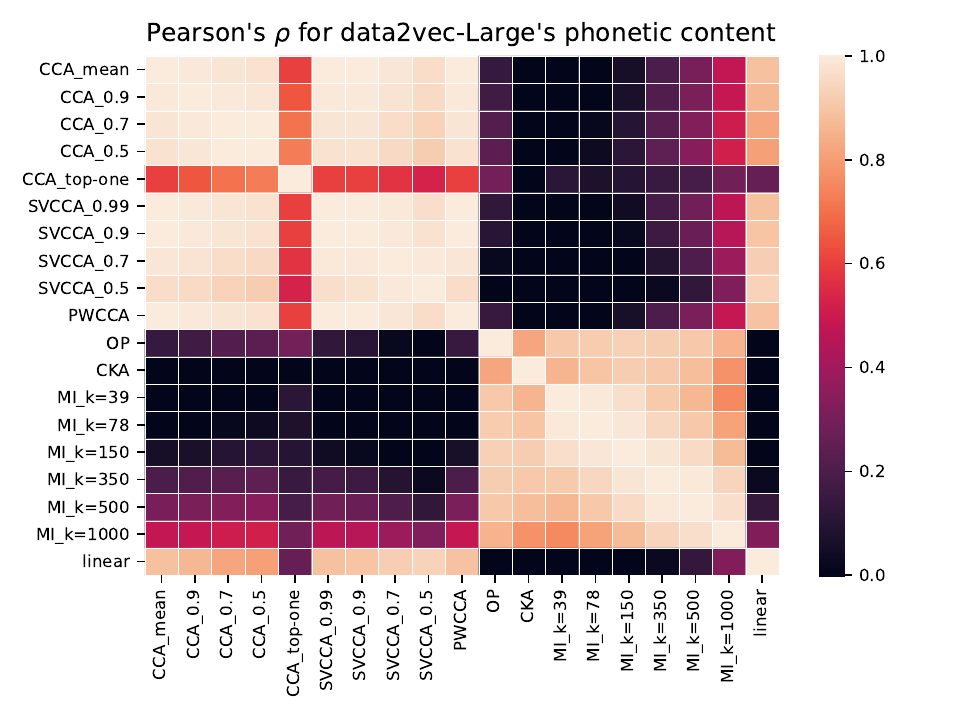}
    \label{fig:data2vecl-phone-pearson}
\end{subfigure}

\caption{Correlation between different analysis tools for phonetic content in \datatovec \ models.}
\label{fig:metric-corr-data2vec-phone}
\end{figure}

\begin{figure}[ht]
\centering

\begin{subfigure}[b]{0.49\textwidth}
    \centering
    \includegraphics[width=\textwidth, trim=0 0 30 0, clip]{images/metric_analysis/wav2vec_small_word_spearman.pdf}
    \label{fig:wav2vecb-word-spearman}
\end{subfigure}
\hfill 
\begin{subfigure}[b]{0.49\textwidth}
    \centering
    \includegraphics[width=\textwidth, trim=0 0 30 0, clip]{images/metric_analysis/wav2vec_small_word_pearson.pdf}
    \label{fig:wav2vecb-word-pearson}
\end{subfigure}

\begin{subfigure}[b]{0.49\textwidth}
    \centering
    \includegraphics[width=\textwidth, trim=0 0 30 0, clip]{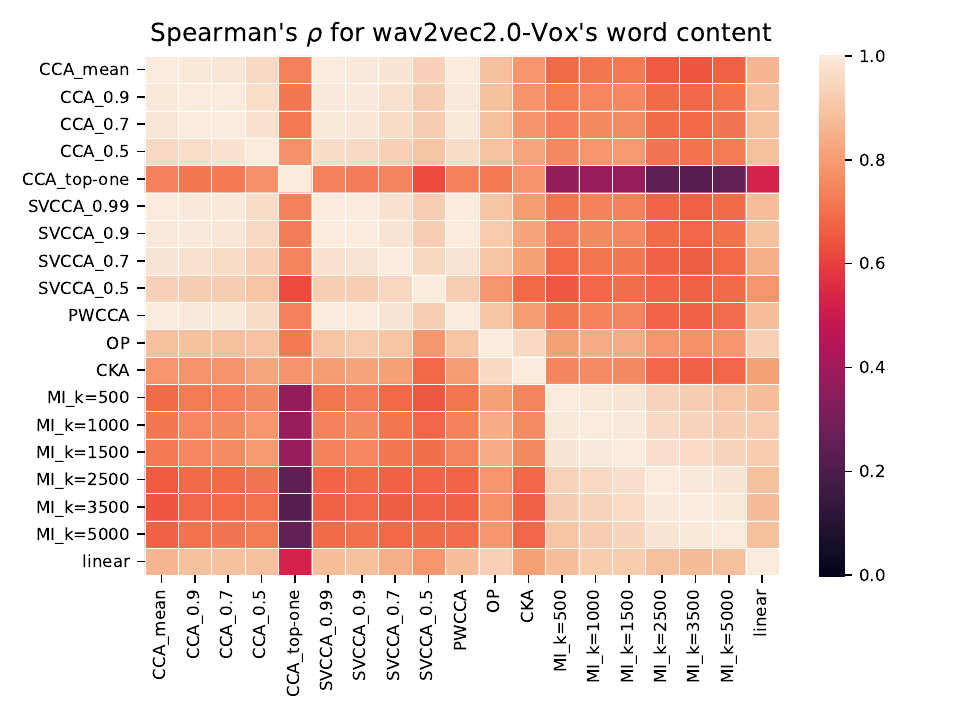}
    \label{fig:wav2vecl-word-spearman}
\end{subfigure}
\hfill 
\begin{subfigure}[b]{0.49\textwidth}
    \centering
    \includegraphics[width=\textwidth, trim=0 0 30 0, clip]{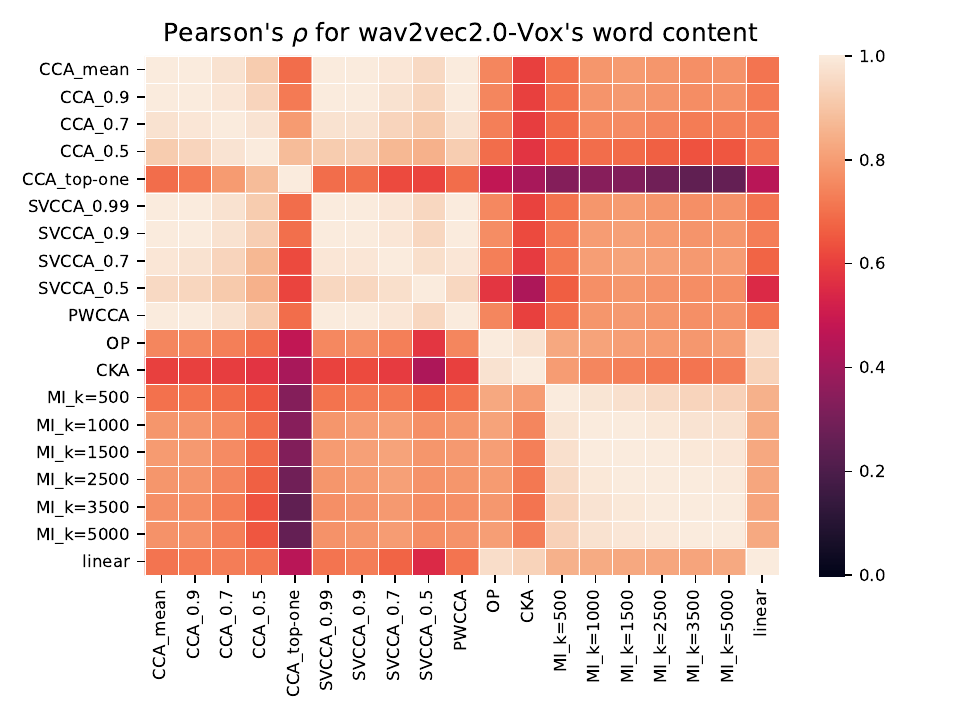}
    \label{fig:wav2vecl-word-pearson}
\end{subfigure}

\caption{Correlation between different analysis tools for word-level content in \wavtovec \ models.}
\label{fig:metric-corr-w2v2-word}
\end{figure}

\begin{figure}[ht]
\centering

\begin{subfigure}[b]{0.49\textwidth}
    \centering
    \includegraphics[width=\textwidth, trim=0 0 30 0, clip]{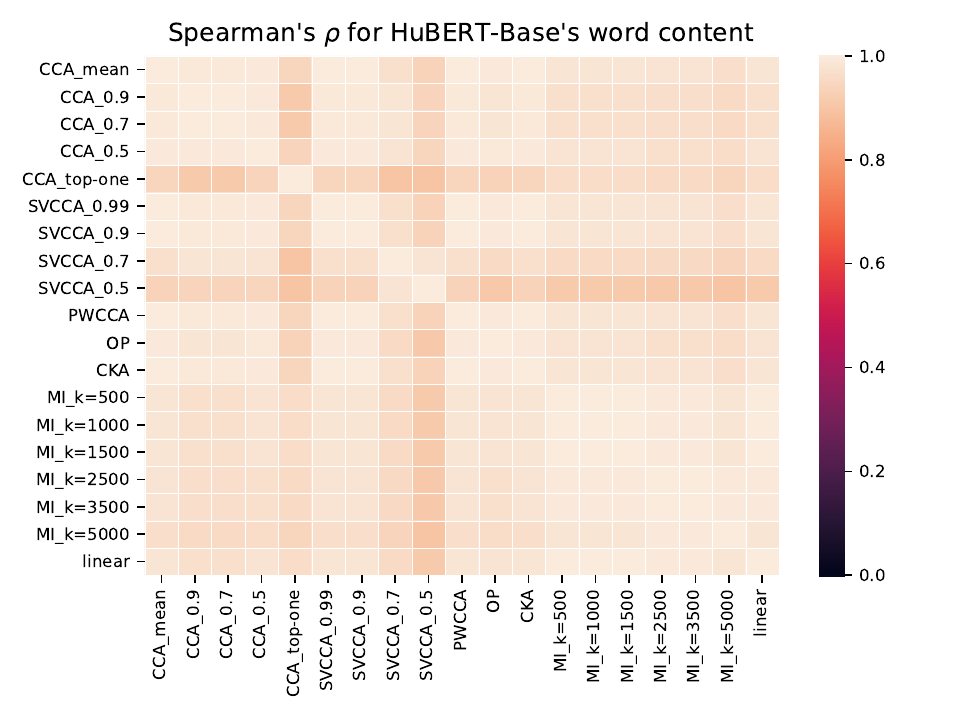}
    \label{fig:hubertb-word-spearman}
\end{subfigure}
\hfill 
\begin{subfigure}[b]{0.49\textwidth}
    \centering
    \includegraphics[width=\textwidth, trim=0 0 30 0, clip]{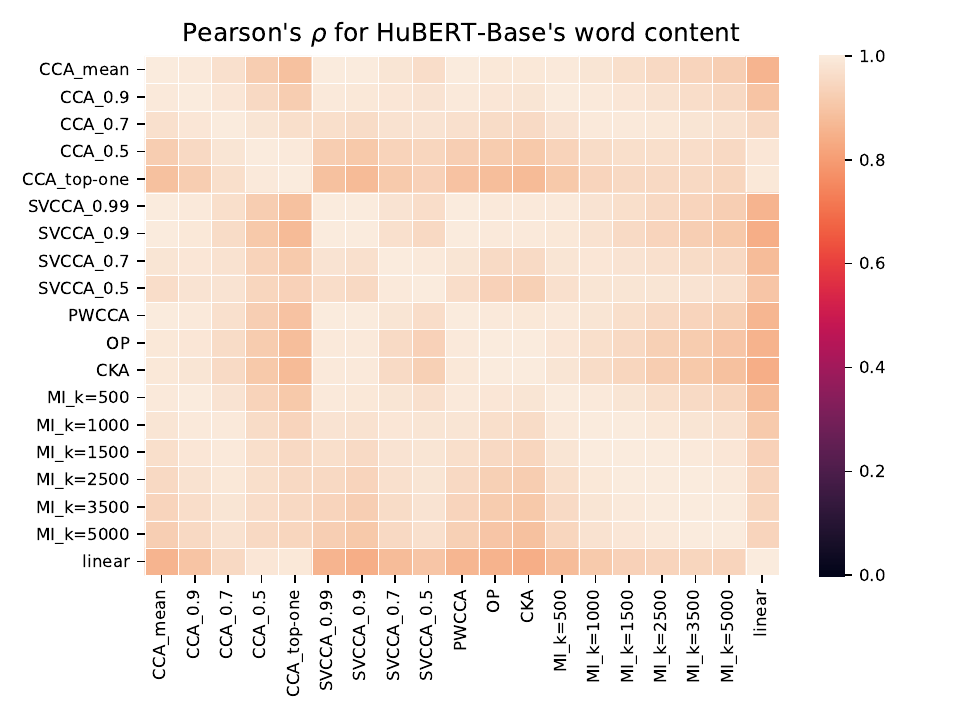}
    \label{fig:hubertb-word-pearson}
\end{subfigure}

\begin{subfigure}[b]{0.49\textwidth}
    \centering
    \includegraphics[width=\textwidth, trim=0 0 30 0, clip]{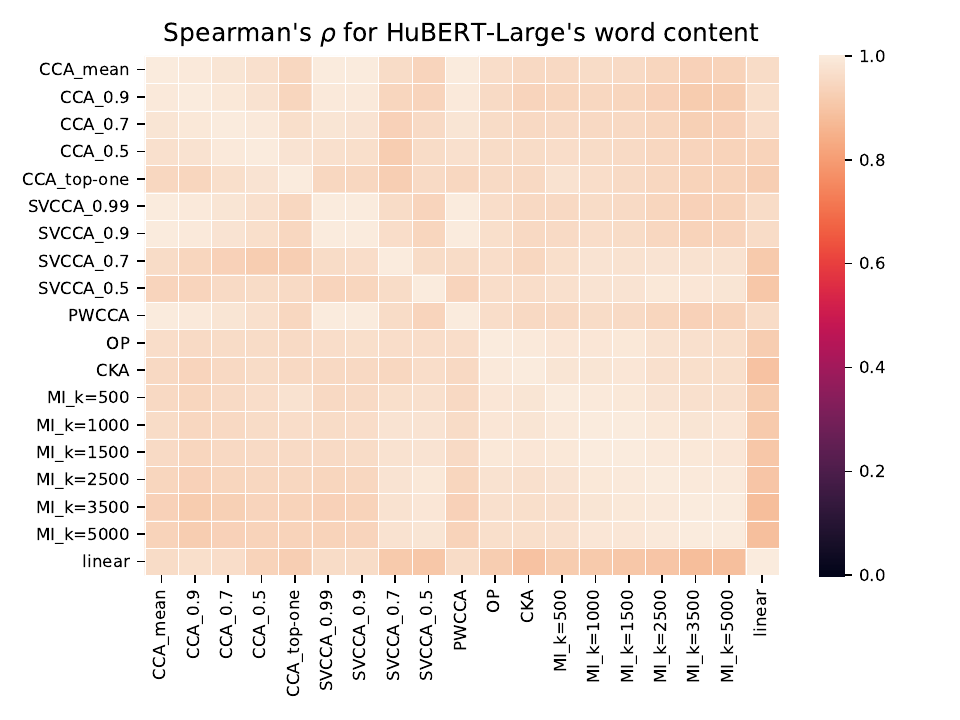}
    \label{fig:hubertl-word-spearman}
\end{subfigure}
\hfill 
\begin{subfigure}[b]{0.49\textwidth}
    \centering
    \includegraphics[width=\textwidth, trim=0 0 30 0, clip]{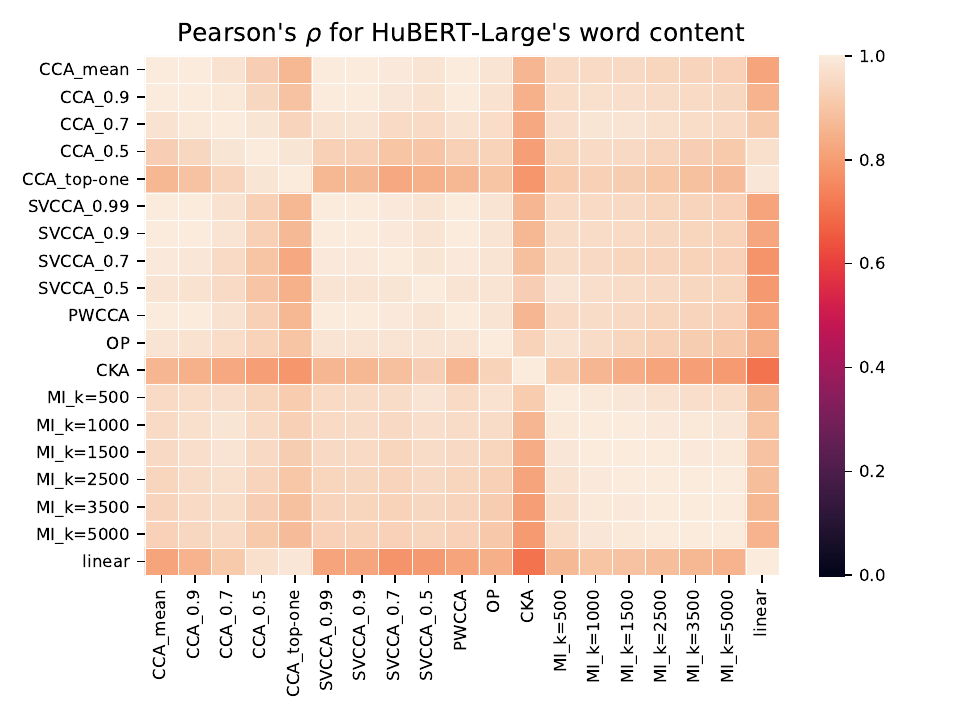}
    \label{fig:hubertl-word-pearson}
\end{subfigure}

\caption{Correlation between different analysis tools for word-level content in \hubert \ models.}
\label{fig:metric-corr-hubert-word}
\end{figure}

\begin{figure}[ht]
\centering

\begin{subfigure}[b]{0.49\textwidth}
    \centering
    \includegraphics[width=\textwidth, trim=0 0 30 0, clip]{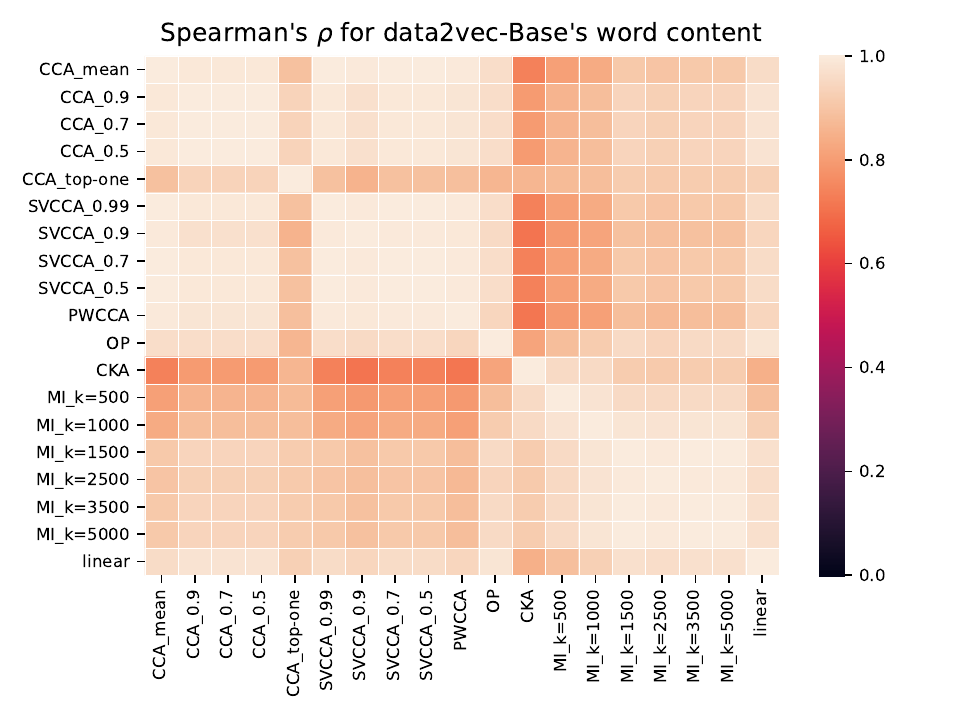}
    \label{fig:data2vecb-word-spearman}
\end{subfigure}
\hfill 
\begin{subfigure}[b]{0.49\textwidth}
    \centering
    \includegraphics[width=\textwidth, trim=0 0 30 0, clip]{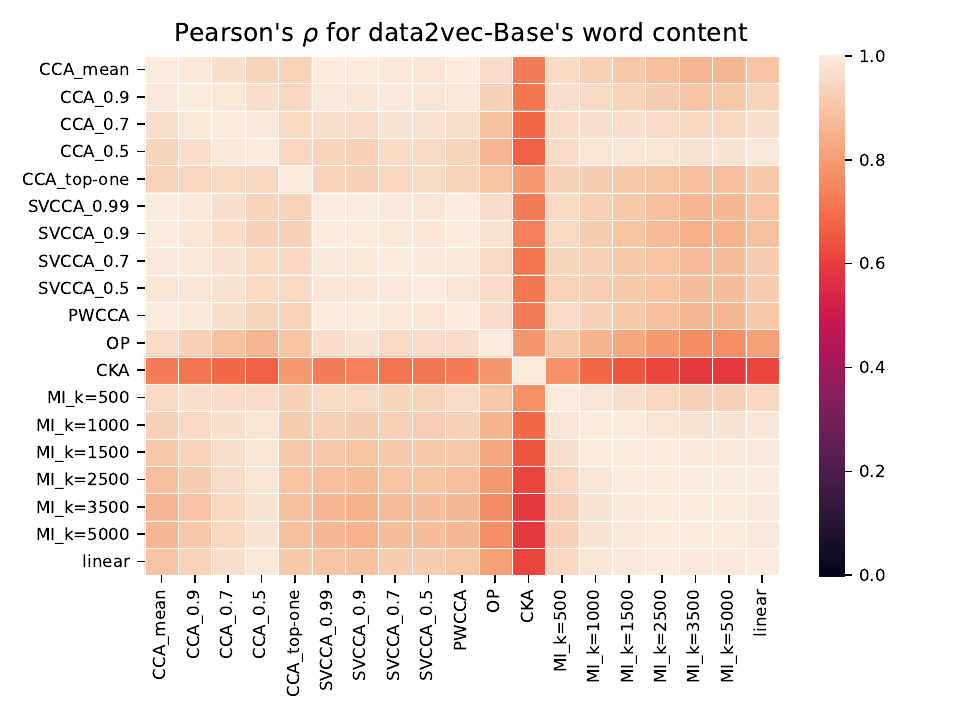}
    \label{fig:data2vecb-word-pearson}
\end{subfigure}

\begin{subfigure}[b]{0.49\textwidth}
    \centering
    \includegraphics[width=\textwidth, trim=0 0 30 0, clip]{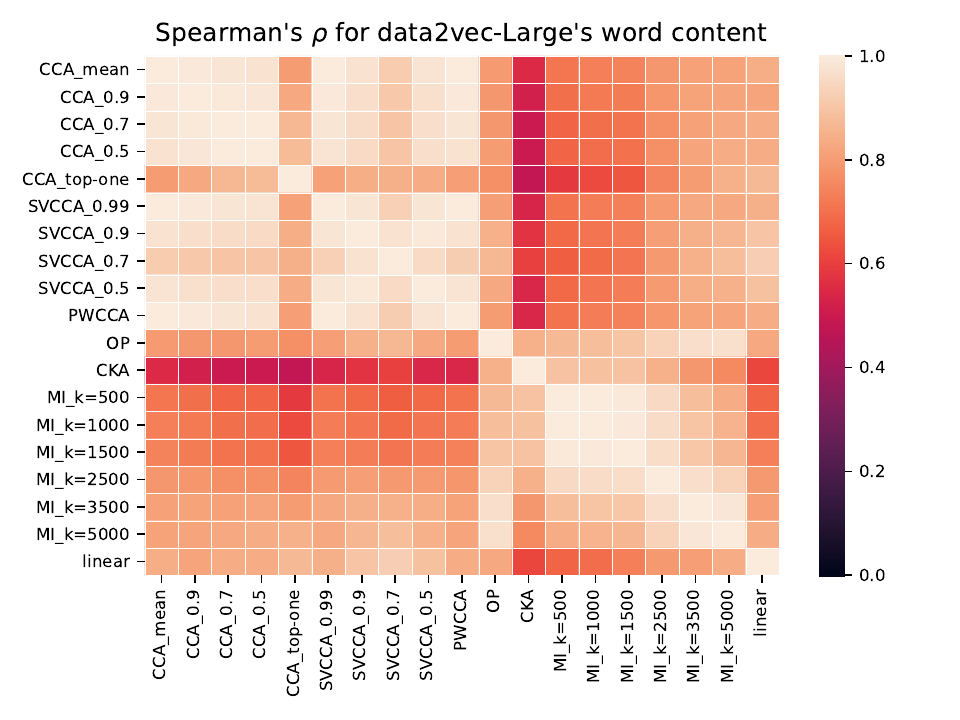}
    \label{fig:data2vecl-word-spearman}
\end{subfigure}
\hfill 
\begin{subfigure}[b]{0.49\textwidth}
    \centering
    \includegraphics[width=\textwidth, trim=0 0 30 0, clip]{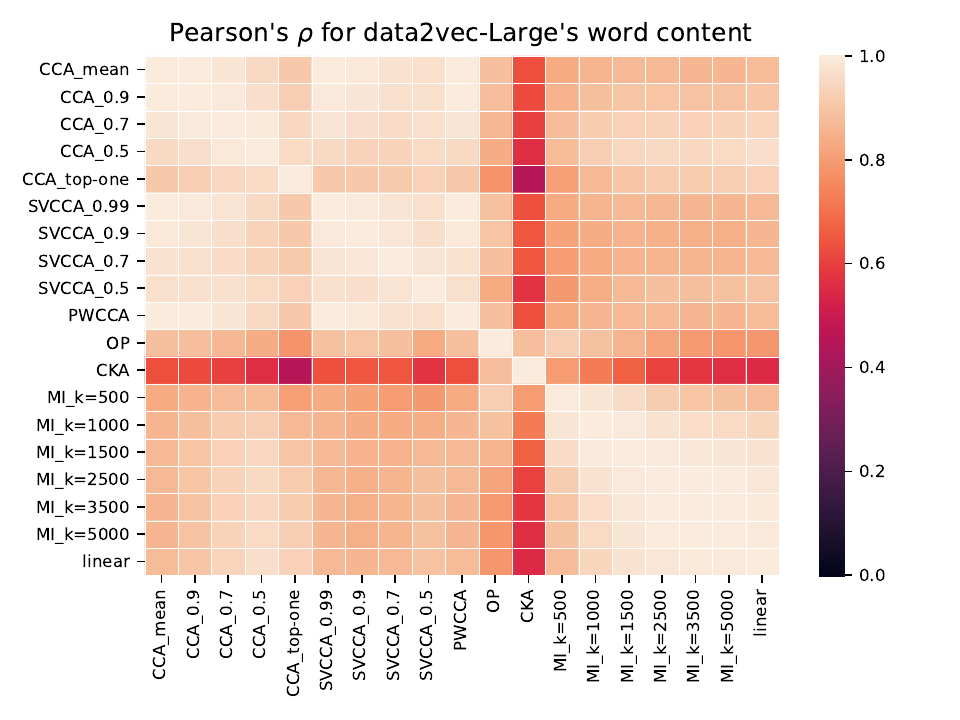}
    \label{fig:data2vecl-word-pearson}
\end{subfigure}

\caption{Correlation between different analysis tools for word-level content in \datatovec \ models.}
\label{fig:metric-corr-data2vec-word}
\end{figure}

\begin{figure}[ht]
\centering

\begin{subfigure}[b]{0.49\textwidth}
    \centering
    \includegraphics[width=\textwidth, trim=0 0 30 0, clip]{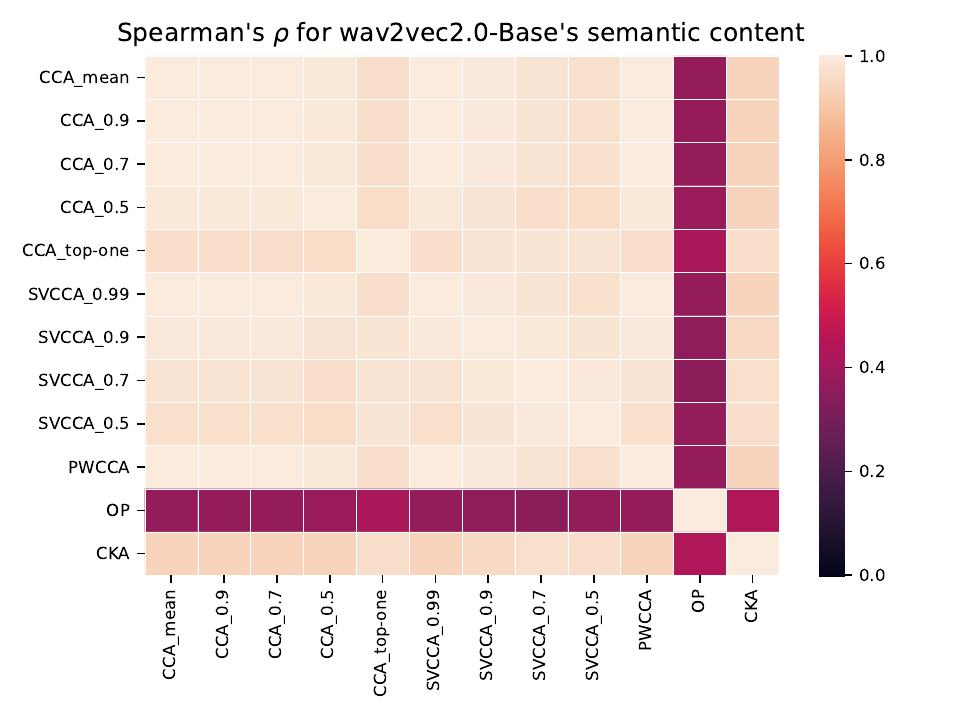}
    \label{fig:wav2vecb-sem-spearman}
\end{subfigure}
\hfill 
\begin{subfigure}[b]{0.49\textwidth}
    \centering
    \includegraphics[width=\textwidth, trim=0 0 30 0, clip]{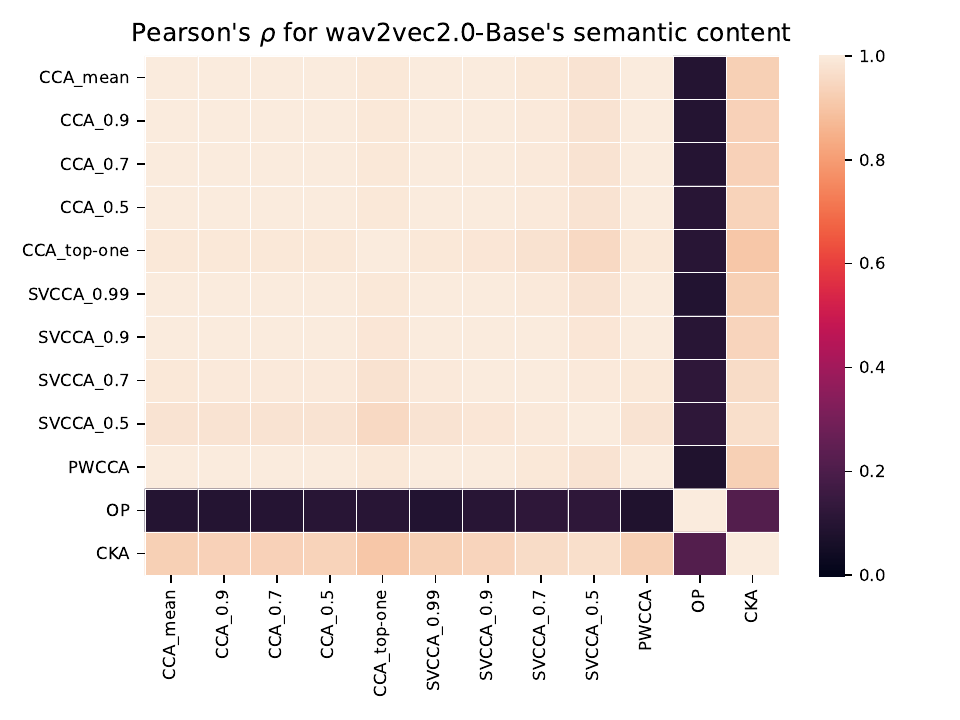}
    \label{fig:wav2vecb-sem-pearson}
\end{subfigure}

\begin{subfigure}[b]{0.49\textwidth}
    \centering
    \includegraphics[width=\textwidth, trim=0 0 30 0, clip]{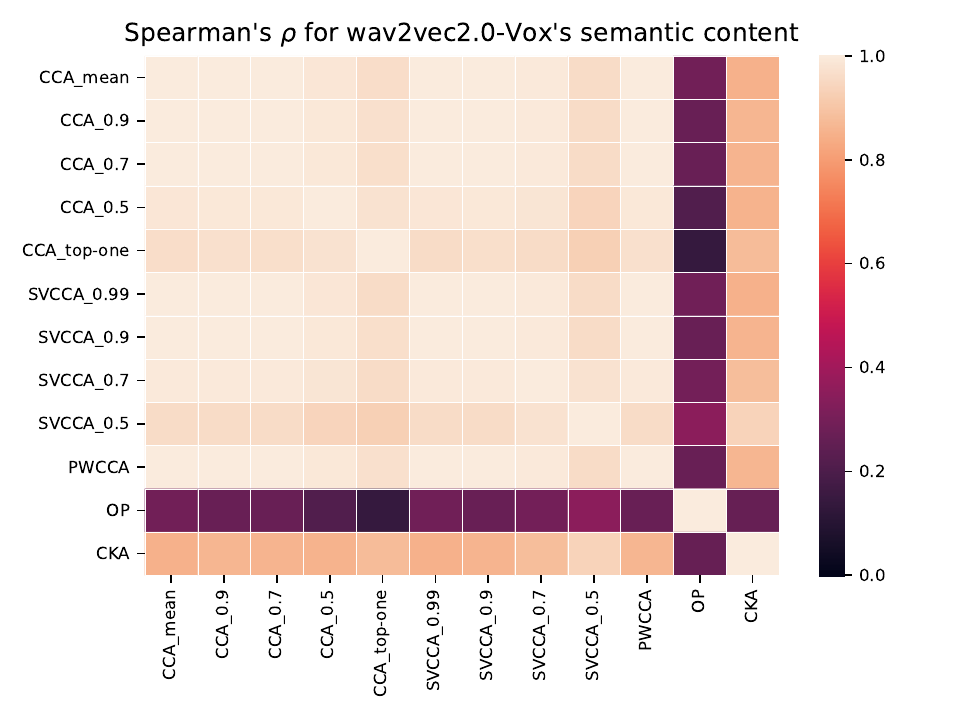}
    \label{fig:wav2vecl-sem-spearman}
\end{subfigure}
\hfill 
\begin{subfigure}[b]{0.49\textwidth}
    \centering
    \includegraphics[width=\textwidth, trim=0 0 30 0, clip]{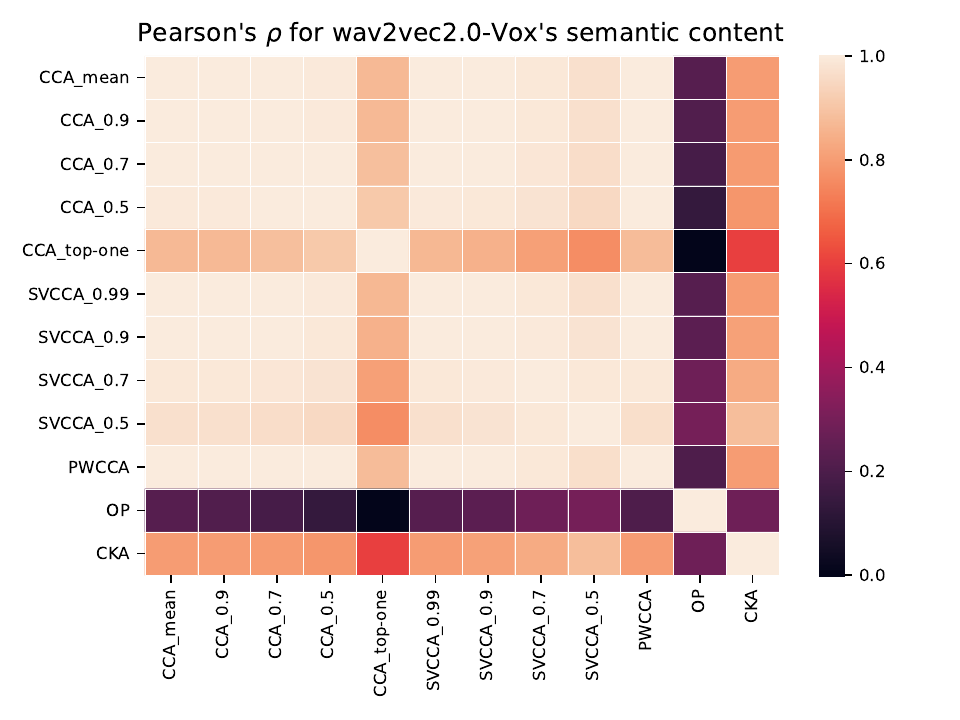}
    \label{fig:wav2vecl-sem-pearson}
\end{subfigure}

\caption{Correlation between different analysis tools for semantic content in \wavtovec \ models.}
\label{fig:metric-corr-w2v2-sem}
\end{figure}

\begin{figure}[ht]
\centering

\begin{subfigure}[b]{0.49\textwidth}
    \centering
    \includegraphics[width=\textwidth, trim=0 0 30 0, clip]{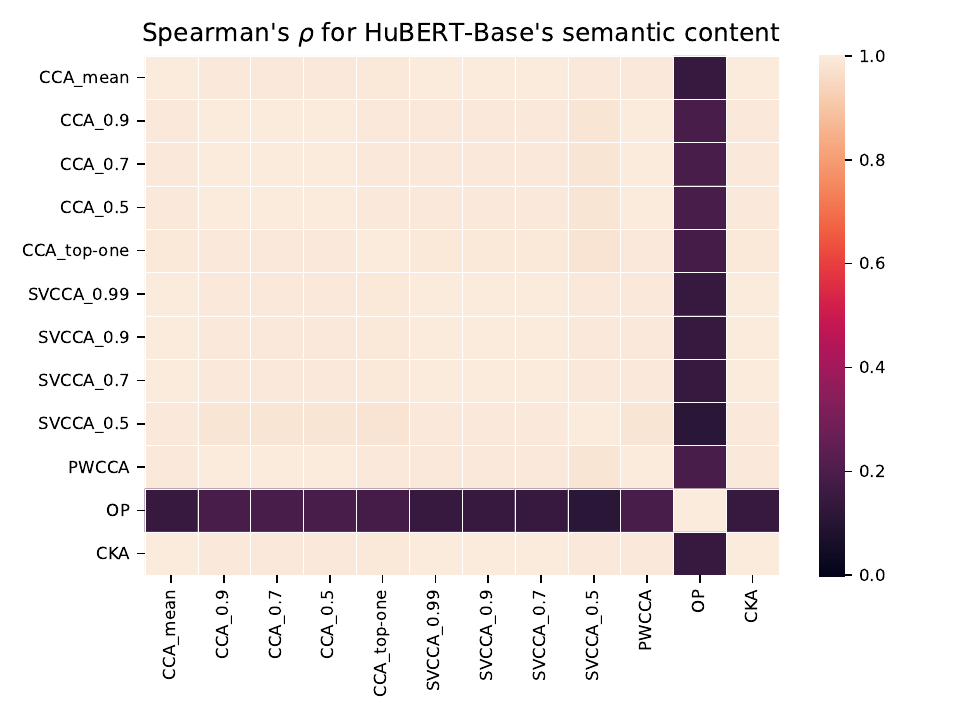}
    \label{fig:hubertb-sem-spearman}
\end{subfigure}
\hfill 
\begin{subfigure}[b]{0.49\textwidth}
    \centering
    \includegraphics[width=\textwidth, trim=0 0 30 0, clip]{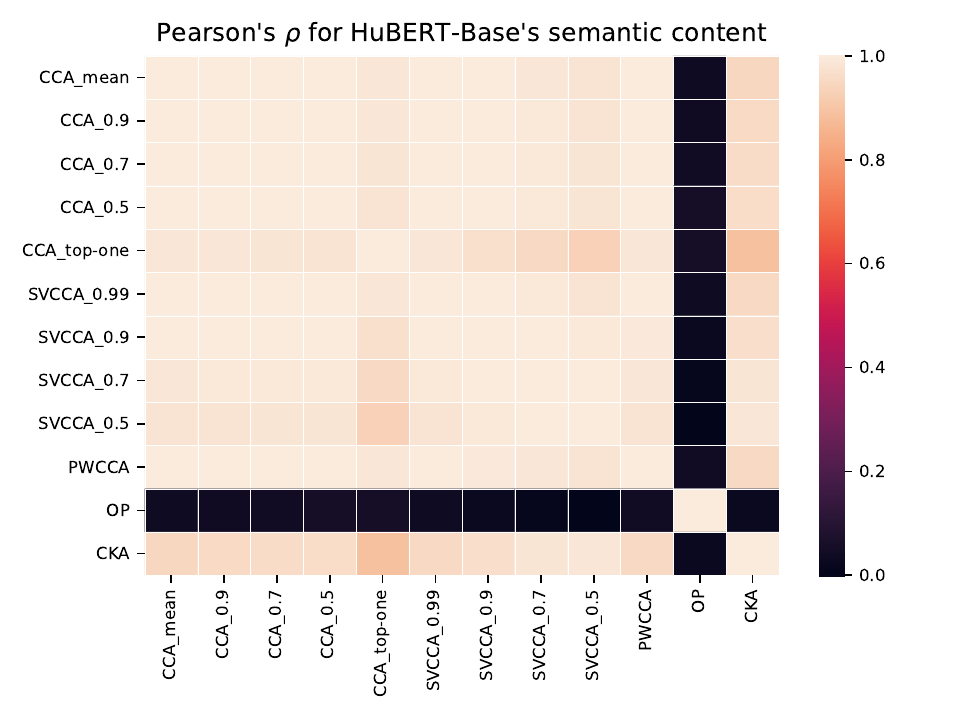}
    \label{fig:hubertb-sem-pearson}
\end{subfigure}

\begin{subfigure}[b]{0.49\textwidth}
    \centering
    \includegraphics[width=\textwidth, trim=0 0 30 0, clip]{images/metric_analysis/hubert_large_sem_spearman.pdf}
    \label{fig:hubertl-sem-spearman}
\end{subfigure}
\hfill 
\begin{subfigure}[b]{0.49\textwidth}
    \centering
    \includegraphics[width=\textwidth, trim=0 0 30 0, clip]{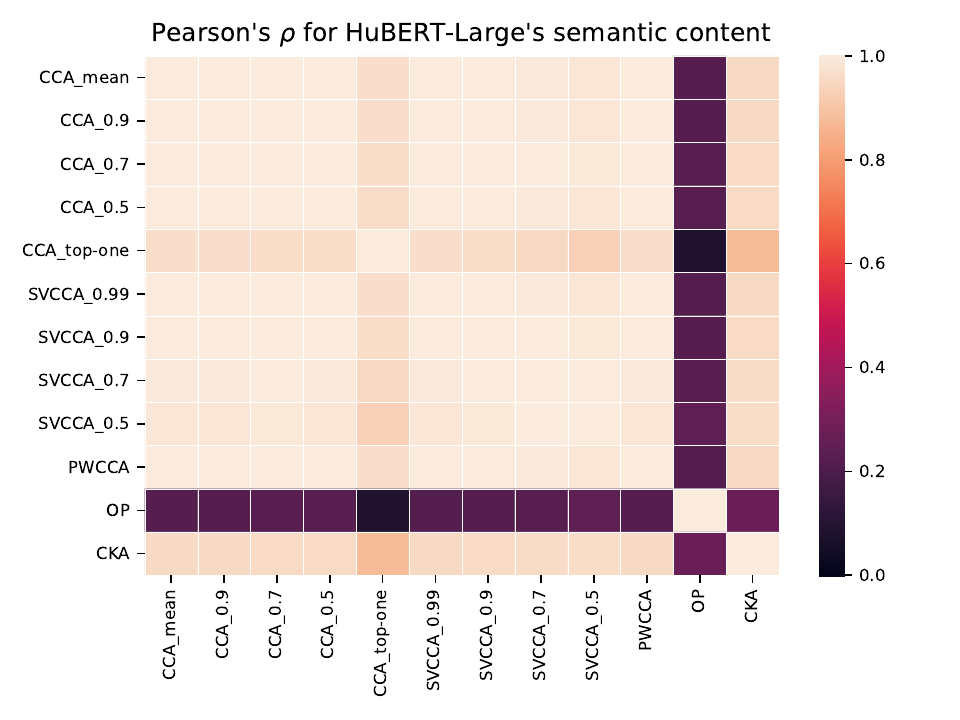}
    \label{fig:hubertl-sem-pearson}
\end{subfigure}

\caption{Correlation between different analysis tools for semantic content in \hubert \ models.}
\label{fig:metric-corr-hubert-sem}
\end{figure}

\begin{figure}[ht]
\centering

\begin{subfigure}[b]{0.49\textwidth}
    \centering
    \includegraphics[width=\textwidth, trim=0 0 30 0, clip]{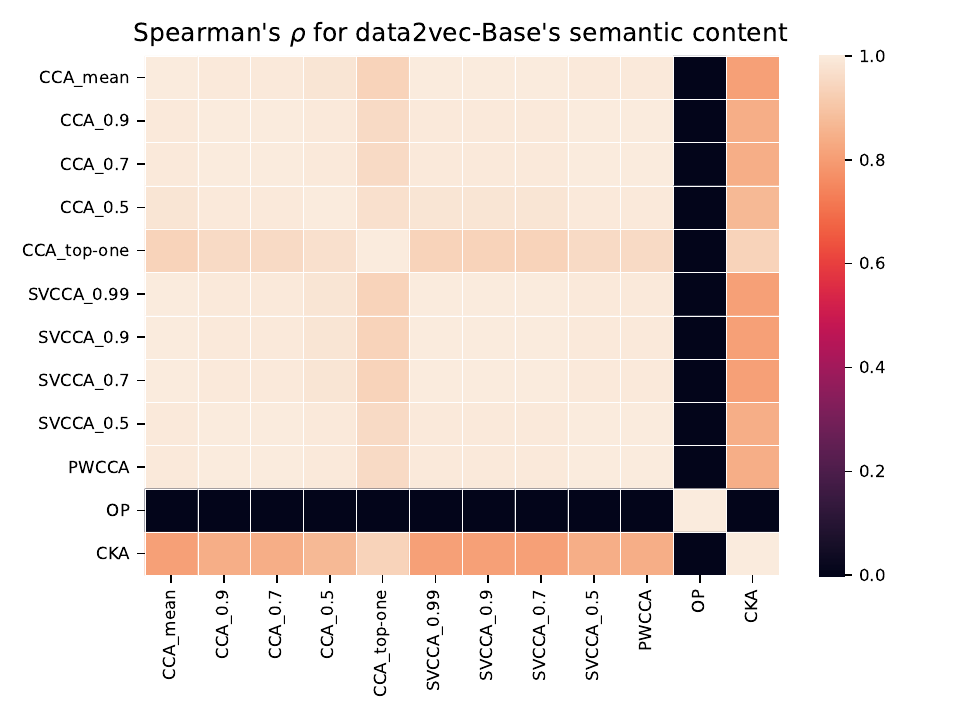}
    \label{fig:data2vecb-sem-spearman}
\end{subfigure}
\hfill 
\begin{subfigure}[b]{0.49\textwidth}
    \centering
    \includegraphics[width=\textwidth, trim=0 0 30 0, clip]{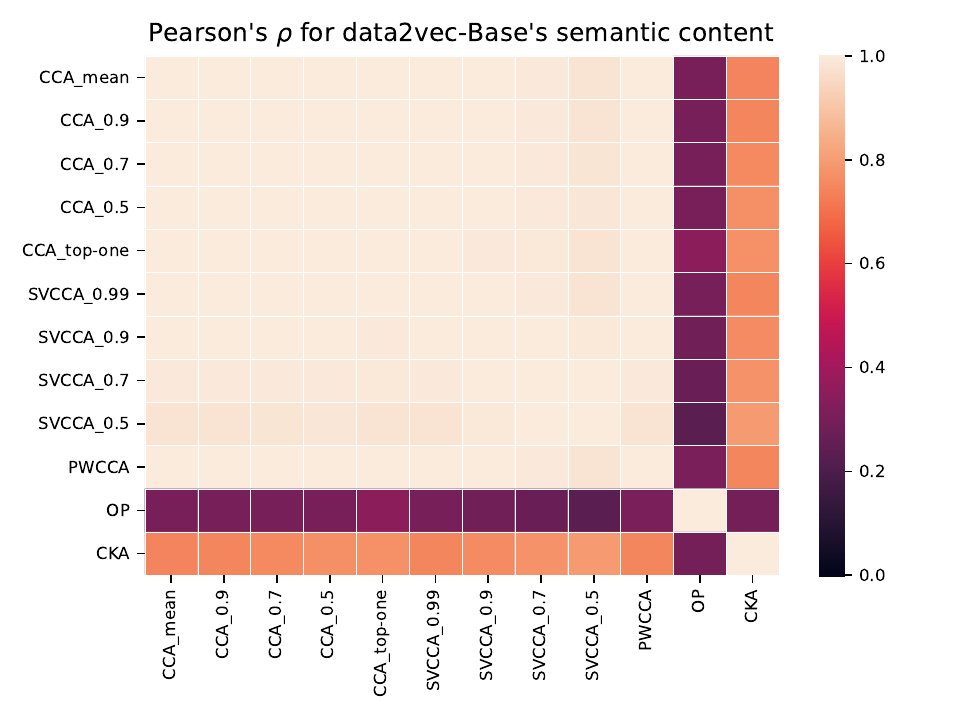}
    \label{fig:data2vecb-sem-pearson}
\end{subfigure}

\begin{subfigure}[b]{0.49\textwidth}
    \centering
    \includegraphics[width=\textwidth, trim=0 0 30 0, clip]{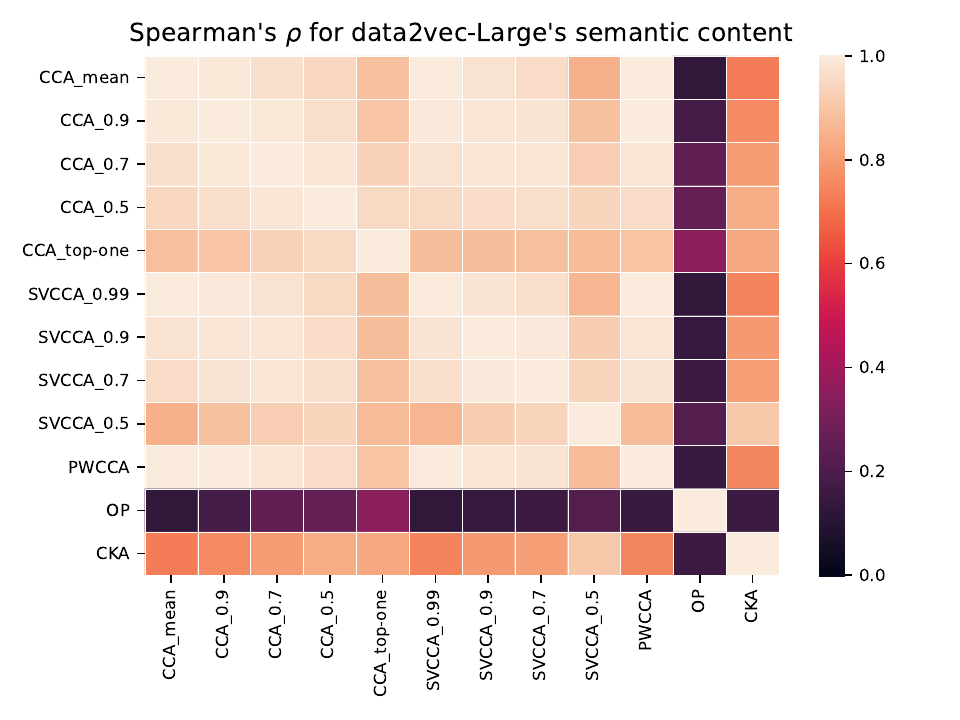}
    \label{fig:data2vecl-sem-spearman}
\end{subfigure}
\hfill 
\begin{subfigure}[b]{0.49\textwidth}
    \centering
    \includegraphics[width=\textwidth, trim=0 0 30 0, clip]{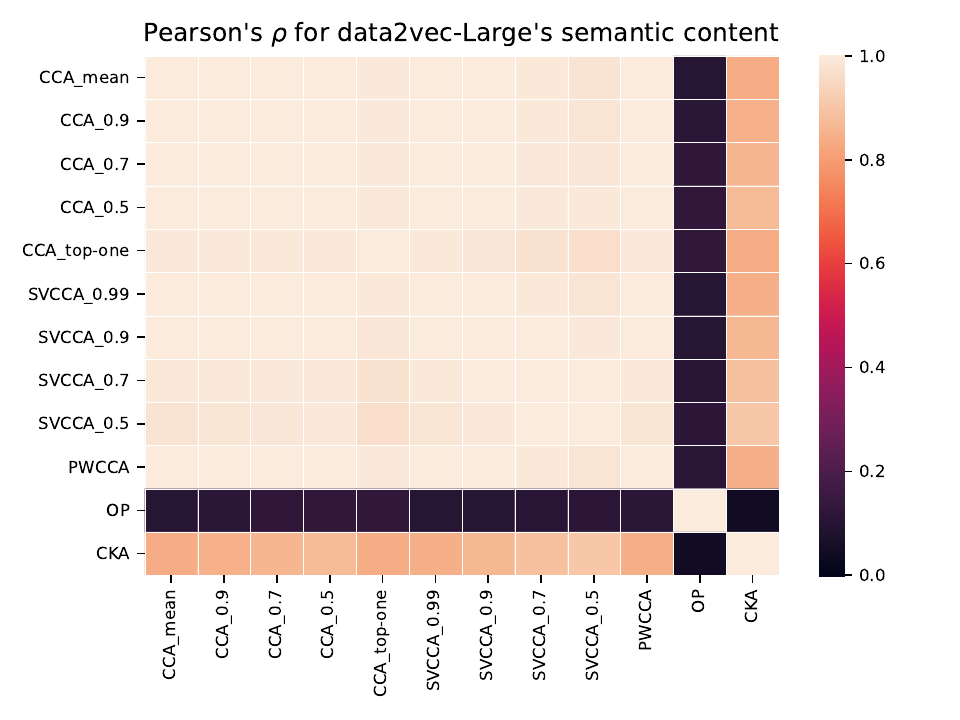}
    \label{fig:data2vecl-sem-pearson}
\end{subfigure}

\caption{Correlation between different analysis tools for semantic content in \datatovec \ models.}
\label{fig:metric-corr-data2vec-sem}
\end{figure}

Figures~\ref{fig:metric-corr-w2v2-phone}-\ref{fig:metric-corr-data2vec-sem} below are confusion maps presenting Spearman's and Pearson's rank correlation scores to compare layer-wise trends from different metrics. These are additional results for discussion in \sect~\ref{sec:res-compare-tools-analysis}.


\section{Transferability to downstream tasks}
\label{sec:appendix-scatter-plots}
\begin{figure}[htb]
\includegraphics[width=0.9\textwidth]{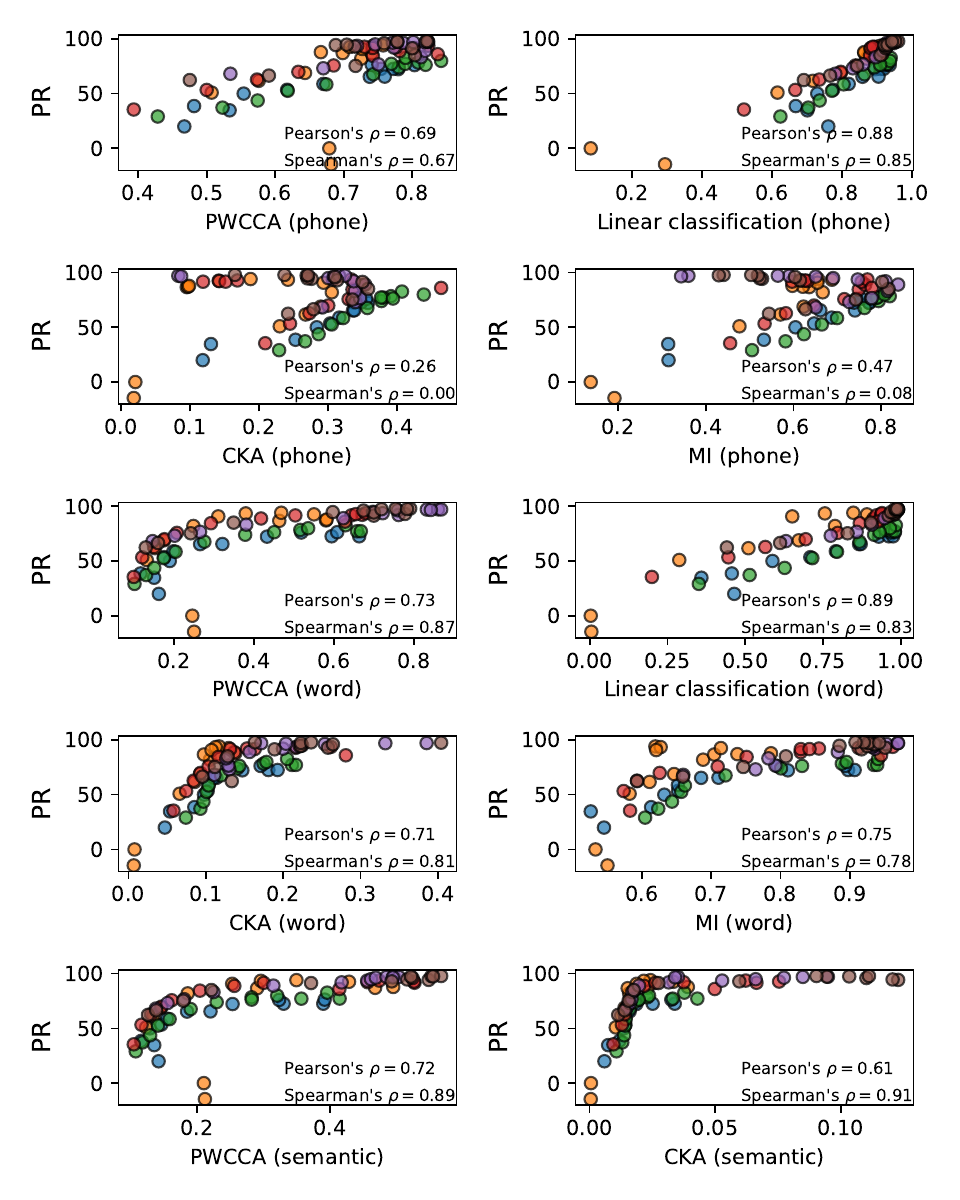}
\caption{Scatter plots comparing PR performance with task-agnostic layer-wise trends for all \sfms.
PR is measured as $100 - $error\_rate (in \%), \pwcca \ and \cka \ shown as similarity scores, \mi \ as normalized MI score, and linear classification as classification accuracy.}
\label{fig:appendix-phone-all}
\end{figure}

\begin{figure}[htb]
\includegraphics[width=0.9\textwidth]{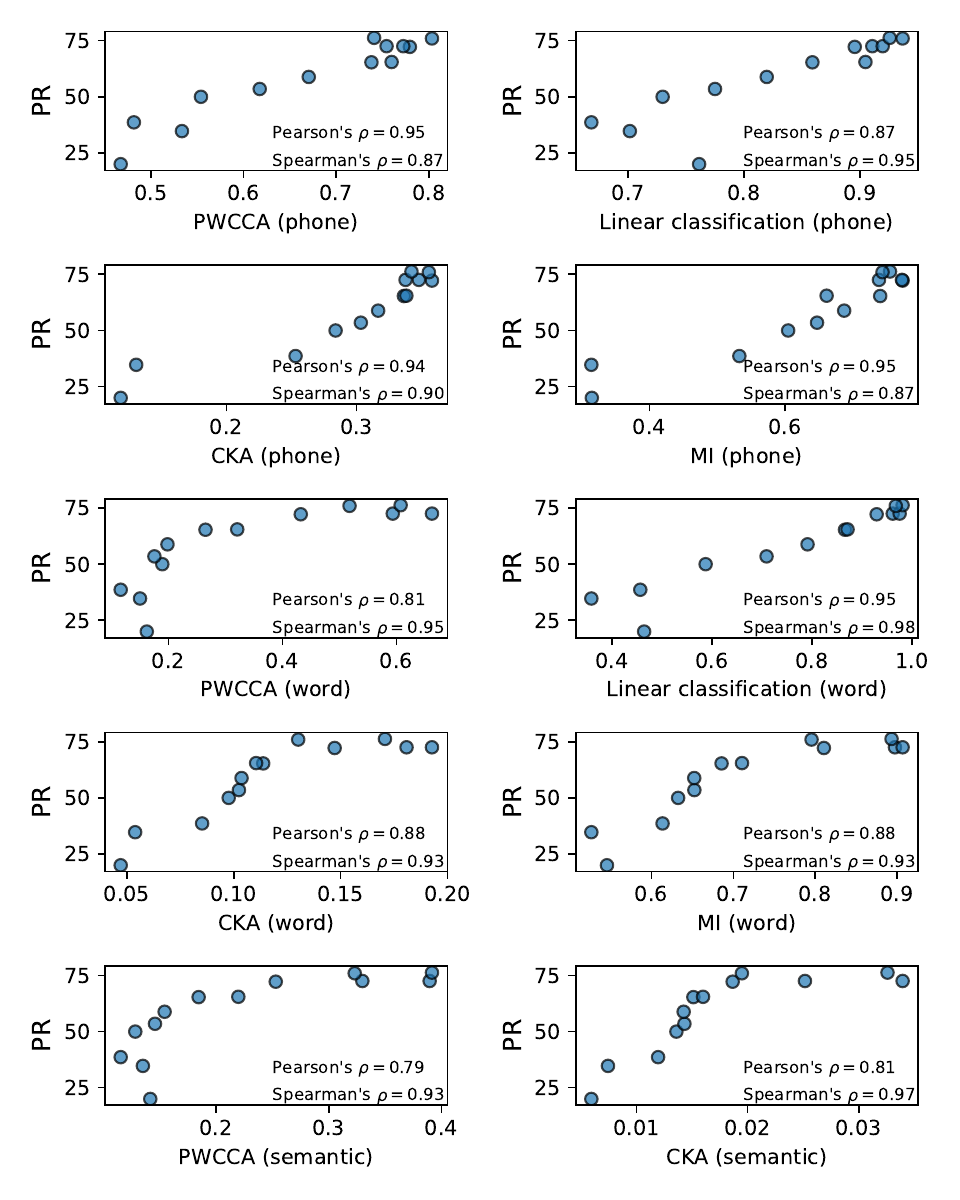}
\caption{Scatter plots comparing PR performance with task-agnostic layer-wise trends for \wavtovec-\baseM.
PR is measured as $100 - $error\_rate (in \%), \pwcca \ and \cka \ shown as similarity scores, \mi \ as normalized MI score, and linear classification as classification accuracy.}
\label{fig:appendix-phone-wav2vec_small}
\end{figure}

\begin{figure}[htb]
\includegraphics[width=0.9\textwidth]{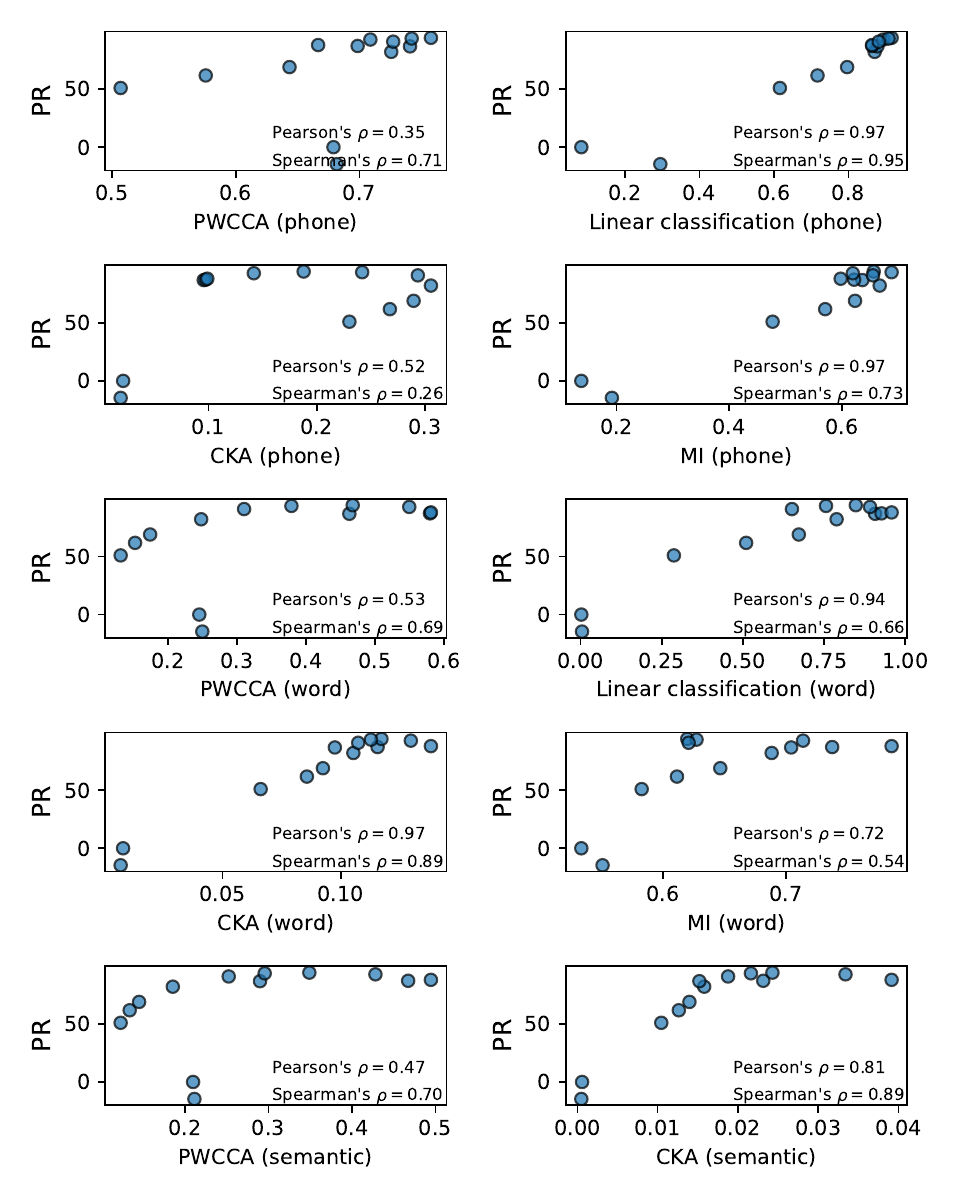}
\caption{Scatter plots comparing PR performance with task-agnostic layer-wise trends for \wavtovec-\voxM.
PR is measured as $100 - $error\_rate (in \%), \pwcca \ and \cka \ shown as similarity scores, \mi \ as normalized MI score, and linear classification as classification accuracy.}
\label{fig:appendix-phone-wav2vec_vox}
\end{figure}

\begin{figure}[htb]
\includegraphics[width=0.9\textwidth]{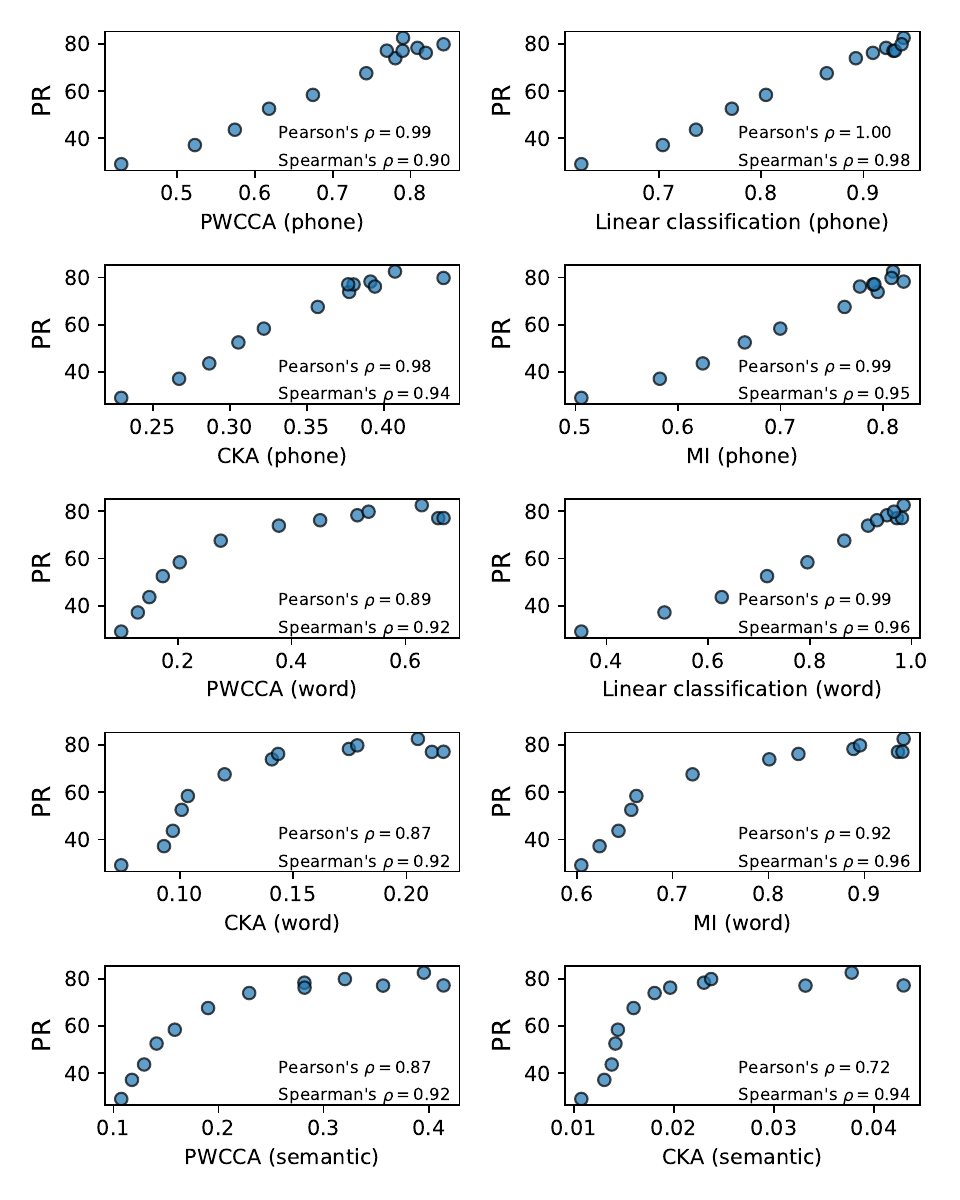}
\caption{Scatter plots comparing PR performance with task-agnostic layer-wise trends for \hubert-\baseM.
PR is measured as $100 - $error\_rate (in \%), \pwcca \ and \cka \ shown as similarity scores, \mi \ as normalized MI score, and linear classification as classification accuracy.}
\label{fig:appendix-phone-hubert_small}
\end{figure}

\begin{figure}[htb]
\includegraphics[width=0.9\textwidth]{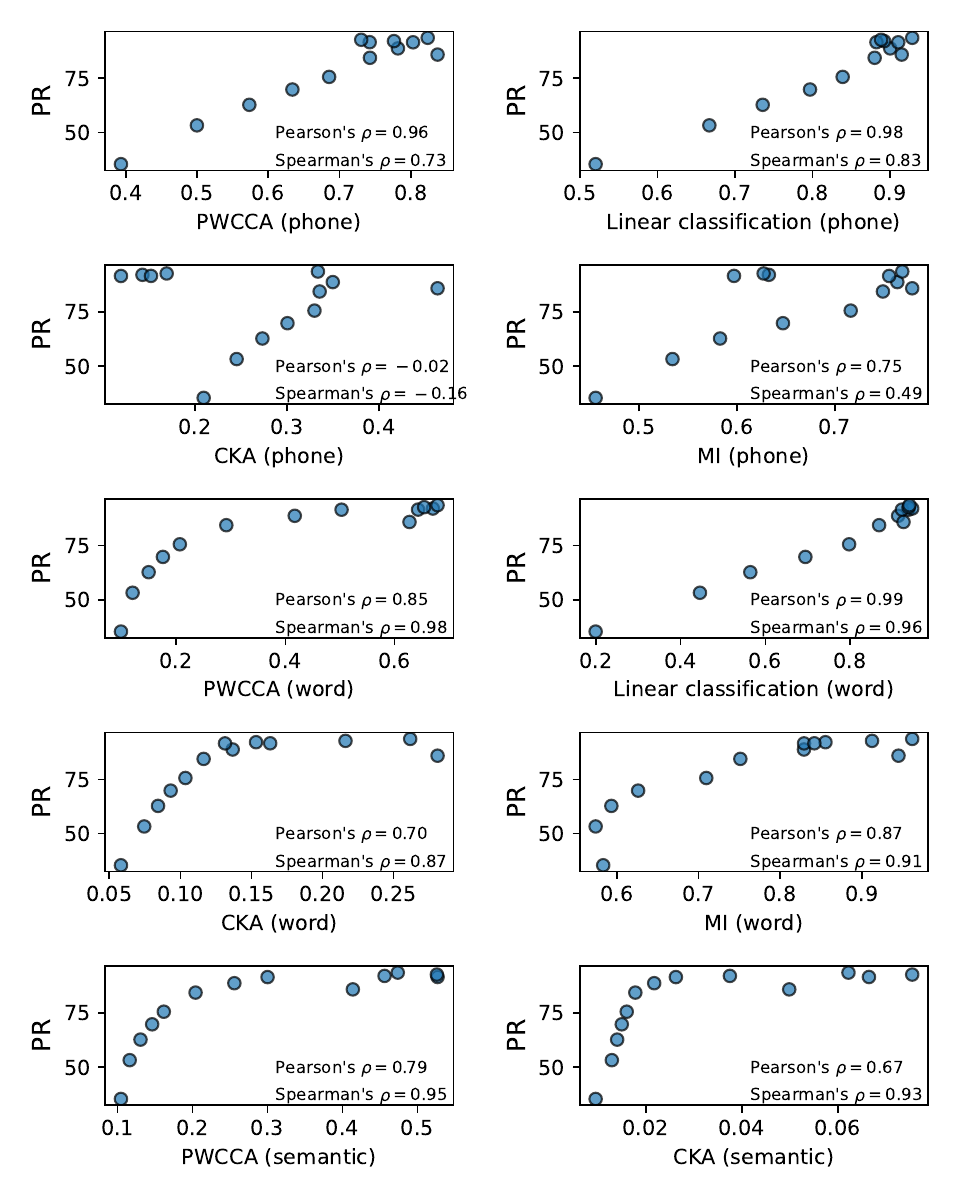}
\caption{Scatter plots comparing PR performance with task-agnostic layer-wise trends for \hubert-\largeM.
PR is measured as $100 - $error\_rate (in \%), \pwcca \ and \cka \ shown as similarity scores, \mi \ as normalized MI score, and linear classification as classification accuracy.}
\label{fig:appendix-phone-hubert_large}
\end{figure}

\begin{figure}[htb]
\includegraphics[width=0.9\textwidth]{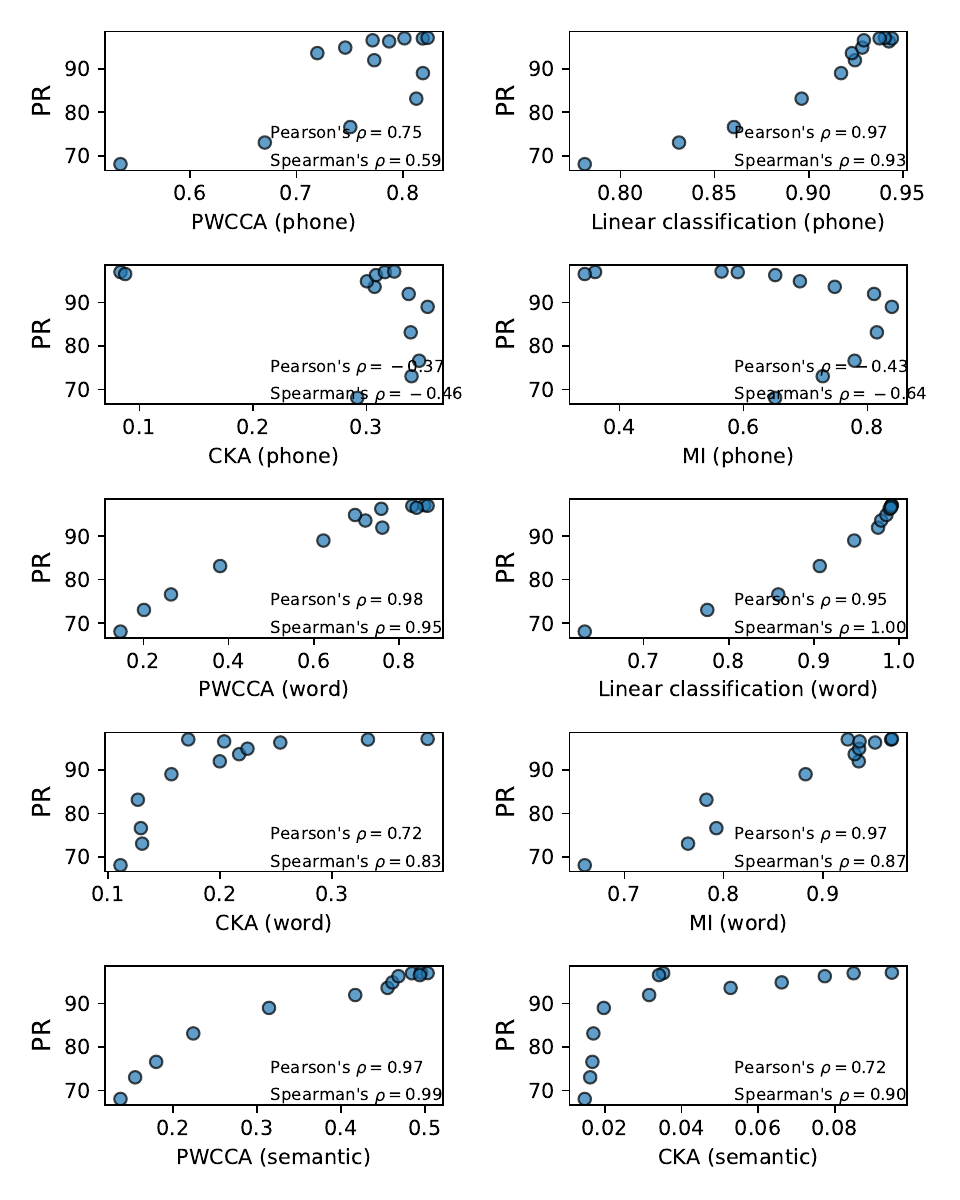}
\caption{Scatter plots comparing PR performance with task-agnostic layer-wise trends for \datatovec-\baseM.
PR is measured as $100 - $error\_rate (in \%), \pwcca \ and \cka \ shown as similarity scores, \mi \ as normalized MI score, and linear classification as classification accuracy.}
\label{fig:appendix-phone-data2vec_small}
\end{figure}

\begin{figure}[htb]
\includegraphics[width=0.9\textwidth]{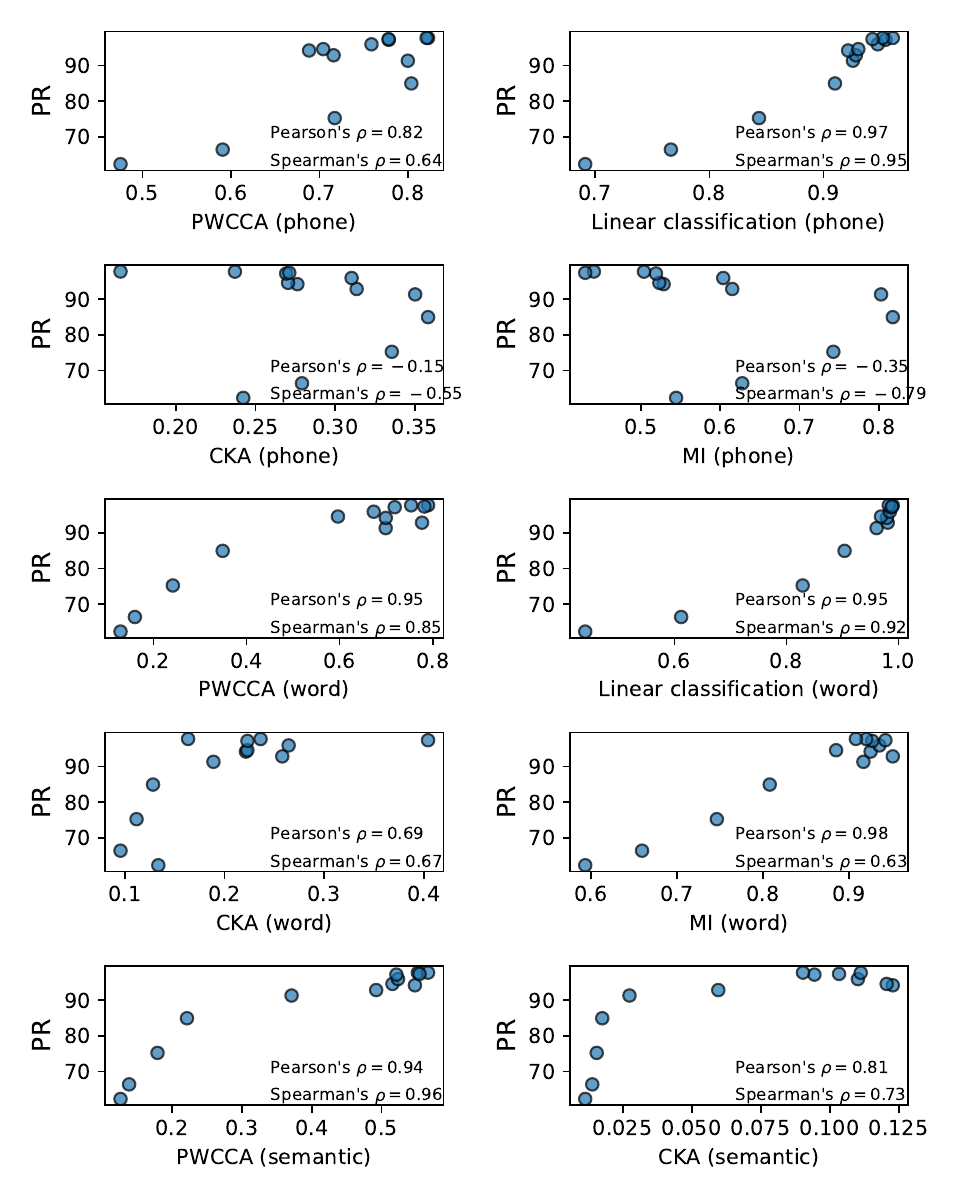}
\caption{Scatter plots comparing PR performance with task-agnostic layer-wise trends for \datatovec-\largeM.
PR is measured as $100 - $error\_rate (in \%), \pwcca \ and \cka \ shown as similarity scores, \mi \ as normalized MI score, and linear classification as classification accuracy.}
\label{fig:appendix-phone-data2vec_large}
\end{figure}

\begin{figure}[htb]
\includegraphics[width=0.9\textwidth]{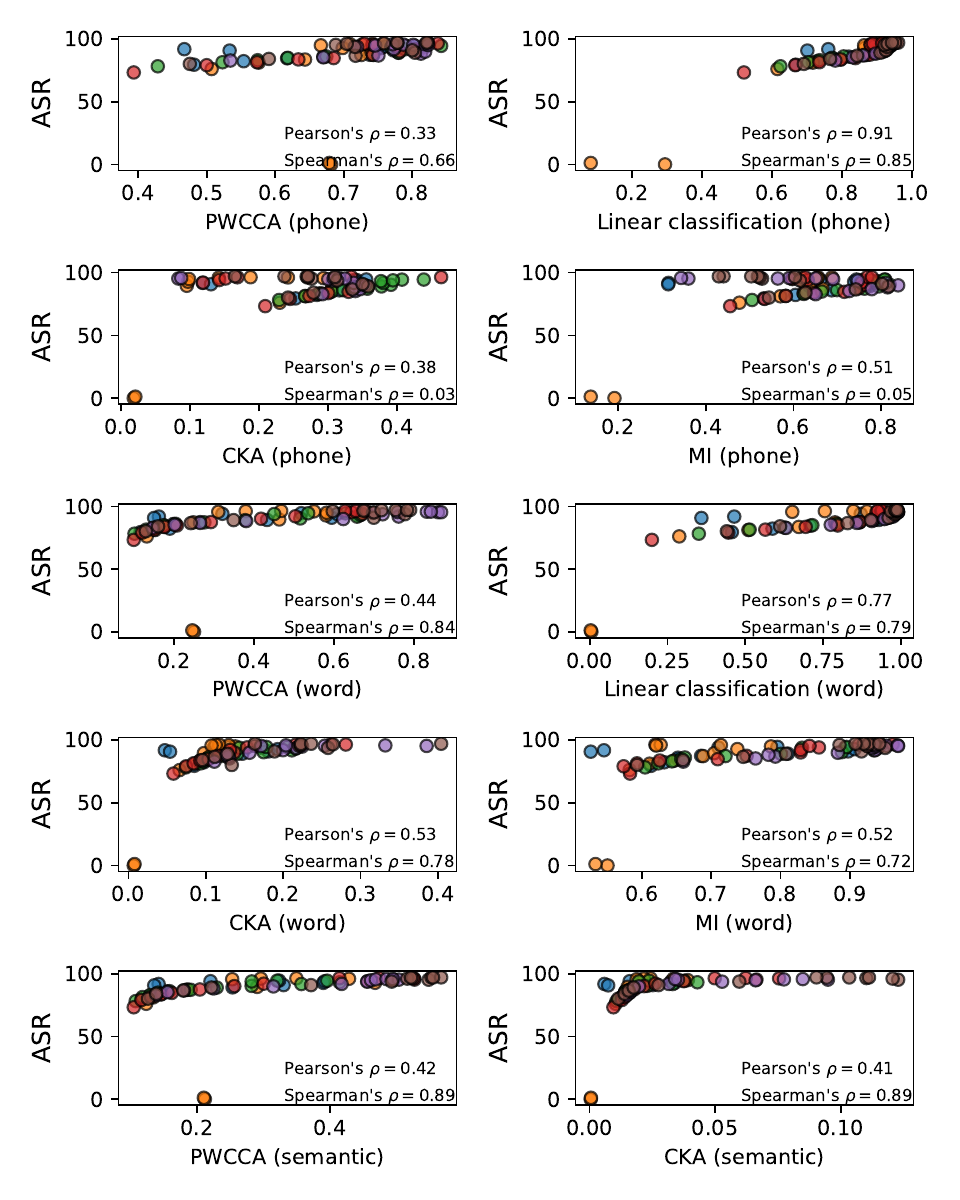}
\caption{Scatter plots comparing ASR performance with task-agnostic layer-wise trends for all \sfms.
ASR is measured as $100 - $error\_rate (in \%), \pwcca \ and \cka \ shown as similarity scores, \mi \ as normalized MI score, and linear classification as classification accuracy.}
\label{fig:appendix-asr-all}
\end{figure}

\begin{figure}[htb]
\includegraphics[width=0.9\textwidth]{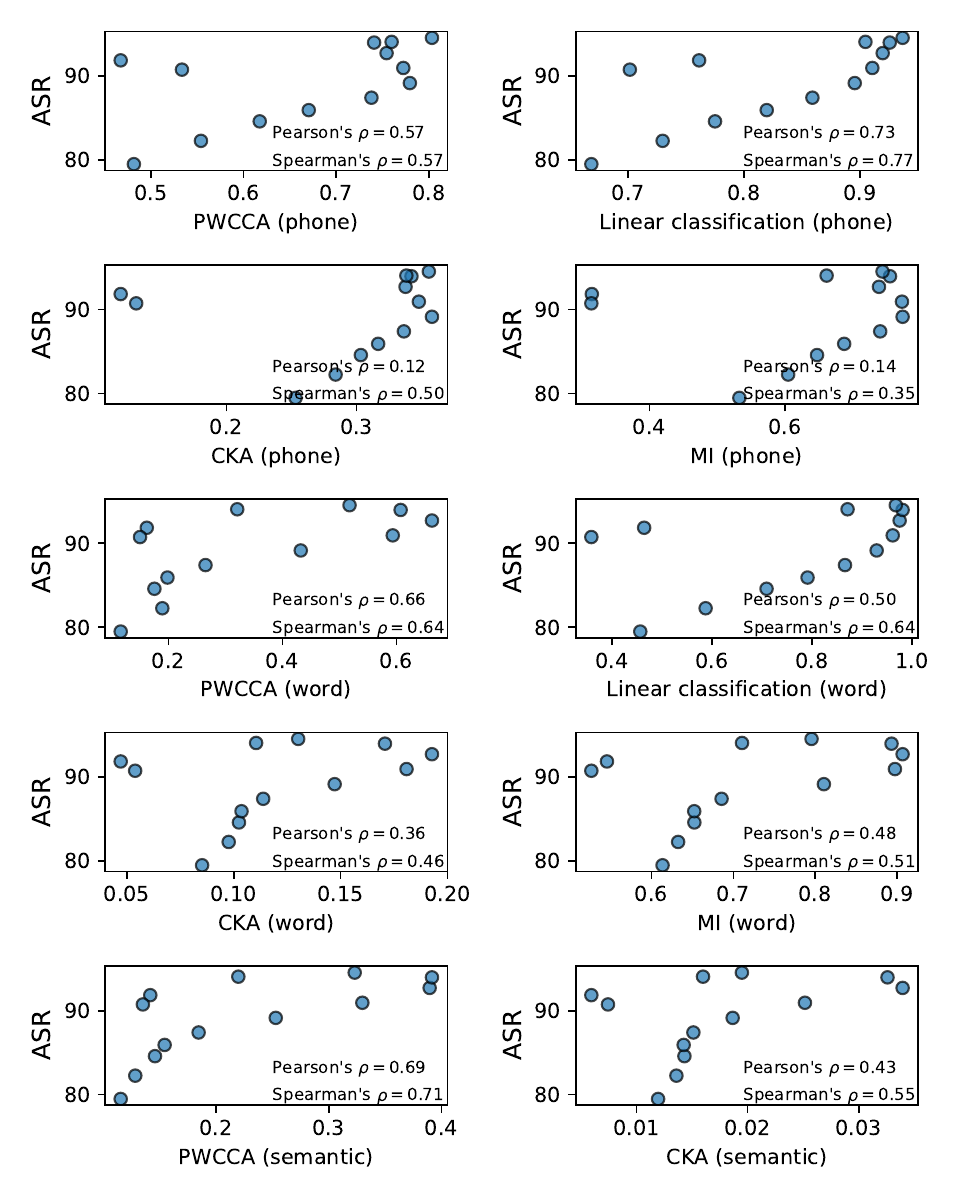}
\caption{Scatter plots comparing ASR performance with task-agnostic layer-wise trends for \wavtovec-\baseM.
ASR is measured as $100 - $error\_rate (in \%), \pwcca \ and \cka \ shown as similarity scores, \mi \ as normalized MI score, and linear classification as classification accuracy.}
\label{fig:appendix-asr-wav2vec_small}
\end{figure}

\begin{figure}[htb]
\includegraphics[width=0.9\textwidth]{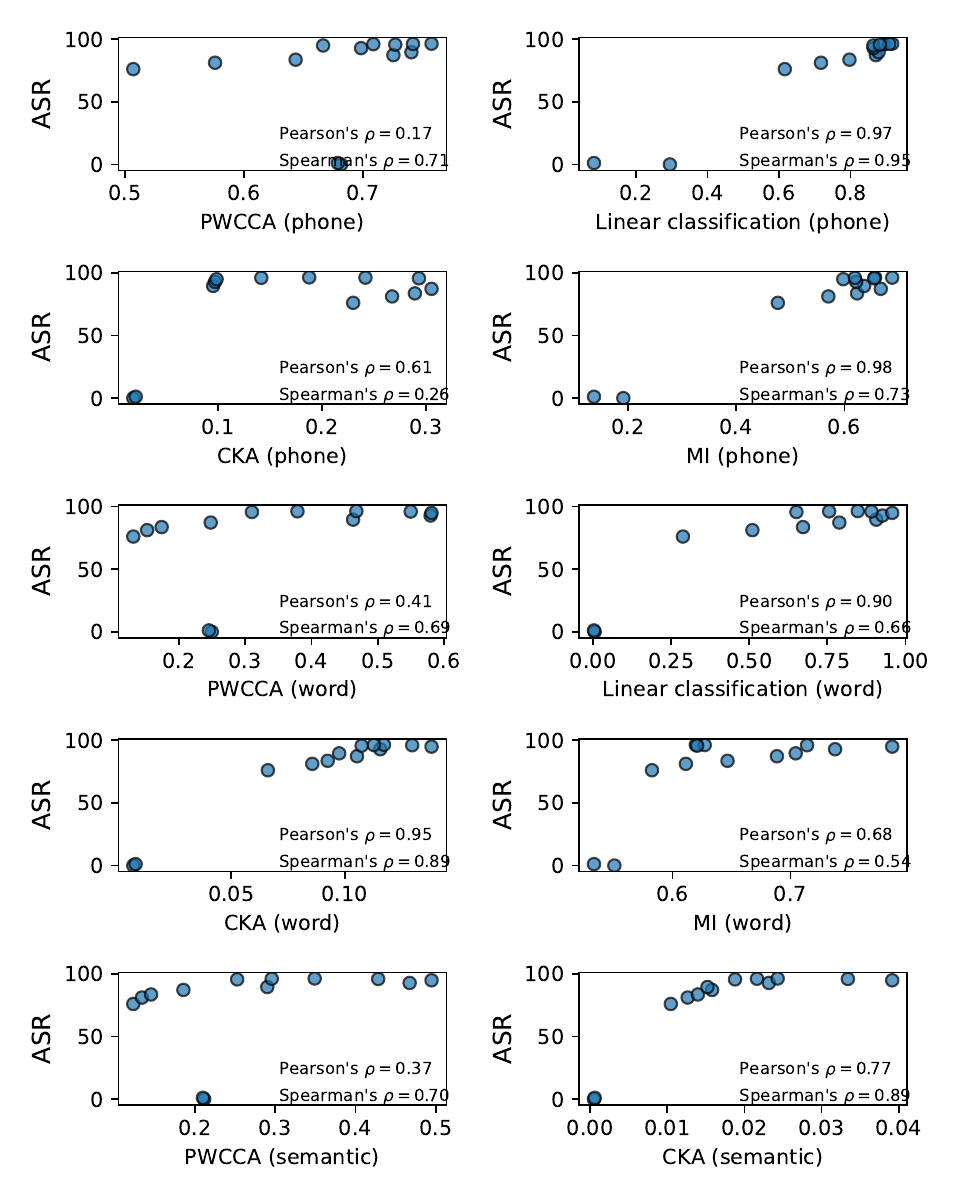}
\caption{Scatter plots comparing ASR performance with task-agnostic layer-wise trends for \wavtovec-\voxM.
ASR is measured as $100 - $error\_rate (in \%), \pwcca \ and \cka \ shown as similarity scores, \mi \ as normalized MI score, and linear classification as classification accuracy.}
\label{fig:appendix-asr-wav2vec_vox}
\end{figure}

\begin{figure}[htb]
\includegraphics[width=0.9\textwidth]{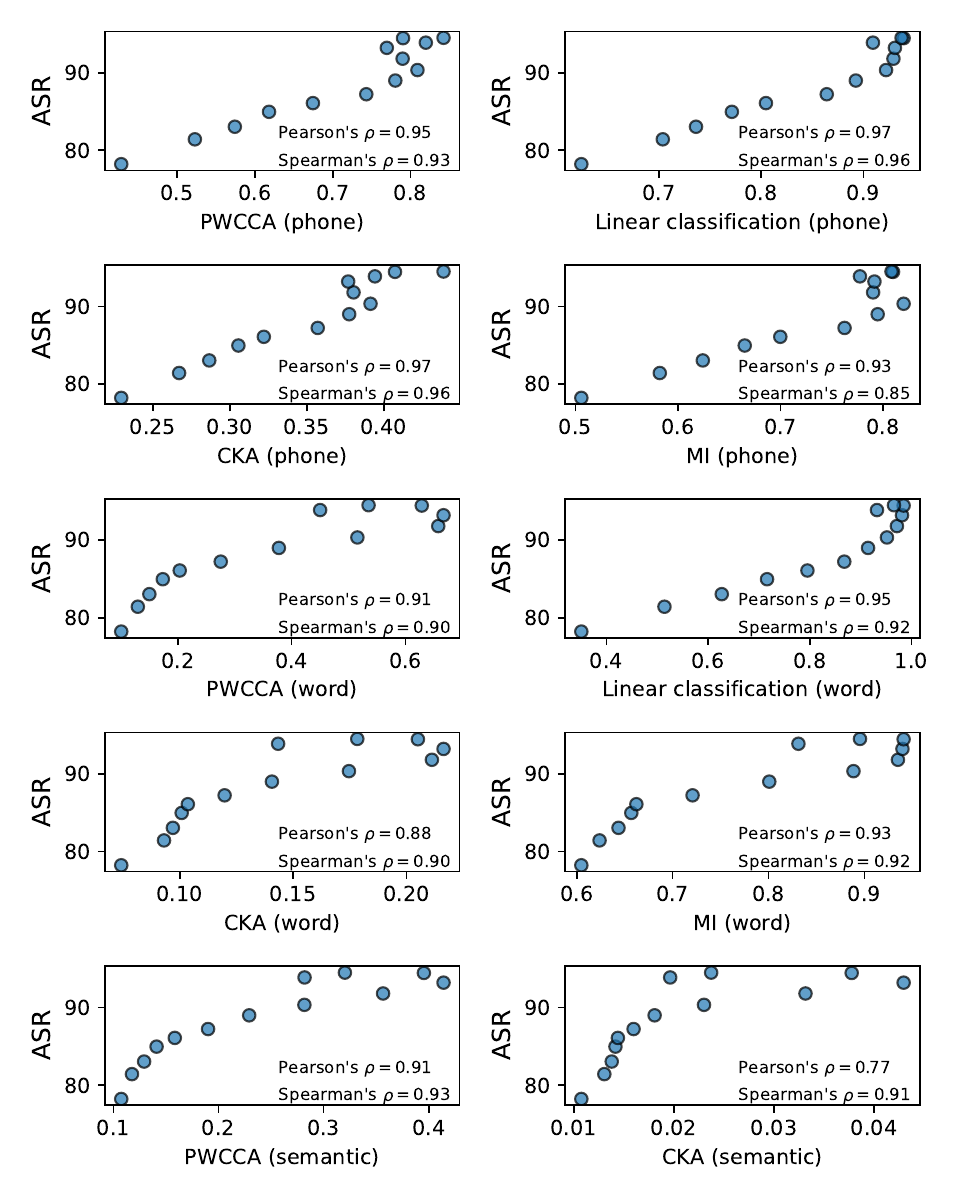}
\caption{Scatter plots comparing ASR performance with task-agnostic layer-wise trends for \hubert-\baseM.
ASR is measured as $100 - $error\_rate (in \%), \pwcca \ and \cka \ shown as similarity scores, \mi \ as normalized MI score, and linear classification as classification accuracy.}
\label{fig:appendix-asr-hubert_small}
\end{figure}

\begin{figure}[htb]
\includegraphics[width=0.9\textwidth]{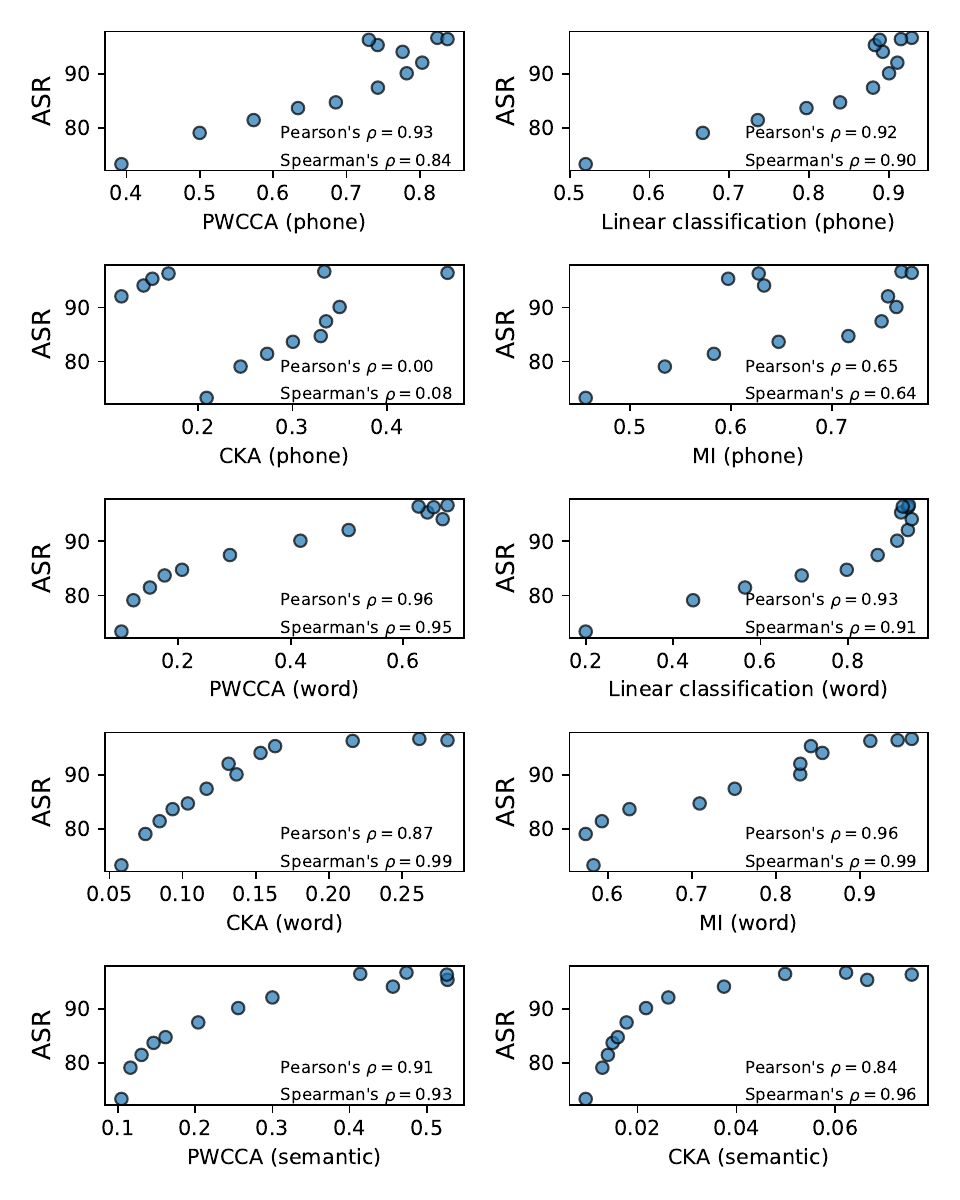}
\caption{Scatter plots comparing ASR performance with task-agnostic layer-wise trends for \hubert-\largeM.
ASR is measured as $100 - $error\_rate (in \%), \pwcca \ and \cka \ shown as similarity scores, \mi \ as normalized MI score, and linear classification as classification accuracy.}
\label{fig:appendix-asr-hubert_large}
\end{figure}

\begin{figure}[htb]
\includegraphics[width=0.9\textwidth]{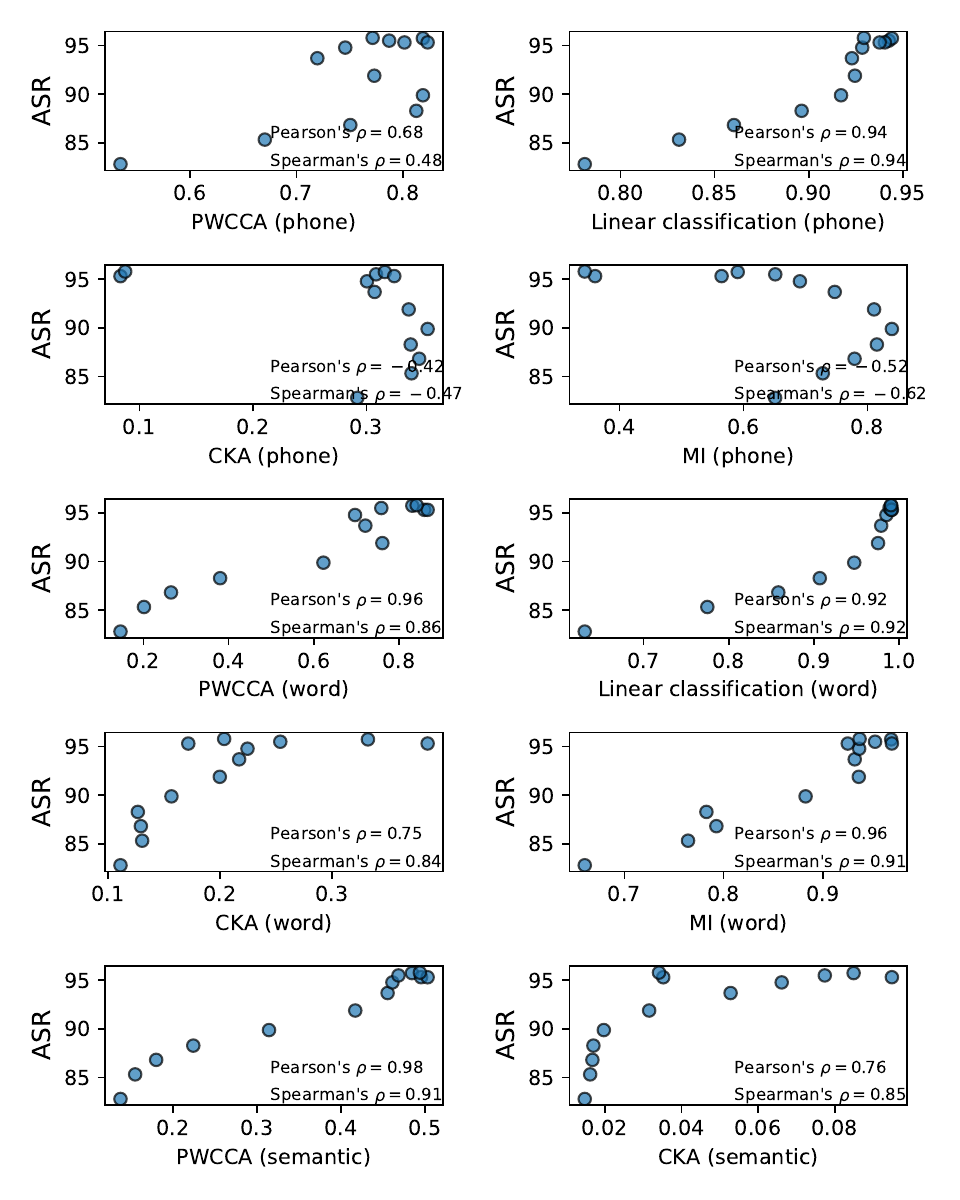}
\caption{Scatter plots comparing ASR performance with task-agnostic layer-wise trends for \datatovec-\baseM.
ASR is measured as $100 - $error\_rate (in \%), \pwcca \ and \cka \ shown as similarity scores, \mi \ as normalized MI score, and linear classification as classification accuracy.}
\label{fig:appendix-asr-data2vec_small}
\end{figure}

\begin{figure}[htb]
\includegraphics[width=0.9\textwidth]{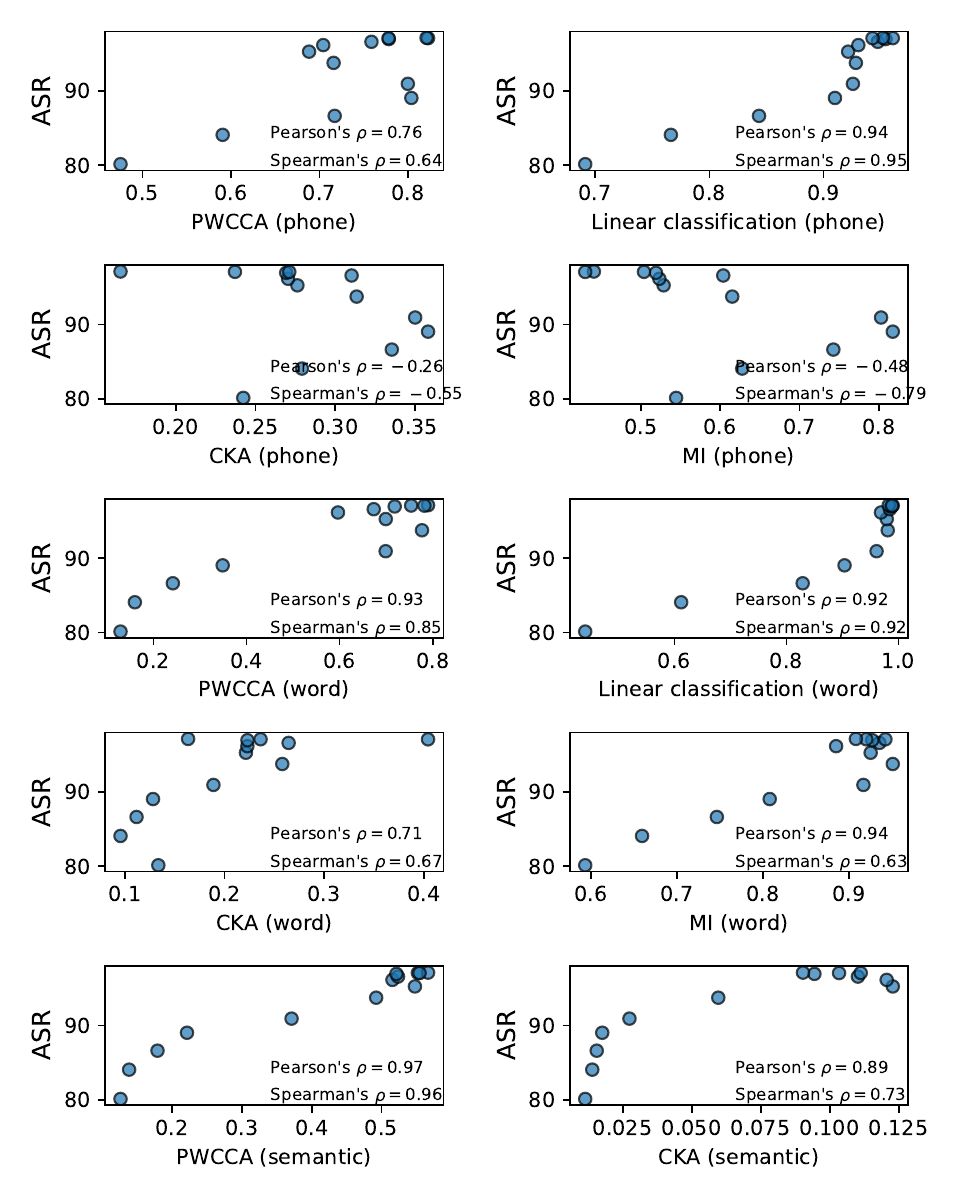}
\caption{Scatter plots comparing ASR performance with task-agnostic layer-wise trends for \datatovec-\largeM.
ASR is measured as $100 - $error\_rate (in \%), \pwcca \ and \cka \ shown as similarity scores, \mi \ as normalized MI score, and linear classification as classification accuracy.}
\label{fig:appendix-asr-data2vec_large}
\end{figure}

\begin{figure}[htb]
\includegraphics[width=0.9\textwidth]{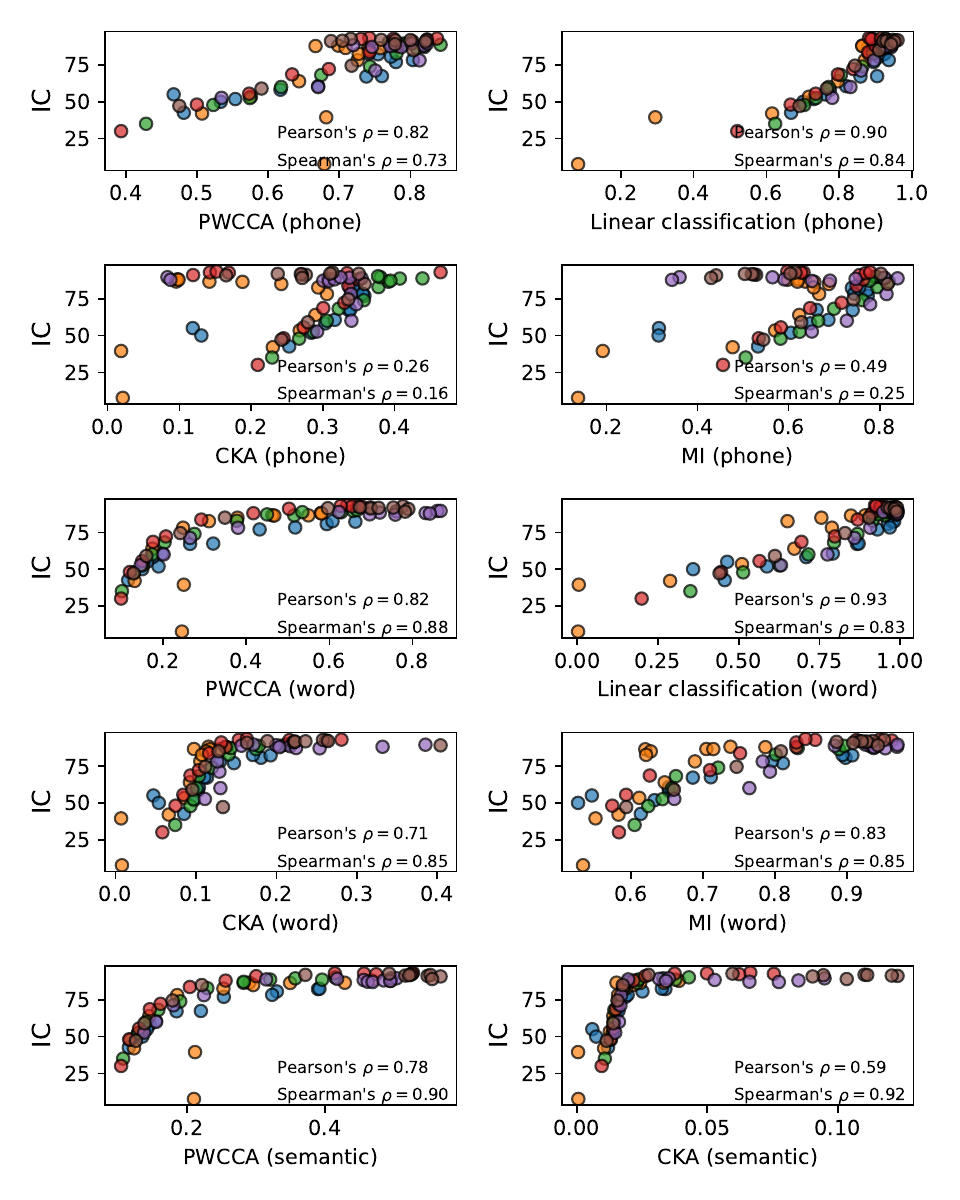}
\caption{Scatter plots comparing IC performance with task-agnostic layer-wise trends for all \sfms.
IC is measured as accuracy (in \%), \pwcca \ and \cka \ shown as similarity scores, \mi \ as normalized MI score, and linear classification as classification accuracy.}
\label{fig:appendix-fsc-all}
\end{figure}

\begin{figure}[htb]
\includegraphics[width=0.9\textwidth]{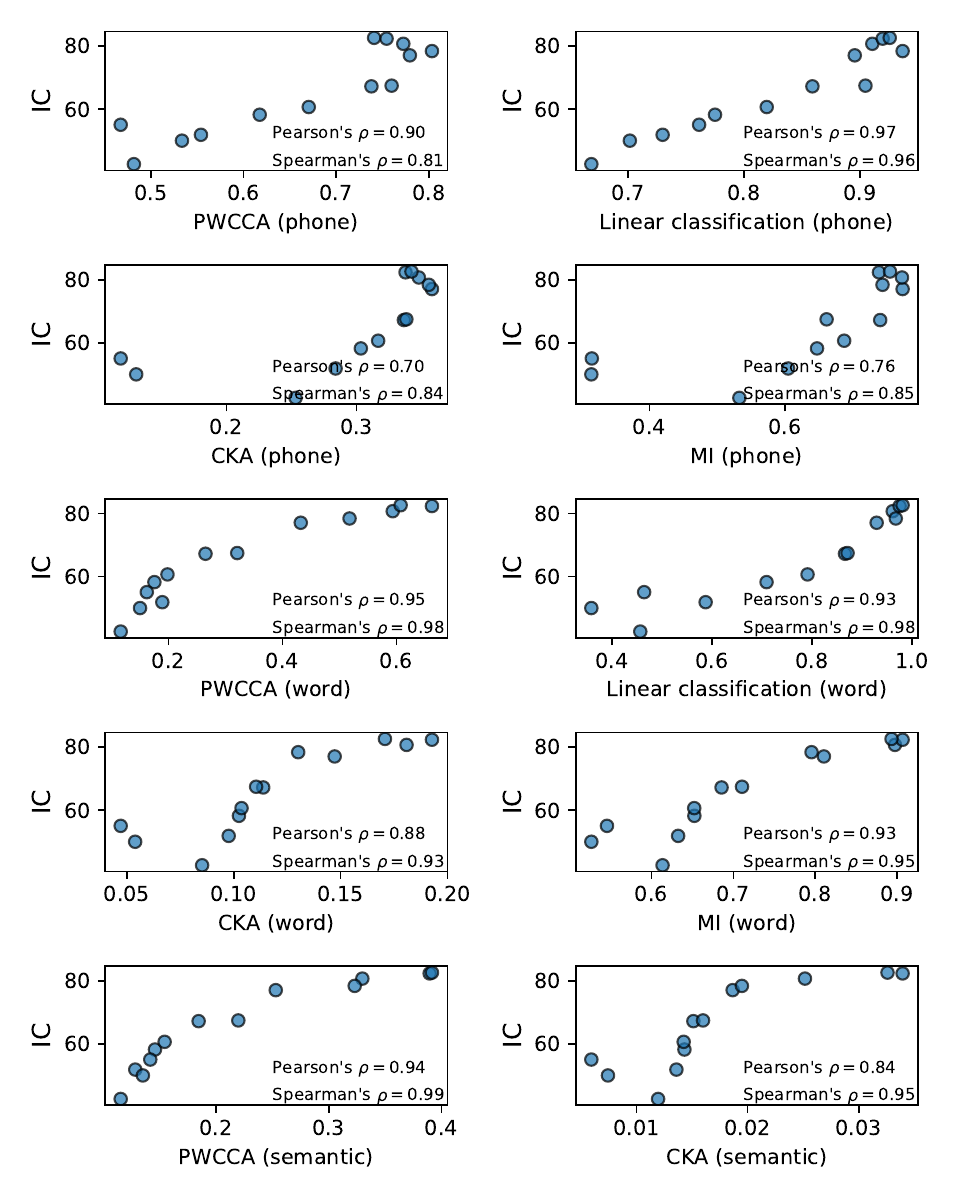}
\caption{Scatter plots comparing IC performance with task-agnostic layer-wise trends for \wavtovec-\baseM.
IC is measured as accuracy (in \%), \pwcca \ and \cka \ shown as similarity scores, \mi \ as normalized MI score, and linear classification as classification accuracy.}
\label{fig:appendix-fsc-wav2vec_small}
\end{figure}

\begin{figure}[htb]
\includegraphics[width=0.9\textwidth]{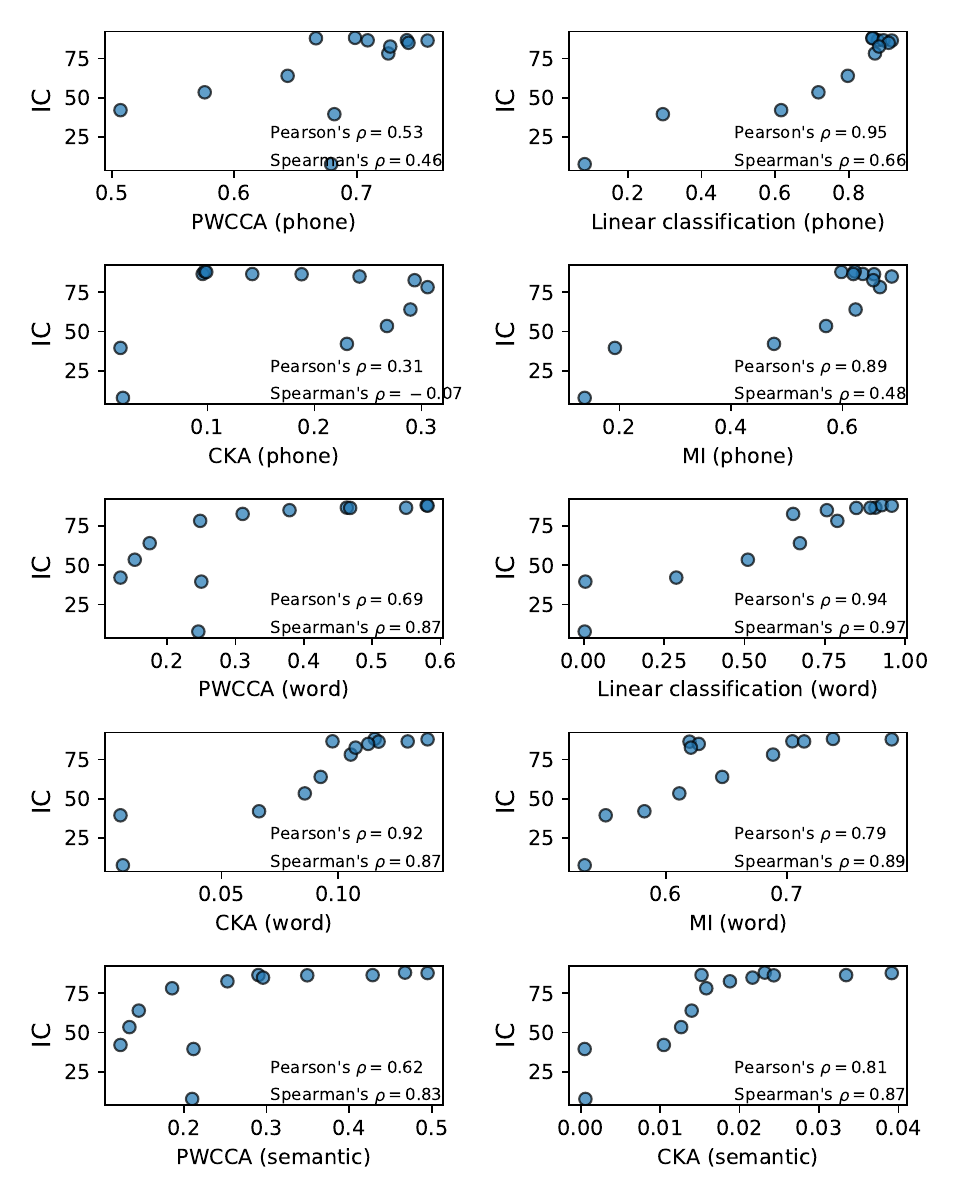}
\caption{Scatter plots comparing IC performance with task-agnostic layer-wise trends for \wavtovec-\voxM.
IC is measured as accuracy (in \%), \pwcca \ and \cka \ shown as similarity scores, \mi \ as normalized MI score, and linear classification as classification accuracy.}
\label{fig:appendix-fsc-wav2vec_vox}
\end{figure}

\begin{figure}[htb]
\includegraphics[width=0.9\textwidth]{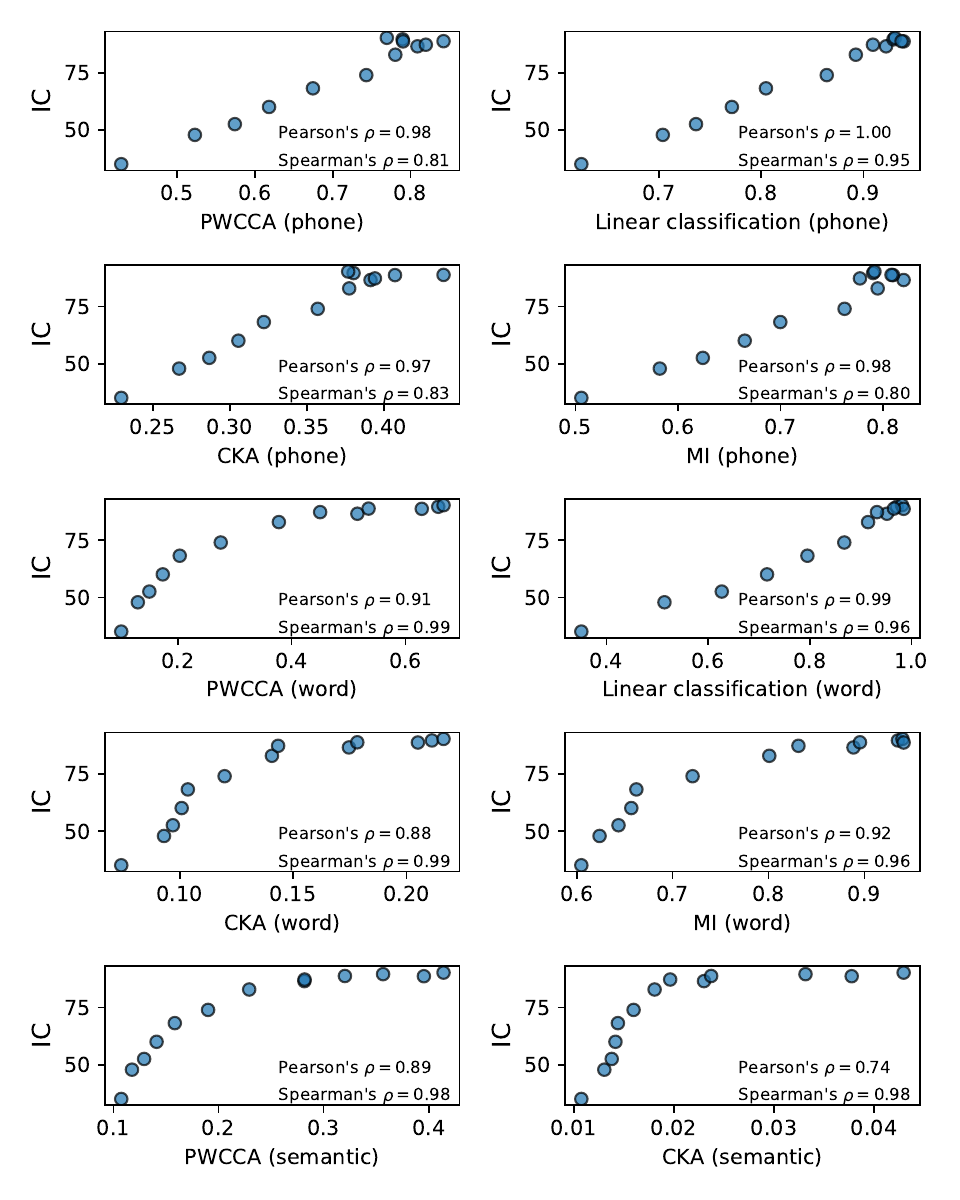}
\caption{Scatter plots comparing IC performance with task-agnostic layer-wise trends for \hubert-\baseM.
IC is measured as accuracy (in \%), \pwcca \ and \cka \ shown as similarity scores, \mi \ as normalized MI score, and linear classification as classification accuracy.}
\label{fig:appendix-fsc-hubert_small}
\end{figure}

\begin{figure}[htb]
\includegraphics[width=0.9\textwidth]{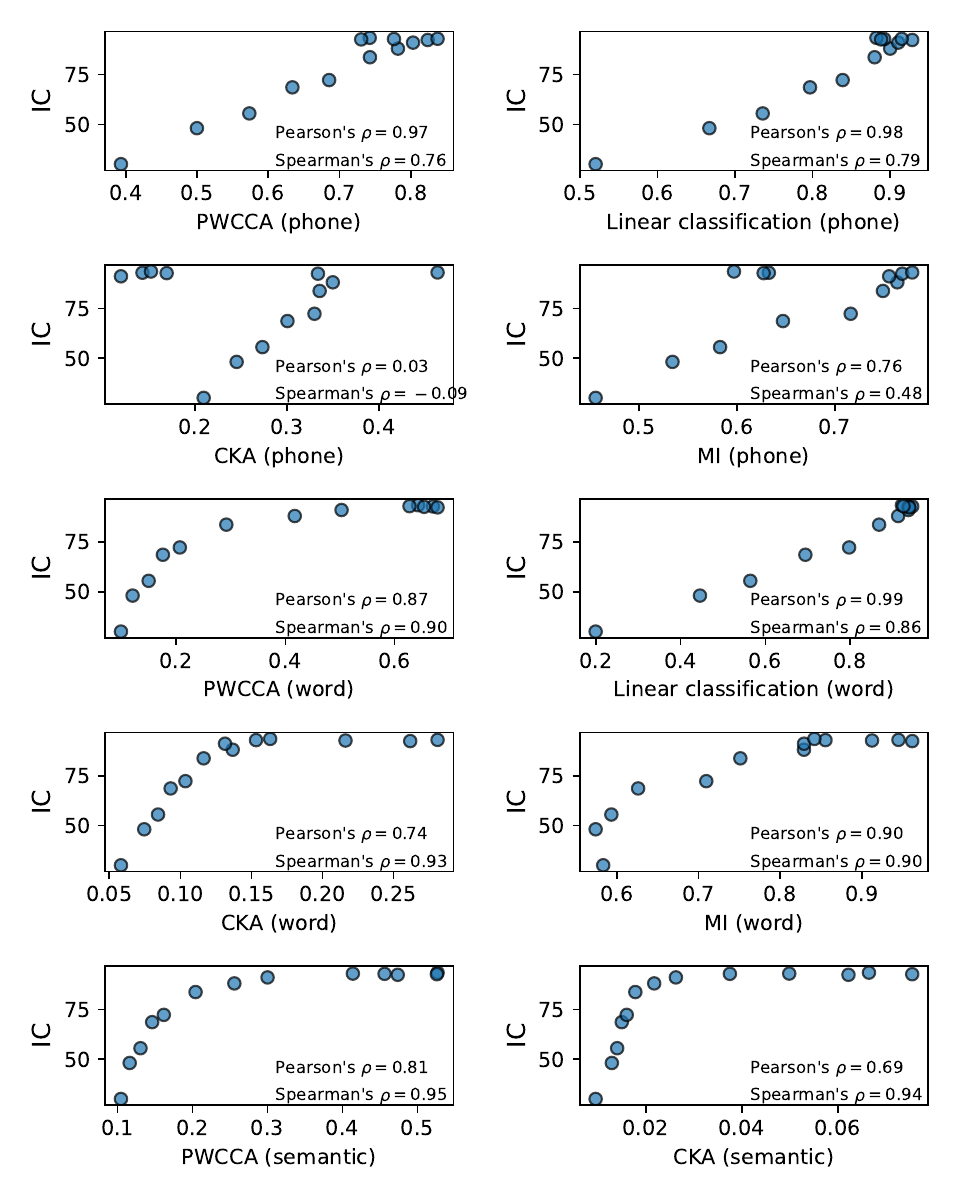}
\caption{Scatter plots comparing IC performance with task-agnostic layer-wise trends for \hubert-\largeM.
IC is measured as accuracy (in \%), \pwcca \ and \cka \ shown as similarity scores, \mi \ as normalized MI score, and linear classification as classification accuracy.}
\label{fig:appendix-fsc-hubert_large}
\end{figure}

\begin{figure}[htb]
\includegraphics[width=0.9\textwidth]{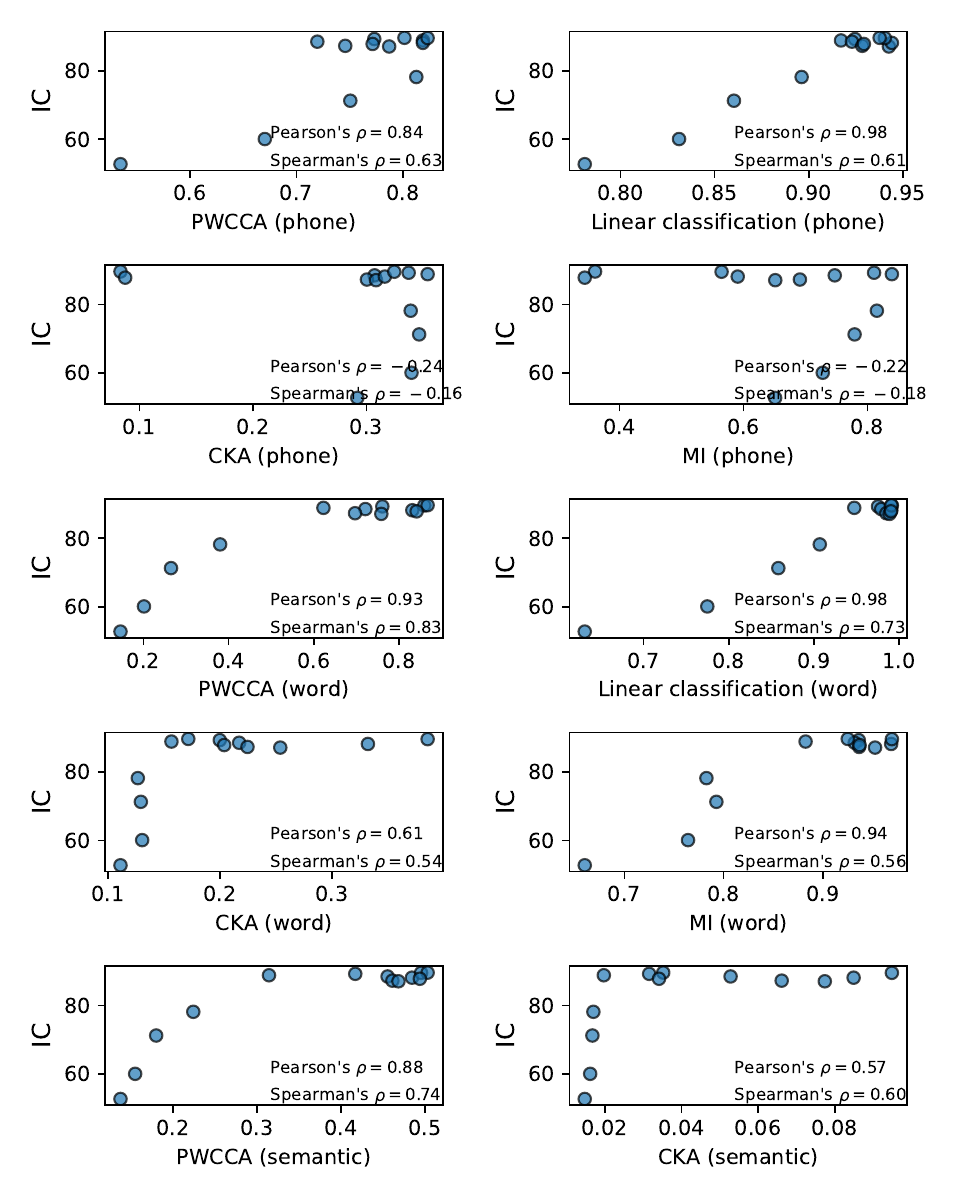}
\caption{Scatter plots comparing IC performance with task-agnostic layer-wise trends for \datatovec-\baseM.
IC is measured as accuracy (in \%), \pwcca \ and \cka \ shown as similarity scores, \mi \ as normalized MI score, and linear classification as classification accuracy.}
\label{fig:appendix-fsc-data2vec_small}
\end{figure}

\begin{figure}[htb]
\includegraphics[width=0.9\textwidth]{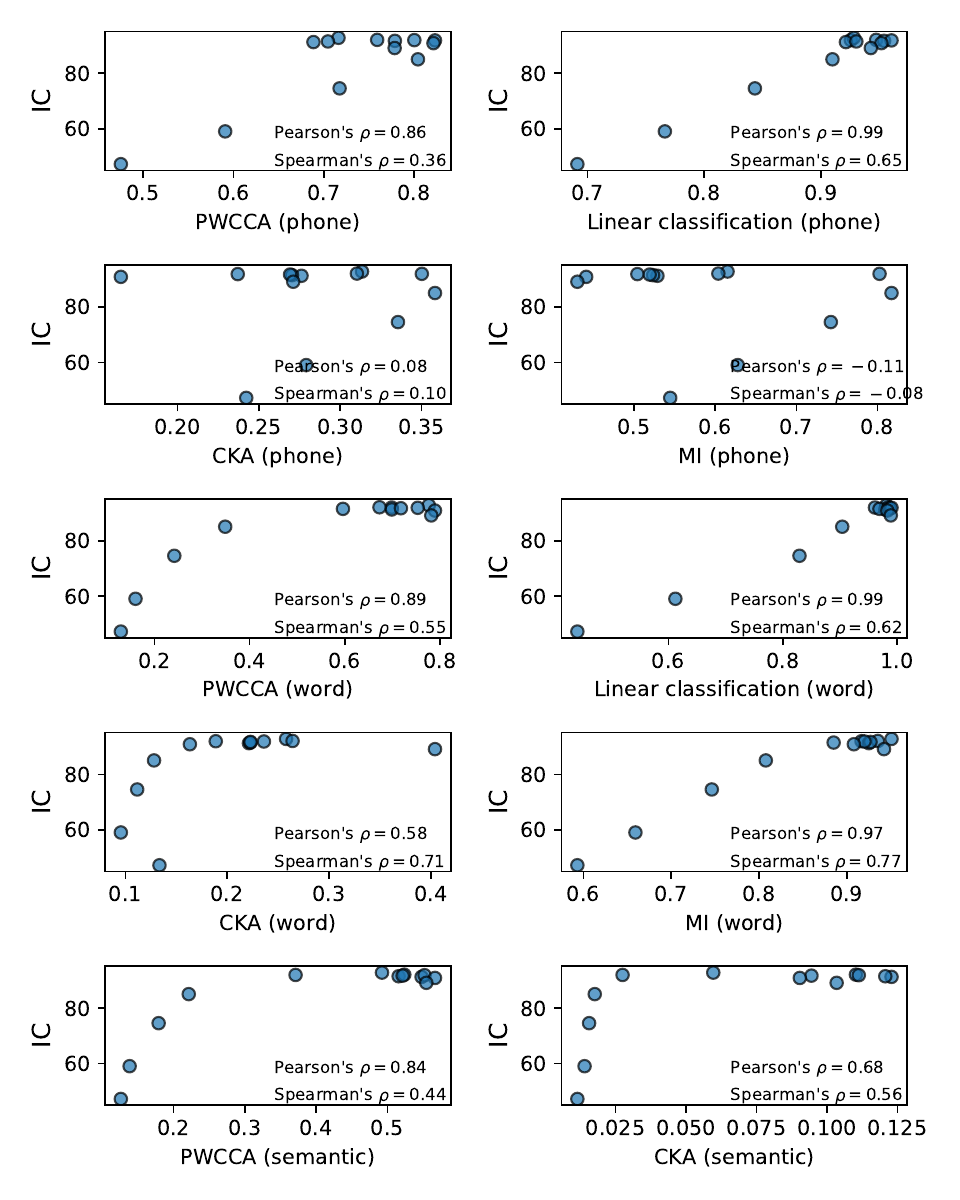}
\caption{Scatter plots comparing IC performance with task-agnostic layer-wise trends for \datatovec-\largeM.
IC is measured as accuracy (in \%), \pwcca \ and \cka \ shown as similarity scores, \mi \ as normalized MI score, and linear classification as classification accuracy.}
\label{fig:appendix-fsc-data2vec_large}
\end{figure}

\begin{figure}[htb]
\includegraphics[width=0.9\textwidth]{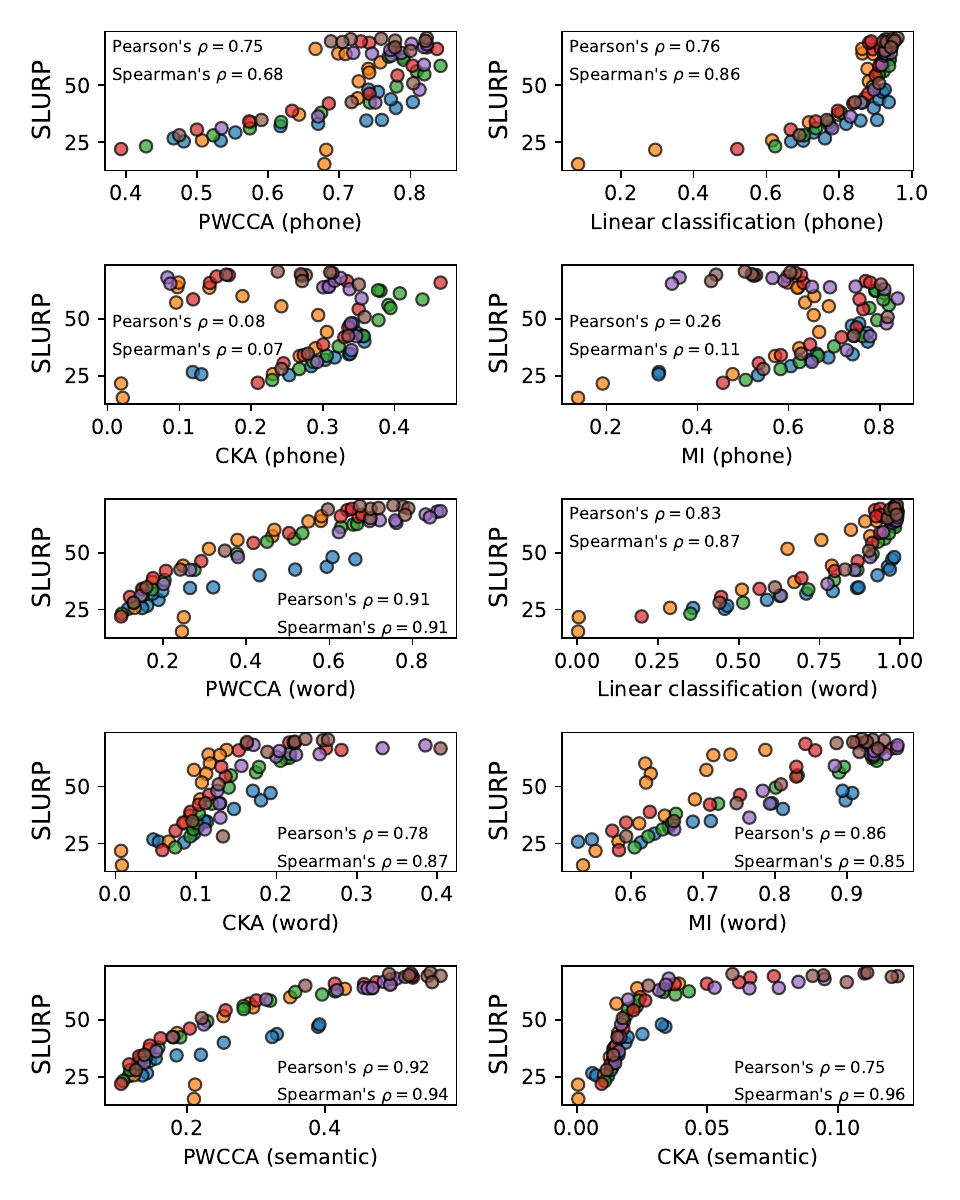}
\caption{Scatter plots comparing SLURP performance with task-agnostic layer-wise trends for all \sfms.
SLURP is measured as accuracy (in \%), \pwcca \ and \cka \ shown as similarity scores, \mi \ as normalized MI score, and linear classification as classification accuracy.}
\label{fig:appendix-slurp_scenario-all}
\end{figure}

\begin{figure}[htb]
\includegraphics[width=0.9\textwidth]{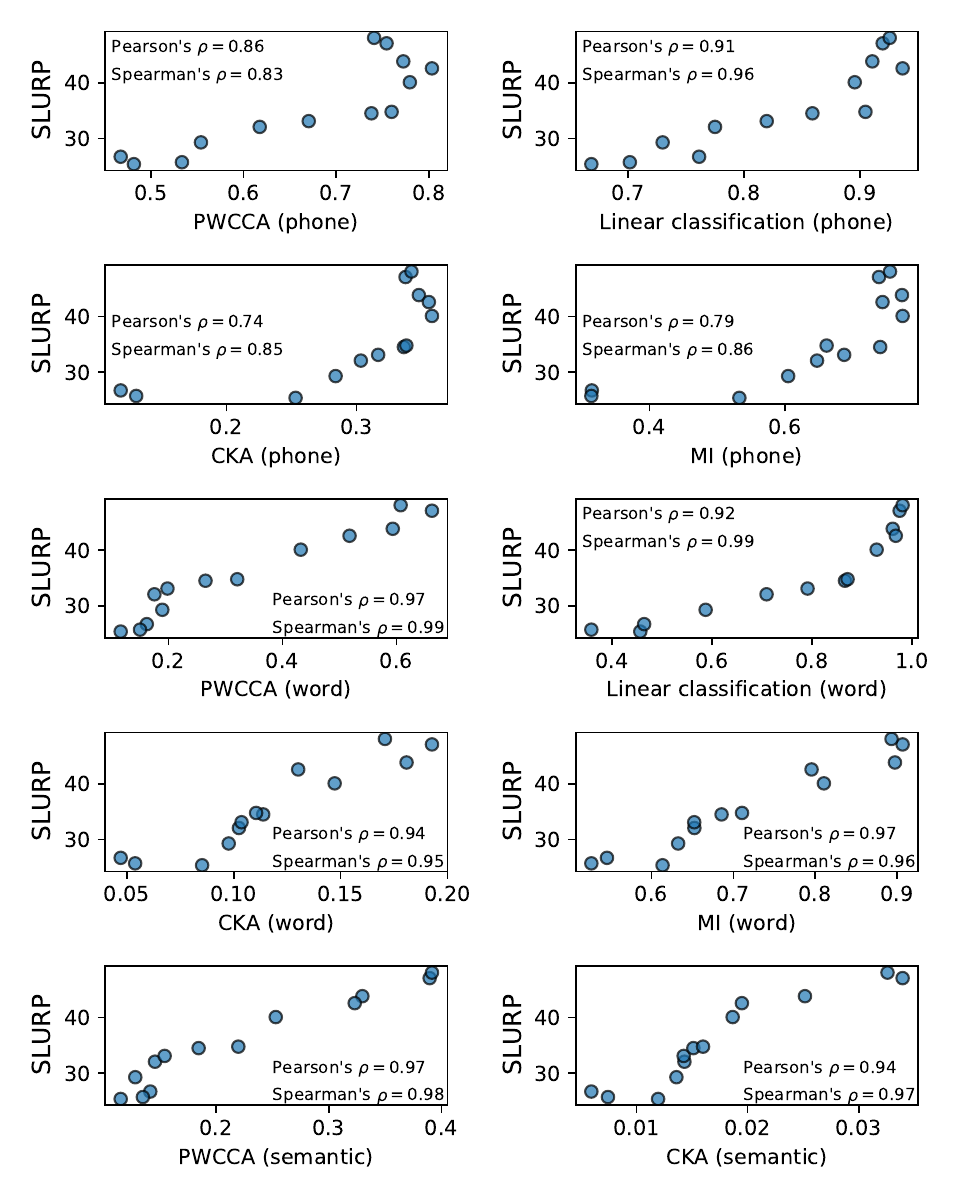}
\caption{Scatter plots comparing SLURP performance with task-agnostic layer-wise trends for \wavtovec-\baseM.
SLURP is measured as accuracy (in \%), \pwcca \ and \cka \ shown as similarity scores, \mi \ as normalized MI score, and linear classification as classification accuracy.}
\label{fig:appendix-slurp_scenario-wav2vec_small}
\end{figure}

\begin{figure}[htb]
\includegraphics[width=0.9\textwidth]{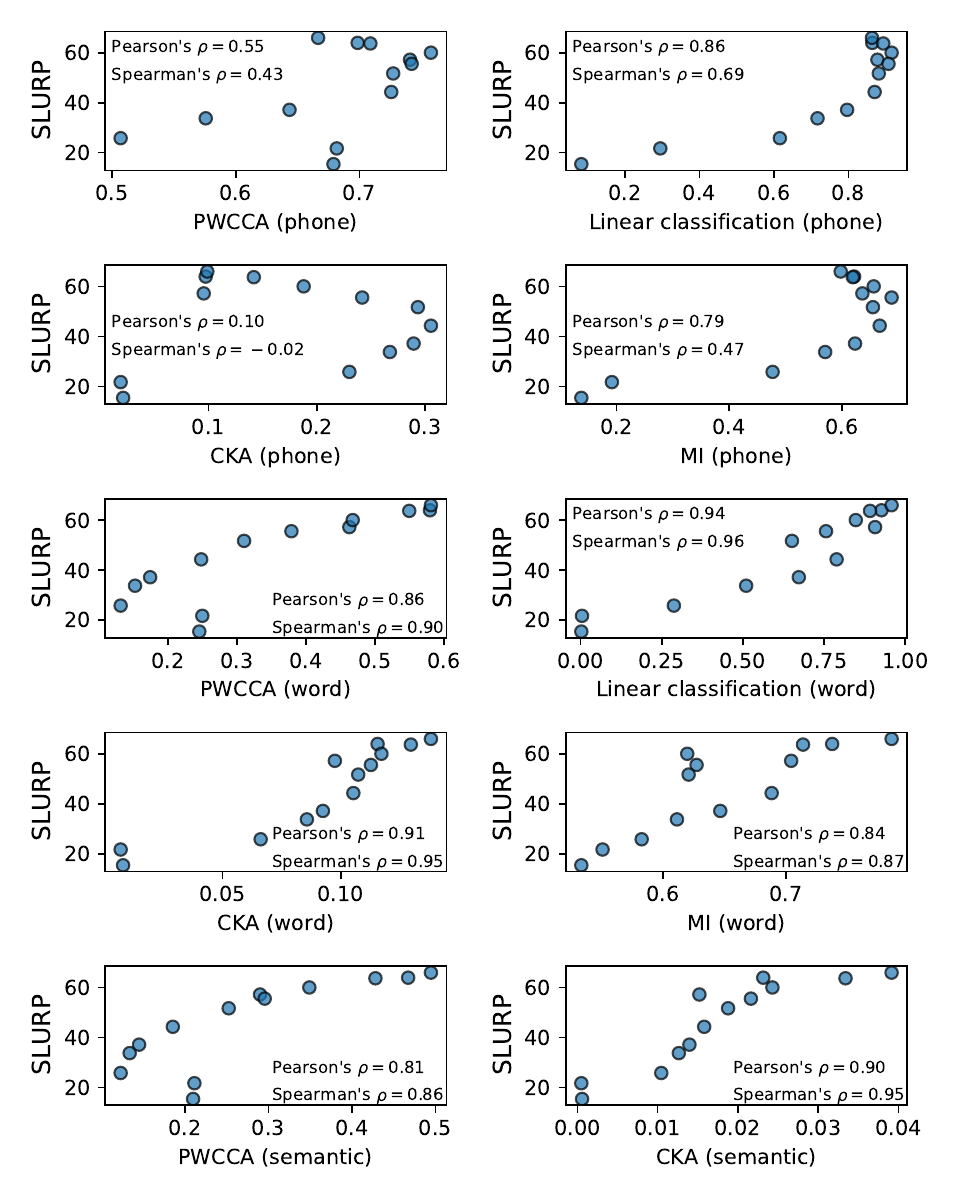}
\caption{Scatter plots comparing SLURP performance with task-agnostic layer-wise trends for \wavtovec-\voxM.
SLURP is measured as accuracy (in \%), \pwcca \ and \cka \ shown as similarity scores, \mi \ as normalized MI score, and linear classification as classification accuracy.}
\label{fig:appendix-slurp_scenario-wav2vec_vox}
\end{figure}

\begin{figure}[htb]
\includegraphics[width=0.9\textwidth]{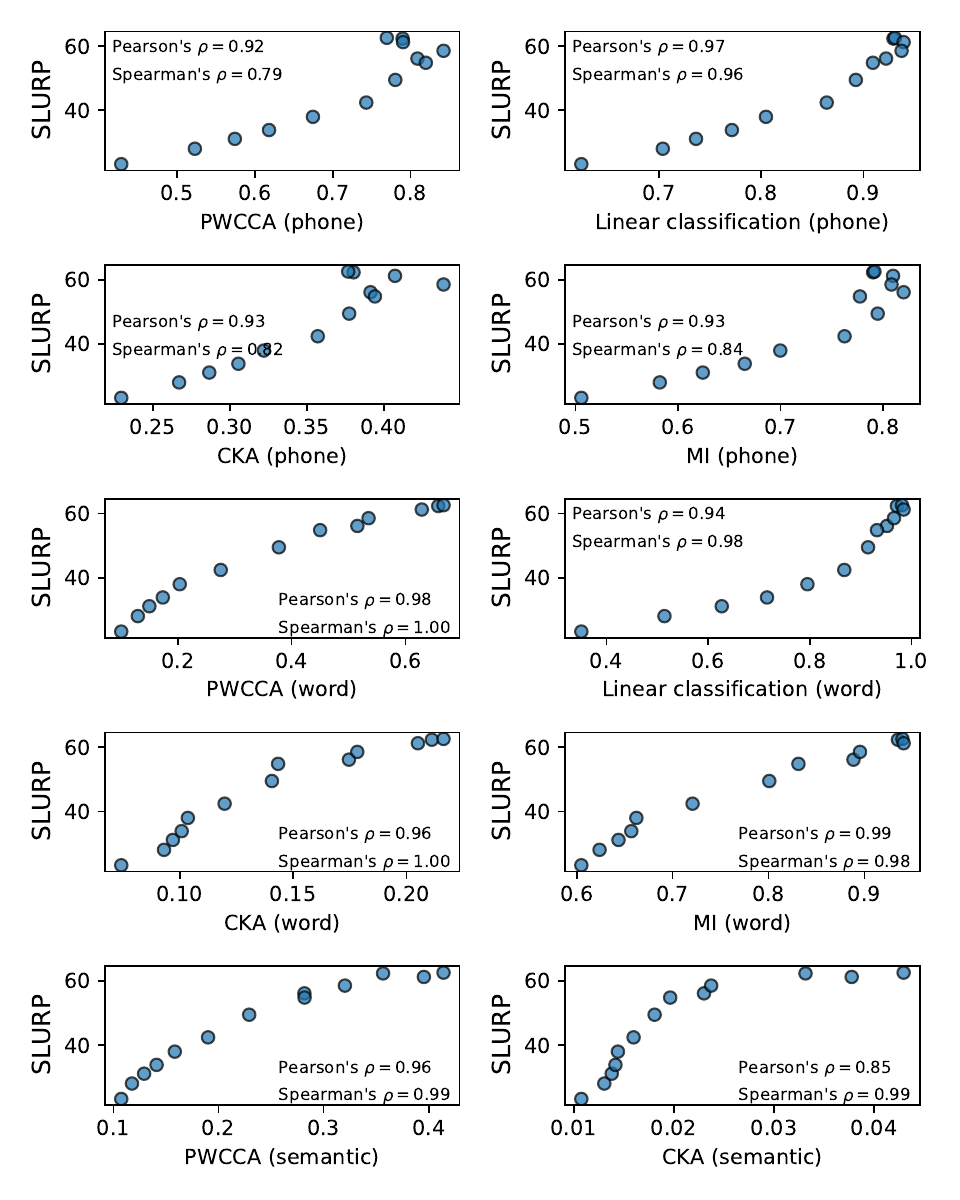}
\caption{Scatter plots comparing SLURP performance with task-agnostic layer-wise trends for \hubert-\baseM.
SLURP is measured as accuracy (in \%), \pwcca \ and \cka \ shown as similarity scores, \mi \ as normalized MI score, and linear classification as classification accuracy.}
\label{fig:appendix-slurp_scenario-hubert_small}
\end{figure}

\begin{figure}[htb]
\includegraphics[width=0.9\textwidth]{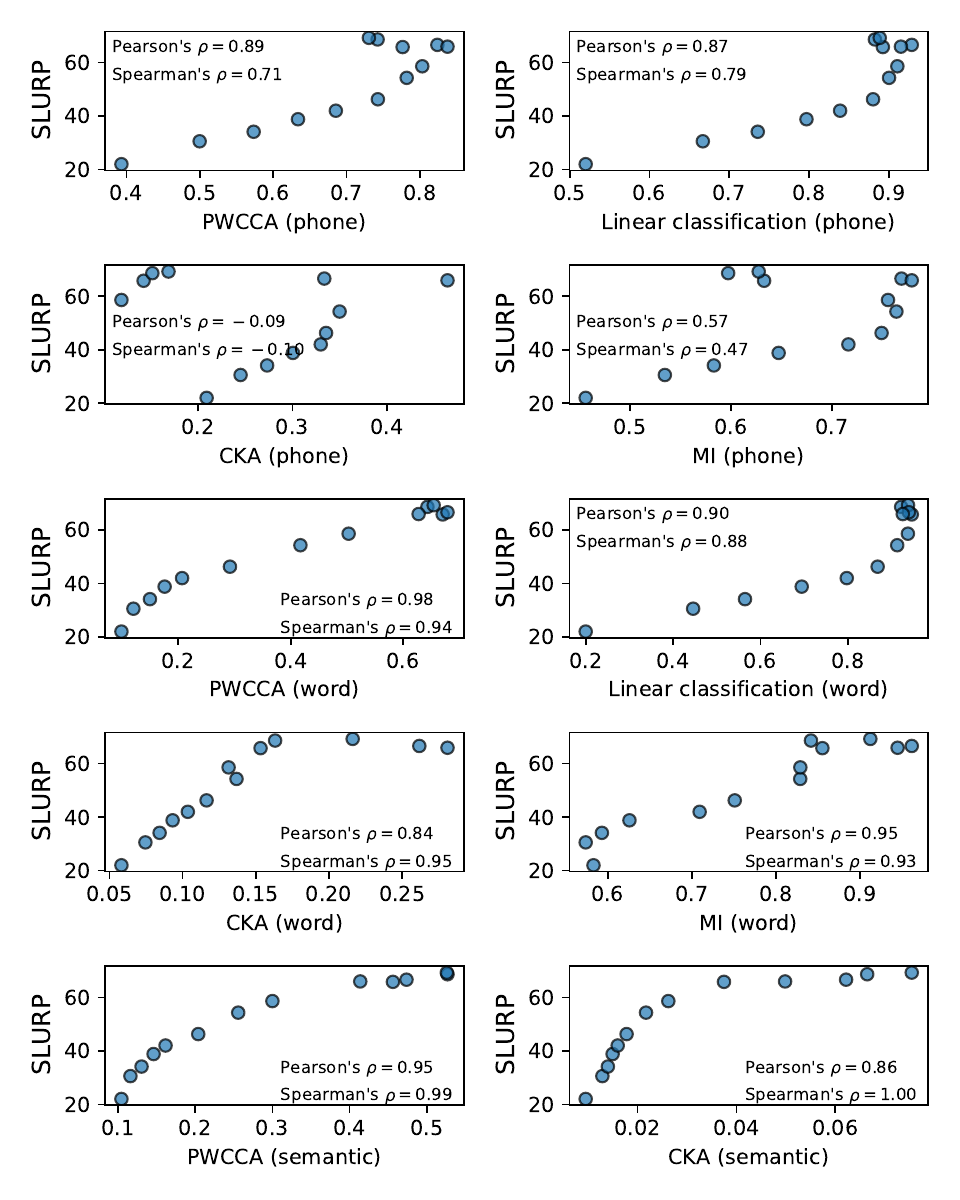}
\caption{Scatter plots comparing SLURP performance with task-agnostic layer-wise trends for \hubert-\largeM.
SLURP is measured as accuracy (in \%), \pwcca \ and \cka \ shown as similarity scores, \mi \ as normalized MI score, and linear classification as classification accuracy.}
\label{fig:appendix-slurp_scenario-hubert_large}
\end{figure}

\begin{figure}[htb]
\includegraphics[width=0.9\textwidth]{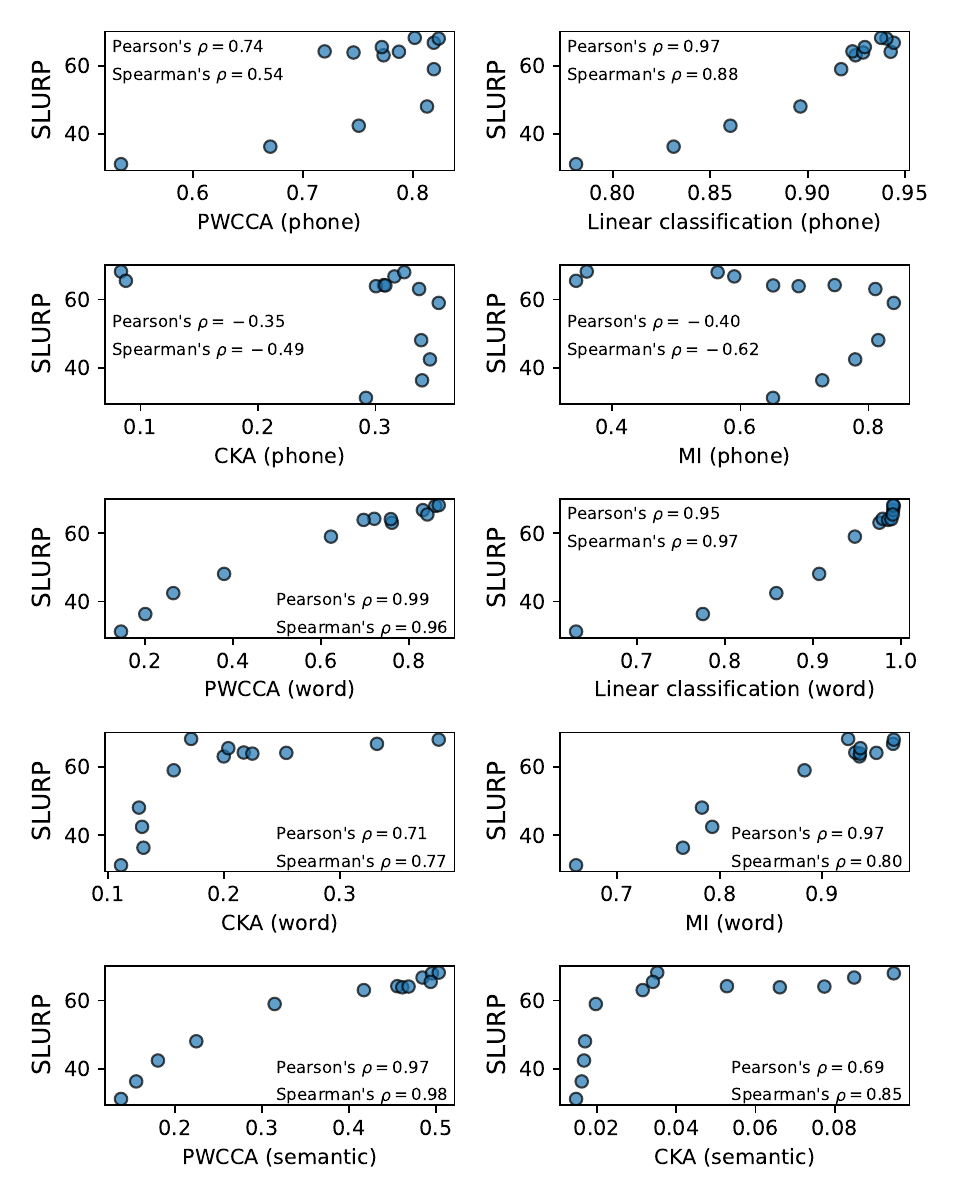}
\caption{Scatter plots comparing SLURP performance with task-agnostic layer-wise trends for \datatovec-\baseM.
SLURP is measured as accuracy (in \%), \pwcca \ and \cka \ shown as similarity scores, \mi \ as normalized MI score, and linear classification as classification accuracy.}
\label{fig:appendix-slurp_scenario-data2vec_small}
\end{figure}

\begin{figure}[htb]
\includegraphics[width=0.9\textwidth]{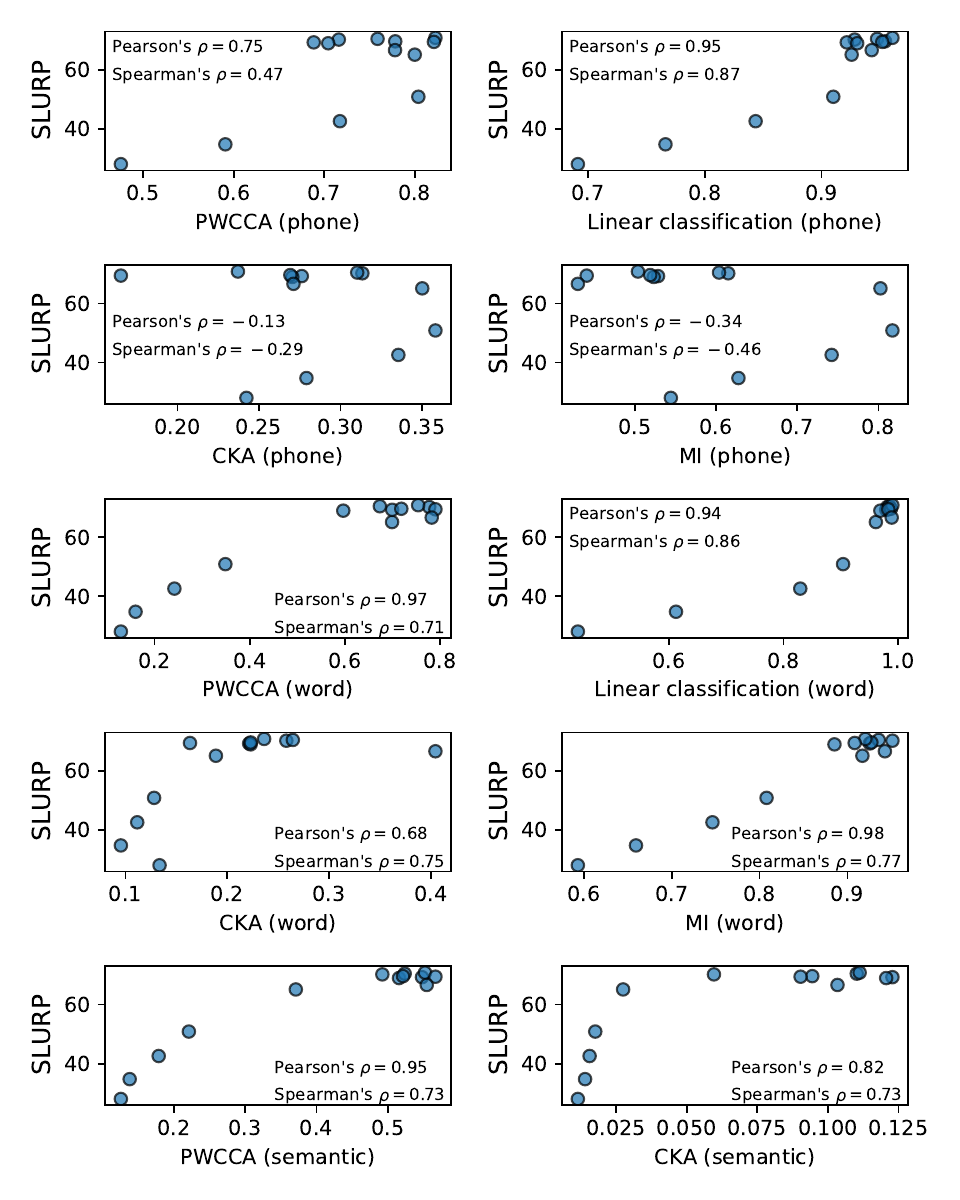}
\caption{Scatter plots comparing SLURP performance with task-agnostic layer-wise trends for \datatovec-\largeM.
SLURP is measured as accuracy (in \%), \pwcca \ and \cka \ shown as similarity scores, \mi \ as normalized MI score, and linear classification as classification accuracy.}
\label{fig:appendix-slurp_scenario-data2vec_large}
\end{figure}

Figures~\ref{fig:appendix-phone-all}-\ref{fig:appendix-slurp_scenario-data2vec_large} present scatter plots between layer-wise task-specific performance and task-agnostic analysis scores. These are additional results for discussion in \sect~\ref{sec:res-compare-tools-tasks}.

\clearpage

\clearpage

\clearpage

\clearpage

\clearpage

\clearpage

\clearpage

\clearpage

\clearpage

\clearpage

\clearpage

\clearpage

\clearpage

\clearpage

\clearpage

\clearpage

\clearpage

\clearpage

\clearpage

\clearpage

\clearpage

\clearpage

\clearpage

\clearpage

\clearpage

\clearpage

\clearpage

\chapter{Implications for Task-Specific Adaptation}
\tab~\ref{tab:lora-res-test} presents DAC and SLURP results on the respective test sets. Relevant discussion and dev set results are presented in \sect~\ref{sec:res-implications-peft} (\tab~\ref{tab:lora-res-dev}).

\begin{table}[htb]
\centering
\small
\caption{Comparison of DAC macro F1 and SLURP scenario accuracy across different LoRA configurations.}
\label{tab:lora-res-test}
\begin{tabular}{l|l|c|l|c|l}
\hlineB{2}
\multicolumn{2}{c|}{} & \multicolumn{2}{c|}{\textbf{DAC}} & \multicolumn{2}{c}{\textbf{SLURP scenario}} \\
\cline{3-6}
\multicolumn{2}{c|}{\multirow{-2}{*}{\textbf{Method}}} & \textbf{macro F1} & \begin{tabular}{c} \textbf{LoRA} \\ \textbf{placement} \end{tabular} & \textbf{Accuracy} & \begin{tabular}{c} \textbf{LoRA} \\ \textbf{placement} \end{tabular} \\
\hline
\multirow{2}{*}{Baselines} 
  & \afl & \cellgradsg{66.3} & N/A & \cellgradsh{62.7} & N/A \\
  & \tff    & \cellgradsg{69.9} & N/A & \cellgradsh{88.4} & N/A \\
\hline
\multirow{6}{*}{\shortstack[l]{\# LoRA\\layers}}
  & 1   & \cellgradsg{67.7} & 12 & \cellgradsh{68.5} & 12 \\
  & 2   & \cellgradsg{68} & 11,12 & \cellgradsh{70.6} & 11,12 \\
  & 3   & \cellcolor{verylightgreen}65.5 & 1,3,10 & \cellgradsh{71.7} & 10,11,12 \\
  & 4   & \cellgradsg{69.5} & 1,7,11,12 & \cellgradsh{75.6} & 2,8,11,12 \\
  & 4+  & \cellgradsg{68.2} & 1,2,3,10,11,12 & \cellgradsh{77.2} & 1,2,3,10,11,12 \\
  & ALL & \cellgradsg{67.9} & ALL & \cellgradsh{78.5} & ALL \\
\hlineB{2}
\end{tabular}
\end{table}

\chapter{Spoken Language Understanding Evaluation Benchmark}

\section{Dataset details}
\label{sec:slue-appendix-data}
\tab~\ref{tab:voxpopuli-stats-detail} presents a distribution of the entity labels, both raw and combined, across train, dev, and test data splits.
Discussed in \sect~\ref{sec:slue-ner-data} in the main text.

\begin{table}[h]
\centering
\small
\caption{SLUE-VoxPopuli NER label statistics}
\begin{tabular}{l|l|l|l}
\hlineB{2}
\begin{tabular}[c]{@{}c@{}}Combined\\ label\end{tabular} & \begin{tabular}[c]{@{}c@{}}Raw label\\ (ontonotes5)\end{tabular} & \begin{tabular}[c]{@{}c@{}}\# of NER phrases\\ (fine-tune/dev/test)\end{tabular} & \begin{tabular}[c]{@{}c@{}}\# of distinct NER phrases\\ (fine-tune/dev/test)\end{tabular} \\\hlineB{2}
\multirow{2}{*}{PLACE} & GPE &1641 / 500 / 560 &162 / 96 / 120 \\
& LOC &371 / 142 / 171 &56 / 27 / 34 \\
\hline
\multirow{5}{*}{QUANT} & CARDINAL &584 / 193 / 171 &137 / 68 / 84 \\
& ORDINAL &267 / 110 / 57 &19 / 14 / 10 \\
& MONEY &60 / 18 / 8 &52 / 16 / 6 \\
& PERCENT &3 / 3 / 2 &3 / 2 / 2 \\
& QUANTITY &9 / 3 / 8 &9 / 2 / 8 \\
\hline
\multirow{1}{*}{ORG} & ORG &864 / 259 / 273 &255 / 100 / 126 \\
\hline
\multirow{2}{*}{WHEN} & DATE &723 / 259 / 179 &327 / 154 / 119 \\
& TIME &39 / 1 / 7 &22 / 1 / 6 \\
\hline
\multirow{1}{*}{NORP} & NORP &647 / 220 / 348 &128 / 60 / 91 \\
\hline
\multirow{1}{*}{PERSON} & PERSON &272 / 51 / 81 &201 / 44 / 59 \\
\hline
\multirow{1}{*}{LAW} & LAW &250 / 60 / 96 &156 / 41 / 75 \\
\hline
\multirow{5}{*}{Discarded tags} & EVENT &73 / 37 / 21 &49 / 23 / 19 \\
& FAC &10 / 2 / 4 &9 / 2 / 4 \\
& LANGUAGE &2 / 0 / 11 &1 / 0 / 7 \\
& PRODUCT &2 / 3 / 6 &2 / 2 / 6 \\
& WORK OF ART &3 / 1 / 3 &3 / 1 / 3 \\
\hline
\end{tabular}%
\label{tab:voxpopuli-stats-detail}
\end{table}

\section{Annotation details}
\label{sec:slue-appendix-annotation}

\subsection{Named entity recognition}
\label{sec:slue-appendix-annotation-ner}
\begin{figure}[h]
    \centering
    \includegraphics[width=0.5\linewidth]{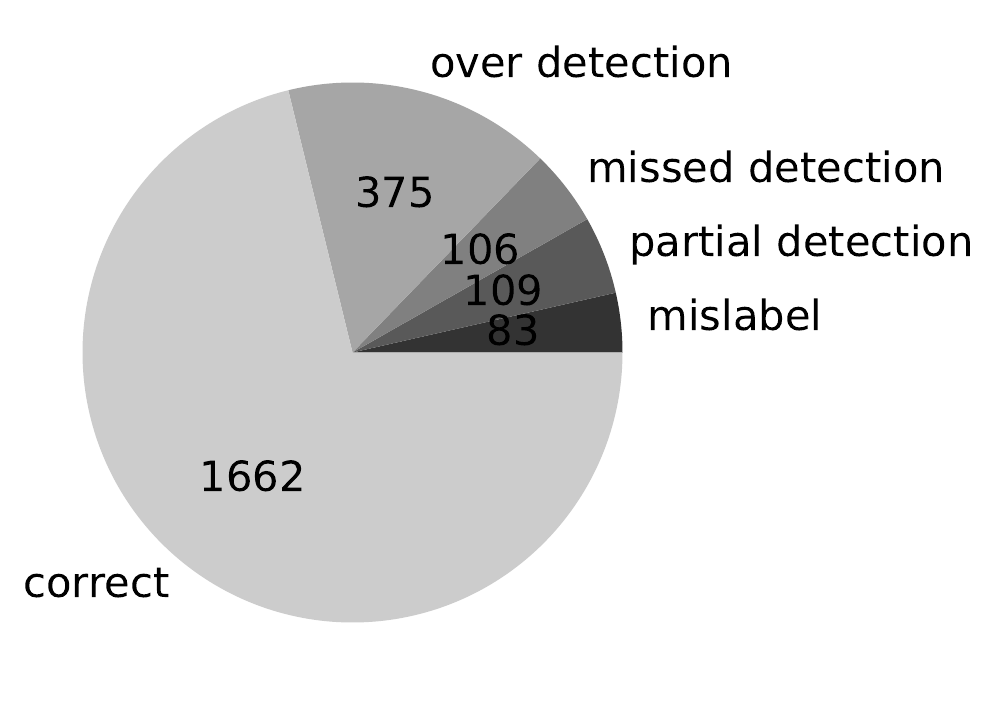}
    \caption{Classification of disagreements between the two annotation passes on the SLUE-VoxPopuli test set. 
    }
    \label{fig:ner-annotation-pie}
\end{figure}

\fig~\ref{fig:ner-annotation-pie} presents a fine-grained comparison between two NER annotation passes.
Discussed in \sect~\ref{sec:slue-ner-data} in the main text.

\subsection{Named entity localization}
\label{sec:slue-appendix-annotation-nel}
As described in \sect~\ref{sec:slue-nel-data}, we use MFA to obtain ground-truth word-level alignments. When we run MFA, it fails to align twenty-six files across dev and test splits. On manual inspection we identify differences in audio utterance and the corresponding text transcript due to incorrect end-pointing for twenty-two of these files. These cases have contiguous words at the end of the transcript that are not a part of the audio utterance. Running MFA after removing these extra words from the transcripts fixes these cases. But, for seven of these files, at least one entity word is a part of the missing words and so, the time alignments don't have all the entity phrases that are a part of the published SLUE-NER annotations. In the interest of utterance-level consistency between SLUE-NER and SLUE-NEL, we skip these files. For the remainder four of the twenty-six files that MFA fails to align, we manually add the word alignments using Praat software~\cite{Boersma2009}.

In order to check the validity of MFA produced alignments, we manually verify the entity alignments for 372 entity phrases across randomly chosen 188 utterances in dev split. This constitutes 20\% of all entity phrases in the dev split and thus our analysis should be representative for the complete split. Our manual pass exposed 51 of 372 phrases to be misaligned and the nature of misalignment varied from a minor offset to being completely off. In order to quantify the effect of the identified misalignments on our evaluation metrics, we manually rectify the alignments for these 51 phrases and report the following scores for this representative set of 188 utterances:
\begin{enumerate}
    \item The frame-F1 between rectified and original timestamps is 96\%,
    \item The relative difference in baseline model scores (evaluating models listed in Table~\ref{tab:nel-baseline}) using these two versions as ground-truths is $<3$\%,
    \item The general trend in baseline model scores is similar across models for the results using these two versions as ground-truths.
\end{enumerate}

Thus, we conclude that the alignments produced by MFA are reliable for robustly comparing between different modeling approaches and can be used as ground-truth despite minor issues in the generated time-stamps.

Additionally, we find that the faulty timestamps are a result of imperfect transcripts in VoxPopuli and not an issue with MFA. The imperfections in these transcripts are expected, since the data is originally curated with 20\% character error rate threshold~\cite{wang2021voxpopuli}.

\section{NEL hyperparameter details}
\label{sec:slue-appendix-nel-hyperparameters}

As described in \sect~\ref{sec:slue-baselines-nel}, NEL evaluation uses two hyperparameters,{\emph offset} and \emph{incl\_blank}. We evaluate the dev set on a range of offset values between -0.3 seconds and 0.3 seconds with an increment of 20 milliseconds. The \emph{incl\_blank} is a Boolean hyperparameter. The best hyperparameter values based on dev set performance are listed in Table~\ref{tab:nel_bestparams}.

The 34 NeMo models have one of the three types of decoding strategies -- (i) character-level CTC, (ii) subword-level CTC, and (iii) subword-level RNN transducer (RNNT).
The character-level CTC models are processed in the same way as the {\it pipeline-w2v2} models, where the \emph{incl\_blank} denotes whether or not the $\epsilon$ tokens before and after the entity phrase, between the word separator tokens, are included in the entity time stamp. 
The subword-level CTC model vocabulary in the NeMo toolkit does not have a word separator token, and instead, the start of the word is characterized by an ``\_'' prepended to a subword. So, the \emph{incl\_blank} denotes whether the trailing $\epsilon$ tokens, before the start of the next word, are included in the entity time stamp.
The RNNT model class in the NeMo toolkit directly gives subword-level start times, so \emph{offset} was the only relevant hyperparameter here. 

\begin{table}[hbp!]
\caption{Best hyperparameters for NEL models}\label{tab:nel_bestparams}\centering
\resizebox{16cm}{!}{%
\begin{tabular}{lllll}\toprule
System & Speech model & \begin{tabular}[c]{@{}c@{}}Training\\ objective\end{tabular} & offset (s) & incl\_blank \\\midrule
E2E-w2v2 & wav2vec2 & char-CTC & 0.00 & True \\\hline
pipeline-w2v2 & wav2vec2 & char-CTC & -0.08 & True \\\hline
\multirow{3}{*}{pipeline-nemo} & QuartzNet15x5Base-En & \multirow{3}{*}{char-CTC} & -0.22 & True \\
  & stt\_en\_jasper10x5dr &   & -0.26 & True \\
  & stt\_en\_quartznet15x5 &   & -0.26 & True \\\hline
\multirow{6}{*}{pipeline-nemo} & stt\_en\_citrinet\_1024 & \multirow{6}{*}{subword-CTC} & -0.10 & True \\
  & stt\_en\_citrinet\_1024\_gamma\_0\_25 &   & -0.10 & True \\
  & stt\_en\_citrinet\_256 &   & -0.10 & True \\
  & stt\_en\_citrinet\_256\_gamma\_0\_25 &   & 0.00 & True \\
  & stt\_en\_citrinet\_512 &   & -0.12 & True \\
  & stt\_en\_citrinet\_512\_gamma\_0\_25 &   & -0.16 & True \\\hline
\multirow{7}{*}{pipeline-nemo} & stt\_en\_conformer\_ctc\_large & \multirow{7}{*}{subword-CTC} & -0.12 & True \\
  & stt\_en\_conformer\_ctc\_large\_ls &   & -0.02 & False \\
  & stt\_en\_conformer\_ctc\_medium &   & -0.12 & True \\
  & stt\_en\_conformer\_ctc\_medium\_ls &   & -0.02 & False \\
  & stt\_en\_conformer\_ctc\_small &   & -0.08 & True \\
  & stt\_en\_conformer\_ctc\_small\_ls &   & 0.00 & False \\
  & stt\_en\_conformer\_ctc\_xlarge &   & -0.08 & True \\\hline
\multirow{6}{*}{pipeline-nemo} & stt\_en\_squeezeformer\_ctc\_large\_ls & \multirow{6}{*}{subword-CTC} & -0.02 & False \\
  & stt\_en\_squeezeformer\_ctc\_medium\_large\_ls &   & -0.02 & False \\
  & stt\_en\_squeezeformer\_ctc\_medium\_ls &   & -0.02 & False \\
  & stt\_en\_squeezeformer\_ctc\_small\_ls &   & -0.02 & False \\
  & stt\_en\_squeezeformer\_ctc\_small\_medium\_ls &   & -0.02 & False \\
  & stt\_en\_squeezeformer\_ctc\_xsmall\_ls &   & -0.02 & False \\\hline
\multirow{6}{*}{pipeline-nemo} & stt\_en\_conformer\_transducer\_large & \multirow{6}{*}{subword-RNNT} & 0.16 & n/a \\
  & stt\_en\_conformer\_transducer\_large\_ls &   & 0.14 & n/a \\
  & stt\_en\_conformer\_transducer\_medium &   & 0.20 & n/a \\
  & stt\_en\_conformer\_transducer\_small &   & 0.20 & n/a \\
  & stt\_en\_conformer\_transducer\_xlarge &   & 0.18 & n/a \\
  & stt\_en\_conformer\_transducer\_xxlarge &   & 0.18 & n/a \\\hline
\multirow{6}{*}{pipeline-nemo} & stt\_en\_contextnet\_1024 & \multirow{6}{*}{subword-RNNT} & 0.22 & n/a \\
  & stt\_en\_contextnet\_1024\_mls &   & 0.30 & n/a \\
  & stt\_en\_contextnet\_256 &   & 0.14 & n/a \\
  & stt\_en\_contextnet\_256\_mls &   & 0.20 & n/a \\
  & stt\_en\_contextnet\_512 &   & 0.22 & n/a \\
  & stt\_en\_contextnet\_512\_mls &   & 0.30 & n/a \\
\bottomrule
\end{tabular}
}
\end{table}

\section{Additional results}
\label{sec:slue-appendix-results}
We report NER and NEL results on the SLUE-VoxPopuli dev set, along with additional results that support analysis presented in \sect~\ref{sec:slue-results}. Corresponding test set results are also presented and discussed in \sect~\ref{sec:slue-results} in the main text.

\subsection{Named entity recognition}
\label{sec:slue-appendix-results-ner}
\tabs~\ref{tab:ner-baseline-dev} reports results on SLUE-VoxPopuli dev set. 

\begin{table}[htbp]
\centering
\small
\caption{Named entity recognition performance on SLUE-VoxPopuli dev set.}
\begin{tabular}{lcc|cc}
\hlineB{2}
\multicolumn{1}{c}{Speech model} & LM & Text model & F1 (\%)   \\
\hlineB{2} 
\textbf{Pipeline-oracle:}  & & &  \\
\hspace{3mm} \multirow{1}{*}{ N/A (GT Text)} & \multirow{1}{*}{ N/A} & DeBERTa-L  & 87.5 \\
\hlineB{2}
\textbf{Pipeline approaches:}  & & &  \\
\hspace{2mm} \wavtovec-\baseM & - & DeBERTa-L & 55.2 \\
\hspace{2mm} \wavtovec-\largeM & - & DeBERTa-L & 65.0 \\
\hspace{2mm} \wavtovec-\baseM & \cmark & DeBERTa-L & 73.8 \\
\hspace{2mm} \wavtovec-\largeM & \cmark & DeBERTa-L & 76.7 \\
\hlineB{2}
\textbf{E2E approaches:}  & & &  \\
\hspace{2mm} \wavtovec-\baseM & - & \multirow{6}{*}{\normalsize N/A} & 55.0 \\
\hspace{2mm} \hubert-\baseM & - &  & 54.5 \\
\hspace{2mm} \wavtovec-\largeM & - &  & 56.6 \\
\hspace{2mm} \wavtovec-\baseM & \cmark &  & 68.1 \\
\hspace{2mm} \hubert-\baseM & \cmark &  & 67.8 \\
\hspace{2mm} \wavtovec-\largeM & \cmark &  & 70.3 \\
\hlineB{2}
\end{tabular}
\label{tab:ner-baseline-dev}
\end{table}

\subsection{Named entity localization}
\label{sec:slue-appendix-results-nel}

Complementing \tab~\ref{tab:nel-baseline} from the main text, \tab~\ref{tab:nel_baseline_all} shows performance of NEL for dev and test sets across different thresholds for word-F1. For word-F1, relaxing the tolerance from $\rho=1$ to $\rho=0.8$ gives a major performance boost -- up to 30\% and 116\% relative for pipeline and E2E models respectively.

\begin{figure}[htbp!]
    \centering
    \includegraphics[width=0.7\linewidth]{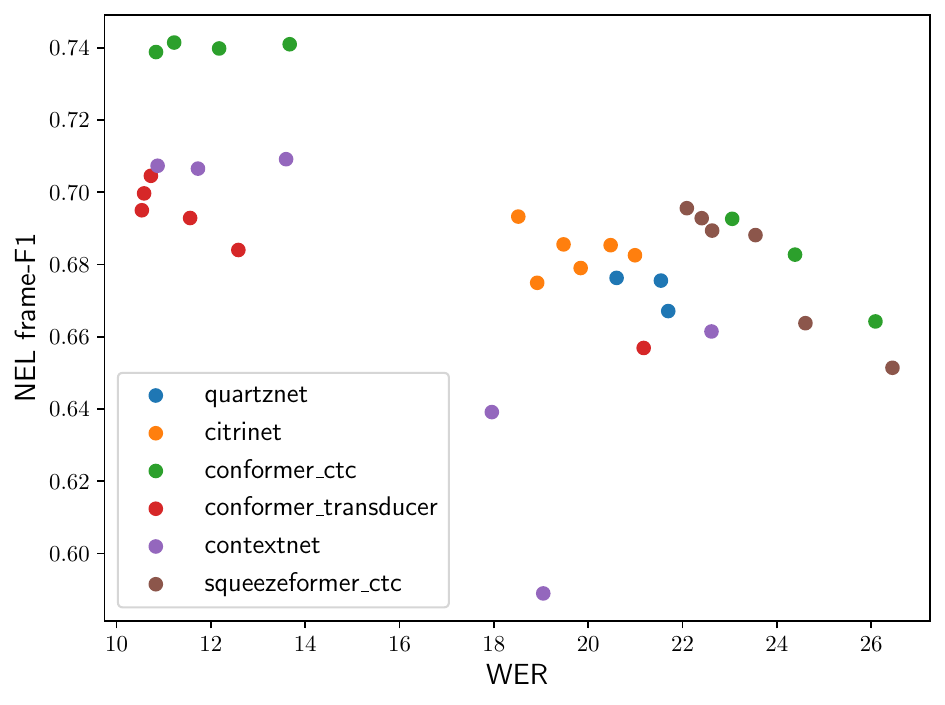}
    \caption{WER and frame-F1 scores on test set for different NeMo models}
    \label{fig:nel_nemo_corr}
\end{figure}

\fig~\ref{fig:nel_nemo_corr} presents a color-coded version of the original scatter plot (\fig~\ref{fig:f1-wer-scatter}) to highlight the groups of models that share the same architecture.

\begin{table}[hbp!]\centering
\small
\caption{NEL task baseline performance. The wav2vec2 models are fine-tuned on slue-voxpopuli data.*the best NeMo model based on NEL frame-f1 score on dev is ``stt\_en\_conformer\_ctc\_small".
}
\label{tab:nel_baseline_all}
\resizebox{\columnwidth}{!}{%
\begin{tabular}{lcc|cc|cc|cc|cc}\toprule
\multirow{2}{*}{System} &\multirow{2}{*}{\makecell{Speech\\model}} & \multirow{2}{*}{\makecell{Text\\model}} & \multicolumn{2}{c|}{frame-F1} & \multicolumn{2}{c|}{word-F1 ($\rho$=1)} & \multicolumn{2}{c|}{word-F1 ($\rho$=0.8)} & \multicolumn{2}{c}{word-F1 ($\rho$=0.5)} \\\cmidrule{4-11}
& & & Dev & Test & Dev & Test & Dev & Test & Dev & Test \\\midrule
pipeline-oracle & x & DeBERTa & 92.3 & 89.0 & 93.6 & 90.0 & 93.6 & 90.0 & 93.6 & 90.0 \\
pipeline-w2v2 & wav2vec2 & DeBERTa & 66.9 & 65.1 & 56.0 & 53.6 & 72.7 & 72.1 & 75.9 & 74.1 \\
E2E-w2v2 & wav2vec2 & x & 63.2 & 56.2 & 30.8 & 25.7 & 66.5 & 59.4 & 71.8 & 64.6 \\
pipeline-nemo & best model* & DeBERTa & 75.5 & 74.1 & 66.9 & 64.0 & 83.4 & 81.4 & 83.7 & 81.0 \\
\bottomrule
\end{tabular}
}
\end{table}

\section{Error analysis}
\label{sec:slue-appendix-error-analysis}
In the main text we discuss that the pipeline and E2E NEL baselines have similar frame-F1 scores (\tab~\ref{tab:nel-baseline}). We dig deeper and study precision and recall values in Table~\ref{tab:nel_prec_recall_analysis}. We notice that E2E and pipeline approaches have complementary strengths and weaknesses.

\begin{table*}[hbp!]
\centering
\small
\caption{NEL task baseline precision and recall performance on dev set. *the best nemo model based on NEL frame-f1 score on dev is ``stt\_en\_conformer\_ctc\_small".}
\label{tab:nel_prec_recall_analysis}
\resizebox{\columnwidth}{!}{%
\begin{tabular}{lcc|cc|cc|cc|cc}\toprule
\multirow{2}{*}{System} &\multirow{2}{*}{\makecell{Speech\\model}} & \multirow{2}{*}{\makecell{Text\\model}} & \multicolumn{2}{c|}{frame-F1} & \multicolumn{2}{c|}{word-F1 ($\rho$=1)} & \multicolumn{2}{c|}{word-F1 ($\rho$=0.8)} & \multicolumn{2}{c}{word-F1 ($\rho$=0.5)} \\\cmidrule{4-11}
& & & Prec. & Recall & Prec. & Recall & Prec. & Recall & Prec. & Recall \\\midrule
pipeline-oracle & x & DeBERTa & 91.7 & 92.8 & 92.4 & 94.7 & 92.4 & 94.7 & 92.4 & 94.7 \\
pipeline-w2v2 & wav2vec2 & DeBERTa & 57.8 & 78.8 & 70.4 & 46.4 & 71.1 & 74.1 & 68.5 & 84.9 \\
E2E-w2v2 & wav2vec2 & x & 81.0 & 51.7 & 71.8 & 19.5 & 83.8 & 55.0 & 83.2 & 63.2 \\
pipeline-nemo & best model* & DeBERTa & 69.2 & 83.2 & 82.4 & 56.4 & 83.7 & 83.1 & 79.7 & 88.1 \\
\bottomrule
\end{tabular}
}
\end{table*}

The E2E model significantly outperforms in {\it precision} (i.e, more predicted regions are named entities), whereas the pipeline models consistently outperforms in {\it recall}. 
We hypothesize that the mismatch in text NER's training (ground-truth text) and inference (ASR output) could lead to higher false positives in the pipeline model. 
\chapter{On the Use of External Data for Spoken Named Entity Recognition}

\section{Results on the test set}
\label{sec:appendix-ner-ext-test-set}
\begin{figure}[h]
    \centering
    \includegraphics[width=0.65\linewidth, trim=0 65 0 85, clip]{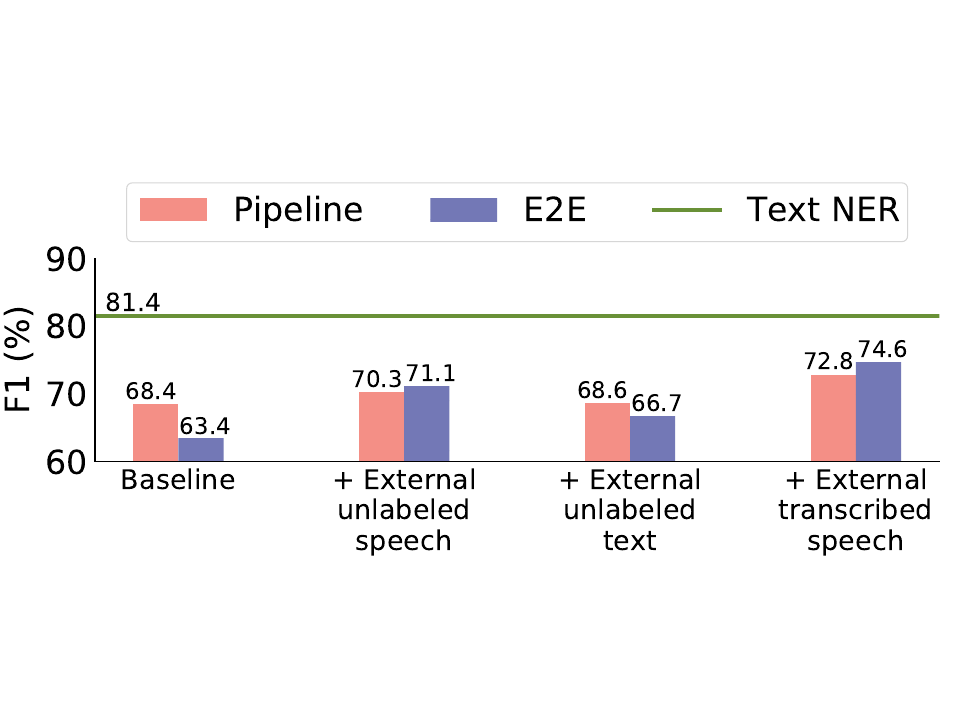}
    \caption{Spoken NER test set results with 100 hours of external data of different types. The ``Baseline'' and ``Text NER'' numbers are from~\cite{shon2022slue}.}
    \label{fig:summary-test}
\end{figure}

We obtain test set results for our best-performing models, by submitting model outputs following the SLUE instructions.\footnote{\href{https://asappresearch.github.io/slue-toolkit/how-to-submit.html}{https://asappresearch.github.io/slue-toolkit/how-to-submit.html}} These results are presented in \fig~\ref{fig:summary-test}. We observe similar trends as on the dev set (see \fig~\ref{fig:ner-ext-summary}).

\begin{figure}[htb]
\begin{minipage}[b]{1.0\linewidth}
\small
 \centering
 \centerline{\includegraphics[width=0.6\linewidth, trim=0 75 0 100, clip]{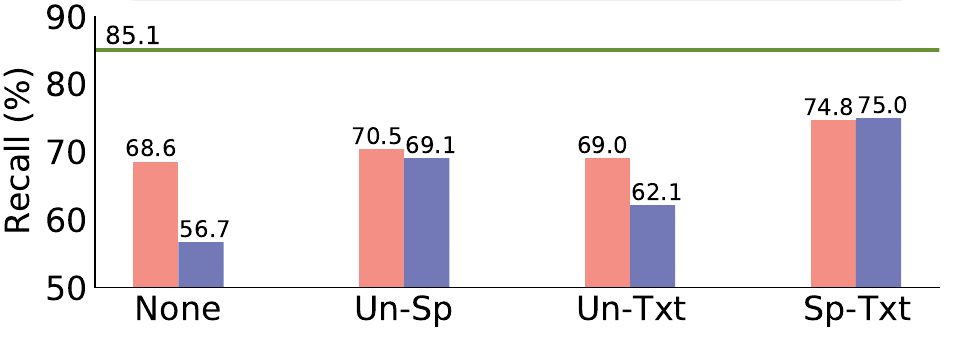}}
\end{minipage}
\begin{minipage}[b]{1.0\linewidth}

\footnotesize
 \centering
  \centerline{\includegraphics[width=0.6\linewidth, trim=0 65 0 75, clip]{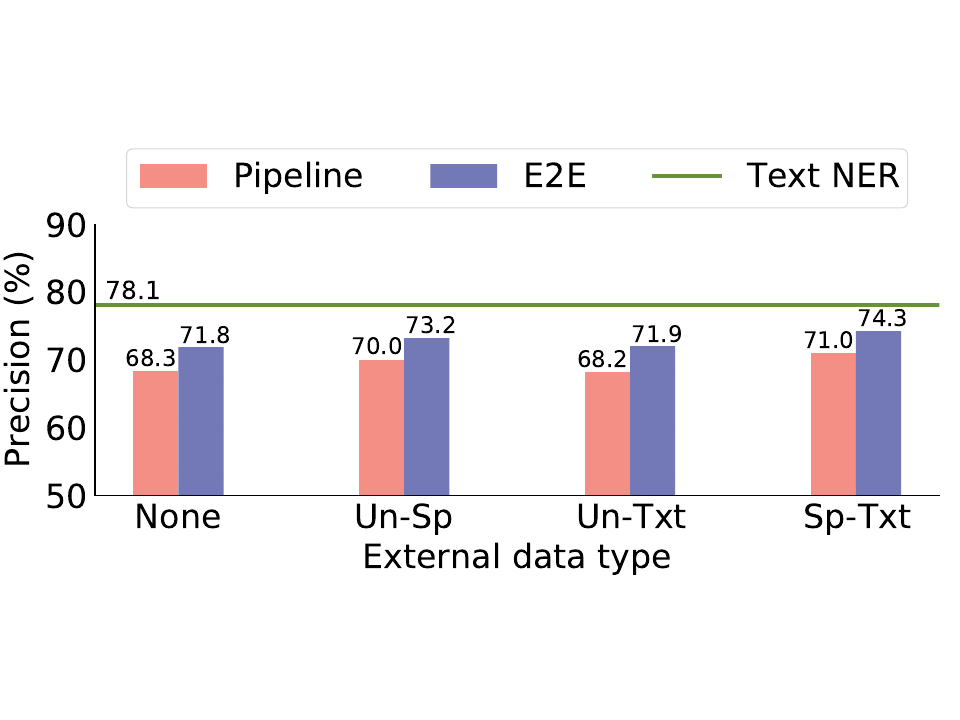}}
\end{minipage}

\caption{{Recall and precision on the test set for the best performing models using 100 hours of external data.}}
  \label{fig:prec-recall-test}
\end{figure}

We can see from the precision and recall scores in  \fig~\ref{fig:prec-recall-test} that our analytical conclusions about the pipeline model performing poorly due to false positives are consistent across these two splits.

\section{Error categories}
\label{sec:appendix-ner-ext-error-analysis}
\begin{figure}[htb]
\begin{minipage}[b]{1.0\linewidth}
\small
 \centering
 \centerline{\includegraphics[width=\linewidth]{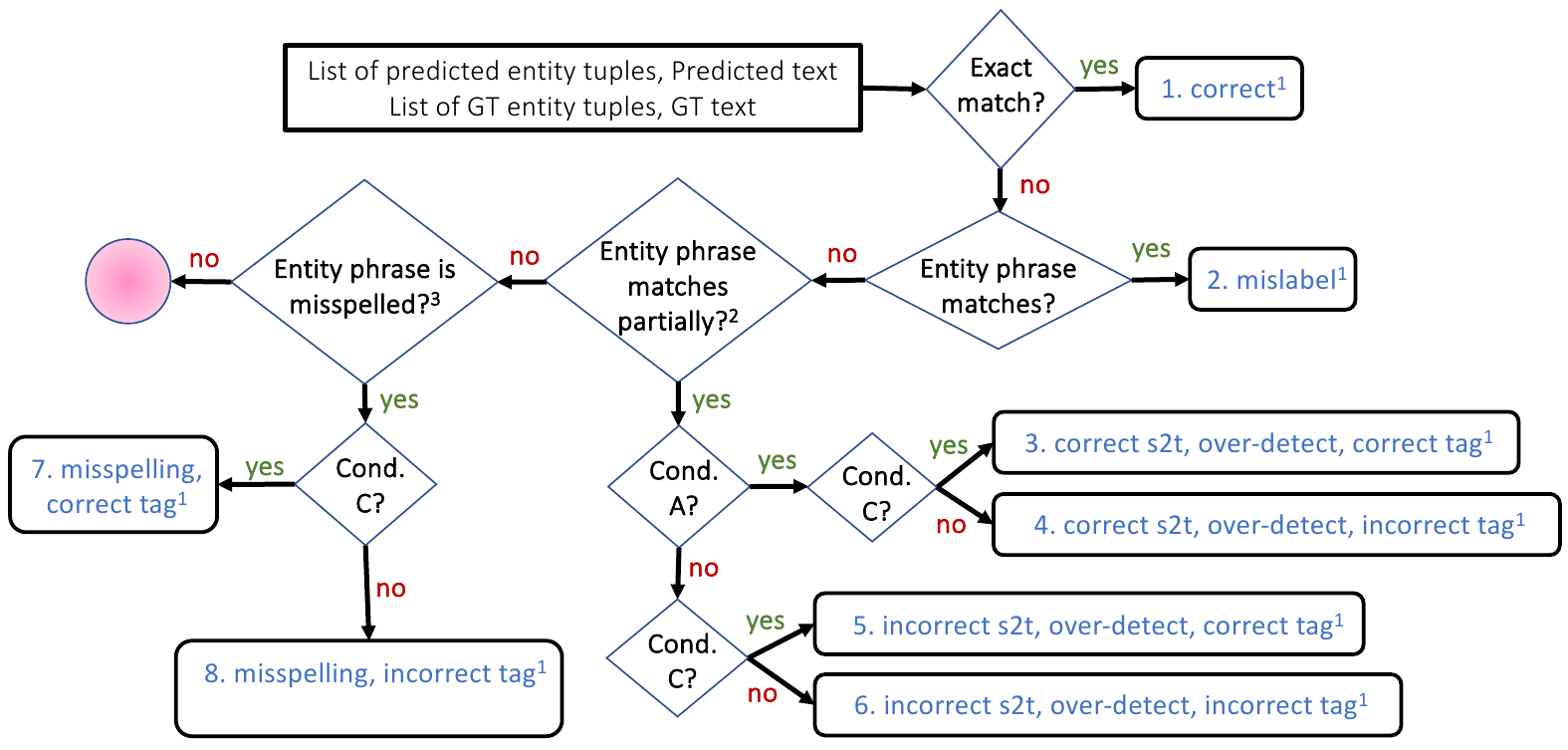}}
\end{minipage}
\begin{minipage}[b]{1.0\linewidth}

\vspace{1cm}
\footnotesize
 \centering
  \centerline{\includegraphics[width=\linewidth]{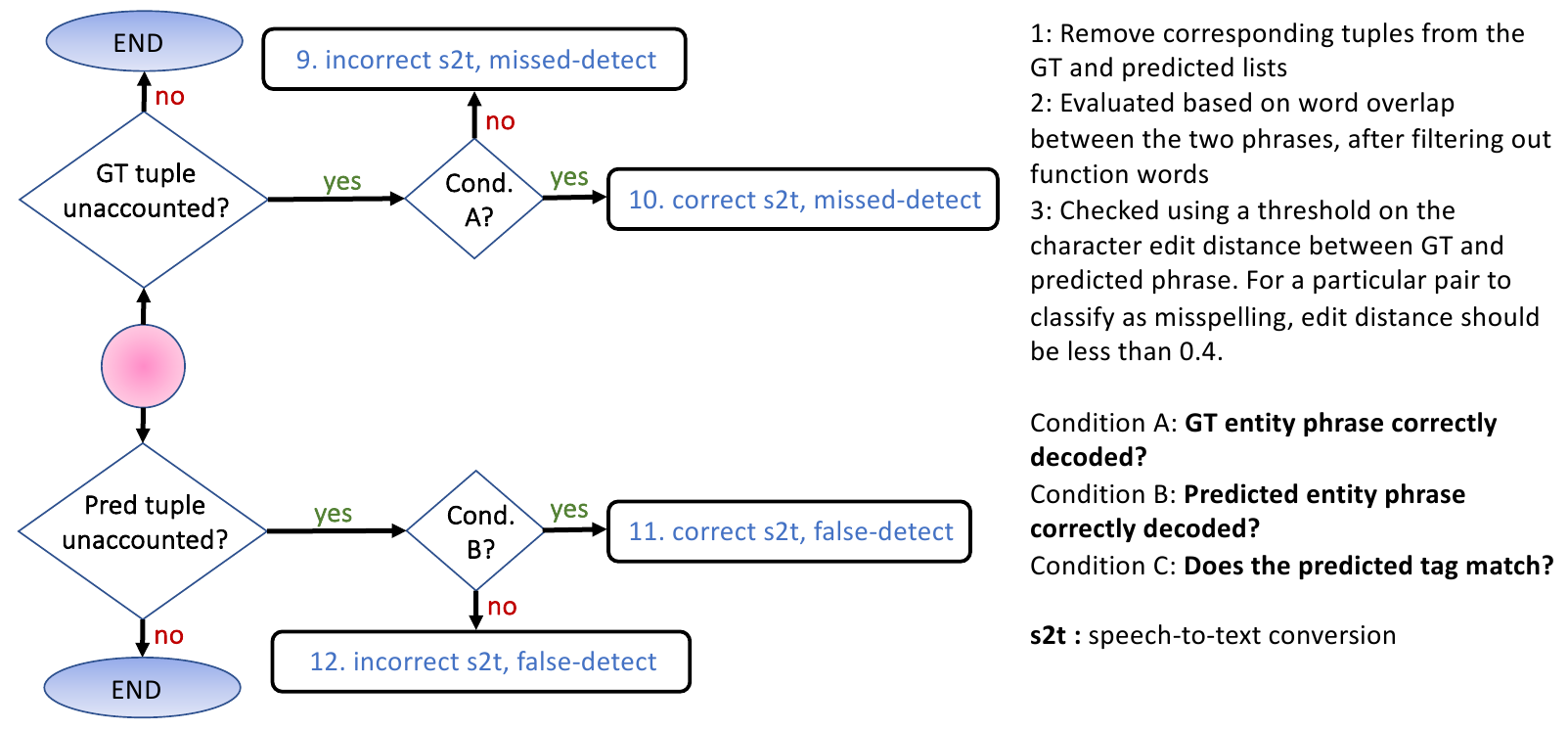}}
\end{minipage}

\caption{{Illustration of algorithm for obtaining error category types for each (entity phrase, entity tag) tuple in ground-truth and predicted outputs.}}
  \label{fig:flowchart}
\end{figure}
\fig~\ref{fig:flowchart} illustrates, via a flowchart, our process of assigning the tuples in ground-truth and predicted outputs into different error categories. The main text discusses our findings in \sect~\ref{sec:ner-ext-analysis}.
\tab~\ref{tab:qualitative-examples} presents examples for the four categories discussed in \sect~\ref{sec:ner-ext-analysis}. These are examples from the dev set, using the {\it Distill-Pipeline} E2E model trained on 100 hours of data.

\newcolumntype{C}[1]{>{\centering\arraybackslash}p{#1}}
\newcolumntype{L}[1]{>{\raggedright\arraybackslash}p{#1}}
\newcolumntype{M}[1]{>{\raggedright\arraybackslash}m{#1}}
\begin{table}[htb]
\small
\centering
\caption{Qualitative examples for different error categories from the output of the E2E model using 100 hours of unlabeled speech (\it Distill-Pipeline).}
\resizebox{\columnwidth}{!}{%
\begin{tabular}{M{3cm}|C{6cm}|C{6cm}}
\toprule 
\multicolumn{1}{c|}{\multirow{2}{*}{\textbf{Error category}}} & \multicolumn{2}{c}{\textbf{Outputs from E2E model}} \\
\cline{2-3}
\multicolumn{1}{c|}{} & \textbf{GT} & \multicolumn{1}{|c}{\textbf{Predicted}} \\
\midrule
\begin{minipage}[c]{\linewidth}
Correct ASR,\\
over detection
\end{minipage}
 & 
and this means that you look and tell us honestly what does it mean if you start @ three years {]} later \par
\bigskip
{[}(`WHEN', `three years'){]} &
this means that you look and tell us honestly what does it mean if you start @ three years later {]} \par
\bigskip
{[}(`WHEN', `three years later'){]} \\
\midrule
\begin{minipage}[c]{\linewidth}
Correct ASR,\\
missed detection
\end{minipage}
&
the situation in the \% drc {]} is indeed terrible and it has been this way for quite a while and i am deeply concerned about the handling of the current issue with regard to the \% kasai {]} province \par
\bigskip
{[}(`PLACE', `drc'), \par
(`PLACE', `kasai'){]}
&
the situation in the drc is indeed terrible and it has been this way for for quite a while and i am deeply concerned about the handling of the current issue with regard to the a province \par
\bigskip
{[}{]} \\
\midrule
\begin{minipage}[c]{\linewidth}
Correct ASR,\\
false detection
\end{minipage}
&
and yet @ one month {]} after we adopted our compromise the council did not put it on the agenda did not even present  it i used this time to talk to the member states and the presidencies \par
\bigskip
{[}(`WHEN', `one month'){]}
&
still @ one month {]} after we voted a compromise the ` council {]} did not put it on the agenda did not even present i use this time to talk with the member states and the presidency \par
\bigskip
{[}(`WHEN', `one month'), \par
(`ORG', `council'){]}
\\
\midrule
\begin{minipage}[c]{\linewidth}
Incorrect ASR,\\
false detection
\end{minipage}
&
it has nothing to do with religion but it has all to do with patriarchy \par
\bigskip
{[}{]}
&
it has nothing to do with religion but it has all to do with \% turkey {]} \par
\bigskip
{[}(`PLACE', `turkey'){]}
\\
\bottomrule
\end{tabular}
}
\label{tab:qualitative-examples}
\end{table}

\chapter{On the Evaluation of Speech Foundation Models for Spoken Language Understanding}
\label{sec:appendix-slueperb}

\begin{table*}[t]
  \centering
  \caption{Performance of various \sfms \ and adaptation strategies on the dev set of SLUE-VoxPopuli for NER, ASR, and NEL tasks; darker shades correspond to better scores and lighter shades correspond to poorer scores. The suffix {\it -L} and {\it -M} for \sfms \ indicate \largeM \ and \mediumM \ sizes respectively. {\it Size} indicates the number of trainable parameters in millions.}
\begin{tabular}{cc|r|cccc}
\toprule
\multirow{2}{*}{\shortstack{Adaptation \\ strategy}} & \multirow{2}{*}{SFM} & \multirow{2}{*}{\centering Size (M)} & \multicolumn{2}{c}{NER} & ASR & NEL \\
\cmidrule(r){4-5}\cmidrule(r){6-6}\cmidrule(r){7-7}
 &  & & label F1 $\uparrow$ & F1 $\uparrow$ & WER $\downarrow$ & frame F1 $\uparrow$ \\
\midrule
\multirow{6}{*}{\shortstack{{\it Frozen} \sfm \\ with a \\ {\it lightweight} \\ prediction head}} & HuBERT-L & 6.5 & \cellgradsc{81.8} & \cellgradsd{64.6} & \cellgradse{13.8} & \cellgradsf{70.9} \\
& wav2vec2.0-L & 6.5 & \cellgradsc{79.9} & \cellgradsd{64.5} & \cellgradse{15.4} & \cellgradsf{68.4} \\
 & WavLM-L & 6.5 & \cellgradsc{87.4} & \cellgradsd{71.4} & \cellgradse{10.2} & \cellgradsf{74.1} \\
& Whisper-M & 9.1 & \cellgradsc{85.8} & \cellgradsd{68.9} & \cellgradse{12.0} & \cellgradsf{73.5} \\
& OWSM 3.1 & 9.1 & \cellgradsc{84.6} & \cellgradsd{69.2} & \cellgradse{12.6} & \cellgradsf{73.1} \\
& Pre-trained SLU & 92.3 & \cellcolor{verylightgreen}{66.6} & \cellcolor{verylightgreen}{50.8} & \cellcolor{verylightgreen}{37.7} & \cellcolor{verylightgreen}{52.2} \\
\midrule
\multirow{6}{*}{\shortstack{{\it Frozen} \sfm \\ with a \\ {\it complex} \\ prediction head}} & HuBERT-L & 32.4 & \cellgradsc{84.6} & \cellgradsd{69.4} & \cellgradse{12.6} & \cellgradsf{72.7} \\
&  wav2vec2.0-L & 32.4 & \cellgradsc{83.1} & \cellgradsd{68.9} & \cellgradse{13.1} & \cellgradsf{74.0} \\
& WavLM-L & 32.4 & \cellgradsc{87.9} & \cellgradsd{74.1} & \cellgradse{9.5} & \cellgradsf{74.7} \\
& Whisper-M & 32.4 & \cellgradsc{86.1} & \cellgradsd{69.9} & \cellgradse{12.7} & \cellgradsf{73.9} \\
& OWSM 3.1 & 35.0 & \cellgradsc{84.8} & \cellgradsd{72.2} & \cellgradse{12.0} & \cellgradsf{70.7} \\
& Pre-trained SLU & 92.3 & \cellcolor{verylightgreen}{73.8} & \cellcolor{verylightgreen}{61} & \cellcolor{verylightgreen}{27.5} & \cellcolor{verylightgreen}{57.8} \\
\midrule
\multirow{6}{*}{\shortstack{{\it Fine-tuned} \sfm \\ with a \\ {\it lightweight} \\ prediction head}} & HuBERT-L & 318.9 & \cellgradsc{84.3} & \cellgradsd{68.2} & \cellgradse{11.6} & \cellgradsf{73.0} \\
&  wav2vec2.0-L & 319.7 & \cellgradsc{84.6} & \cellgradsd{70.4} & \cellgradse{11.3} & \cellgradsf{71.1} \\
& WavLM-L & 317.8 & \cellgradsc{88.3} & \cellgradsd{73.5} & \cellgradse{9.3} & \cellgradsf{73.9} \\
& Whisper-M & 314.8 & \cellgradsc{82.3} & \cellgradsd{65.5} & \cellgradse{16.7} & \cellcolor{verylightgreen}{56.3} \\
& OWSM 3.1 & 569.9 & \cellgradsc{83.7} & \cellgradsd{68.3} & \cellgradse{13.7} & \cellgradsf{66.9} \\
& Pre-trained SLU & 92.3 & \cellcolor{verylightgreen}{67.5} & \cellcolor{verylightgreen}{54.1} & \cellcolor{verylightgreen}{35.3} & \cellcolor{verylightgreen}{54.8} \\
\bottomrule
\end{tabular}
\label{tab:slueperb-dev-results}
\end{table*}

\tab~\ref{tab:slueperb-dev-results} presents results on development set, complementing test set results presented in \tab~\ref{tab:slueperb-test-results}. While the performance on development sets is a bit better, the takeaways remain the same as presented in \sect~\ref{sec:slueperb-results}.

\addtocontents{toc}{\protect\setcounter{tocdepth}{2}}

\end{appendices}



\end{document}